\documentclass[runningheads]{llncs}

\usepackage{graphicx}
\usepackage{amssymb}
\usepackage{amsmath}
\usepackage{mathtools}
\usepackage{subcaption}
\usepackage{hyperref}
\usepackage{xcolor}
\usepackage[inline]{enumitem}
\usepackage[colorinlistoftodos,prependcaption,textsize=small,disable]{todonotes}

\title{Phase Transition Behavior in Knowledge Compilation}

\author{Rahul Gupta\inst{1} \and 
Subhajit Roy\inst{1} \and
Kuldeep S. Meel\inst{2}}
\authorrunning{Gupta, Roy, and Meel}
\institute{Indian Institute of Technology Kanpur \and
School of Computing, National University of Singapore}

\DeclarePairedDelimiter\ceil{\lceil}{\rceil}
\DeclarePairedDelimiter\floor{\lfloor}{\rfloor}

\newcommand{\dDNNF}{\text{d-DNNF}}

\newcommand{\D}{\ensuremath{\mathsf{D4}}}

\newcommand{\CUDD}{\ensuremath{\mathsf{CUDD}}}
\newcommand{\sddpackage}{\ensuremath{\mathsf{The SDD Package}}}

\newcommand{\expect}{\ensuremath{\mathsf{E}}}
\newcommand{\lang}{\mathcal{L}}
\newcommand{\bigo}{\mathcal{O}}
\newcommand{\N}{\mathbb{N}}
\newcommand{\R}{\mathbb{R}}

\newcommand{\shortcite}[1]{\cite{#1}}

\begin{document}

\maketitle

\begin{abstract}

    The study of phase transition behaviour in SAT has led to deeper understanding and algorithmic improvements of modern SAT solvers. Motivated by these prior studies of phase transitions in SAT, we seek to study the behaviour of size and compile-time behaviour for random $k$-CNF formulas in the context of \textit{knowledge compilation}. 

    We perform a rigorous empirical study and analysis of the size and runtime behavior for different knowledge compilation forms (and their corresponding compilation algorithms): d-DNNFs, SDDs and OBDDs across multiple tools and compilation algorithms. We employ instances generated from the random $k$-CNF model with varying generation parameters to empirically reason about the expected and median behavior of size and compilation-time  for these languages. Our work is similar in spirit to the early work in CSP community on phase transition behavior in SAT/CSP. In a similar spirit, we identify the interesting behavior with respect to different parameters: clause density and \textit{solution density}, a novel  control parameter that we identify for the study of phase transition behavior in the context of knowledge compilation. Furthermore, we summarize our empirical study in terms of two concrete conjectures; a rigorous study of these conjectures will possibly require new theoretical tools.  

\end{abstract}

\section{Introduction}

Phase transition is concerned with a sudden change in the behavior of a property of interest of an object pertaining to variations of a parameter of interest. In the context of combinatorial problems, the phase transition behavior was first demonstrated in random graphs in the seminal work of Erdos and Renyi~\shortcite{ER60}. With the advent of SAT as a modeling language, the initial studies observed phase transition in the satisfiability of random $k-$CNF formulas and the seminal work of Mitchell, Selman, and Levesque~\shortcite{MSL92} demonstrated empirical hardness around the phase transition region for modern SAT solvers. 
Theoretical investigations into determining the location of the phase transition region have led to several exciting results that yield insights into the algorithmic behavior of modern SAT heuristics~\cite{Achlioptas09}. In a significant theoretical breakthrough, the existence of the phase transition behavior for random $k-$CNF for large $k$ was theoretically proved; the question for small $k$ ($> 2$) is still open~\cite{DSS15}.

The success of SAT solvers has led the development of tools and techniques for problems and sub-fields broadly classified under the umbrella of {\em Beyond NP}. This has led to an interest in the behavior of solvers through the lens of phase transition~\cite{DMV16,DMV17,PJM19}.
Motivated by the success of these studies in uncovering surprising insights into the solution space structure and regions of hardness for the modern SAT solvers, we turn our focus to another sub-field in {\em Beyond NP} that has found several practical applications: {\em Knowledge Compilation}.

Knowledge compilation broadly refers to the approaches that seek to compile propositional formulae into tractable representations; tractability is defined with respect to queries that can be performed in polynomial time over the compiled form. As the runtime of queries such as counting, sampling, equivalence is often polytime in the size of a representation, efficient compilation gains importance. Several compilations forms have been proposed showcasing the tradeoff between tractability and succinctness:  OBDDs (Ordered Binary Decision Diagrams)~\cite{Bryant86}, SDDs (Sentential Decision Diagrams)~\cite{Dar11} and d-DNNFs (deterministic-Decomposable Negation Normal Forms)~\cite{Darwiche01} and others.  We refer the reader to \cite{DM02} for a detailed survey on the size and tractability for several target languages. 
While every tractable language known so far has exponential size complexity in the worst case when the input is presented in CNF form, the size of compiled forms can often be only polynomially larger than the input CNF formula, which has highlighted the need for more detailed study of the size and runtime complexity of compilation procedures. 

In this work, we undertake a rigorous empirical study and analysis of phase transition behavior in knowledge compilation. Our experimental setup employs instances generated from the random $k$-CNF model with varying generation parameters (number of variables and clauses as well as length of clauses). Our study is multidimensional, comparing and contrasting phase transition behaviors spanning:
\begin{itemize}
	\item knowledge compilations: d-DNNFs, SDDs and OBDDs;
	\item properties of interest: size  and compilation times;
	\item compilation algorithms and tools: 
	\begin{itemize}
		\item C2D~\cite{c2d},  D4~\cite{LM17}, DSharp~\cite{Dsharp12} for d-DNNF, 
		\item MiniC2D~\cite{OD15} and TheSDDPackage~\cite{Sdd-package} for SDD, 
		\item CUDD with different variable ordering heursitics~\cite{CUDD,dd} for BDD.
	\end{itemize}
\end{itemize}

A primary contribution of the seminal work of Mitchell et al~\cite{MSL92} was the establishment of clause density as a popular parameter of interest in the study of SAT solving. In a similar spirit, one of the key contributions of this work is the proposal of \textit{solution density} as a new control parameter, along with clause density, for studying knowledge compilations. Solution density is defined as the ratio of the logarithm of the number of satisfying assignments to the number of variables. We show that while clause density is linked with solution density in expectation, the size and compilation-time varies significantly with varying solution density for a fixed clause density. We discover that for low clause densities, varying solution density has minimal effect on the size of compilations. In contrast, for high clause densities, solution density dictates the size as there is a minimal variation with clause density for a given fixed solution density.

Based on our experiments, we make two conjectures for compiled structures in a language $L$ that is a subset of DNNF:
\begin{enumerate}
	\item Over a population of k-CNF formulas, $F_k(n,\ceil{rn})$ of $k$-clauses over $n$ variables with a clause density $r$, for all integers $k \geq 2$, there exists a clause density $r$ that witnesses a phase transition on the size of the compiled structures in a language $L$; that is, there always exists a clause density $r_t$ such that:
	\begin{enumerate}
		\item for each pair $(r_1 , r_2)$, such that $r_1 < r_2 < r_t$, the expected size of the compiled structure in $L$ for $F_k(n,{r_1n})$ is strictly smaller than that for clause density $r_2$;
		\item for each pair $(r_1 , r_2)$, such that $r_t < r_1 < r_2$, the expected size of the compiled structure in $L$ for $F_k(n,{r_1n})$ is strictly larger than that for clause density $r_2$.
	\end{enumerate}
	
	\item Over a population of k-CNF formulas, $G_k(n,\ceil{2^{\alpha n}})$ of $k$-clauses over $n$ variables and having $\ceil{2^{\alpha n}}$ solutions, for all integers $k \geq 2$, there exists a solution density $\alpha_k$ that witnesses a phase transition on the size of the compiled structures in a language $L$; that is, there always exists a solution density $\alpha_t$ such that:
	\begin{enumerate}
		\item for each pair $(\alpha_1, \alpha_2)$ such that $0 \leq \alpha_1 < \alpha_2 < \alpha_k$, the expected size of the compiled structure in $L$ for $G_k(n,2^{\alpha_1 n})$ is strictly smaller than that for solution density $\alpha_2$;
		\item for each pair $(\alpha_1, \alpha_2)$ such that $\alpha_k < \alpha_1 < \alpha_2 \leq 1$, the expected size of the compiled structure in $L$ for $G_k(n,2^{\alpha_1 n})$ is strictly larger than that for solution density $\alpha_2$.
	\end{enumerate} 
\end{enumerate}

The phase-transition behavior for the satisfiability of CNF constraint has been instrumental in driving several breakthroughs in the design of new solvers and better understanding the problem structure.~\cite{AC08,BMZ05}. We hope that our experimental study of phase transitions for knowledge compilations would lead to similar developments in the knowledge compilation community. Our results in this work are similar in spirit to the seminal work by the CSP community in empirical identification of phase transition phenomenon in the early 1990s and their summarization in the form of \todo{check if the citations are fine} conjectures~\cite{CKT91,GMPSW01,GW94,GW99}. It is worth emphasizing that theoretical proofs of these conjectures were presented nearly  20 years since the first empirical studies~\cite{DSS15}, and the efforts to establishing these conjectures contributed to the development of several theoretical tools of widespread applicability~\cite{Achlioptas09}. We hope our empirical results will inspire similar efforts in the context of knowledge compilation.

The organization of the rest of the paper is as follows: Section~\ref{sec:notations} describes the notations and preliminaries, along with a survey of prior work.  Section~\ref{sec:design} describes the design of our experiments while Sections~\ref{sec:CDexp} and \ref{sec:SDexp} provide detailed observations with respect to clause density and solution density respectively.  Due to space restrictions, this article is limited to experiments using the following tools: D4 for d-DNNF, TheSDDPackage for SDD, and CUDD (with SIFT variable reordering) for BDD. We, however, observed similar behaviors across all the other tools and summarise them in Section~\ref{sec:tools}. Furthermore, throughout the article, we present only representative plots and an extended collection of corresponding plots is deferred to Appendix.\todo{Remove the text from full version}

\section{Notations and Preliminaries}
\label{sec:notations}
Let $X = \{x_1, x_2, \dots , x_n\}$ be a set of propositional variables. A literal is a propositional variable ($x_i$) or its negation ($\lnot x_i$). For a \textit{formula} $F$ defined over $X$, a satisfying assignment or \textit{witness} of $F$ is an assignment of truth values to the variables in $X$ such that $F$ evaluates to \texttt{true}. The total number of witnesses for a formula $F$ is denoted by $\#F$.

 Let a $k$-clause be a disjunction of $k$ literals drawn over $X$ without repetition. Given $k\in\N$, $n\in\N$ and $r\in\R_{>0}$, let the random variable $F_k(n,\ceil{rn})$ denote a Boolean formula consisting of the conjunction of $\ceil{rn}$ $k$-CNF clauses selected uniformly and independently from ${n\choose k}2^k$ possible $k$-clauses over $n$ variables. For a given $k$, $k$-CNF denotes the set of all possible boolean formulas $F_k(n, \ceil{rn})$. For any given formula $F$, we denote the \textit{clause density}~($r$) as the ratio of number of clauses to variables and the \textit{solution density}~($\alpha$) as the ratio of logarithm of the number of satisfying assignments to the number of variables, i.e. $\alpha = \log(\#F)/n$. We denote the SAT phase transition clause density for random $k$-CNF by $r_k^p$. Given an $\alpha \in [0,1]$, we define another random variable $G_k(n, \ceil{2^{\alpha n}})$ that denotes a randomly chosen boolean formula from $k$-CNF with $n$ variables and $\ceil{2^{\alpha n}}$ satisfying assignments.

 \subsection{Target Compilation  Languages}
  We briefly describe some of the prominent target compilation languages that we study---d-DNNFs, OBDDs and SDDs~\cite{DM02}.

 \begin{definition}{\cite{DM02}}
 	Let V be the set of propositional variables. A formula in
 	NNF    is a rooted, directed acyclic graph (DAG) where each leaf node
 	is labelled with true, false, $x$ or $\neg x$, $x \in V$; and each internal node is labelled with $\vee$ or $\wedge$ and can have arbitrarily many children.
 \end{definition}

 \textit{Deterministic Decomposable Negation Normal Form} (d-DNNFs), a subset of NNF, satisfies \textit{determinism} (operands of $\lor$ are mutually inconsistent) and \textit{decomposition} (the operands of $\land$ are expressed on a mutually disjoint set of variables).

 \textit{Ordered Binary Decision Diagrams} (OBDD) is a subset of d-DNNFs where the root node is a decision node and the order of decision variables is same for  all paths from the root to a leaf; OBDDs are canonicalised by the variable ordering.  A decision node is either a constant (\texttt{T}/\texttt{F}), or of the form $(X\wedge \alpha)\vee (\neg X \wedge \beta)$, on decision nodes $\alpha$ and $\beta$, and decision variables $X$. 

 \textit{Sentential Decision Diagrams} (SDDs) are a subset of d-DNNFs that hold the properties of \textit{structured decomposability} and \textit{strongly deterministic decompositions}. The idea of structured decomposability is captured by the notion of \textit{vtrees}; vtrees are binary trees whose leaves correspond to variables of the formula and internal nodes mark the decomposition into variables given by their left and right child. Vtrees seek to generalise variable ordering and precisely specify a decomposition scheme to be followed by the corresponding SDD. Strongly deterministic decompositions seek to generalise Shannon's expansion in which decision is made on a single variable. For a formula $F$ where $\forall\,U,V \subseteq vars(F)$ such that $vars({F}) = {U} \cup {V}$ and ${U}\cap {V} = \phi$, representing $F=\left(p_{1}({U}) \wedge s_{1}({V})\right) \vee \ldots \vee\left(p_{n}({U}) \wedge s_{n}({V})\right)$ where $p_{i}$'s and $q_{i}$'s are boolean functions and $p_{i}({U}) \wedge p_{j}({U})= \bot$ for $i \neq j$ captures the idea of {strongly deterministic decompositions}. Here, the decomposition of $vars(F)$ into ${U}$ and ${V}$ is the decomposition scheme.
 
 We denote a target language by $\lang \in $ \{d-DNNF, OBDD, SDD\}. For a formula $F$ compiled to $\lang$, let $\mathcal{N}_{\lang}(F)$ denote the size (in the number of nodes) of the target representation and $\mathcal{T}_{\lang}(F)$ denote the compilation time. The expected size for $k$-CNF with clause density $r$ and target language $\lang$ can, therefore, be represented as $\expect(\mathcal{N}_{\lang}(F_k(n,rn)))$ and the expected time as $\expect(\mathcal{T}_{\lang}(F_k(n,rn)))$.

\subsection{Related Work}
\label{sec:related}

Cheeseman et al.~\cite{CKT91} studied the phase transition behavior for several CSP problems, noting an easy-hard-easy behavior with respect to an order parameter for these problems. For Hamiltonian circuits, graph coloring and k-SAT, the average connectivity of the graph (which translates to clause density for k-SAT) was found to be a good order parameter demonstrating phase transition behavior. Subsequently, exploring phase transition behaviors with various random generative models has drawn considerable attention.~\cite{AKKKMS97,GMPSW01,GW94,GW99,WH94,MZ08}.

Aguirre and Vardi~\shortcite{AV01a} identified an ``easy-hard-less hard" behaviour in OBDD compilations with clause density and discovered a phase transition from polynomial running time to exponential running time for SAT.  Huang and Darwiche~\shortcite{HD05} proposed a new top-down algorithm for OBDD compilation based on the DPLL algorithm typically used for SAT solving and observed that the expected size of OBDD peaks around clause density equal to $2$. 
Later, Gao, Yin, and Xu~\shortcite{GYX11} conducted an experimental study for phase transitions in random $k$-CNF formulas taking d-DNNFs, OBDDs and DFAs (Deterministic Finite Automata) as the target compilation languages. They observe a phase transition behavior with respect to size of compilations and draw a conjecture stating that all subsets of DNNF show a ``small-large-small" behavior with a unique peak. Our study is more comprehensive and we discuss in Section~\ref{sec:CDexp} several behaviors that were not explored by Gao et al. 

 Birnbaum and Lozinskii~\shortcite{BL99} presented a procedure called as Counting Davis-Putnam (CDP) for \#SAT whose trace lies in FBDD, a stricter subset of d-DNNF language.  They found that the median number of recursive calls of CDP reaches its peak when for clause density is $1.2$ for random 3-CNF formulas. A different DPLL extension for solving \#SAT, called Decomposing Davis-Putnam (DDP) leveraging component decomposition, was developed by Bayardo and Pehoushek~\shortcite{BP00}. The trace of DPP-model counter lies in d-DNNF language and the authors observed phase transition around the clause density of $1.5$. 
\section{Design of experiments}\label{sec:design}

We aim to dive deeper into the size and compile-time behavior for d-DNNF, SDD and OBDD compilations owing to their wide range of applications. 
We measure the size of compilations by the number of nodes (edges also display a very similar behavior). To study the phase transition behavior, we identify a new parameter, \textit{solution density}, the ratio of log of the number of satisfying assignments to the number of variables. While it is trivial to generate a random instance $F_k(n,rn)$ given $r$, a direct way to generate $G_k(n,2^{\alpha n})$ given $\alpha$ is unknown. Therefore, to analyze the effect of varying $\alpha$ in compilations, we study the variation in size and compile-time against $\alpha$ for individual instances generated by varying $r$. Our experimental observations with varying clause density are given in Section~\ref{sec:CDexp} while those with solution density follow in Section~\ref{sec:SDexp}.

Note that, in case of SAT which is a decision problem, satisfiability phase transition is often characterized by existence of a unique \textit{cross-over point}, at the intersection of the curves representing probability of \textit{true} (SAT) decision against clause density for different number of variables. 
However, in our case, we are dealing with a functional problem and therefore, we characterize phase transition with respect to the gradient of the function, i.e., when the underlying function achieves a local maxima. More concretely, we will be concerned with the size and runtime of the underlying knowledge compilation form and its corresponding compiler. 

We carried out our experiments on a high performance compute cluster whose each node is an Intel E5-2690 v3 CPU with 24 cores and 96GB of RAM. We utilize a single core per benchmark instance. Note that knowledge compilation is significantly harder and memory intensive than satisfiability, which restricts the scale of our experiments. Therefore, we have conducted experiments up to the largest number of variables for which we could gather all the needed statistics. We utilized more than 40,000 computational hours for our experimentation.

\subsection{The variables of our study}
We conducted our experiments on a large number of random $k$-CNF formulas with a varying number of variables and clauses. For studying variations with clause density, we aggregated the results over $1000$ instances of the random variable $F_k(n,rn)$ for each $r$ given a fixed $n$ and $k$. Here, $r$ is incremented with step size of at least $0.1$ in a range containing 0 to $r_k^p$. For studying variations with solution density, we aggregated the results over at least $5$ instances of the random variable $F_k(n,rn)$ for each $r$ given a fixed $n$ and $k$. Here, $r$ is incremented with step size of $1/n$ in a range from 0 to $r_k^p$. As clause density and solution density are linked in expectation, we vary $r$ finely for study with solution density to facilitate uniform distribution of instances with $\alpha$.
Specifically, a representative subset of the following variations are a part of our study:
\begin{itemize}
	\item number of variables($n$): $20 - 70$ in general,
	upto an exponential number ($\floor{1.3^{37}}$) for small $r$
	\item length of clauses($k$): $2 - 7$
	\item target languages: d-DNNFs, OBDDs, SDDs
	\item knowledge compilers: $\D$ \cite{LM17}, $\CUDD$ \cite{CUDD,dd}, $\sddpackage$ \cite{Sdd-package}
\end{itemize}

\section{Phase transition with clause density}
\label{sec:CDexp}
In this section, we present our observations for variations in size and compile times with d-DNNFs, SDDs, and OBDDs as target languages. We also dive deeper into the complexity of compilations. As noted earlier, for exposition, we will first present plots for a representative subset of the experimental results, and a comprehensive collection of plots appears in the Appendix. \todo{Change for the final version}

\subsection{Observing the phase transitions}
\label{subsec:ObservingPhaseTransitions}

\subsubsection{Size of compilations} 

\begin{figure}[!th]
	\centering
	\captionsetup[subfigure]{labelformat=empty}
	\begin{subfigure}[b]{0.495\linewidth}
		\centering
		\includegraphics[width=\linewidth]{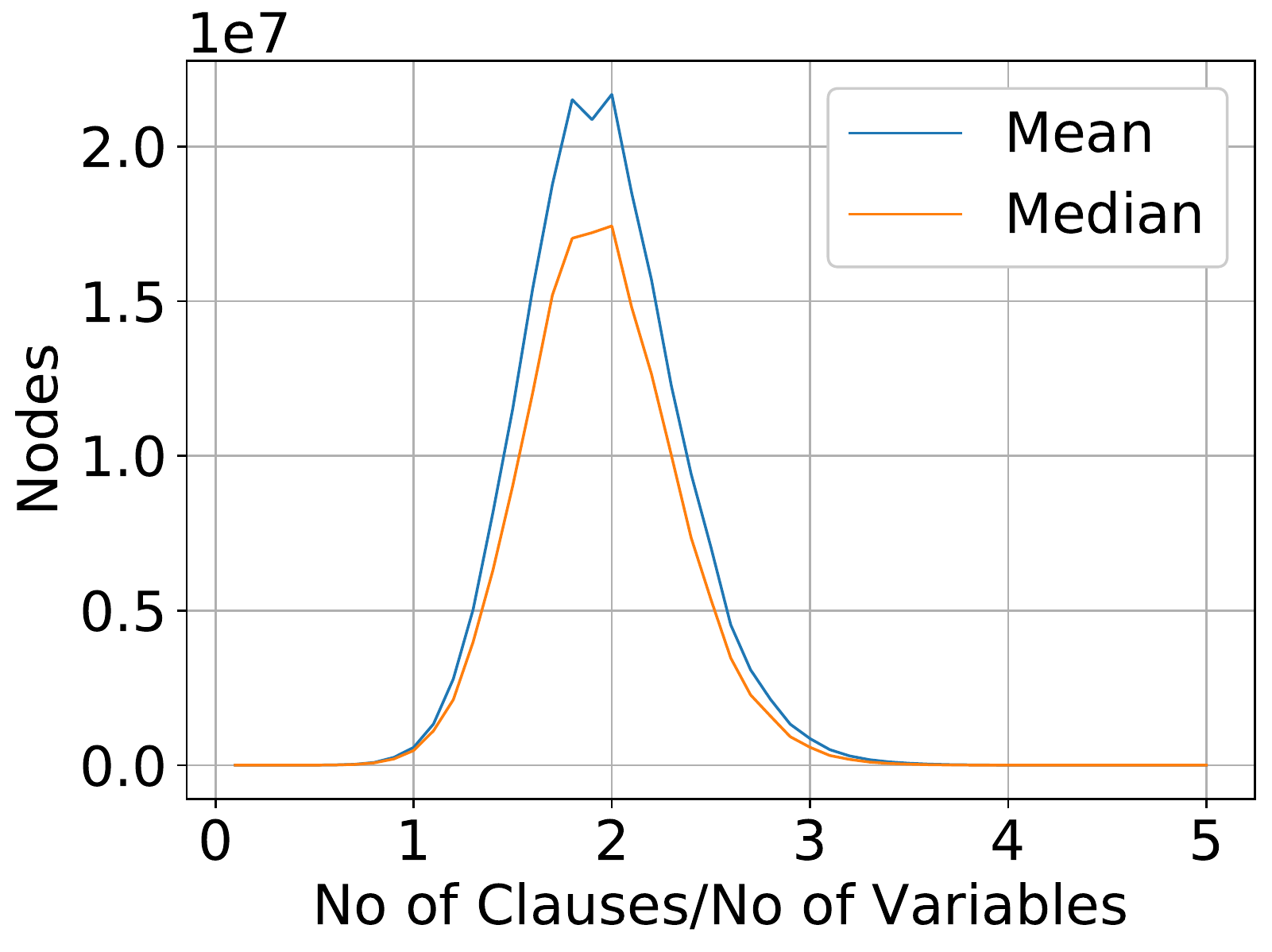}
		\caption{Nodes in d-DNNF , 70 vars}
		\label{fig:d4_nVc_3CNF_70vars}
	\end{subfigure}
	\hfill
	\begin{subfigure}[b]{0.495\linewidth}
		\centering
		\includegraphics[width=\linewidth]{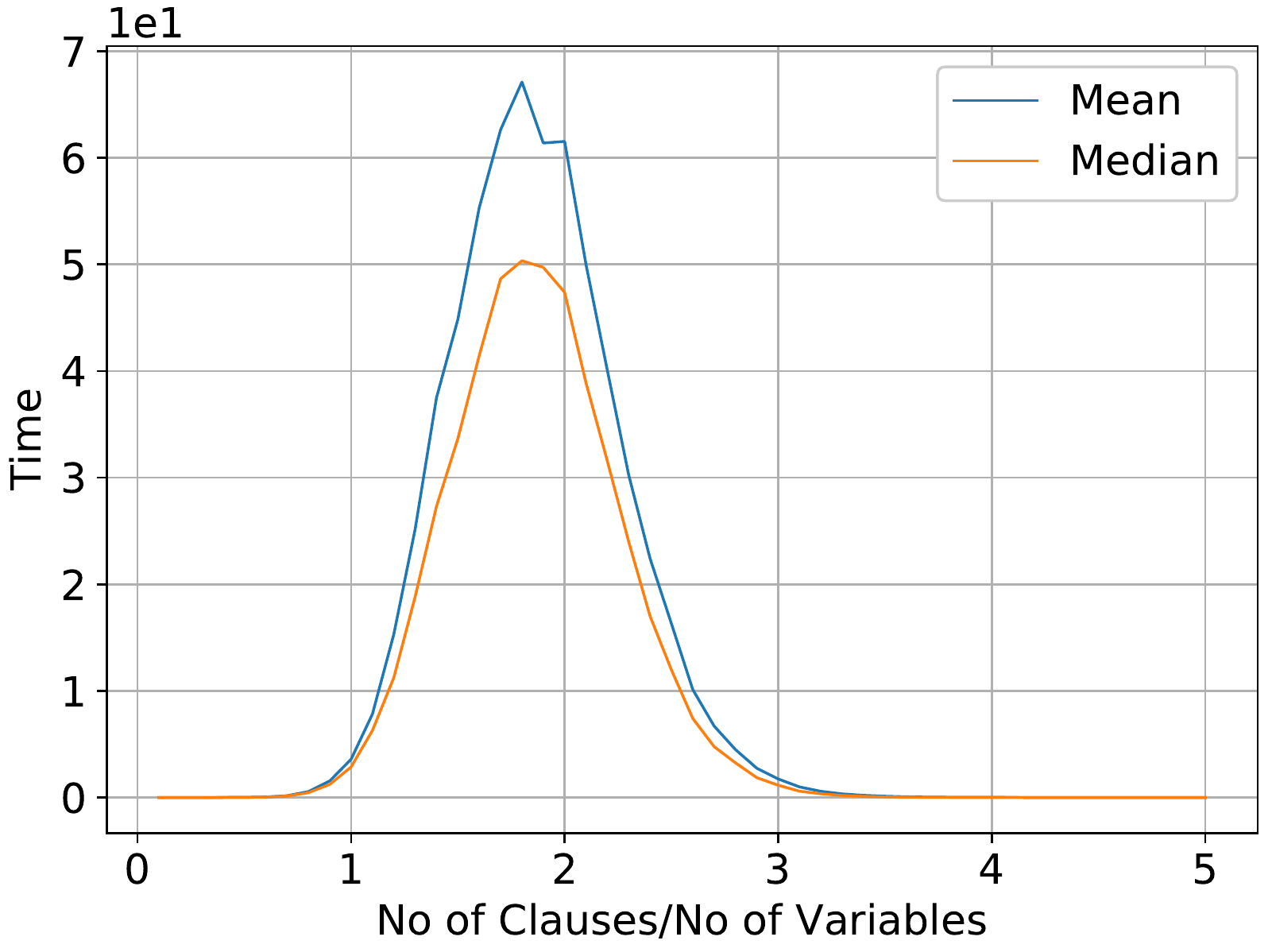}
		\caption{d-DNNF compile-time, 70 vars}
		\label{fig:d4_tVc_3CNF_70vars}
	\end{subfigure}
	\\
	\begin{subfigure}[b]{0.495\linewidth}
		\centering
		\includegraphics[width=\linewidth]{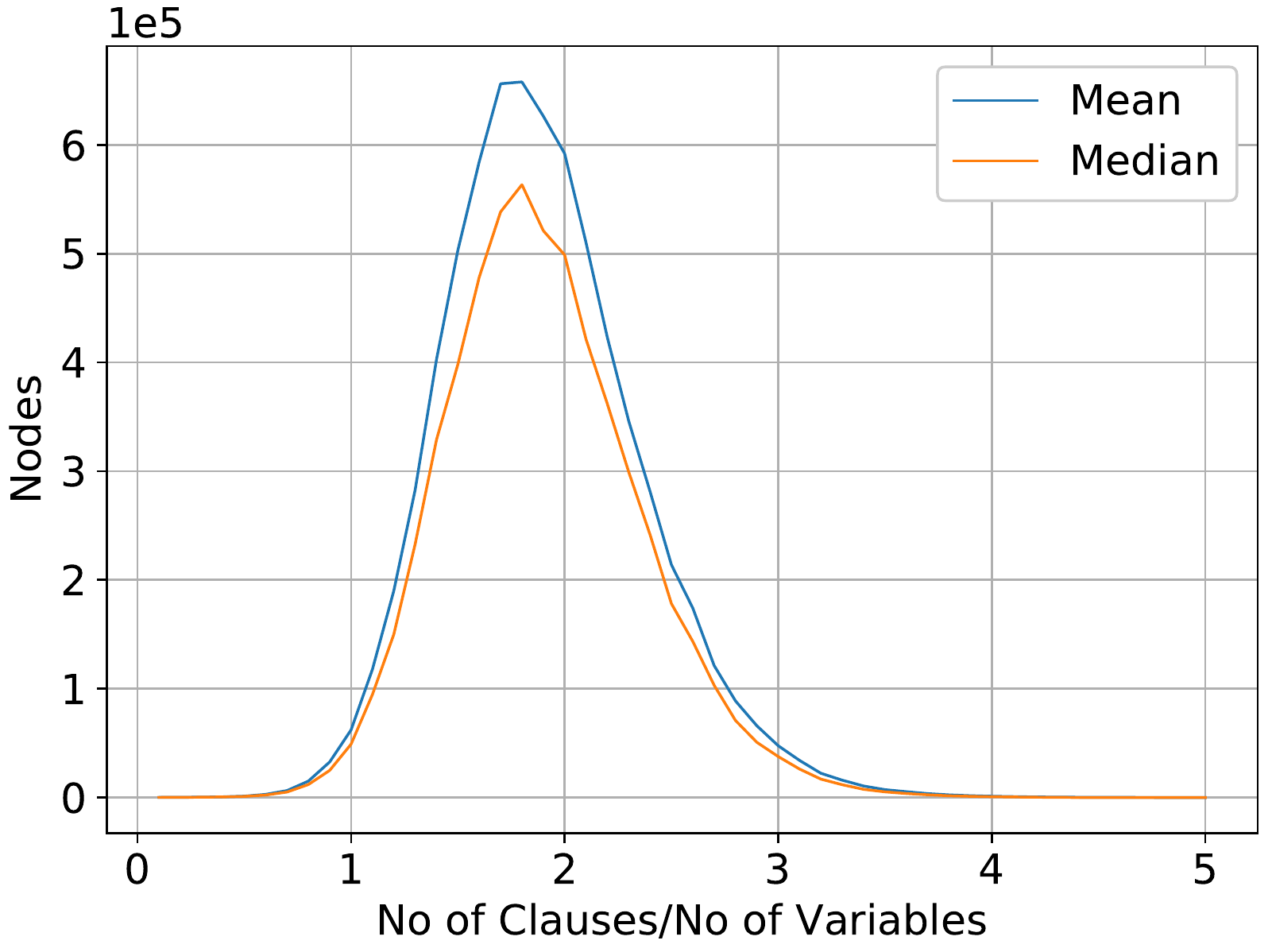}
		\caption{Nodes in SDD , 50 vars}
		\label{fig:sdd_nVc_50vars_3cnf}
	\end{subfigure}
	\hfill
	\begin{subfigure}[b]{0.495\linewidth}
		\centering
		\includegraphics[width=\linewidth]{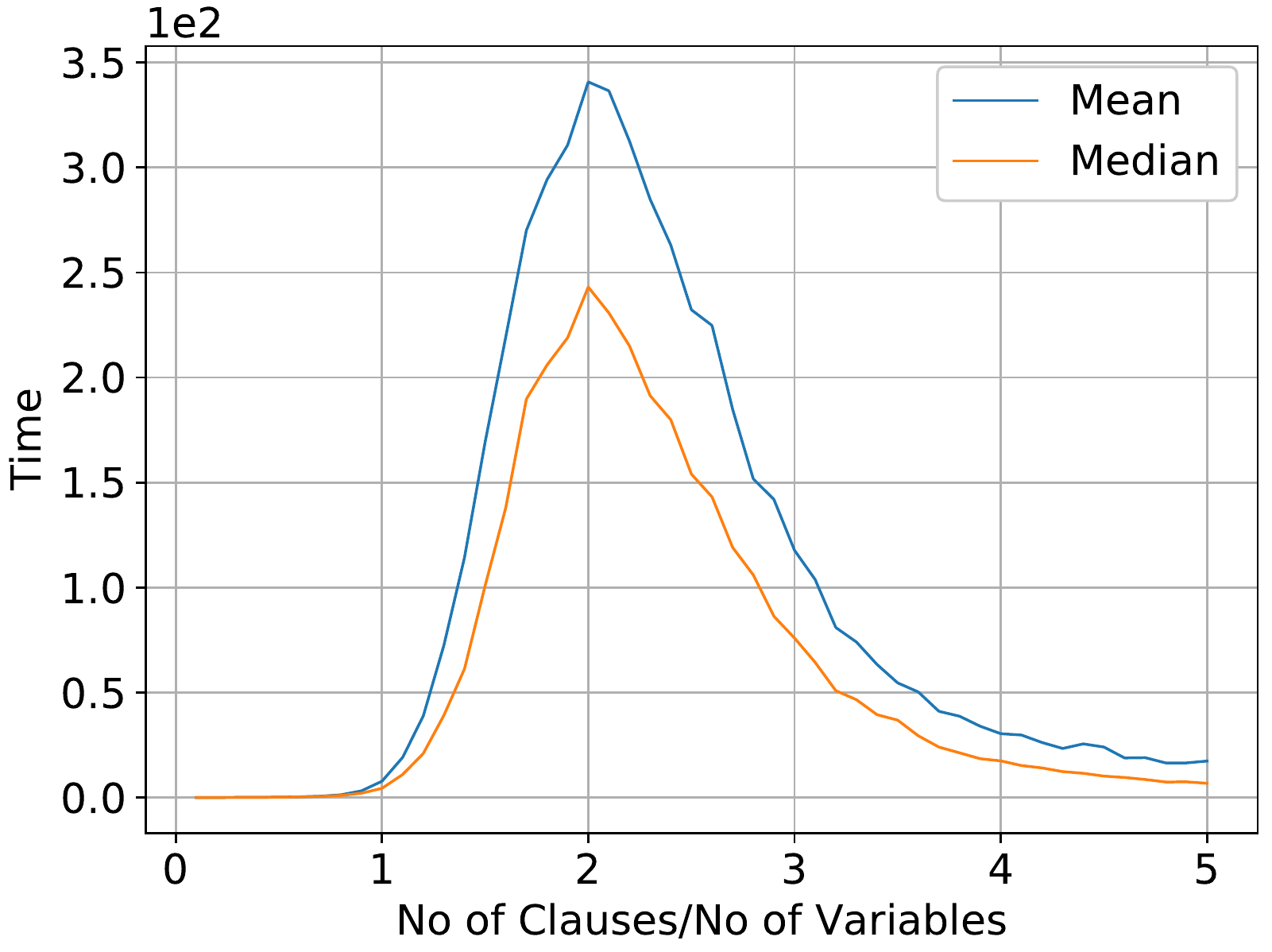}
		\caption{SDD compile-time, 50 vars}
		\label{fig:sdd_tVc_50vars_3cnf}
	\end{subfigure}
	\\    
	\begin{subfigure}[b]{0.495\linewidth}
		\centering
		\includegraphics[width=\linewidth]{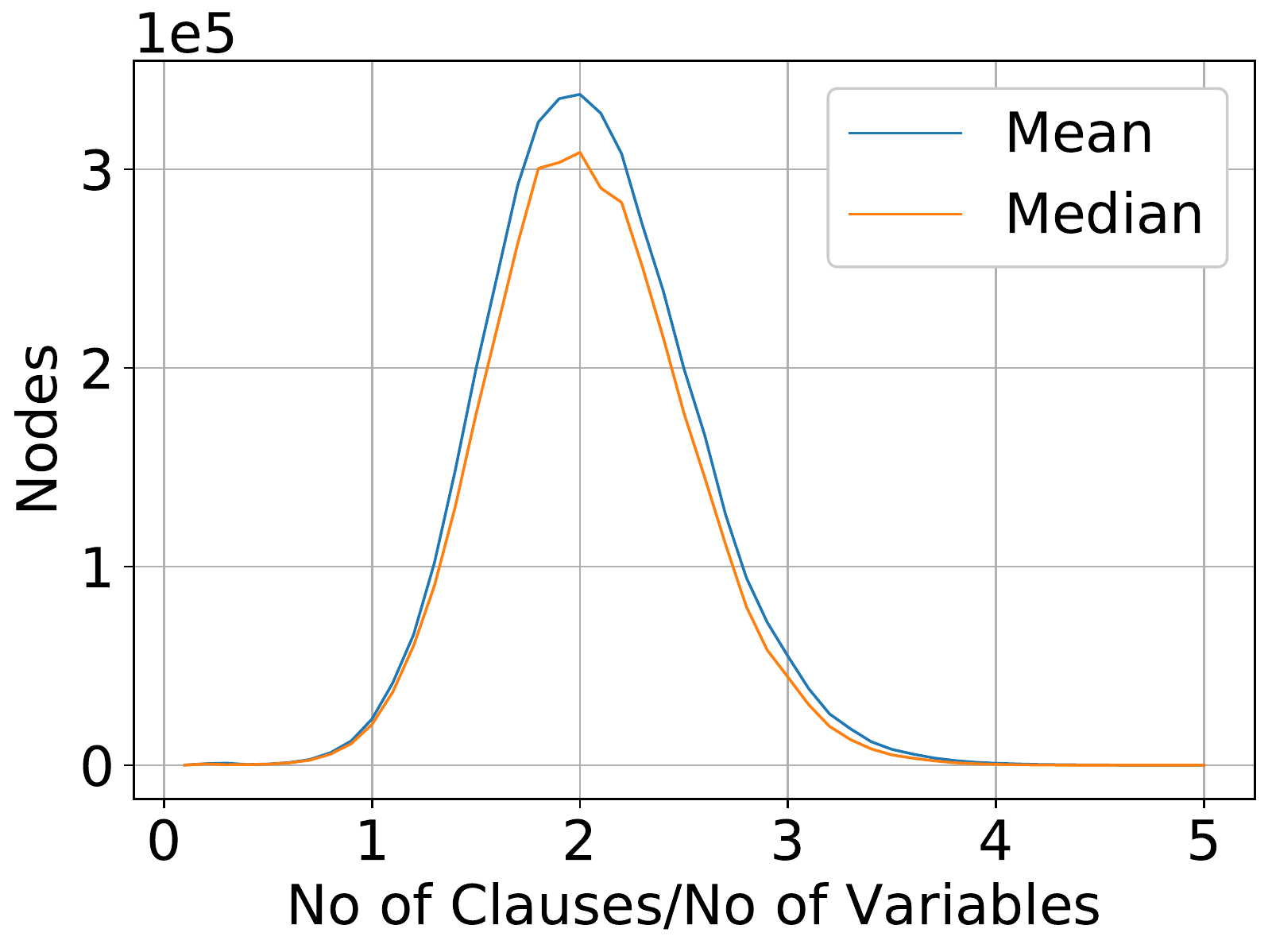}
		\caption{Nodes in OBDD, 50 vars}
		\label{fig:CUDD_nVc_3CNF_50vars}
	\end{subfigure}
	\hfill
	\begin{subfigure}[b]{0.495\linewidth}
		\centering
		\includegraphics[width=\linewidth]{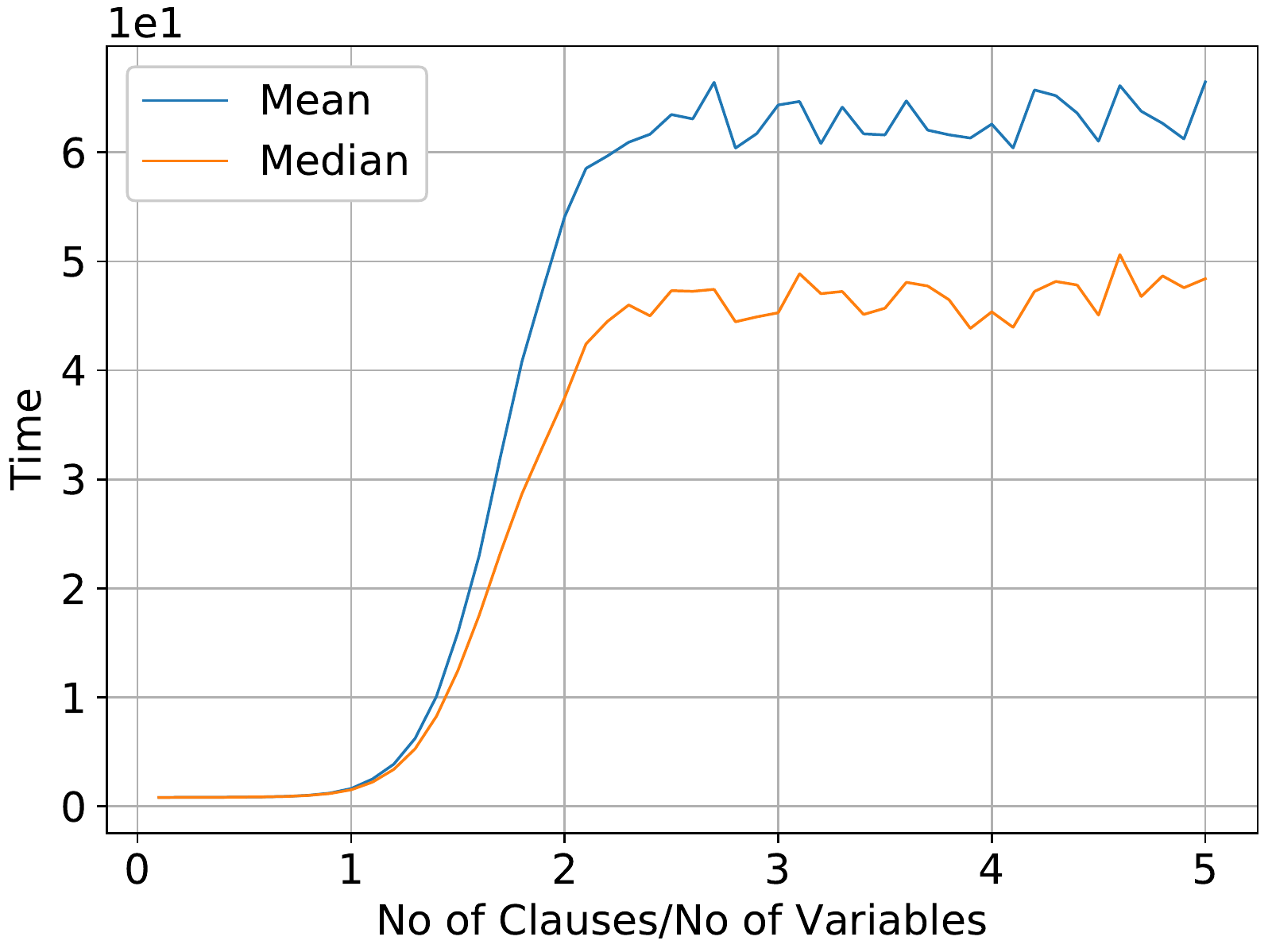}
		\caption{OBDD compile-time, 50 vars}
		\label{fig:CUDD_tVc_3CNF_50vars}
	\end{subfigure}
	\caption{Mean number of nodes, compile-times for 3-CNF}
	\label{fig:nVc_3CNF}
\end{figure}

\begin{figure}[!th]
	\centering
	\includegraphics[width=0.49\linewidth]{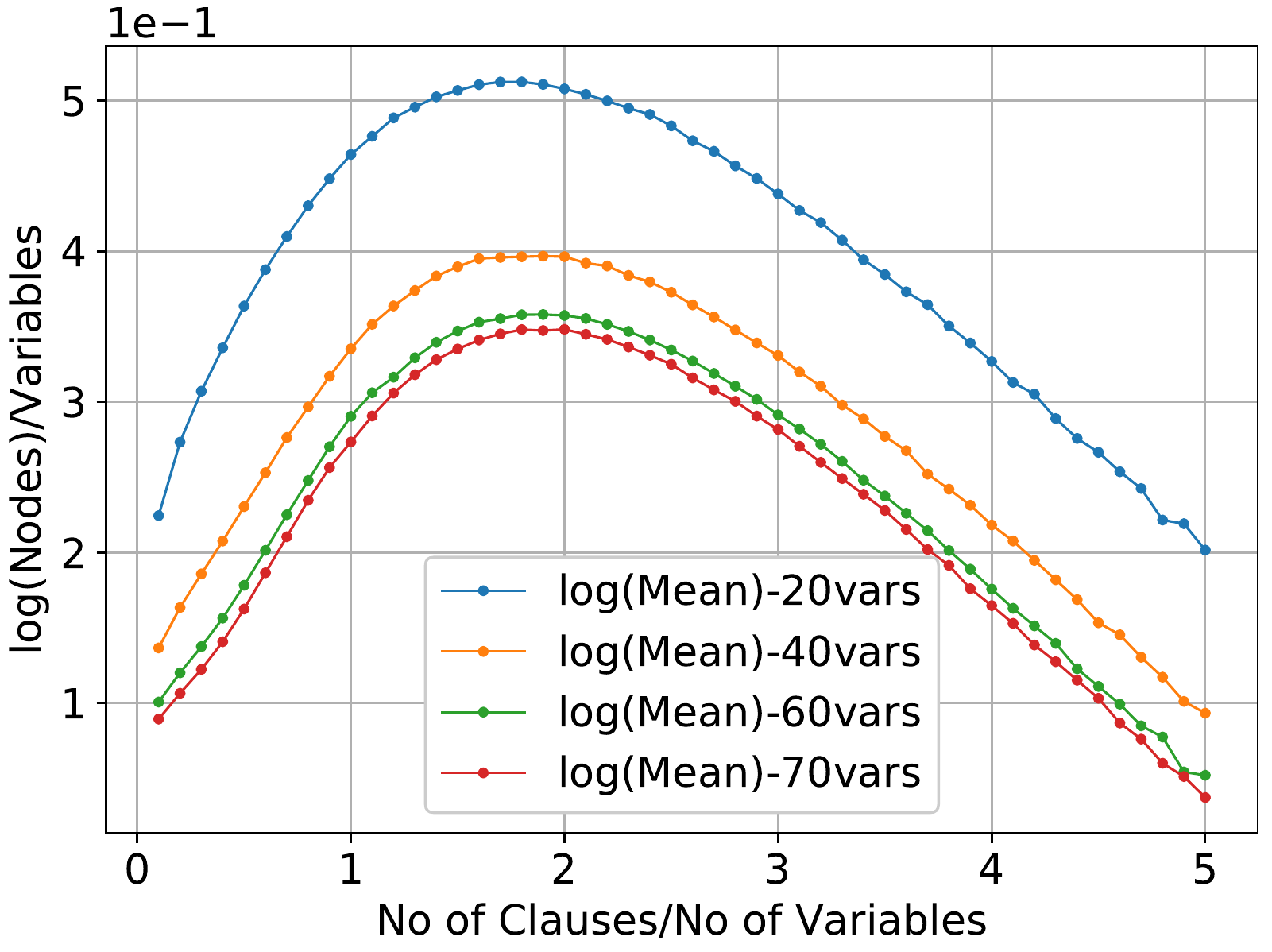}
	\caption{ $\log(\expect(\mathcal{N}_{\lang}(F_k(n,rn))))/n$ vs $r$ for different $n$ for $\lang=\dDNNF$}
	\label{fig:d4_log_nVc_3CNF_vars}
\end{figure}

Figure~\ref{fig:nVc_3CNF} shows the variations in the mean and median of the number of nodes for d-DNNF, SDD, and OBDD with respect to clause density. Figure~\ref{fig:d4_log_nVc_3CNF_vars} shows variations in mean number of nodes for d-DNNF on a $log$ scale versus number of variables. The small-large-small pattern in the size of compilations is reminiscent of the empirical hardness of SAT near the phase transition, which is marked by an exponential increase in SAT solvers' runtimes near a specific (phase transition) clause density. 

\subsubsection{Runtime of compilation}
From an empirical usage perspective, the runtime of compilation assumes paramount importance. It is worth emphasizing that the size of the compiled form does not solely determine the runtime. In particular, the  runtime of compilation broadly depends upon:
\begin{enumerate*}
	\item[(1)] the search space itself,
	\item[(2)] the time spent in heuristics.
\end{enumerate*}
In state-of-the-art knowledge compilers, newer heuristics increasingly attempt to prune the search space leading to reduced memory and runtime for subsequent exploration. However, they incur an increased overhead of applying heuristics. 
Therefore, to provide a holistic view of the hardness of compilations, Figure \ref{fig:nVc_3CNF} also shows the variation in average runtimes against clause density for d-DNNF, SDD, and OBDD. We observe that the location of phase transition appears slightly shifted (within $\pm 0.3$) compared to that for the size of compilations in the case of d-DNNF and SDD. However, in the case of OBDD, we have a starkly different behavior due to the different compilation procedures, even though both SDDs and OBDDs are compiled by repeatedly conjoining the clauses one by one using the polytime APPLY operation. 

A plausible explanation is that while SDD compilation using $\sddpackage$ involves dynamic vtree search with clause reordering, OBDD compilation using $\CUDD$ supports only dynamic variable ordering. This can lead to an easy-hard-less hard behavior for SDD compilations as clause reordering facilitates early pruning of intermediate SDDs \cite{Sdd-package}. On the other hand, OBDD compilation involves large intermediate OBDDs near phase transition clause density irrespective of the total number of clauses as clauses are selected randomly for the APPLY operation. 
This can lead to a sharp step transition in runtimes as the addition of more clauses after phase transition prunes the OBDD, translating to reduced cost of successive APPLY operations.

\begin{figure}[!th]
	\centering
	\begin{subfigure}[b]{0.49\linewidth}
		\centering
		\includegraphics[width=\linewidth]{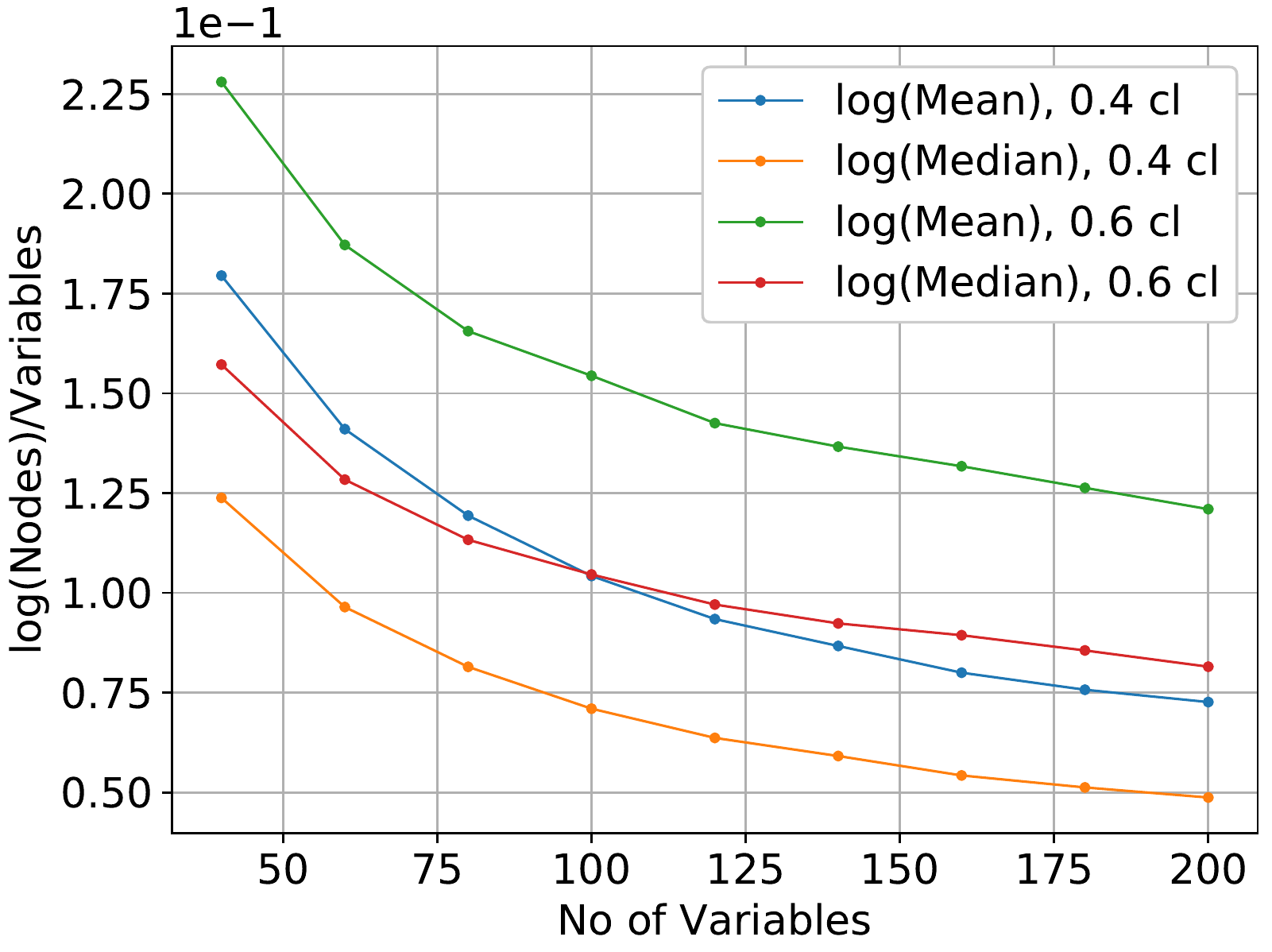}
		\caption{clause densities 0.6 and 0.4}
		\label{fig:d4_log_nVv_0.4-0.6cl_3cnf}
	\end{subfigure}
	\hfill
	\begin{subfigure}[b]{0.49\linewidth}
		\centering
		\includegraphics[width=\linewidth]{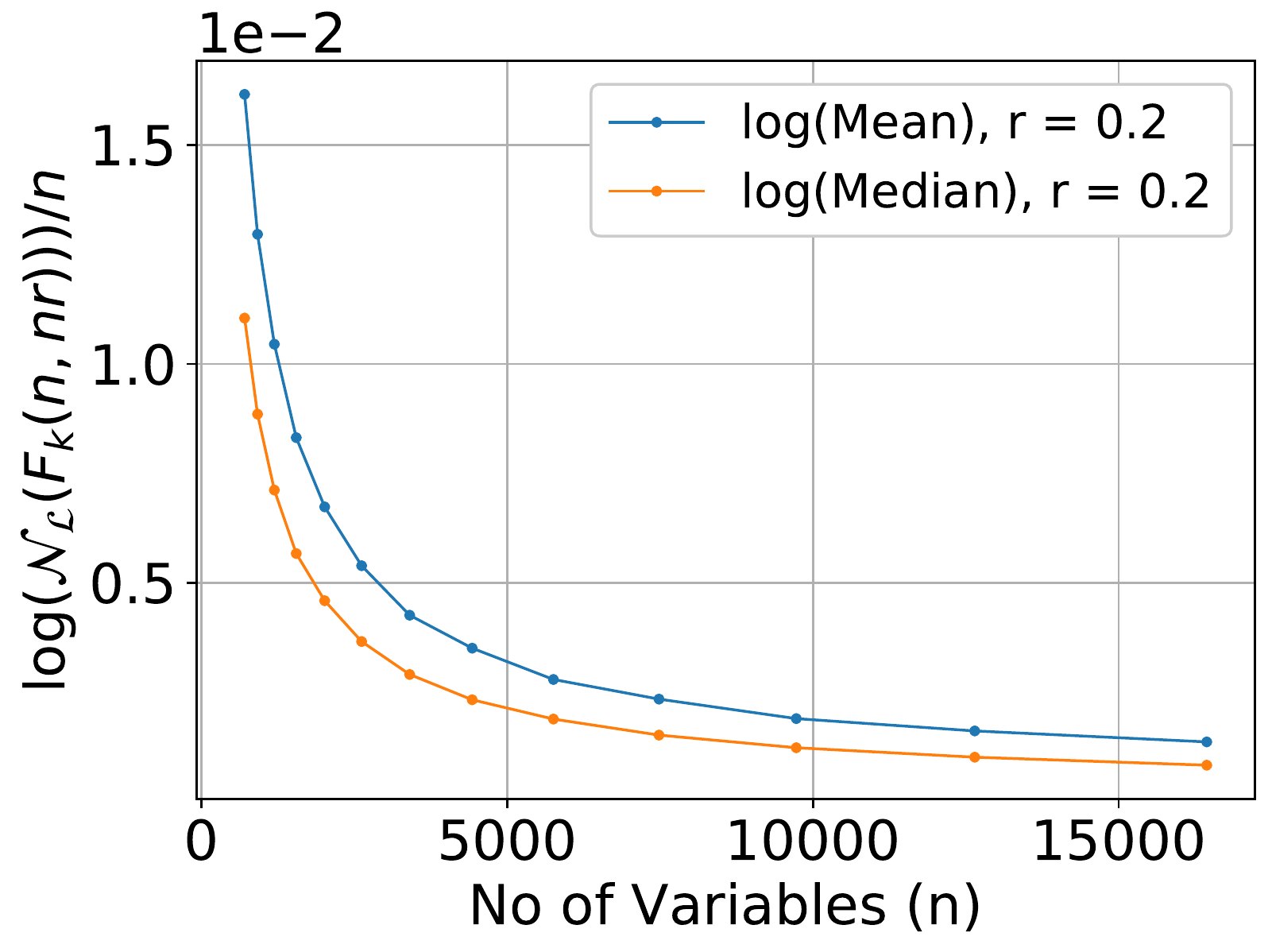}
		\caption{clause density 0.2}
		\label{fig:d4_log_nVv_0.2cl_3cnf}
	\end{subfigure}
	\caption{$\log(\mathcal{N}_{\lang}(F_k(n,rn)))/n$ vs $n$ for $\lang=\dDNNF$}
	\label{fig:d4_log_nVv_3CNF}
\end{figure}

\begin{figure}
	\centering
	\begin{subfigure}[b]{0.495\linewidth}
		\centering
		\includegraphics[width=\linewidth]{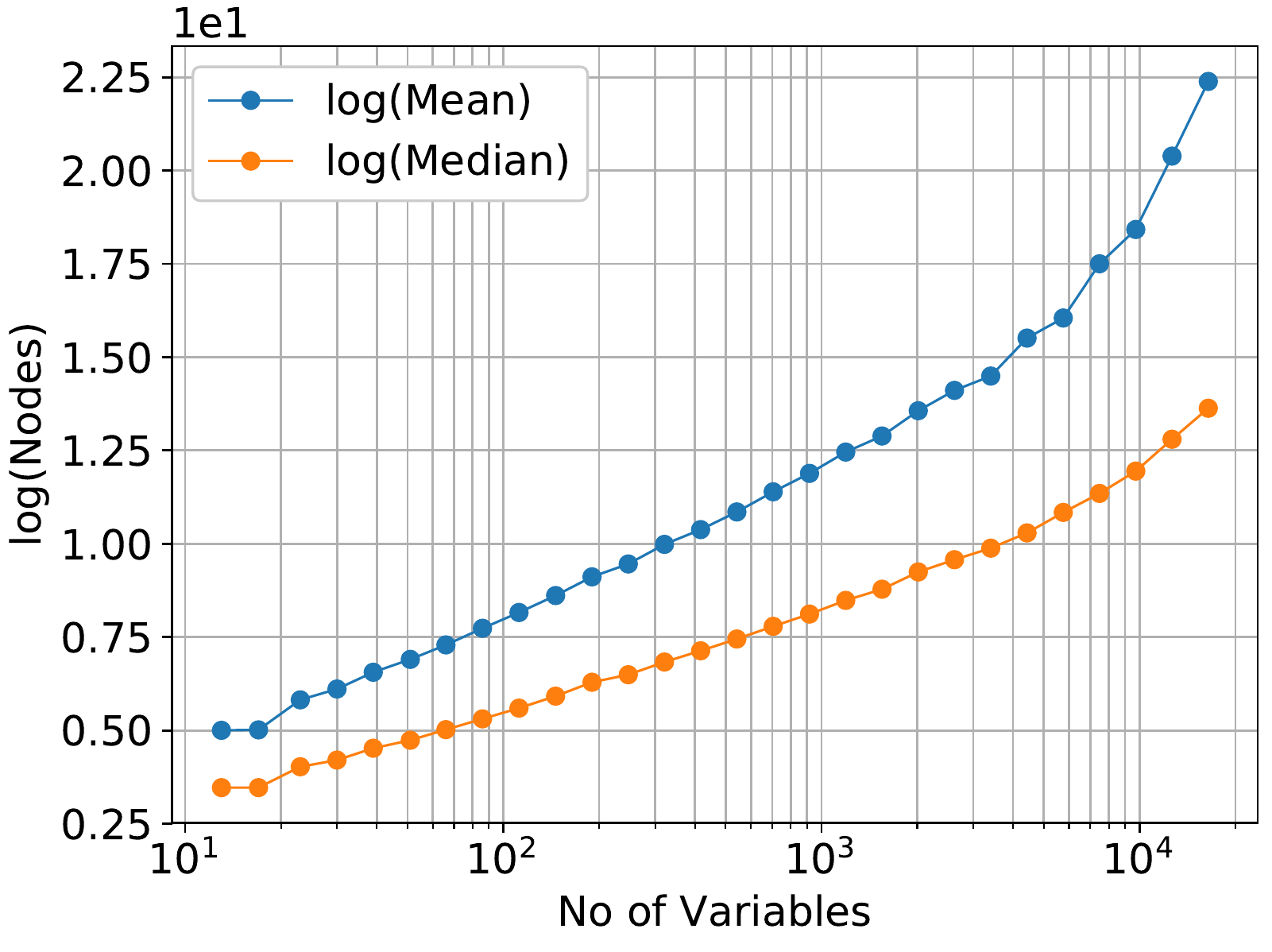}
		\caption{$\log(\mathcal{N}_{\lang}(F_k(n,rn)))$ vs $log(n)$  for $\lang=\dDNNF$}
		\label{fig:d4_loglog_nVv_0.2cl_3cnf}
	\end{subfigure}
	\hfill
	\begin{subfigure}[b]{0.495\linewidth}
		\centering
		\includegraphics[width=\linewidth]{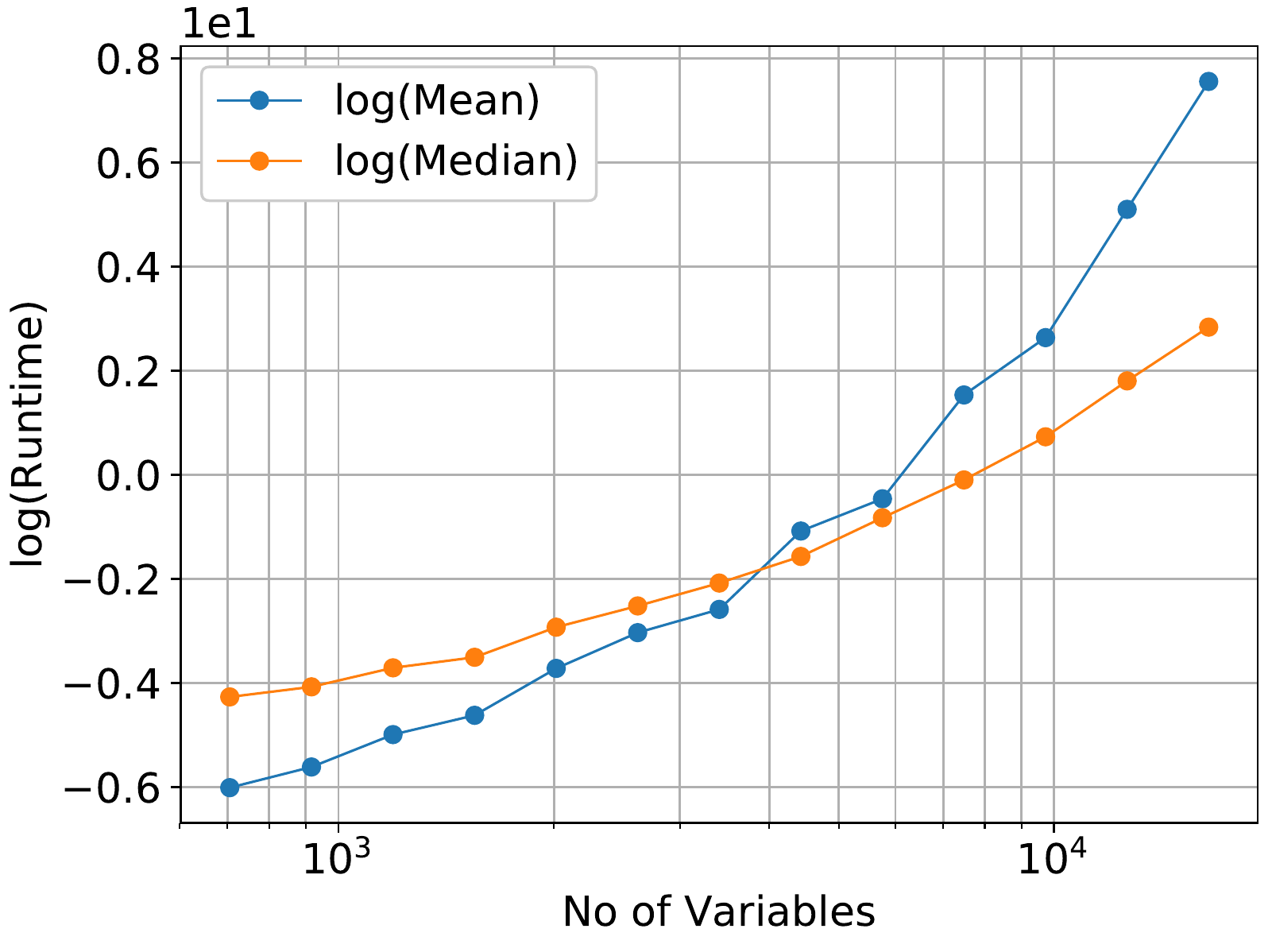}
		\caption{$\log(\mathcal{T}_{\lang}(F_k(n,rn)))$ vs $log(n)$ for $\lang=\dDNNF$}
		\label{fig:d4_loglog_tVv_0.2cl_3cnf}
	\end{subfigure}
	\caption{log-log graph for $r=0.2$}
	\label{fig:d4_log_log_tVv_3CNF}
\end{figure}

We refer the reader to Appendix for variation with respect to clause length.  \todo{Change to full version for camera ready}
We sum up our observations via the following conjecture 
\begin{conjecture}\label{conjecture1}
	For every integer $k \geq 2$, given the number of variables $n$ and a target language $\lang$ that is a subset of DNNF, there exists a positive real number $r_k$ such that\\
	For each pair $(r_1, r_2)$, if $r_1 < r_2 < r_k$ then,
	$$\expect(\mathcal{N}_{\lang}(F_k(n,r_1n))) < \expect(\mathcal{N}_{\lang}(F_k(n,r_2n)))$$
	For each pair $(r_1, r_2)$, if $r_k < r_1 < r_2$ then,
	$$\expect(\mathcal{N}_{\lang}(F_k(n,r_1n))) > \expect(\mathcal{N}_{\lang}(F_k(n,r_2n)))$$
\end{conjecture}

\begin{figure}[!th]
	\centering
	\begin{subfigure}[b]{0.495\linewidth}
		\centering
		\includegraphics[width=\linewidth]{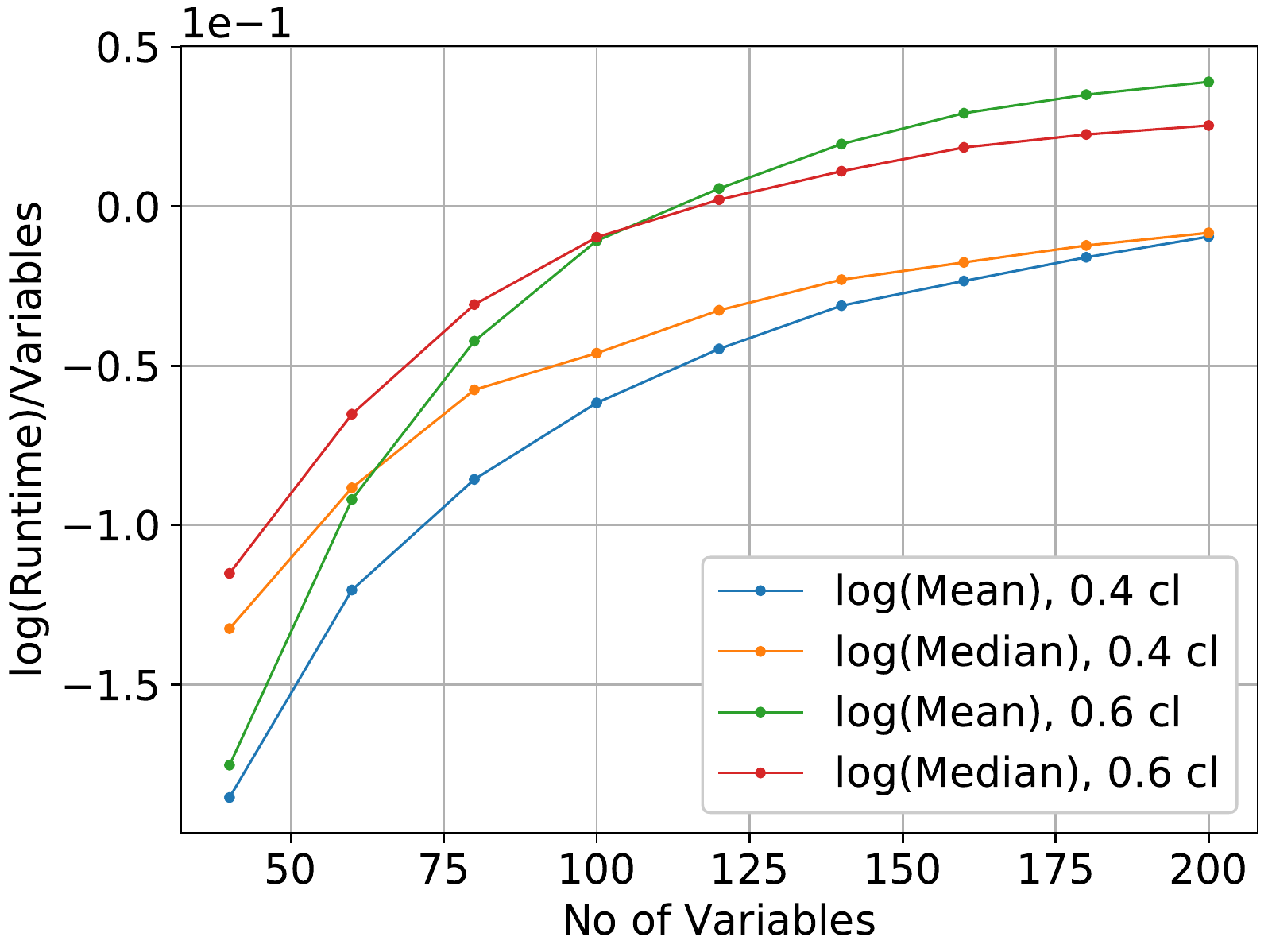}
		\caption{clause densities 0.6 and 0.4}
		\label{fig:d4_log_tVv_0.6cl_3cnf}
	\end{subfigure}
	\hfill
	\begin{subfigure}[b]{0.495\linewidth}
		\centering
		\includegraphics[width=\linewidth]{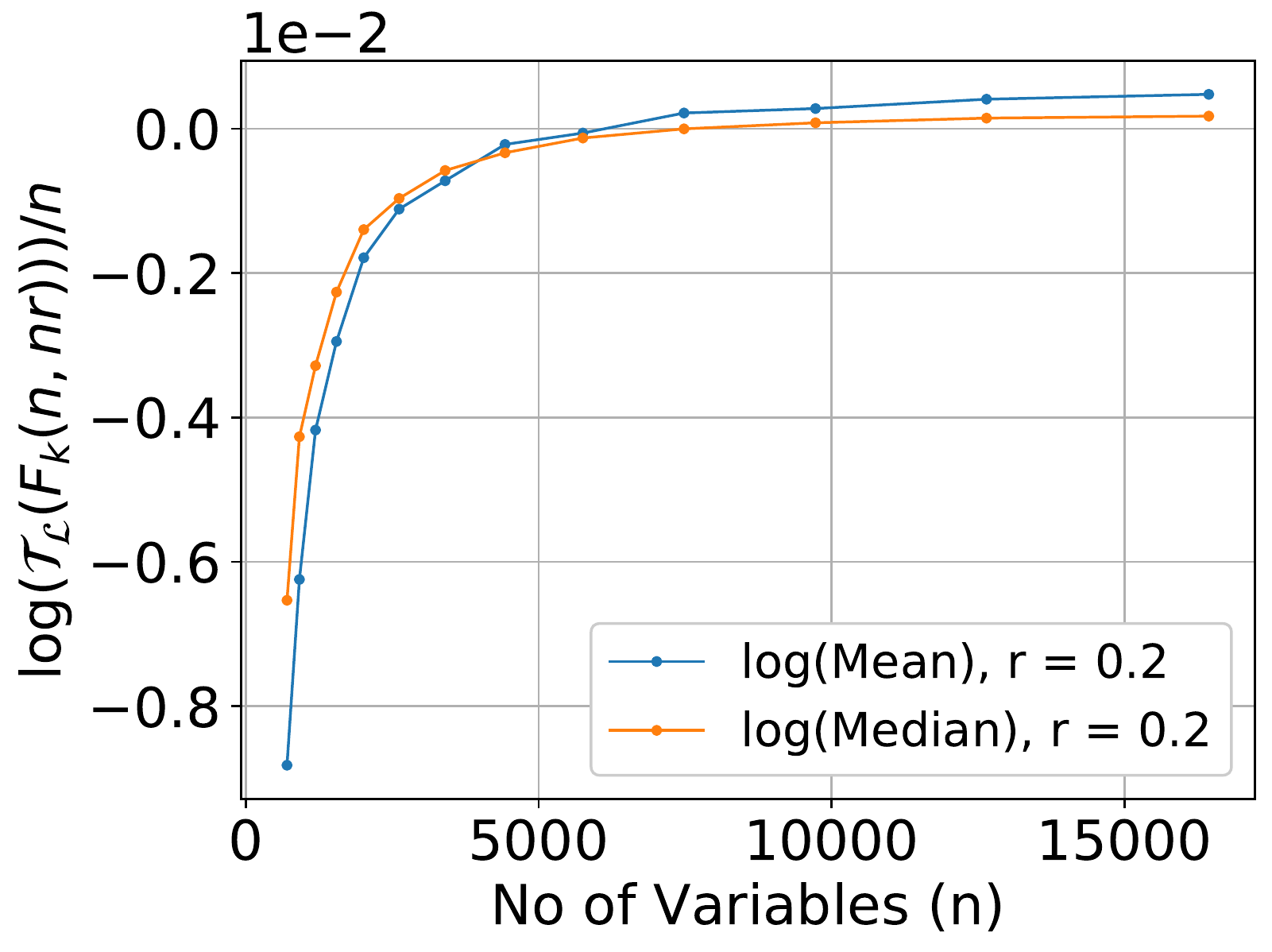}
		\caption{clause density 0.2}
		\label{fig:d4_log_tVv_0.2cl_3cnf}
	\end{subfigure}
	\caption{$\log(\mathcal{T}_{\dDNNF}(F_k(n,rn)))/n$ vs $n$ for d-DNNF}
	\label{fig:d4_log_tVv_3CNF}
\end{figure}

\subsection{Diving deep into the asymptotic complexity}

The size and runtime of compilations are exponential in the number of variables in the worst case. We are, however, interested in more precise relationship as size and runtime play crucial roles in many applications. While it is known that the exponent increases towards phase transition clause density \cite{GYX11}, it is not clear if the exponent is linear or sublinear in the number of variables and whether this behavior changes with changing clause density.

Since, the size of OBDDs (and d-DNNFs and SDDs being more succinct~\cite{B16}) is bounded by $\bigo(2^n)$, we plot $\log(\expect(\mathcal{N}_{\dDNNF}(F_k(n,rn))))/n$ for different $n$ against varying $r$ in Figure \ref{fig:d4_log_nVc_3CNF_vars}~and~\ref{fig:d4_log_nVv_3CNF} for further investigation. From Figure~\ref{fig:d4_log_nVc_3CNF_vars}, we observe phase transition with respect to clause density even though the location of the phase transition point seems to shift slightly with different $n$. Next,  we turn our attention to Figure~\ref{fig:d4_log_nVv_3CNF}, wherein we  observe that $\log(\expect(\mathcal{N}_{\dDNNF}(F_k(n,rn))))/n$ decreases while showing signs of possible convergence with increasing $n$ for a given $r$ for all compilations in our study. In order to understand it better, we scale our experiments to a larger $n$ for small $r$ in Figure~\ref{fig:d4_log_nVv_0.2cl_3cnf}. However, even in this case, the answer remains unclear if $\log(\expect(\mathcal{N}_{\dDNNF}(F_k(n,rn))))/n$ shall converge to a constant $> 0$. Therefore, we can only say that for a constant, $c \geq 0$, $\lim\limits_{n\rightarrow\infty}\frac{\log(\expect(\mathcal{N}_{\lang}(F_k(n,rn))))}{n} = c$.

While the question if $c =0$ remains open for size, the case for runtime of compilation is a different story: 
Figure~\ref{fig:d4_log_tVv_3CNF} shows that $\frac{\expect(\mathcal{T}_{\lang}(F_k(n,rn))))}{n}$ increases with $n$. Knowing that the runtime in worst case is $poly(2^n)$ and extrapolating the observation, we conjecture that $\lim\limits_{n\rightarrow\infty}\frac{\expect(\mathcal{T}_{\lang}(F_k(n,rn))))}{n} = c $ where $c>0$ and depends on~$r$. In other words, $\expect(\mathcal{T}_{\lang}(F_k(n,rn)))) = \theta(2^{cn})$ for state-of-the-art compilers.

\subsubsection{Polynomial to Exponential size phase transition.}

Gao et al.~\cite{GYX11} had shown the existence of a polynomial to exponential phase transition for size of compilations around $r=0.3$ by showing increase in slope for $r>0.3$ and near constant slope for $r<0.3$ on a $\log(\expect(\mathcal{N}_{\dDNNF}(F_k(n,rn))))$-$\log(n)$ graph. Relationships of the form $y=ax^{k}$ appear as straight lines in a log-log graph. 
Our observations in Figure~\ref{fig:d4_log_nVv_3CNF} and \ref{fig:d4_loglog_nVv_0.2cl_3cnf} show that while the instances with  $r = 0.2$ are indeed very easy compared to $r = 0.4$, the behavior is still exponential or quasi-polynomial for $r=0.2$ which becomes dominant for large enough number of variables.
In Figure~\ref{fig:d4_loglog_nVv_0.2cl_3cnf}, the behavior appears to change from a straight line to a line with increasing slope around $n=7482$.  On extrapolating the behavior for even smaller $r$, we can conjecture that $\forall \,r$, $\expect(\mathcal{N}_{\mathcal{\dDNNF}}(F_k(n,rn)))$ is at least quasi-polynomial in $n$.

\section{Phase transitions with solution density} 
\label{sec:SDexp}
\begin{figure}[t]
	\centering            
	\includegraphics[width=0.48\linewidth]{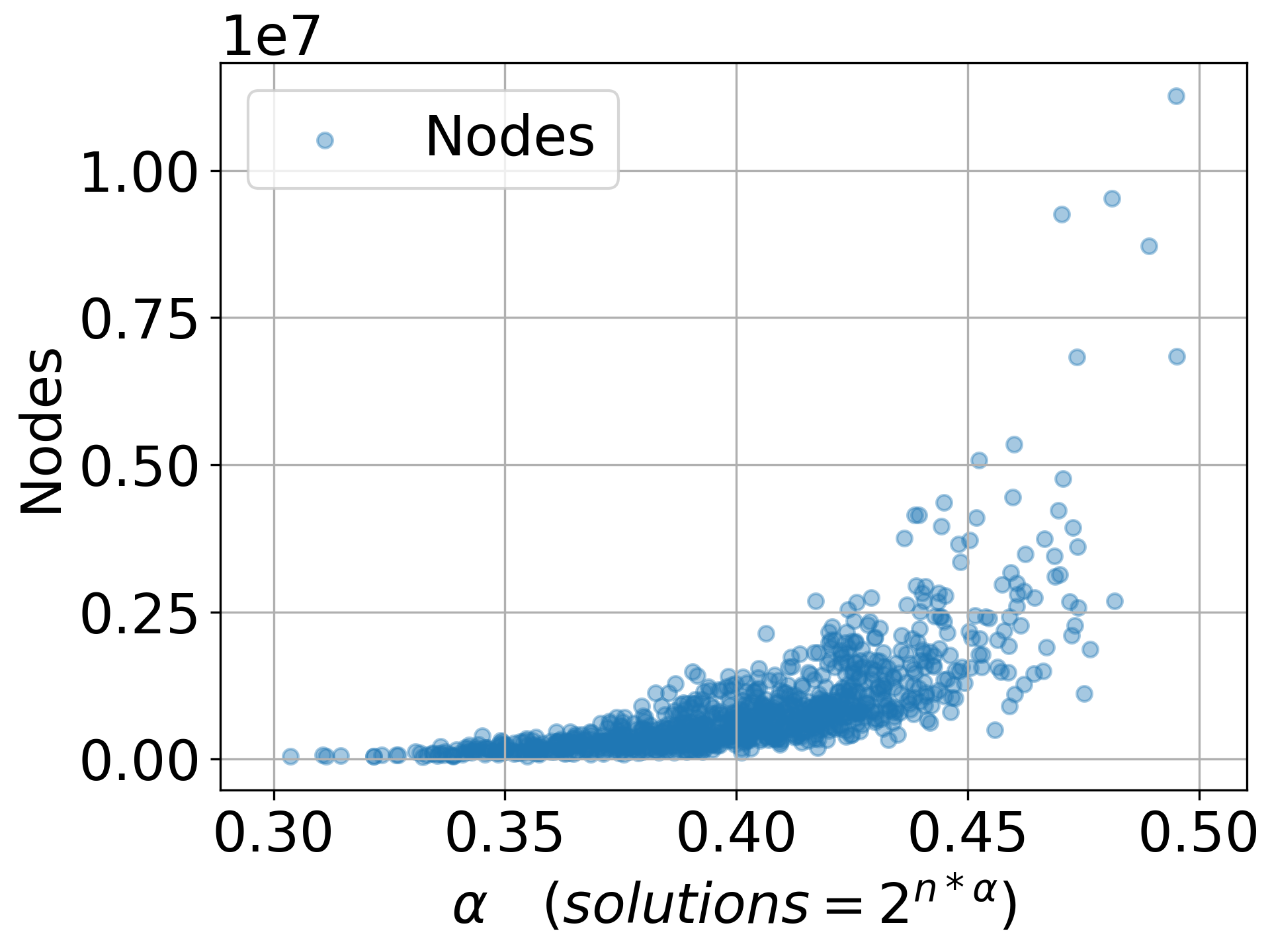}
	\caption{Variation in size with solution density for individual instances with $r=3.0$ for d-DNNF}
	\label{fig:d4_fixCl_nVc_3CNF_60vars}
\end{figure}

Since knowledge compilations are a compact way of representing solutions, one can expect that they show variations in sizes with respect to the solution density as well. The solution density, however, is not independent of clause density for random $k$-CNF as given a clause density, expected solution density is fixed, and vice versa. Notwithstanding, we observe (Figure~\ref{fig:d4_fixCl_nVc_3CNF_60vars}) that solution density also appears to be a fundamental parameter given a fixed clause density, instances with different solution density have marked changes in their size of d-DNNF compilations. We, now, look at the size and compile-time behavior with respect to solution density.

\subsection{Observing the phase transition}
\begin{figure}[!th]
	\centering
	\captionsetup[subfigure]{labelformat=empty}
	\begin{subfigure}[b]{0.475\linewidth}
		\centering
		\includegraphics[width=\linewidth]{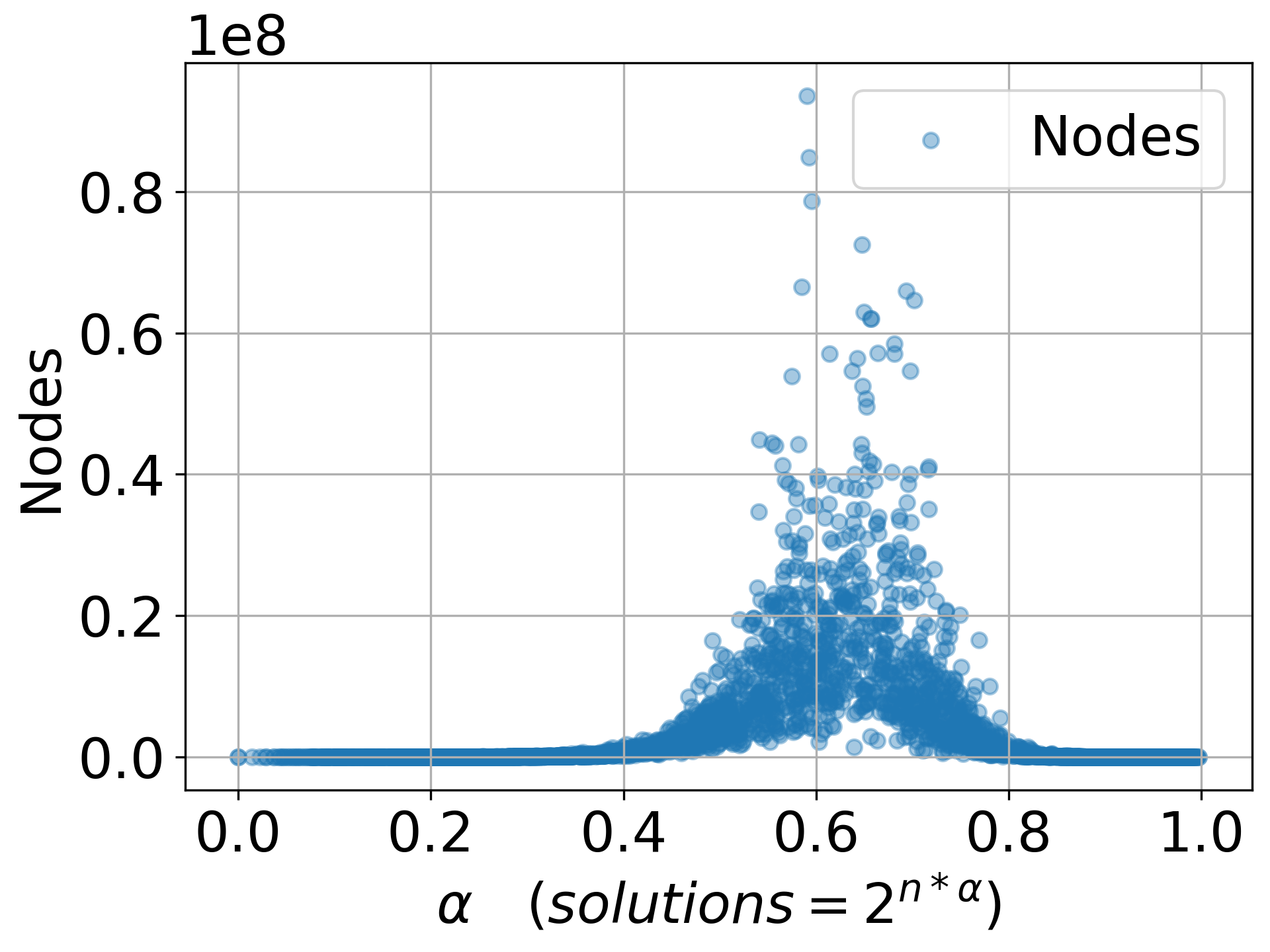}
		\caption{Nodes in d-DNNF, 70 vars}
		\label{fig:d4__70vars}
	\end{subfigure}
	\hfill
	\begin{subfigure}[b]{0.475\linewidth}
		\centering
		\includegraphics[width=\linewidth]{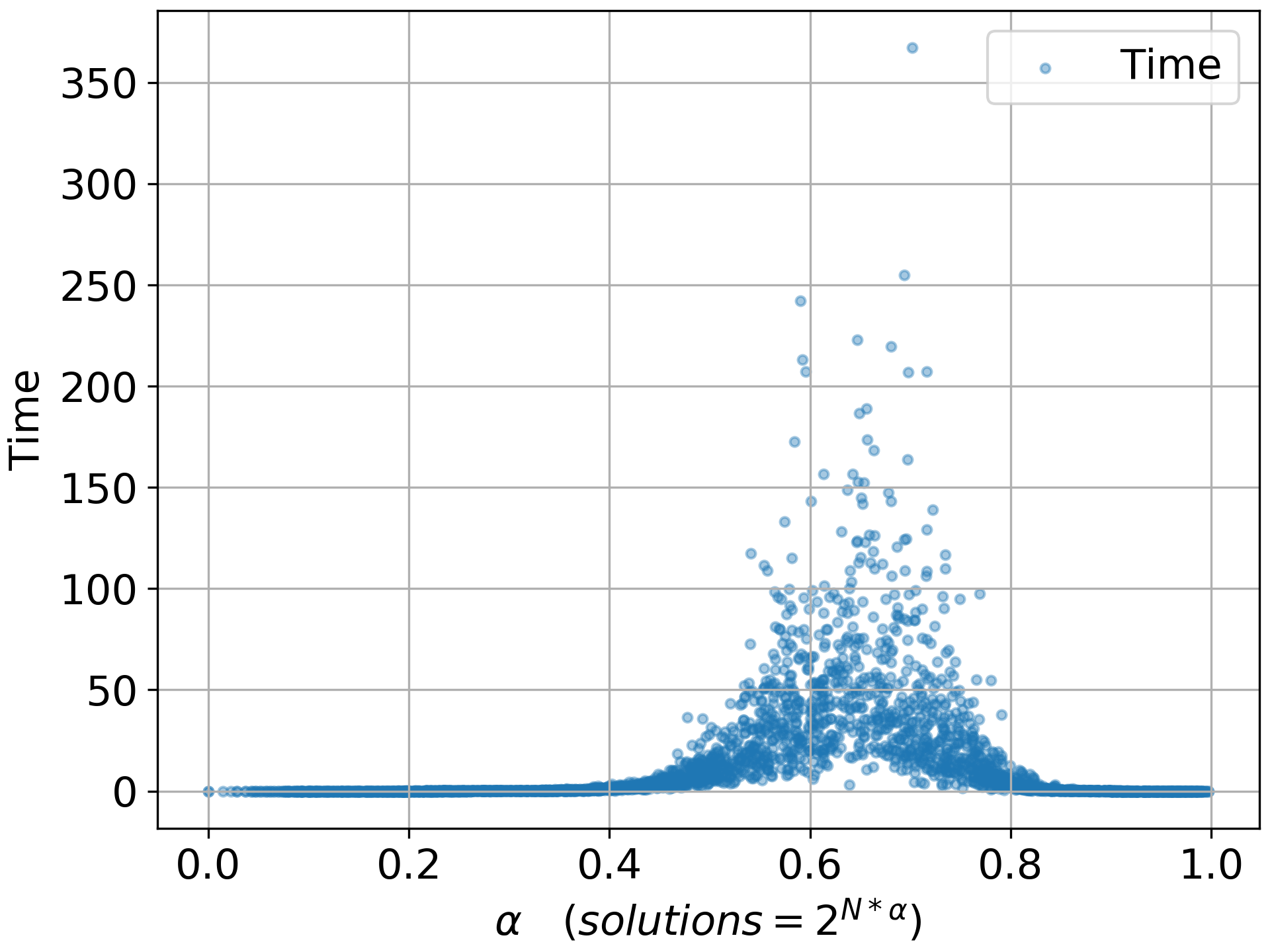}
		\caption{d-DNNF compile-time, 70 vars}
		\label{fig:d4_tVa_3CNF_70vars}
	\end{subfigure}
	\\
	\begin{subfigure}[b]{0.475\linewidth}
		\centering
		\includegraphics[width=\linewidth]{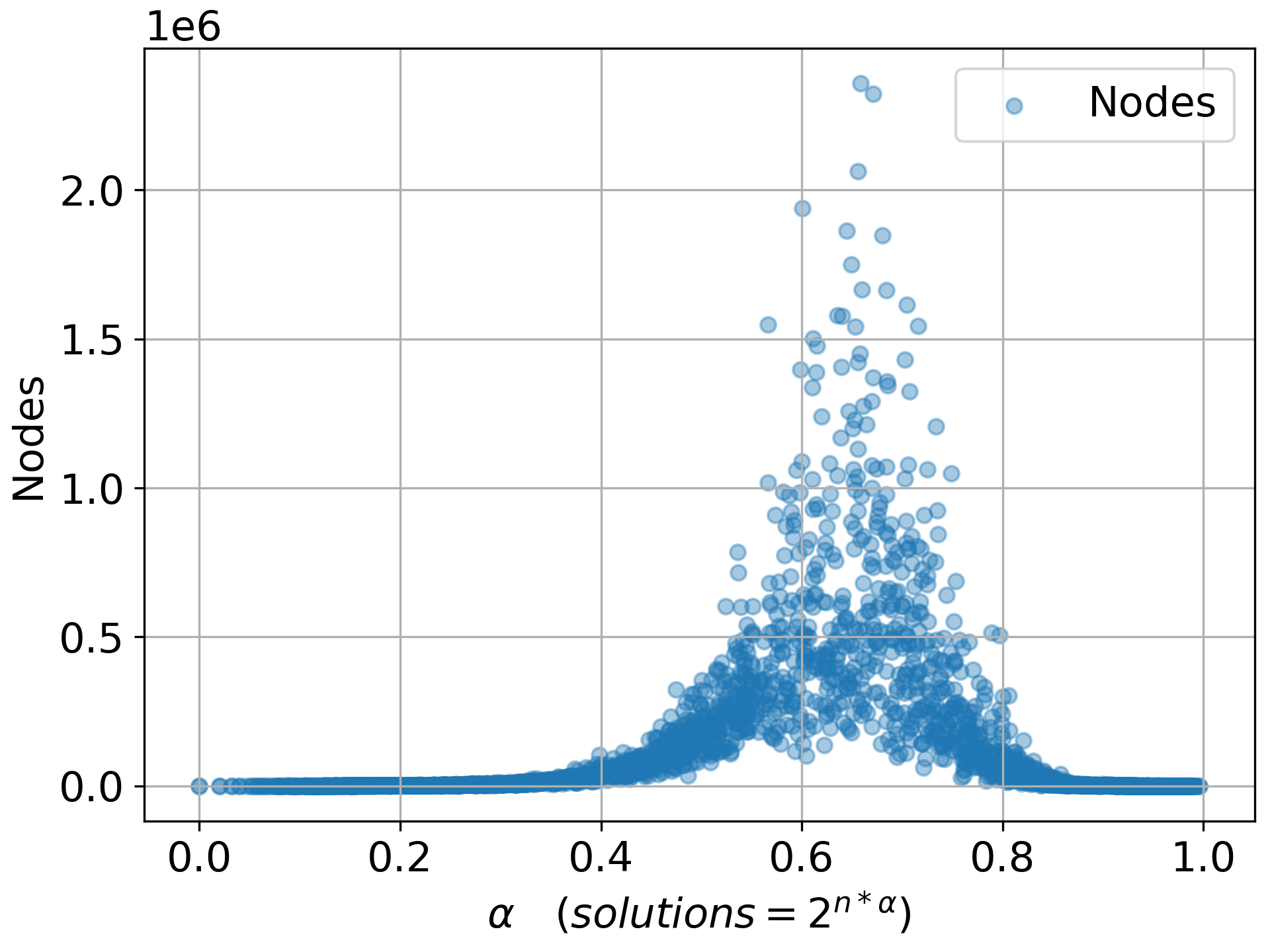}
		\caption{Nodes in SDD, 50 vars}
		\label{fig:sdd_nVa_50vars_3cnf}
	\end{subfigure}
	\hfill
	\begin{subfigure}[b]{0.475\linewidth}
		\centering
		\includegraphics[width=\linewidth]{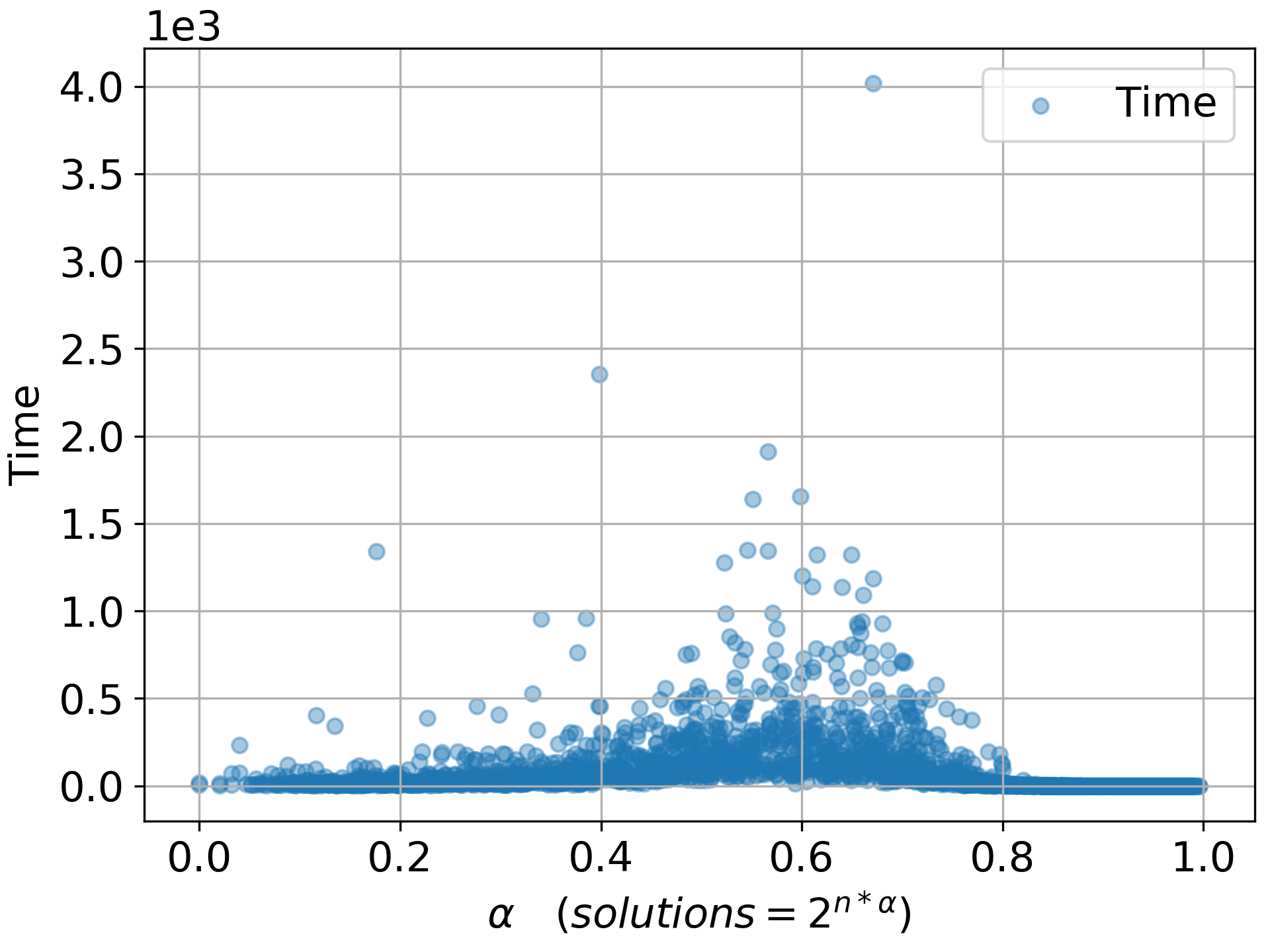}
		\caption{SDD compile-time, 50 vars}
		\label{fig:sdd_tVa_50vars_3cnf}
	\end{subfigure}
	\\    
	\begin{subfigure}[b]{0.475\linewidth}
		\centering
		\includegraphics[width=\linewidth]{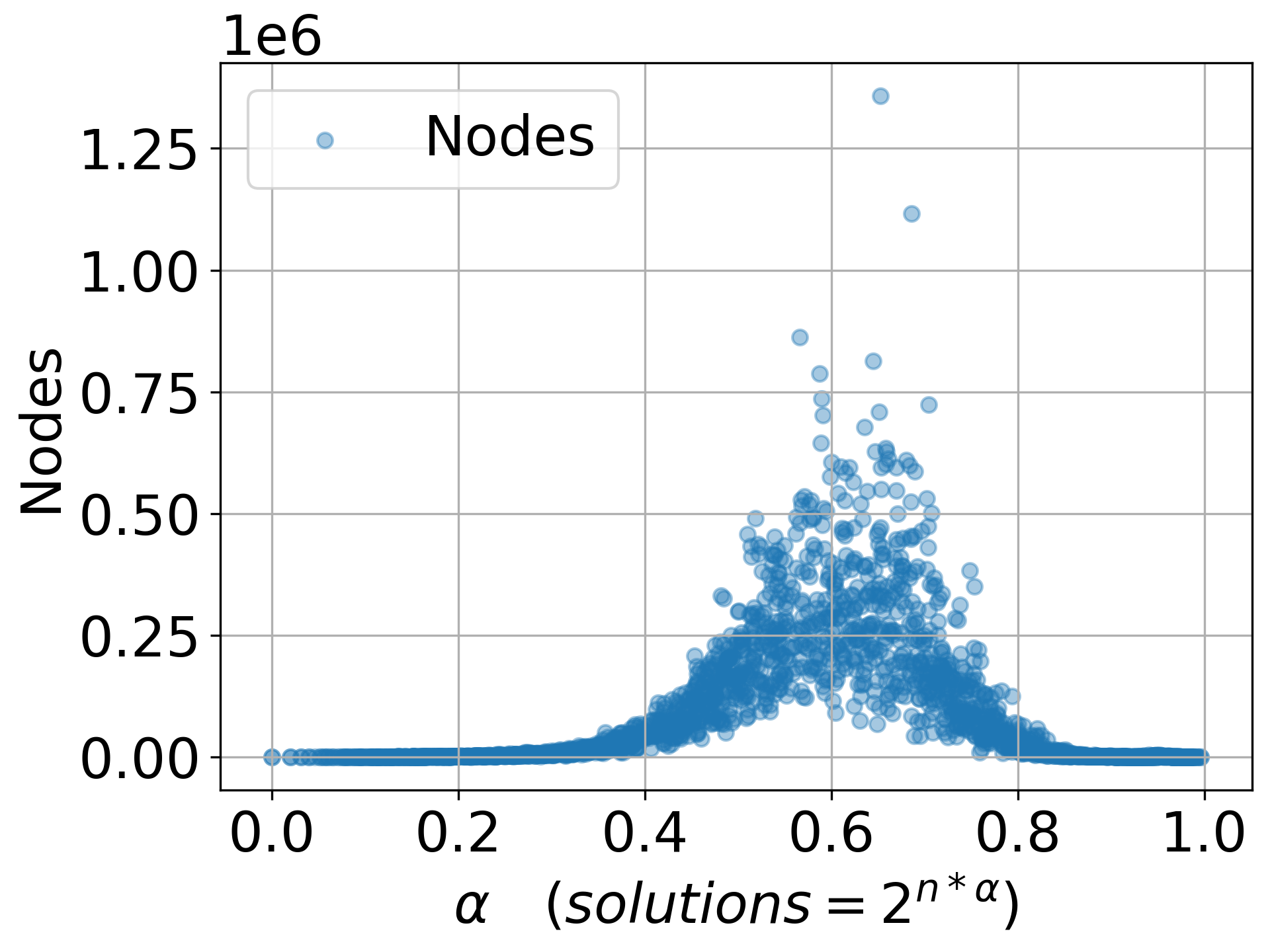}
		\caption{Nodes in OBDD, 50 vars}
		\label{fig:CUDD_nVa_3CNF_50vars}
	\end{subfigure}
	\hfill
	\begin{subfigure}[b]{0.475\linewidth}
		\centering
		\includegraphics[width=\linewidth]{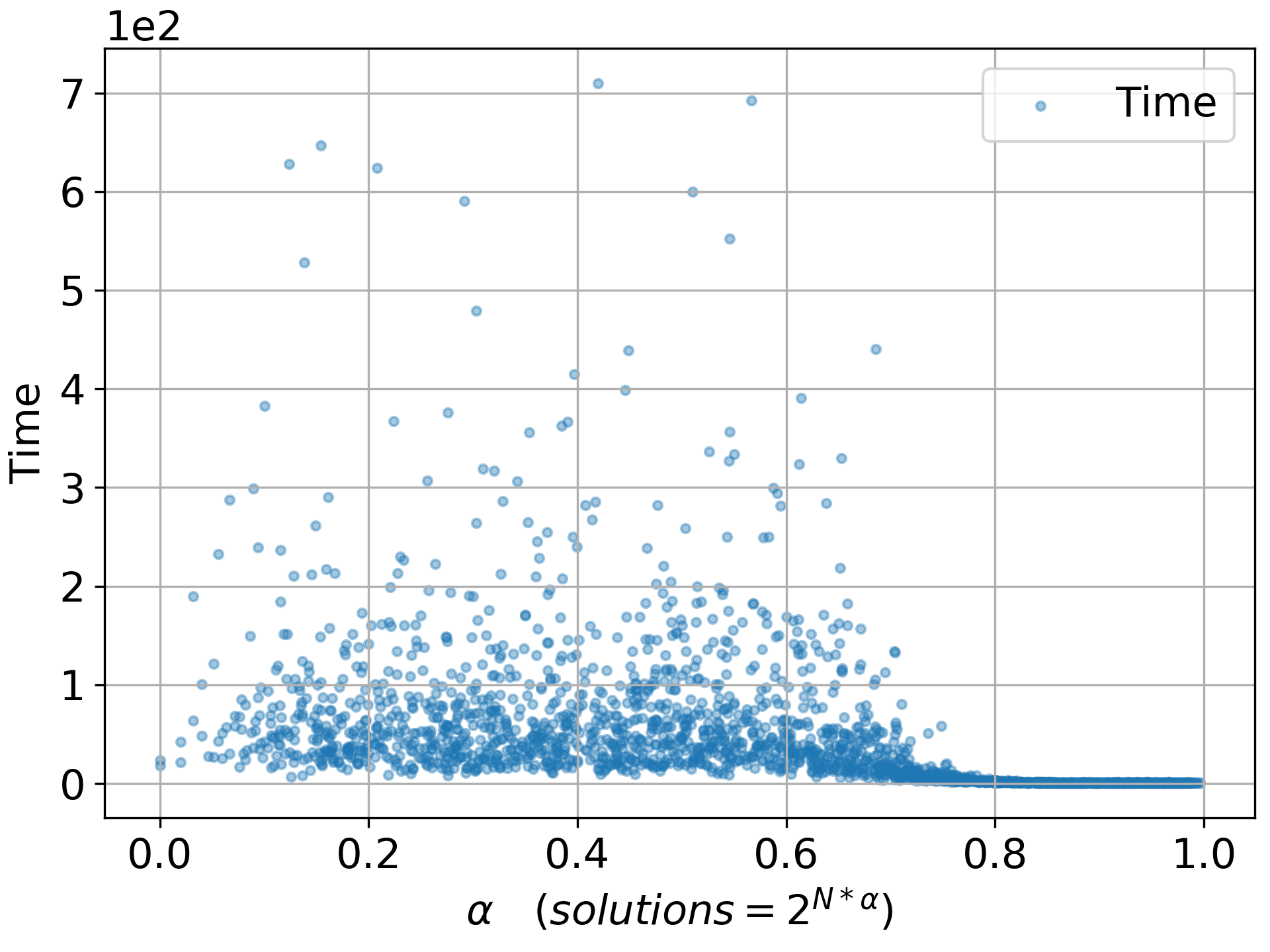}
		\caption{OBDD compile-time, 50 vars}
		\label{fig:CUDD_tVa_3CNF_50vars}
	\end{subfigure}
	\caption{Number of Nodes and compile-times for individual instances of 3-CNF against solution density}
	\label{fig:nVa_3CNF}
\end{figure}
Figure~\ref{fig:nVa_3CNF} shows the small-large-small variation in size of compilations with respect to solution density. We can observe that there exists a region of critical solution density for each target language around which the size of instances are very large. The location of phase transition appears to depend upon the target compilation  but it is difficult to comment upon the precise location and an extended study is required to generate independent instances for a given solution density.

\begin{figure}[!th]
	\centering
	\begin{subfigure}[b]{0.475\linewidth}
		\centering
		\includegraphics[width=\textwidth]{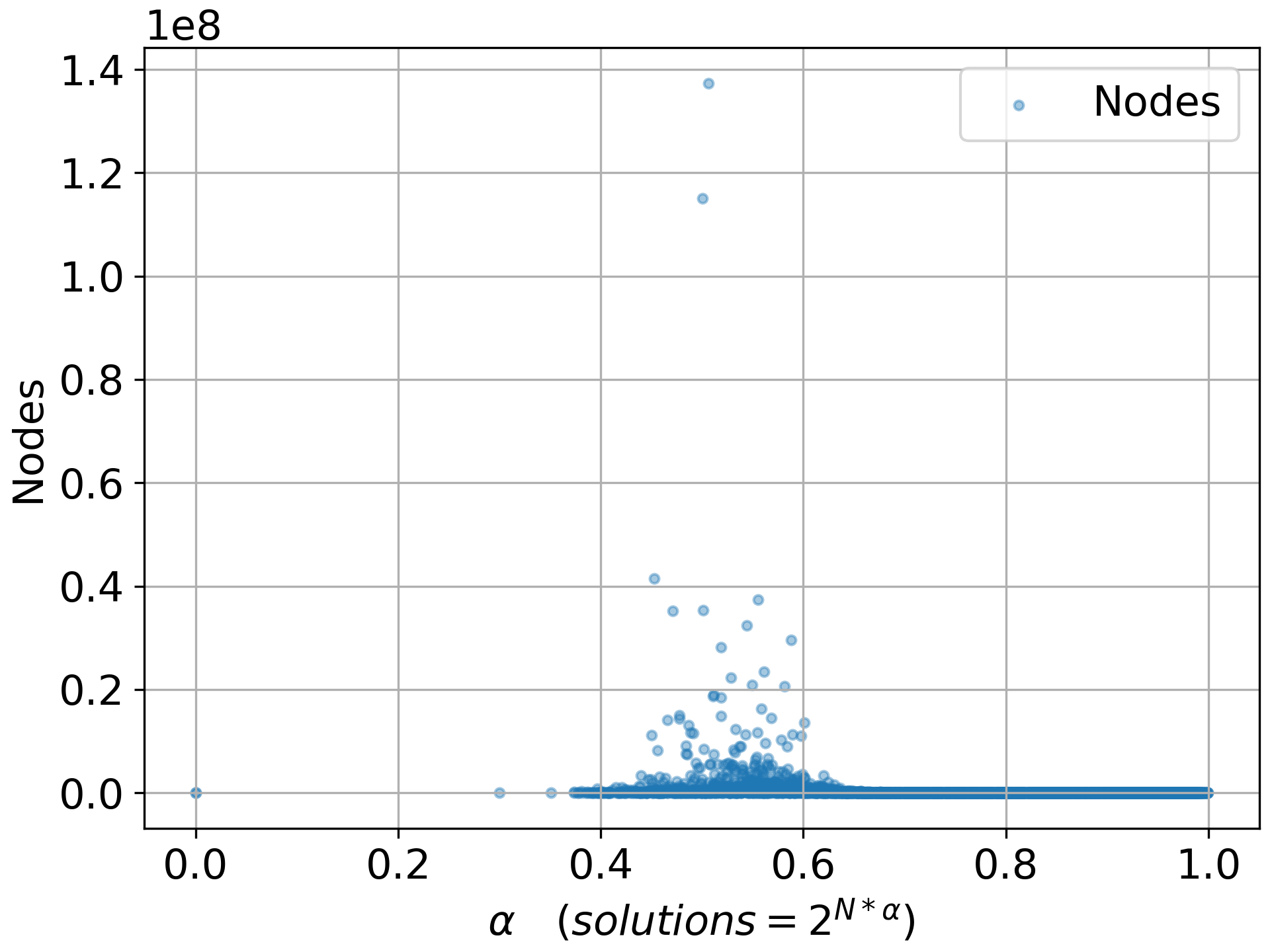}
		\caption{2-CNF and 600 vars}
		\label{fig:d4_nVa_2CNF_600vars}
	\end{subfigure}
	\hfill
	\begin{subfigure}[b]{0.475\linewidth}
		\centering
		\includegraphics[width=\textwidth]{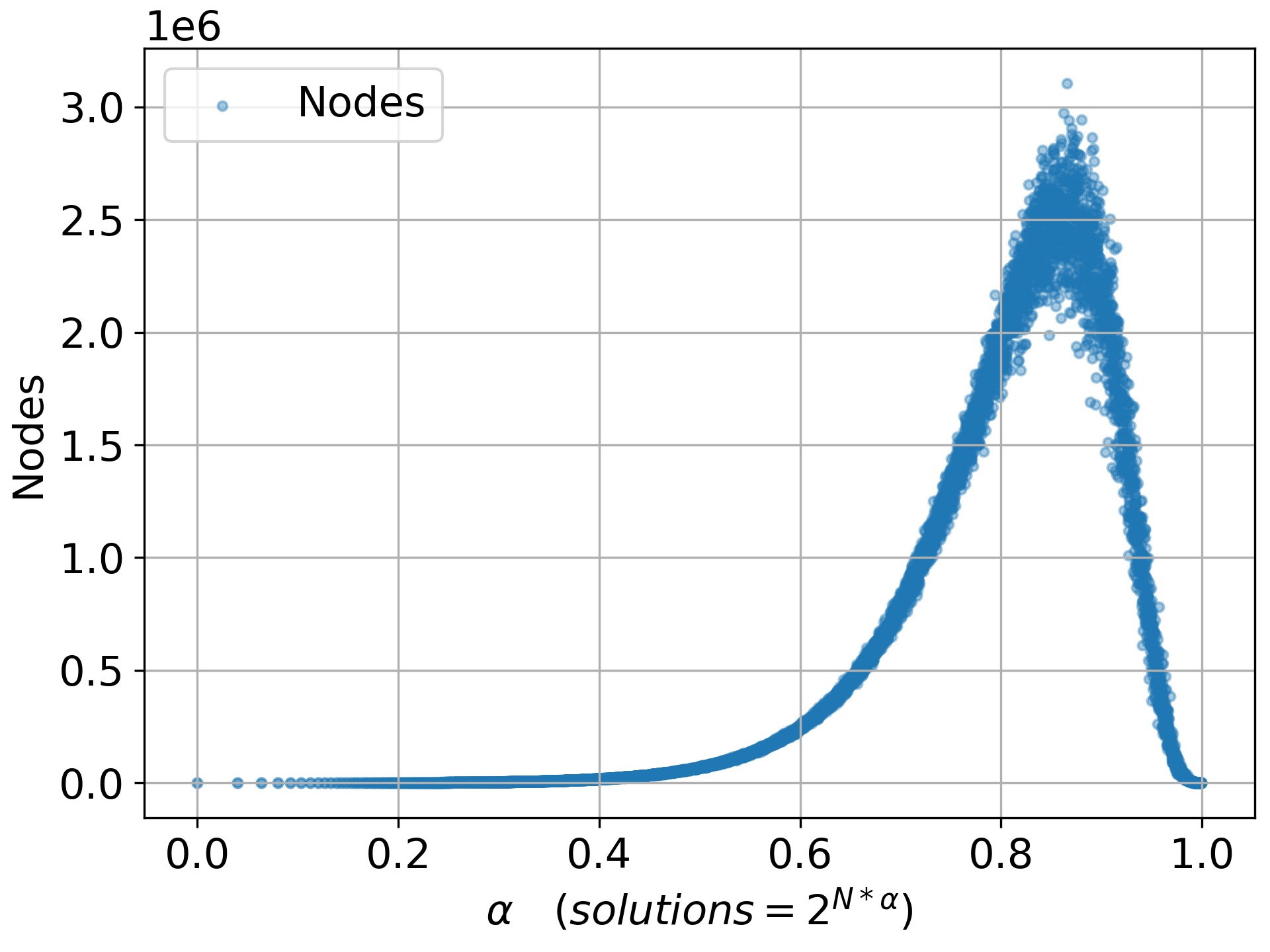}
		\caption{7-CNF and 30 vars}
		\label{fig:d4_nVa_7CNF_30vars}
	\end{subfigure}    
	\caption{\label{fig:d4_nVa_CNF}Nodes in d-DNNF vs solution density for different clause lengths($k$)}
\end{figure}

\noindent We have seen that solution density is a major parameter affecting the size and runtime of compiled instances. In this context, we seek to understand the impact of other parameters on the phase transition location with respect to solution density. We focus on two such parameters: clause length and the number of variables.  

\subsubsection{Impact of clause length} Figure~\ref{fig:d4_nVa_CNF} shows that as we increase the clause length, the location of phase transition point with respect to solution density moves closer to $1$.  It is worth remarking that in the context of the satisfiability, the location of phase transition, albeit with respect to clause densityk, is known to depend on the clause length, so a similar behavior in the context of knowledge compilation is indeed not surprising.  

\subsubsection{Size with number of variables} 
\begin{figure}[!th]
	\centering
	
	\begin{subfigure}[b]{0.48\linewidth}
		\centering
		\includegraphics[width=\textwidth]{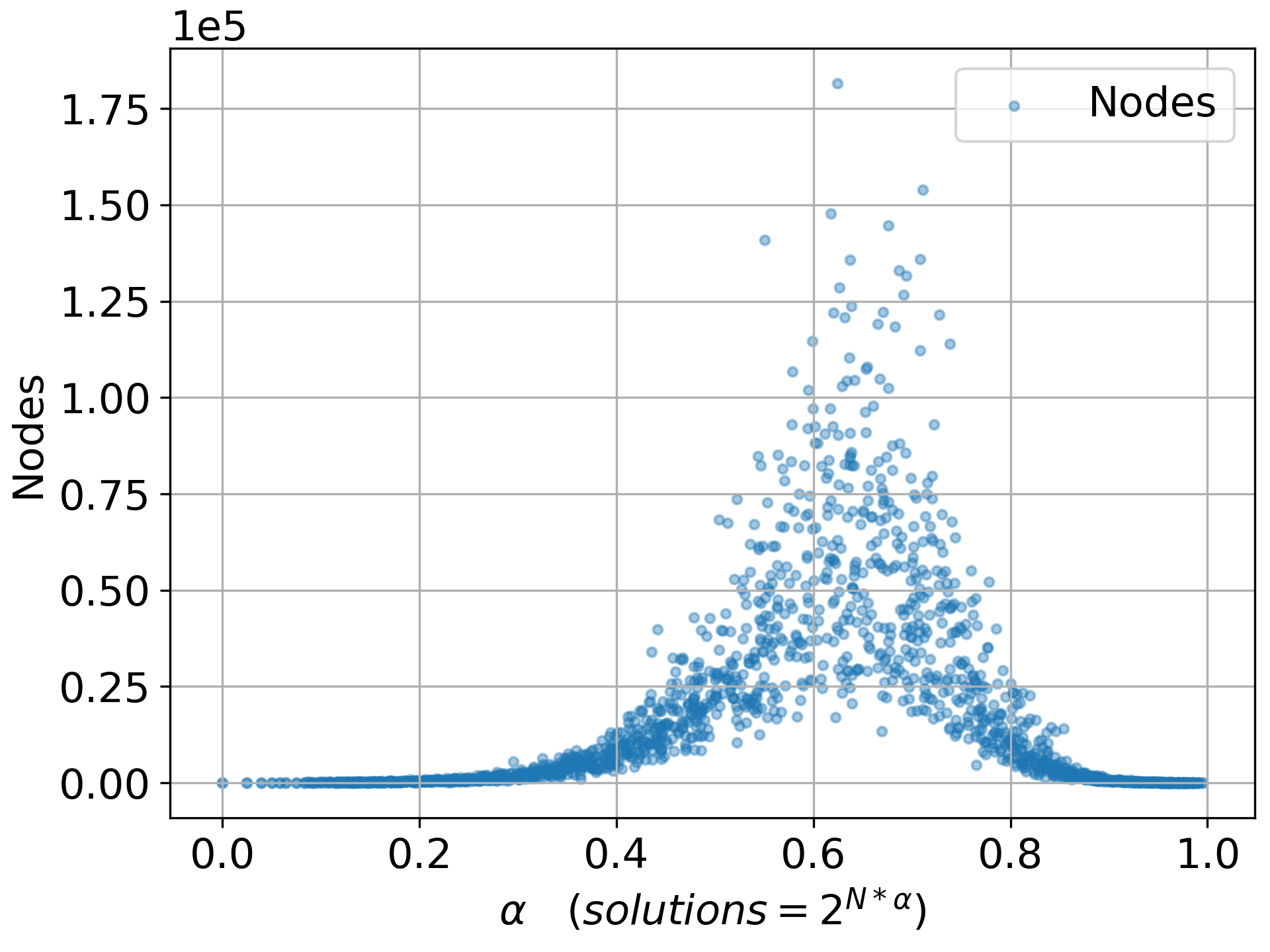}
		\caption{40 vars}
		\label{fig:d4_nVa_3CNF_40vars}
	\end{subfigure}
	\hfill
	\begin{subfigure}[b]{0.48\linewidth}
		\centering
		\includegraphics[width=\textwidth]{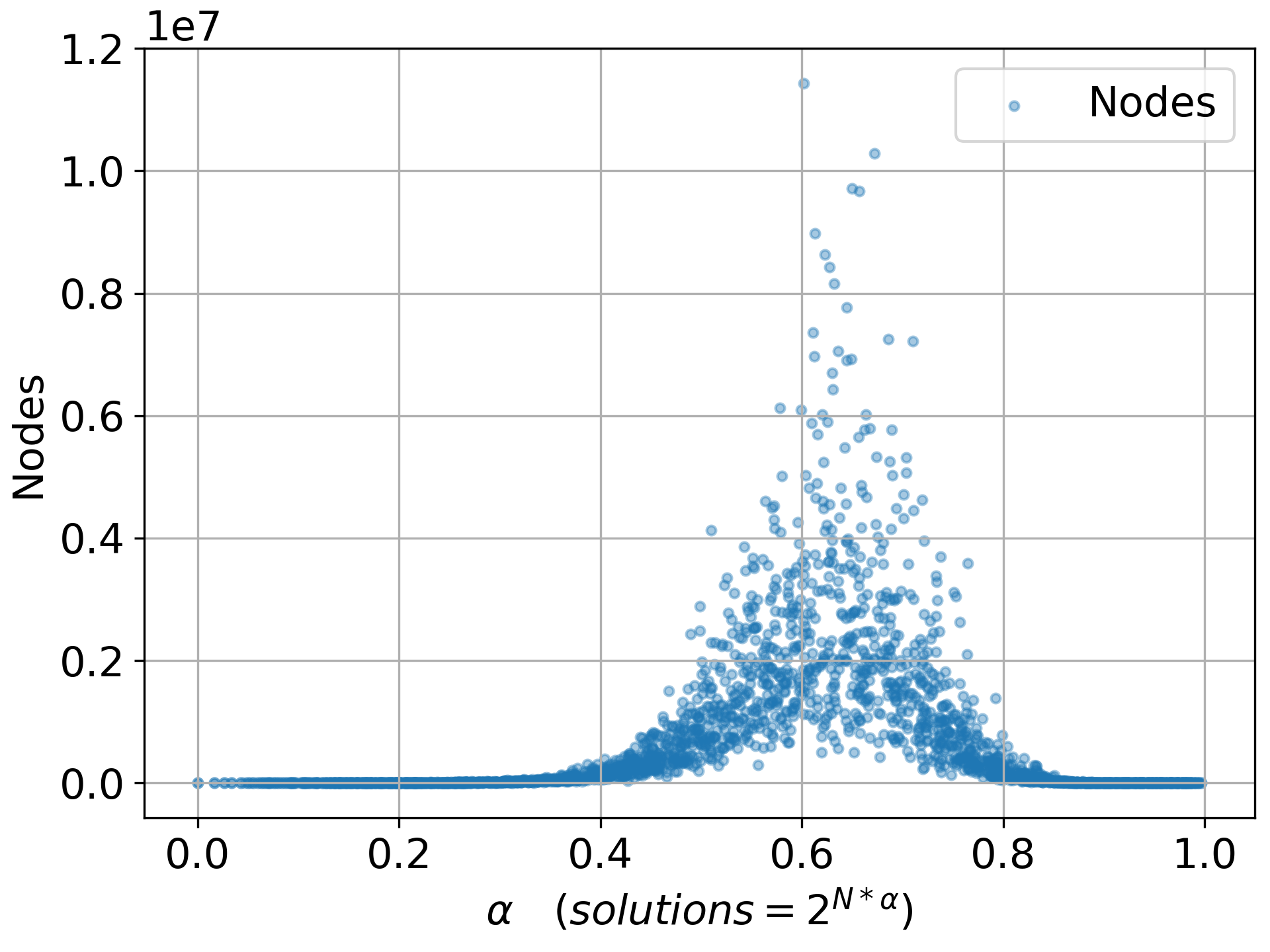}
		\caption{60 vars}
		\label{fig:d4_nVa_3CNF_60vars}
	\end{subfigure}
	\caption{Nodes in d-DNNF vs solution density($\alpha$) for different number of variables}
	\label{fig:d4_nVa_3CNF_vars}
\end{figure}

From Figure~\ref{fig:d4_nVa_3CNF_vars} we observe that the distribution of Nodes in d-DNNF compilation becomes sharper around the phase transition solution density with increasing number of variables.

\subsubsection{Runtime of compilation} From Figure~\ref{fig:nVa_3CNF}, we observe that the distribution of runtimes follows a similar easy-hard-easy pattern for d-dNNFs as well as SDDs but not OBDDs. The runtime behavior is similar to that of distribution of the number of nodes, as discussed in section~\ref{sec:CDexp}. 

We sum up our observations in the following conjecture:

\begin{conjecture}\label{conjecture3}
	For every integer $k\geq 2$, given the number of variables $n$ and a target language $\lang$ that is a subset of DNNF, there exists a positive real number $\alpha_k$ such that \\For each pair $(\alpha_1, \alpha_2)$,
	if $0 \leq \alpha_1 < \alpha_2 < \alpha_k$ then,
	$$\expect(\mathcal{N}_{\lang}(G_k(n,2^{\alpha_1 n}))) < \expect(\mathcal{N}_{\lang}(G_k(n,2^{\alpha_2n})))$$
	For each pair $(\alpha_1, \alpha_2)$, if $\alpha_k < \alpha_1 < \alpha_2 \leq 1$ then,
	$$\expect(\mathcal{N}_{\lang}(G_k(n,2^{\alpha_1n}))) > \expect(\mathcal{N}_{\lang}(G_k(n,2^{\alpha_2n})))$$
\end{conjecture}

\begin{figure}[t]
	\centering
	\begin{subfigure}[b]{0.5\textwidth}
		\includegraphics[width=0.95\textwidth]{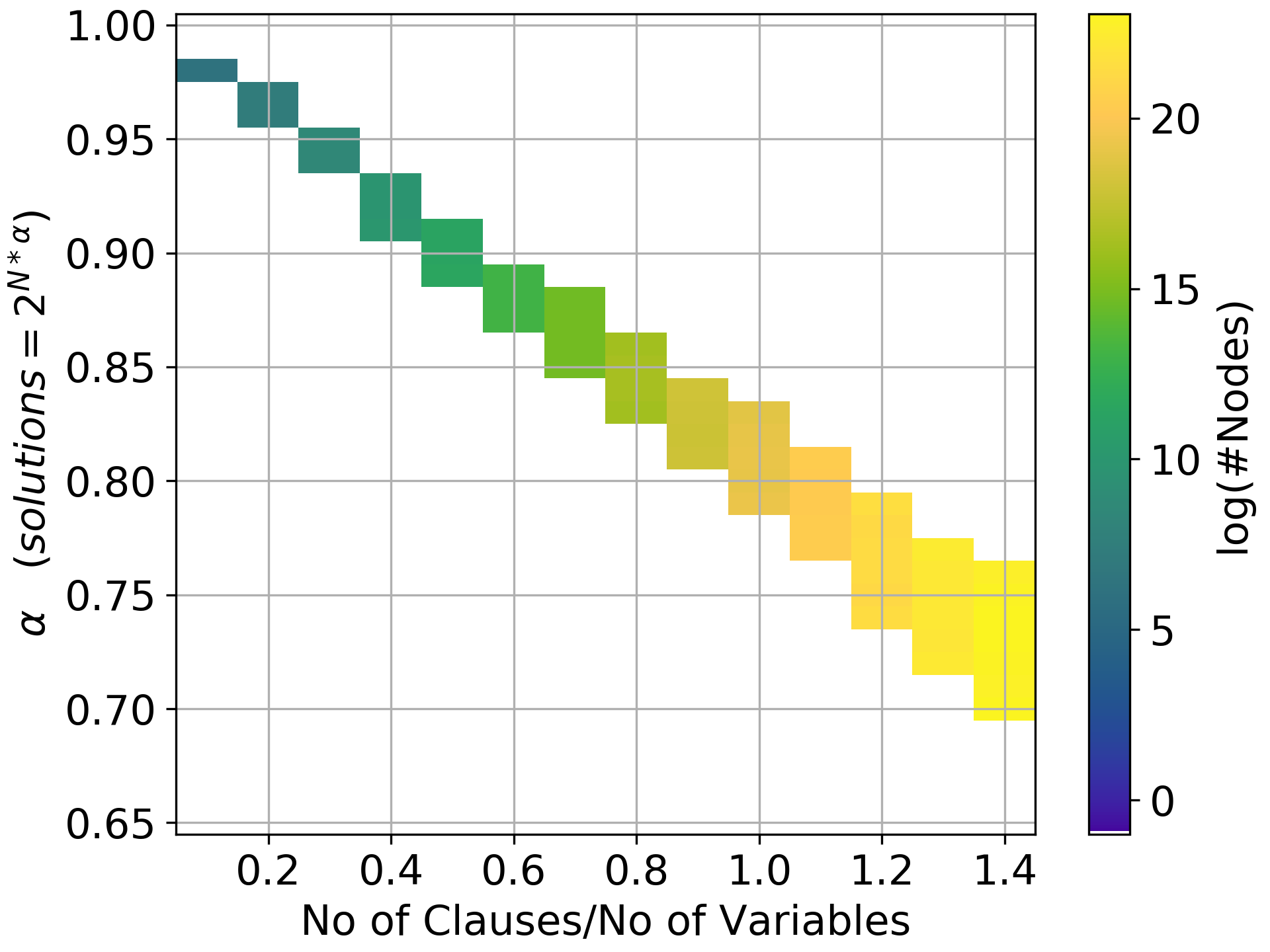}
		\caption{$r\in [0.1,1.4]$}
		\label{fig:d4_log_nVac_1d4}
	\end{subfigure}%
	\hspace*{-15pt}%
	\begin{subfigure}[b]{0.5\textwidth}
		\includegraphics[width=\textwidth]{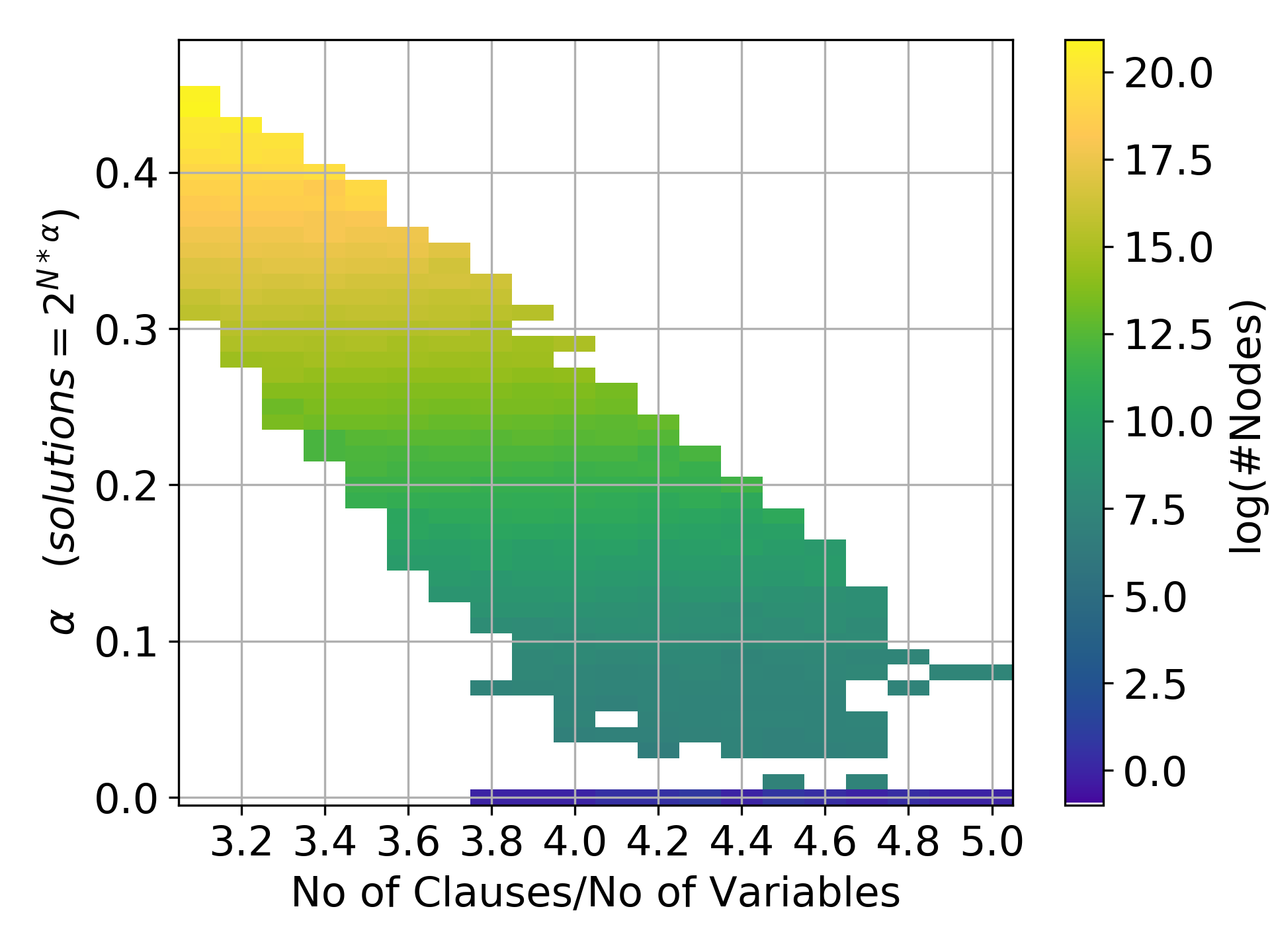}
		\caption{$r \in [3.1,5.0]$}
		\label{fig:d4_log_nVac_3d1}
	\end{subfigure}
	\caption{$\log$(Nodes) in d-DNNF compilation in solution vs clause density grid for 3-CNF}
	\label{fig:d4_log_nVac}
\end{figure}%

\subsection{Combined effect of clause and solution density}
We have seen that the density of solutions and clauses play a pivotal role, affecting the size and runtime. Also, given a clause density, the expected solution density is fixed and vice versa. This makes one wonder if there is a more complex relationship at play that affects the phase transition behavior of knowledge compilations. To investigate, we plot a heatmap (Figure~\ref{fig:d4_log_nVac}) on $\alpha \times r$ grid where the colours indicate the size of compilations. We employ random $3$-CNF instances with 70 variables used for comparisons with clause density, as described earlier in section \ref{sec:design}.  For each cell in the grid, we take an average of instances with the corresponding clause density that lie within the interval of solution density marked by the cell. Since the number of such instances can differ across the cells, we mark the average for a cell only if the number of instances is greater than $5$ to minimize the variance to a feasible extent. From Figure~\ref{fig:d4_log_nVac}, we observe that for low clause densities, varying solution density has minimal effect on the size of compilations. In contrast, for high clause densities, solution density has a dominant effect on the size as there is a minimal variation with clause density. 
On the other hand, near phase transition, both the parameters play a significant role, and the precise relationship is murkier.

\section{Effect of different tools}
\label{sec:tools}

Heuristics and compilation algorithms play a crucial role in perceived hardness as elaborated in Section~\ref{subsec:ObservingPhaseTransitions}. 
We experimented with bottom up (TheSDDPackage) as well as top down (MiniC2D) compilation strategies for SDDs, predefined total variable ordering against dynamic ordering for OBDDs (CUDD) and different decomposition techniques for d-DNNFs (C2D, Dsharp and D4). Notably, target language for MiniC2D is decision-SDD and CUDD with predefined total variable order is OBDD$_>$ \cite{OD18,DM02}, which are less succinct than SDD and OBDD respectively. We state our representative observations for 3-CNF here. We observed that both MiniC2D and The SDD Package show maximum number of nodes around clause density 1.8 and solution density around 0.62. However, runtime for MiniC2D peaks around clause density 1.8 while TheSDDPackage peaks around clause density 2.0. In case of BDDs, we observed that disabling dynamic variable reordering shifts the peak (number of nodes) clause density from 2.0 to 1.5 and peak (number of nodes) solution density from 0.62 to 0.75. The observations for runtimes of OBDDs are much more involved due to reasons discussed in Section~\ref{subsec:ObservingPhaseTransitions}. For d-DNNFs, we observe that the peak (number of nodes) clause density stays around $1.8$ and peak solution density stays around $0.62$ irrespective of the hypergraph partitioning algorithm. Similar observations were recorded for runtime as well in case of d-DNNF. Summing up, we observe that while the precise behaviour of phase transition (for example, its location) can depend upon the heuristics employed in the process, the general behaviour persists irrespectively. 
\section{Conclusion}
\label{sec:conclusion}

Our study provides evidence of phase transition behavior with respect to clause as well as solution density. While, both these parameters are linearly linked in expectation, it is interesting that varying the number of solutions on a fixed clause density leads to significant variation in the expected size of compilations. %
In terms of the complexity of compilations, we found the expected size is at least quasi-polynomial and expected runtime is exponential in the number of variables with varying clause density for state-of-the-art knowledge compilers. 
We believe that this paper opens up new directions for theoretical studies in an attempt to explain our empirical obeservations and conjectures.

\paragraph{Acknowledgments}
This work was supported in part by National Research Foundation Singapore under its NRF Fellowship Programme[NRF-NRFFAI1-2019-0004 ] and AI Singapore Programme [AISG-RP-2018-005],  and NUS ODPRT Grant [R-252-000-685-13]. The computational work for this article was performed on resources of the National Supercomputing Centre, Singapore \url{https://www.nscc.sg}. Any opinions, findings and conclusions or recommendations expressed in this material are those of the author(s) and do not reflect the views of National Research Foundation, Singapore.

\clearpage
\appendix
\section*{APPENDIX}
In this Appendix, we provide some detailed results from our experiments. To aid readability, some of the MainFigures from the main paper are also included here. 
\section{Phase transition with clause density}
\label{sec:appendix-CDexp}
In this section, we show the variations in size and compile times for d-DNNFs, SDDs and OBDDs with clause density from our experiments.

\subsection{Observing the phase transitions}

\subsubsection{Size with number of variables} 
\begin{figure*}[!h]
	\centering
	\begin{subfigure}[b]{0.30\textwidth}
		\centering
		\includegraphics[width=\linewidth]{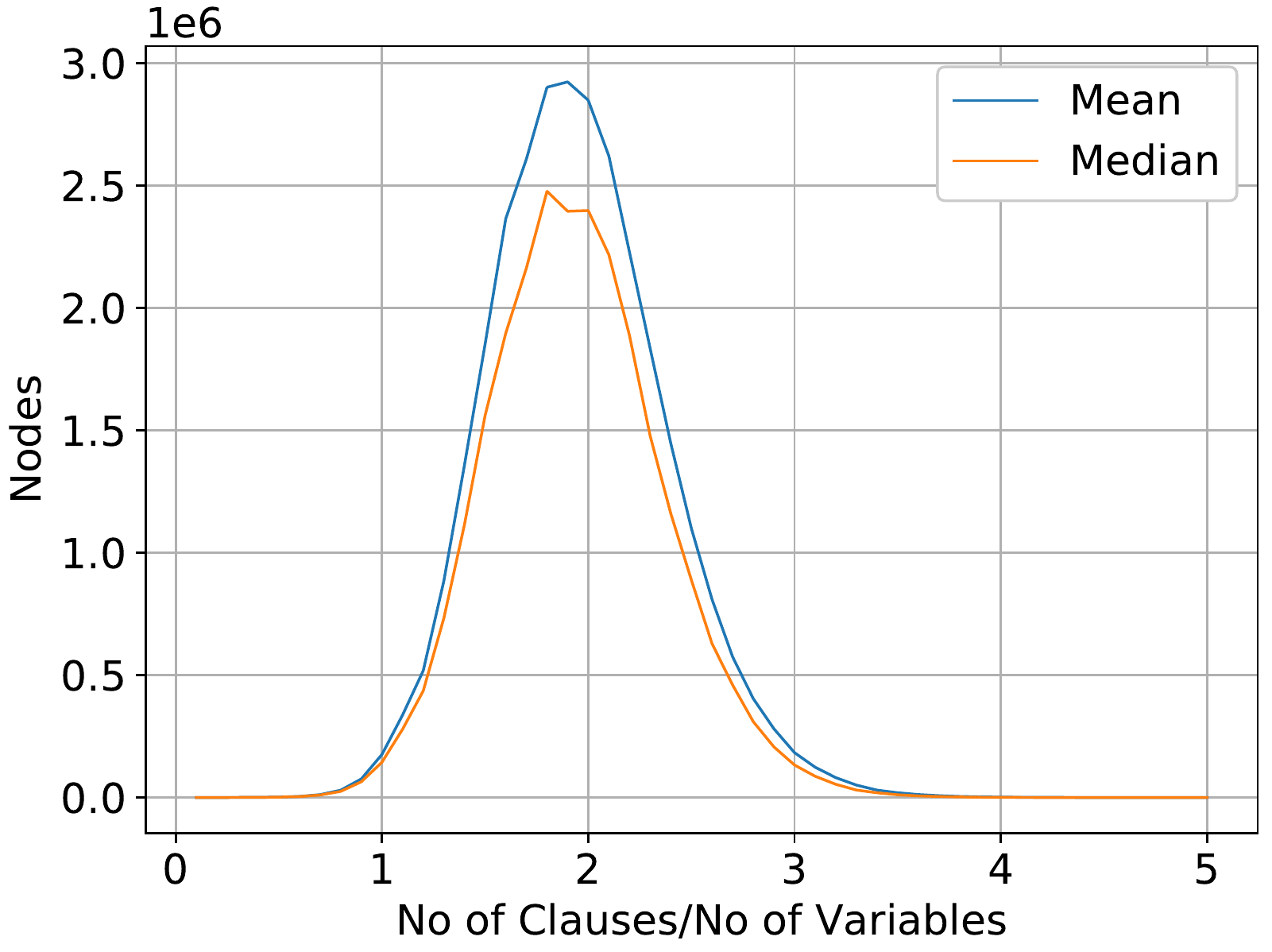}
		\caption{60 vars}
		\label{fig:appendix:d4_nVc_3CNF_60vars}
	\end{subfigure}
	\hfill
	\begin{subfigure}[b]{0.30\textwidth}
		\centering
		\includegraphics[width=\linewidth]{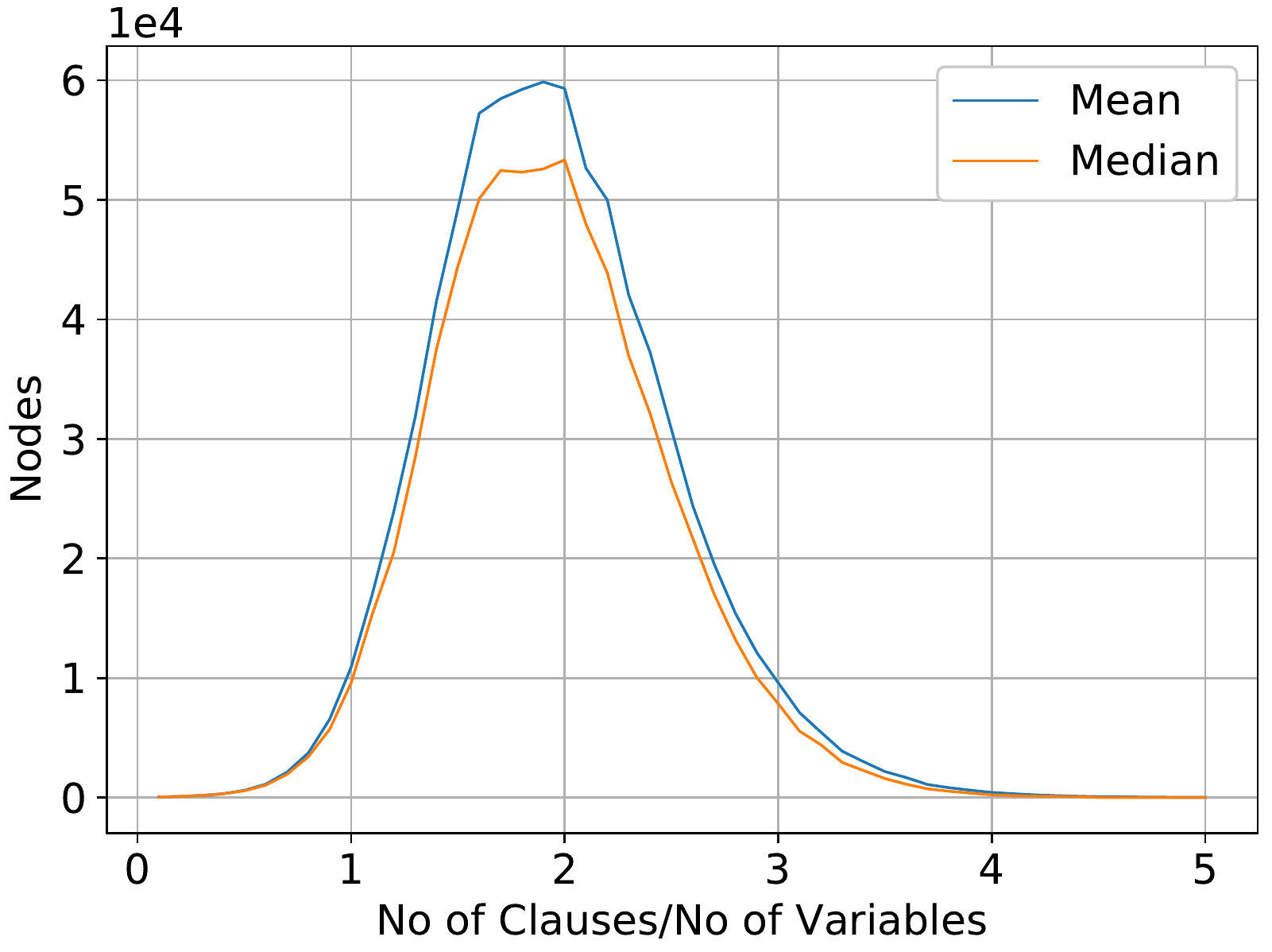}
		\caption{40 vars}
		\label{fig:appendix:d4_nVc_3CNF_40vars}
	\end{subfigure}
	\hfill
	\begin{subfigure}[b]{0.30\textwidth}
		\centering
		\includegraphics[width=\linewidth]{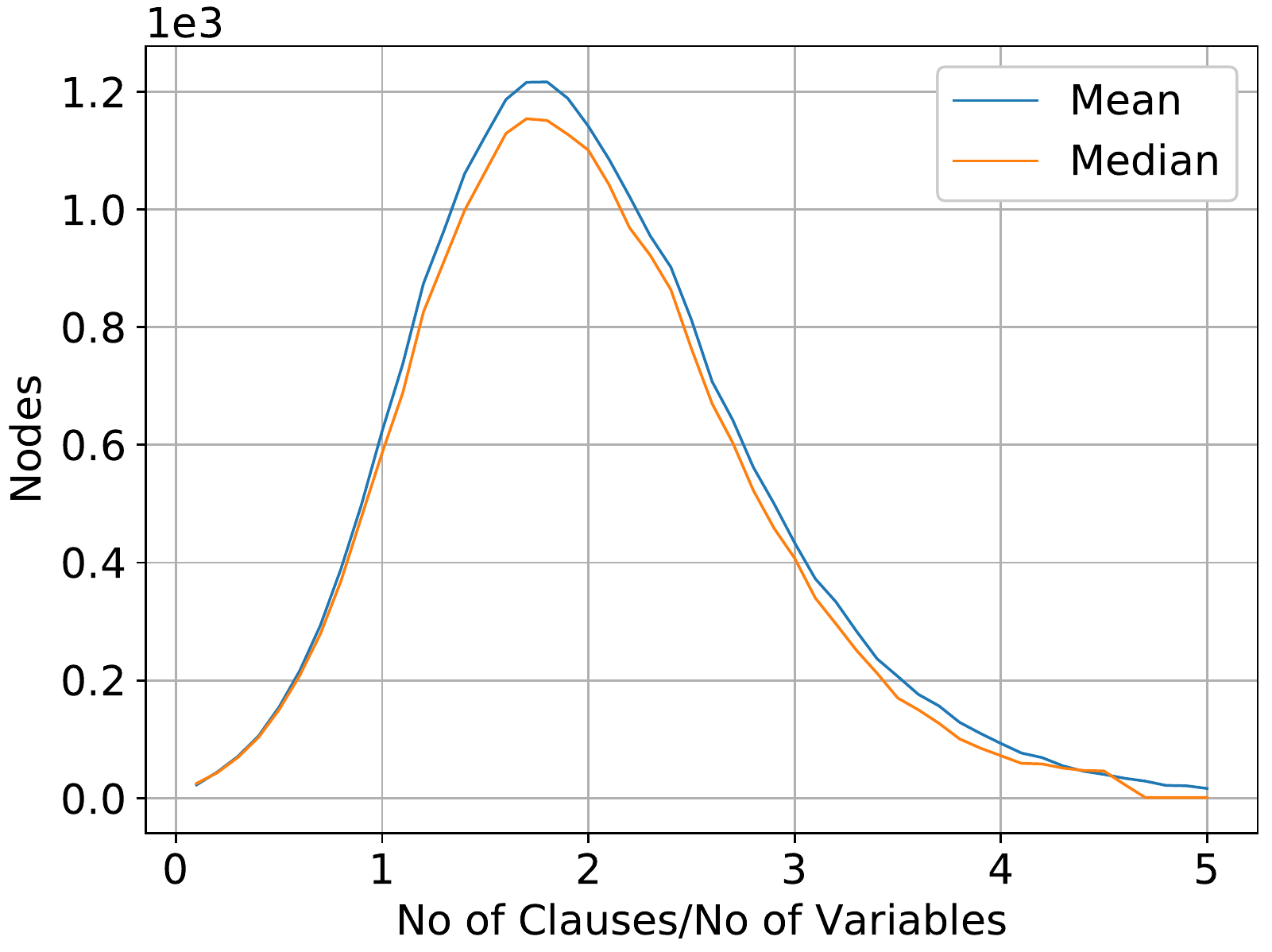}
		\caption{20 vars}
		\label{fig:appendix:d4_nVc_3CNF_20vars}
	\end{subfigure}
	\caption{Nodes in d-DNNF vs clause density with different number of variables}
	\label{fig:appendix:d4_nVc_3CNF}
\end{figure*}

\begin{figure*}[!h]
	\centering
	\begin{subfigure}[b]{0.30\textwidth}
		\centering
		\includegraphics[width=\linewidth]{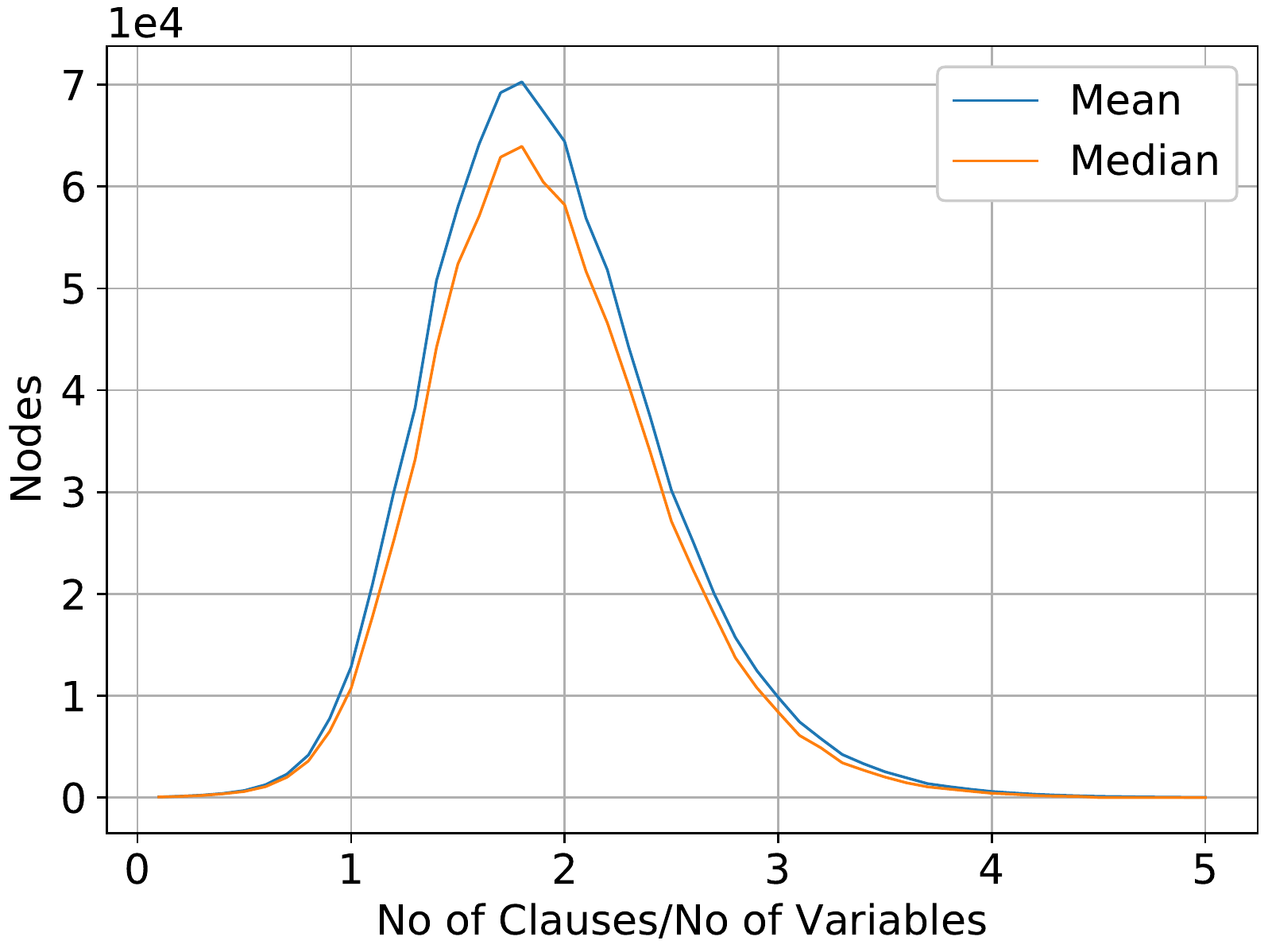}
		\caption{40 vars}
		\label{fig:appendix:sdd_nVc_3CNF_40vars}
	\end{subfigure}
	\hfill
	\begin{subfigure}[b]{0.30\textwidth}
		\centering
		\includegraphics[width=\linewidth]{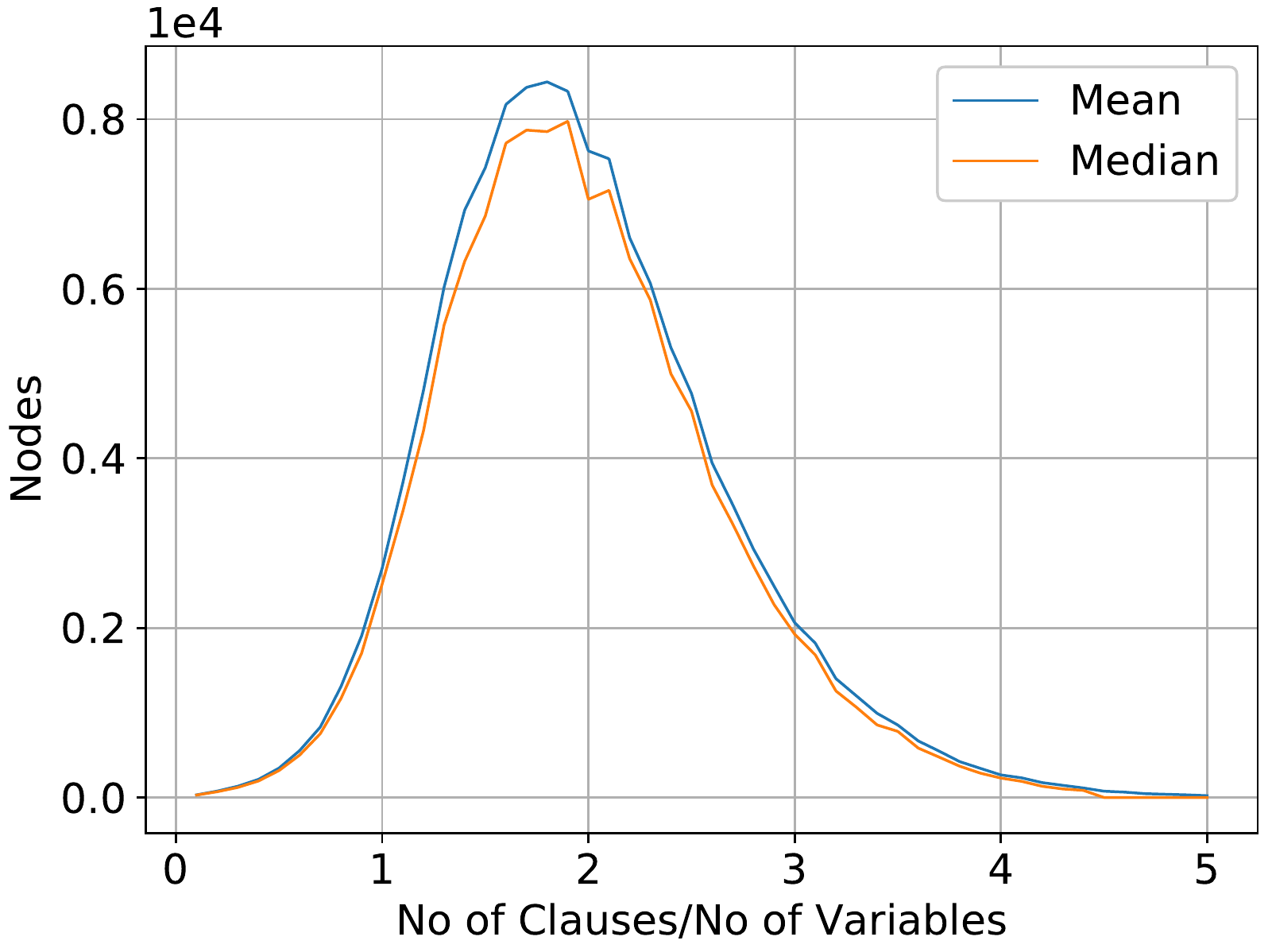}
		\caption{30 vars}
		\label{fig:appendix:sdd_nVc_3CNF_30vars}
	\end{subfigure}
	\hfill
	\begin{subfigure}[b]{0.30\textwidth}
		\centering
		\includegraphics[width=\linewidth]{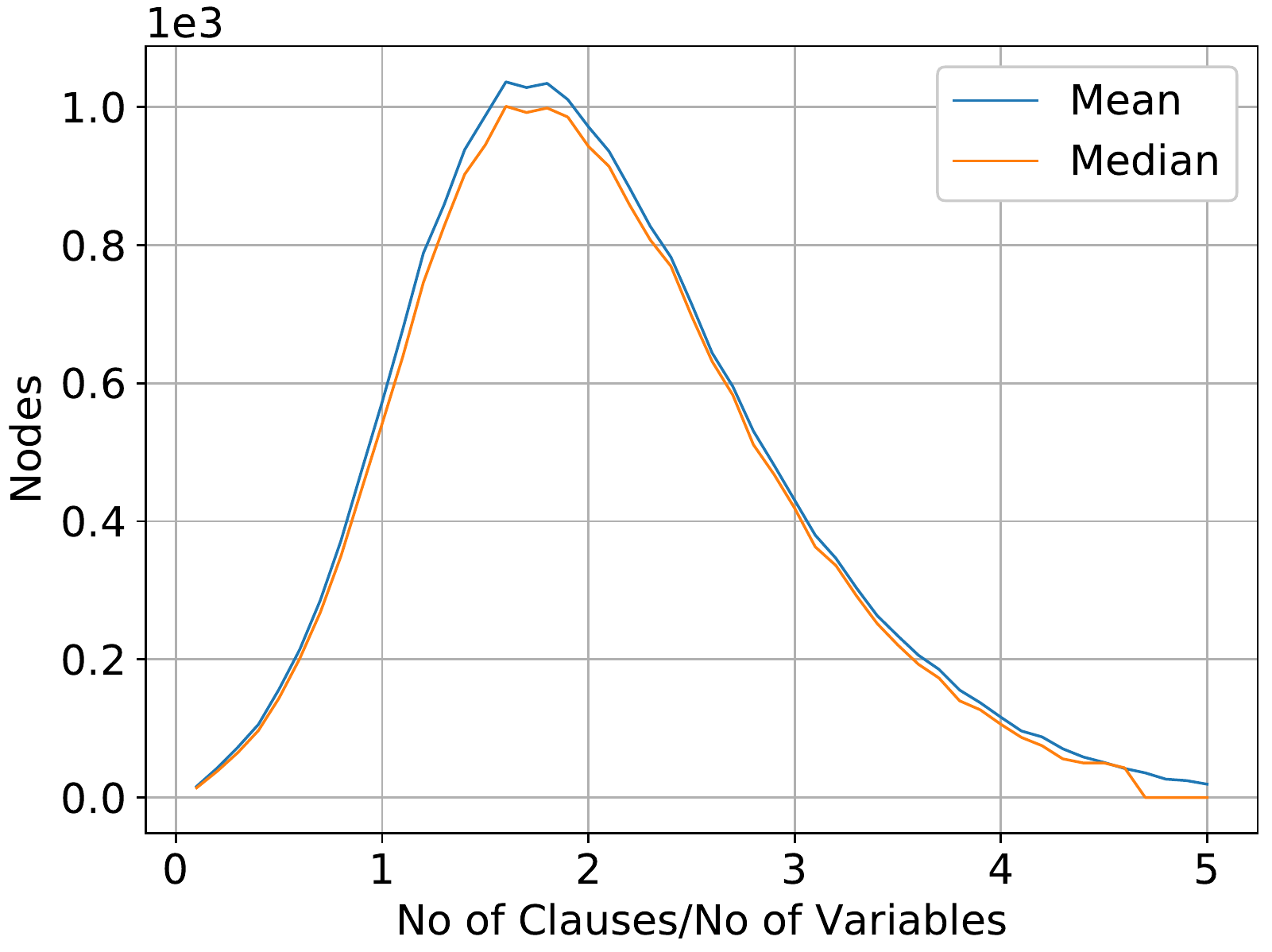}
		\caption{20 vars}
		\label{fig:appendix:sdd_nVc_3CNF_20vars}
	\end{subfigure}
	\caption{Nodes in SDD vs clause density with different number of variables}
	\label{fig:appendix:sdd_nVc_3CNF}
\end{figure*}

\begin{figure*}[!h]
	\centering
	\begin{subfigure}[b]{0.30\textwidth}
		\centering
		\includegraphics[width=\linewidth]{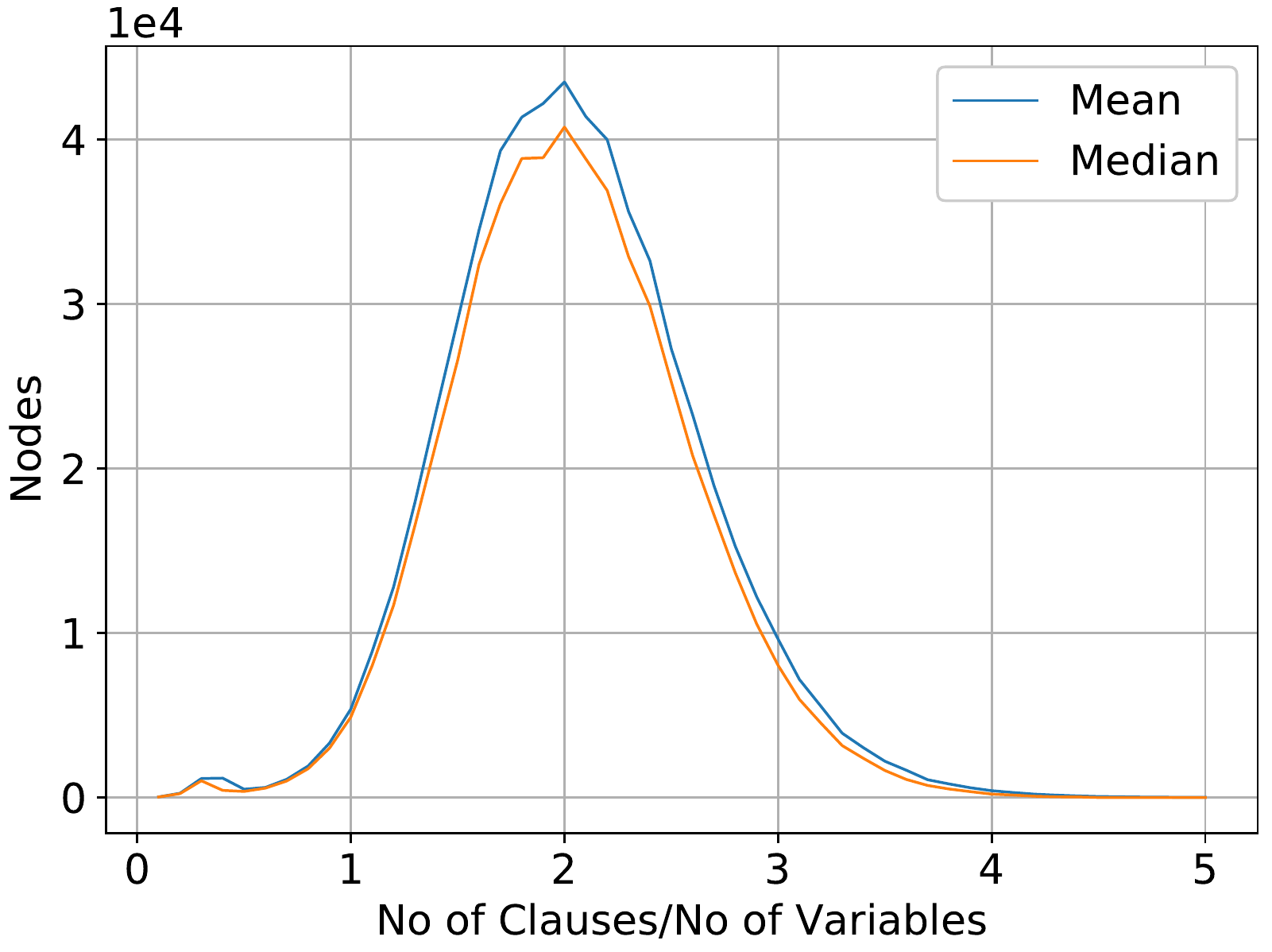}
		\caption{40 vars}
		\label{fig:appendix:CUDD_nVc_3CNF_40vars}
	\end{subfigure}
	\hfill
	\begin{subfigure}[b]{0.30\textwidth}
		\centering
		\includegraphics[width=\linewidth]{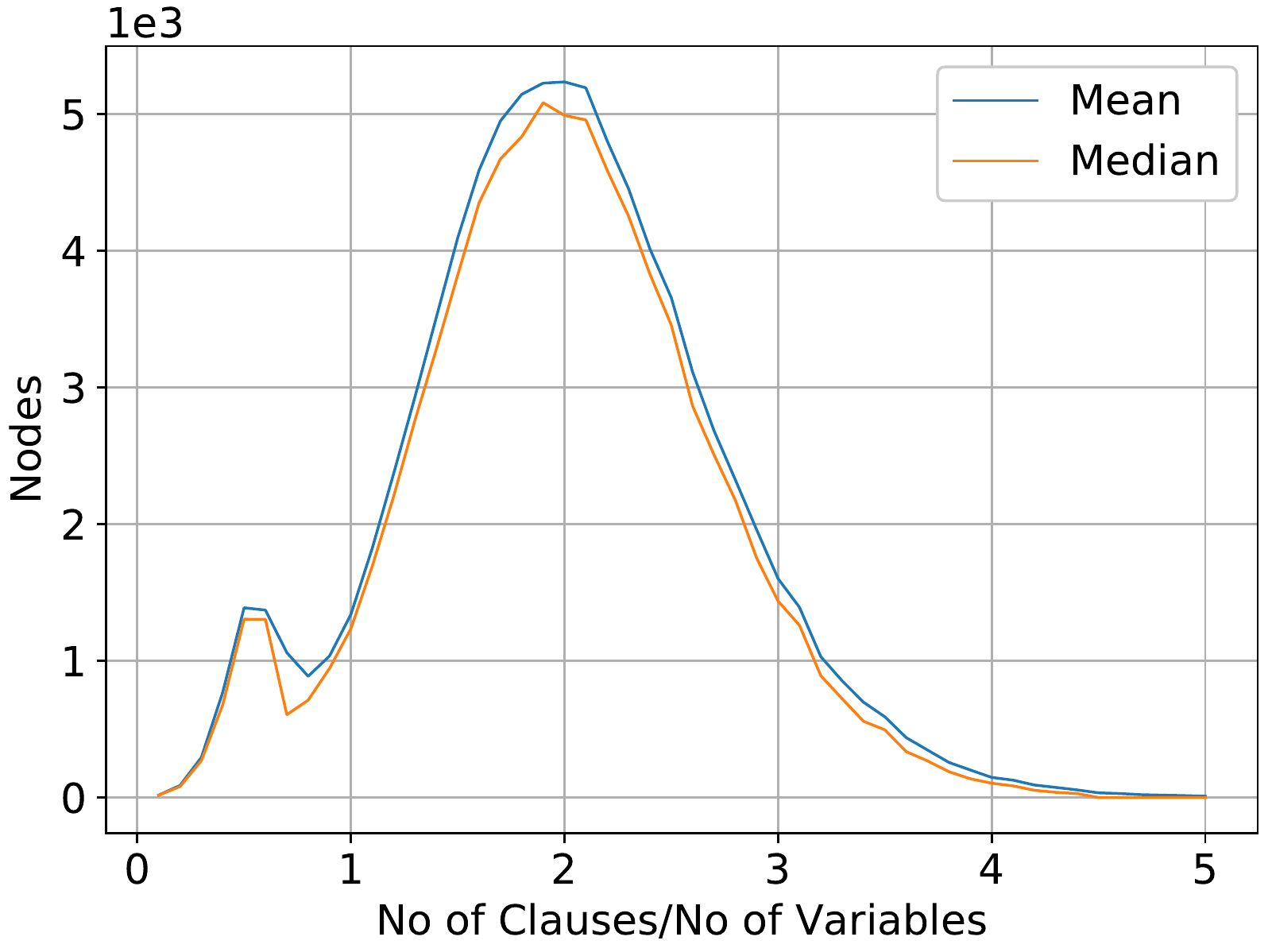}
		\caption{30 vars}
		\label{fig:appendix:CUDD_nVc_3CNF_30vars}
	\end{subfigure}
	\hfill
	\begin{subfigure}[b]{0.30\textwidth}
		\centering
		\includegraphics[width=\linewidth]{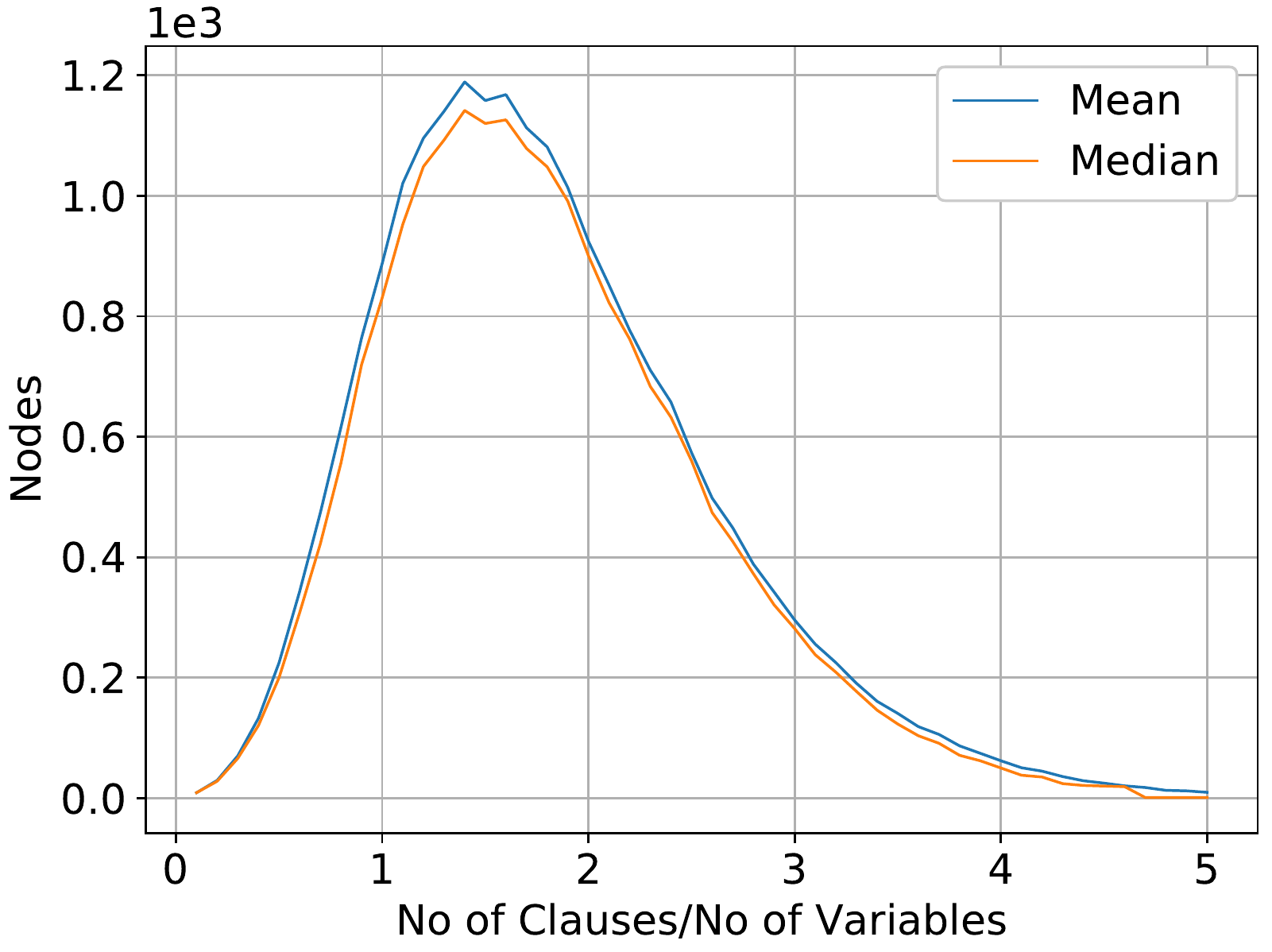}
		\caption{20 vars}
		\label{fig:appendix:CUDD_nVc_3CNF_20vars}
	\end{subfigure}
	\caption{Nodes in OBDD vs clause density with different number of variables}
	\label{fig:appendix:CUDD_nVc_3CNF}
\end{figure*}

\begin{figure*}[!h]
	\centering
	\begin{subfigure}[b]{0.30\linewidth}
		\centering
		\includegraphics[width=\linewidth]{MainFigures/dDNNF/d4act_log_mean_nodesVclauses_3cnf1d0exp}
		\caption{d-DNNF, 70 vars}
		\label{fig:appendix:d4_log_nVc_3CNF}
	\end{subfigure}
	\hfill
	\begin{subfigure}[b]{0.30\linewidth}
		\centering
		\includegraphics[width=\linewidth]{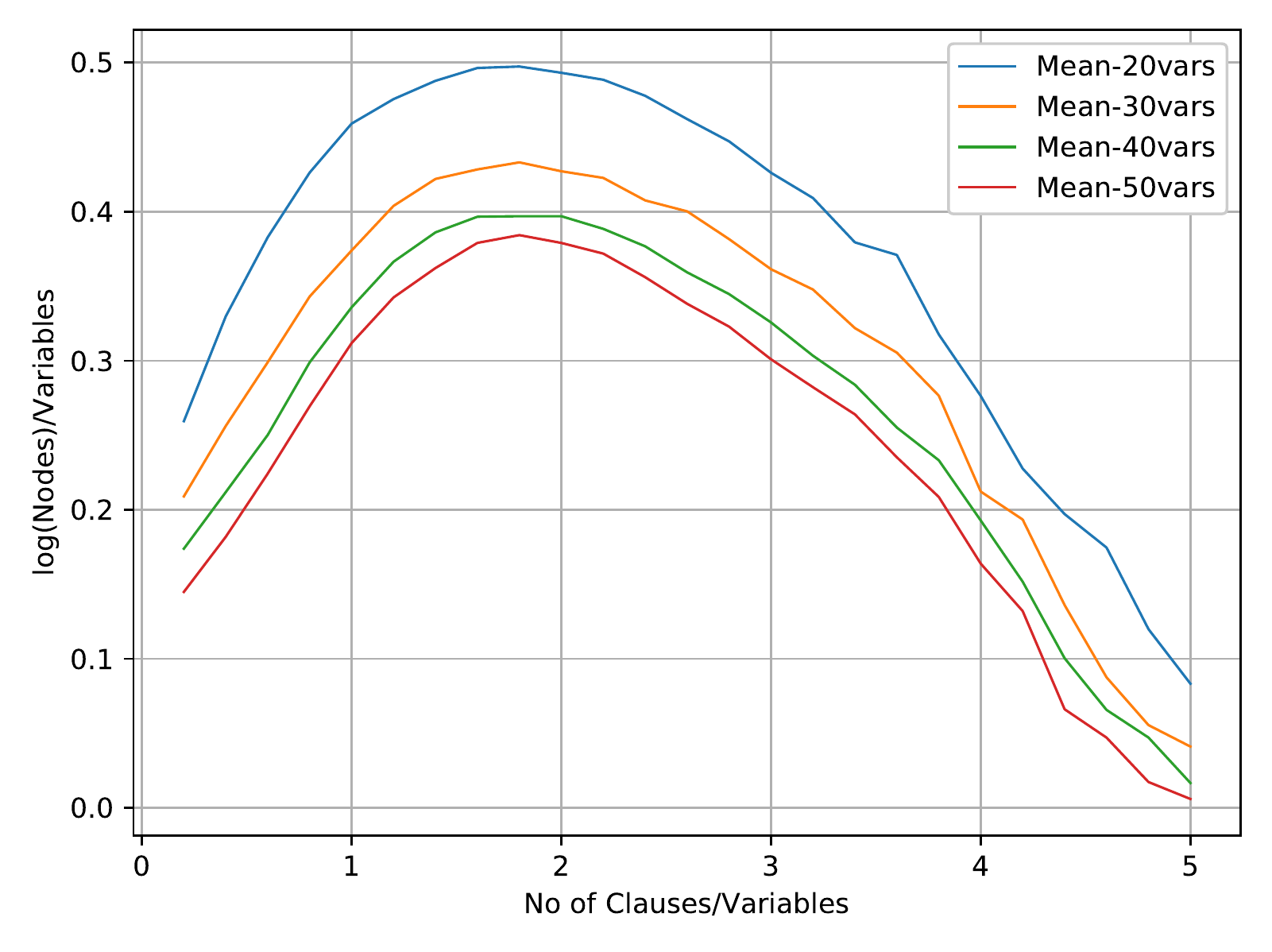}
		\caption{SDD, 50 vars}
		\label{fig:appendix:sdd_log_nVc_3CNF}
	\end{subfigure}
	\hfill
	\begin{subfigure}[b]{0.30\linewidth}
		\centering
		\includegraphics[width=\linewidth]{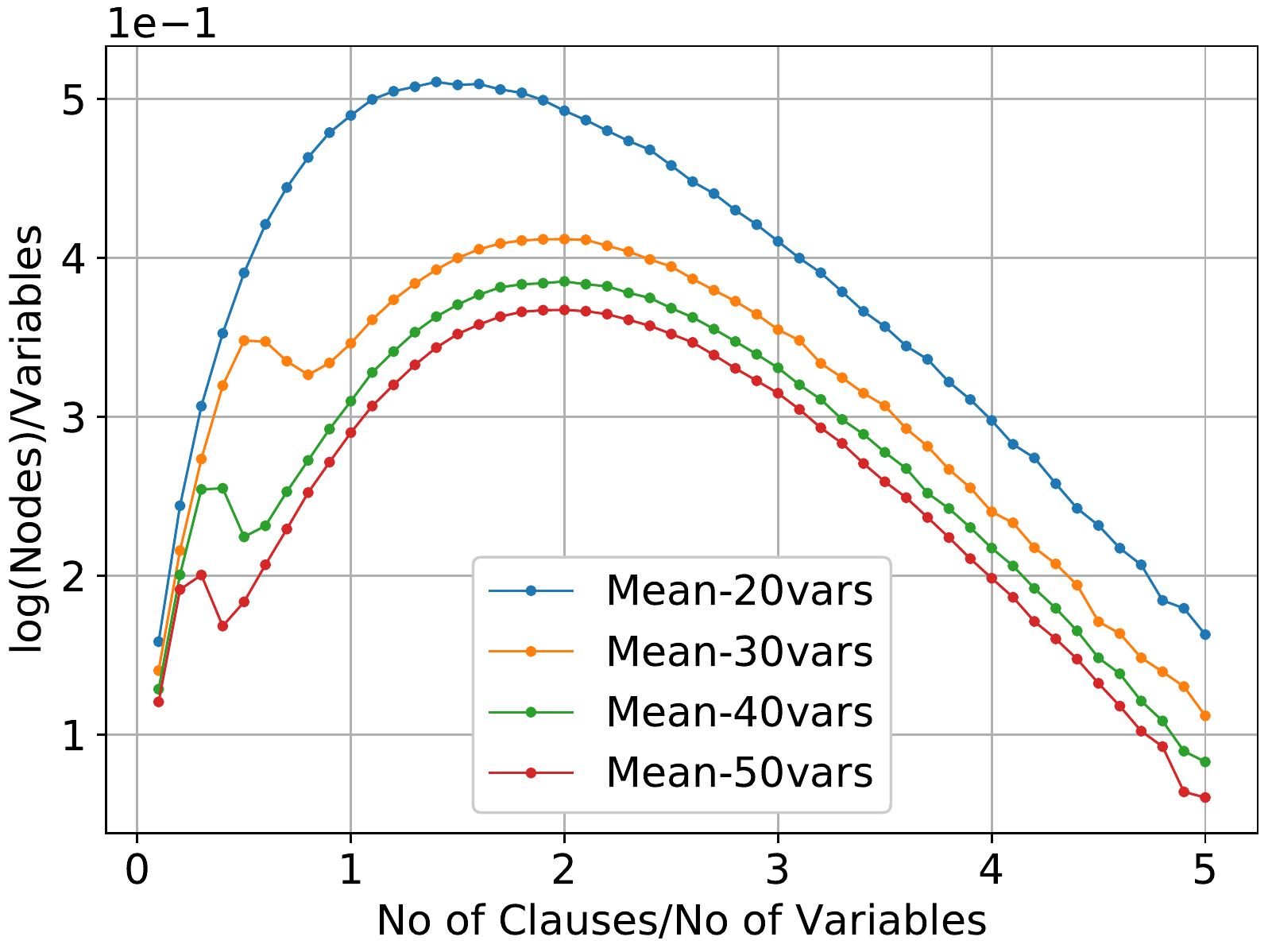}
		\caption{OBDD, 50 vars}
		\label{fig:appendix:CUDD_log_nVc_3CNF}
	\end{subfigure}
	\caption{log(nodes)/variables vs clause density for different target languages}
	\label{fig:appendix:log_nVc_3CNF}
\end{figure*}

We look at the effect of variation in the number of variables on phase transition in Figures \ref{fig:appendix:d4_nVc_3CNF}, \ref{fig:appendix:sdd_nVc_3CNF}, \ref{fig:appendix:CUDD_nVc_3CNF}.
Overall, the transition appears sharper with increase in number of variables. Figure~\ref{fig:appendix:log_nVc_3CNF} compares $\log(nodes)/variables$ for different number of variables on same graph.

\subsubsection{Runtime of compilation}
\begin{figure*}[!h]
	\centering
	\begin{subfigure}[b]{0.30\textwidth}
		\centering
		\includegraphics[width=\linewidth]{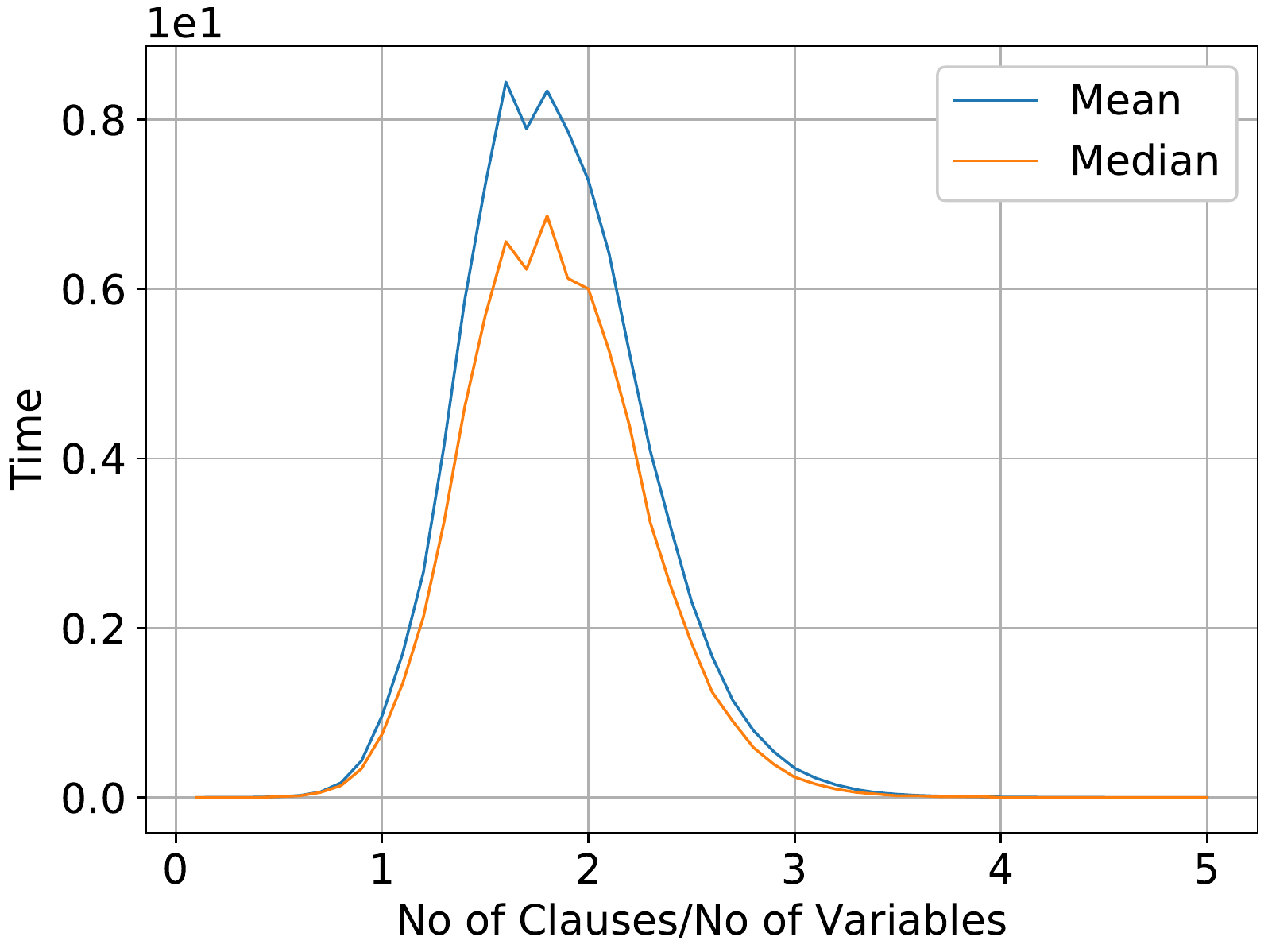}
		\caption{60 vars}
		\label{fig:appendix:d4_tVc_3CNF_60vars}
	\end{subfigure}
	\hfill
	\begin{subfigure}[b]{0.30\textwidth}
		\centering
		\includegraphics[width=\linewidth]{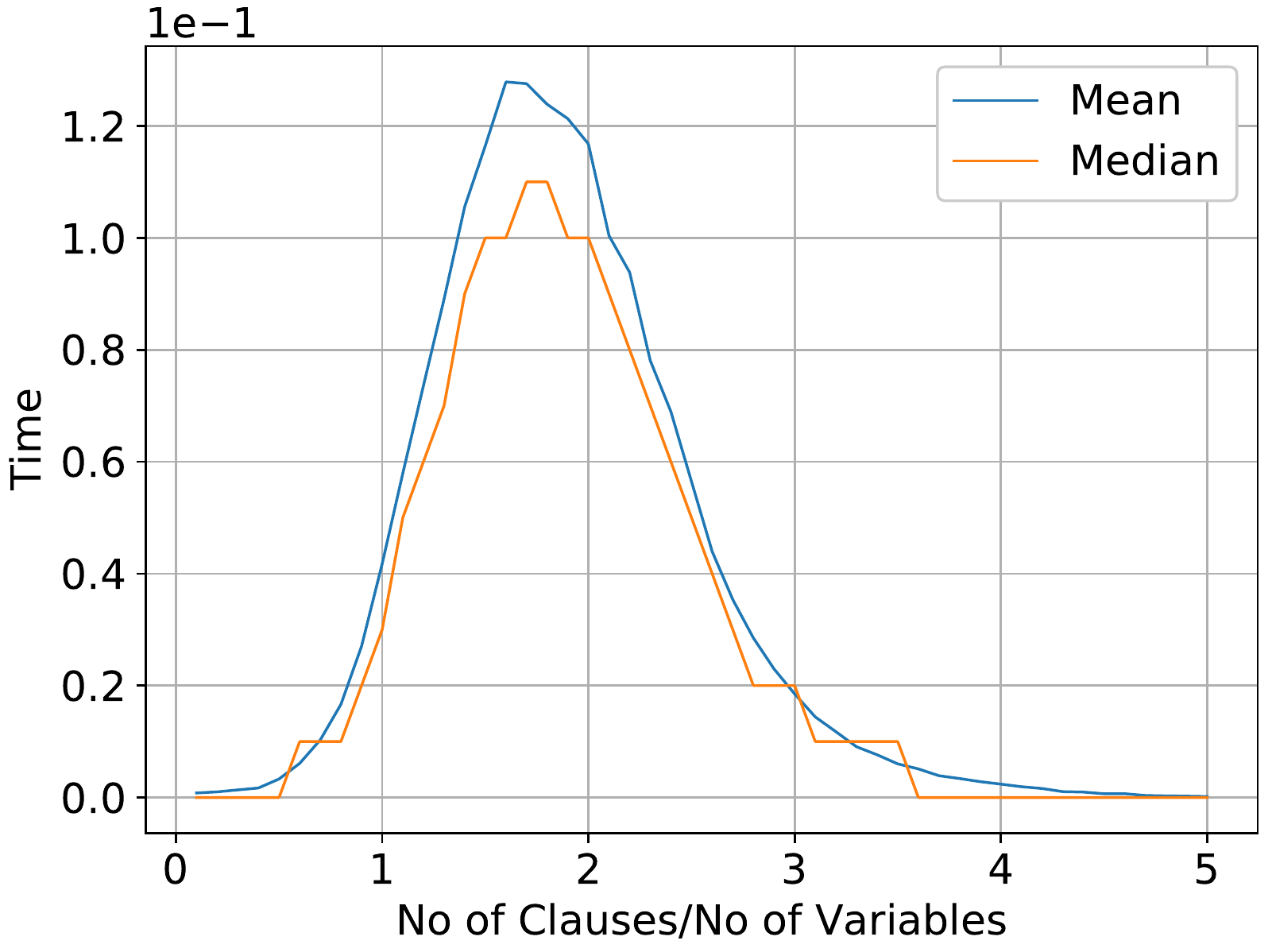}
		\caption{40 vars}
		\label{fig:appendix:d4_tVc_3CNF_40vars}
	\end{subfigure}
	\hfill
	\begin{subfigure}[b]{0.30\textwidth}
		\centering
		\includegraphics[width=\linewidth]{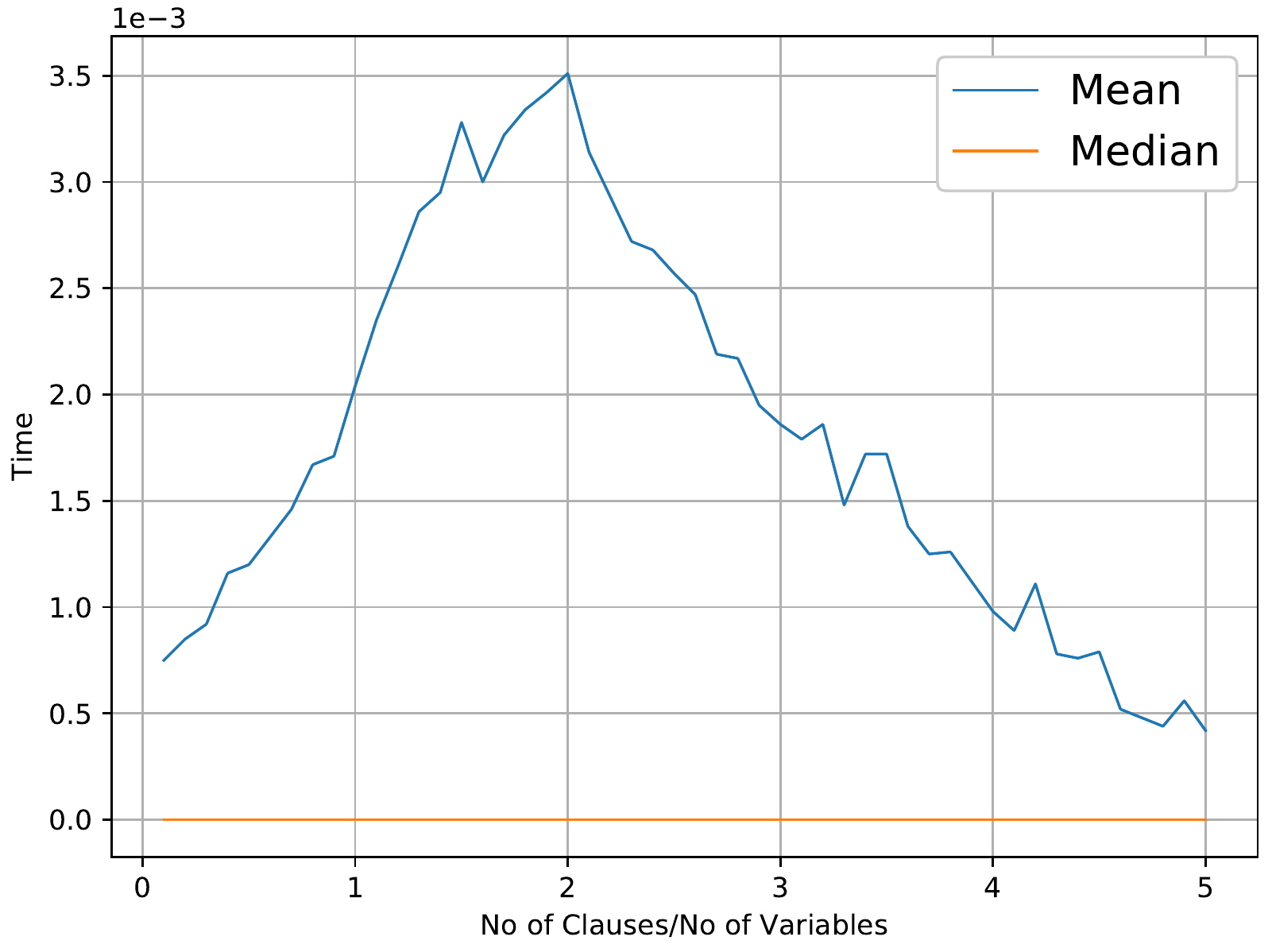}
		\caption{20 vars}
		\label{fig:appendix:d4_tVc_3CNF_20vars}
	\end{subfigure}
	\caption{Compile-time for d-DNNF vs clause density with different number of variables}
	\label{fig:appendix:d4_tVc_3CNF}
\end{figure*}

\begin{figure*}[!h]
	\centering
	\begin{subfigure}[b]{0.30\textwidth}
		\centering
		\includegraphics[width=\linewidth]{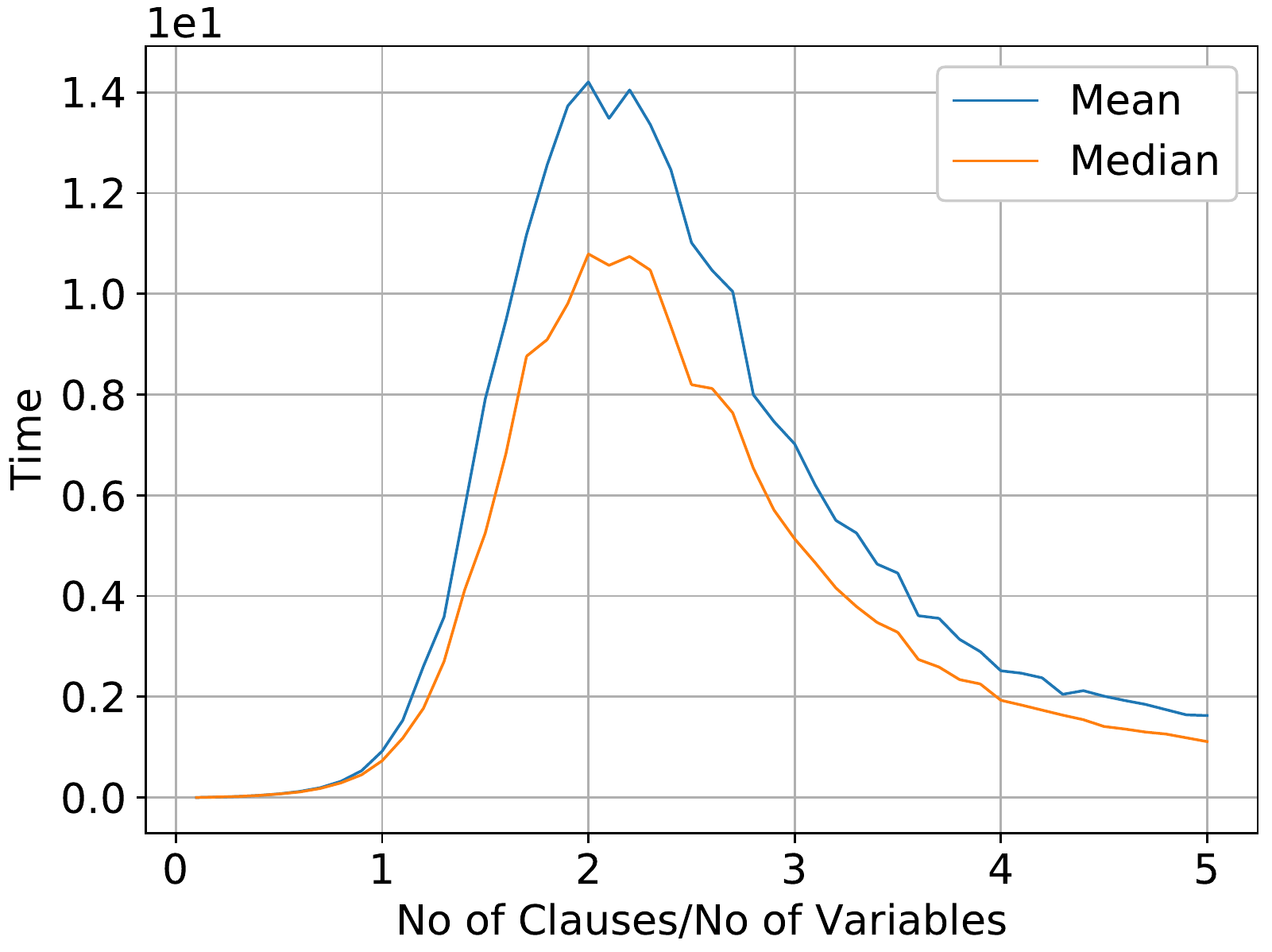}
		\caption{40 vars}
		\label{fig:appendix:sdd_tVc_3CNF_40vars}
	\end{subfigure}
	\hfill
	\begin{subfigure}[b]{0.30\textwidth}
		\centering
		\includegraphics[width=\linewidth]{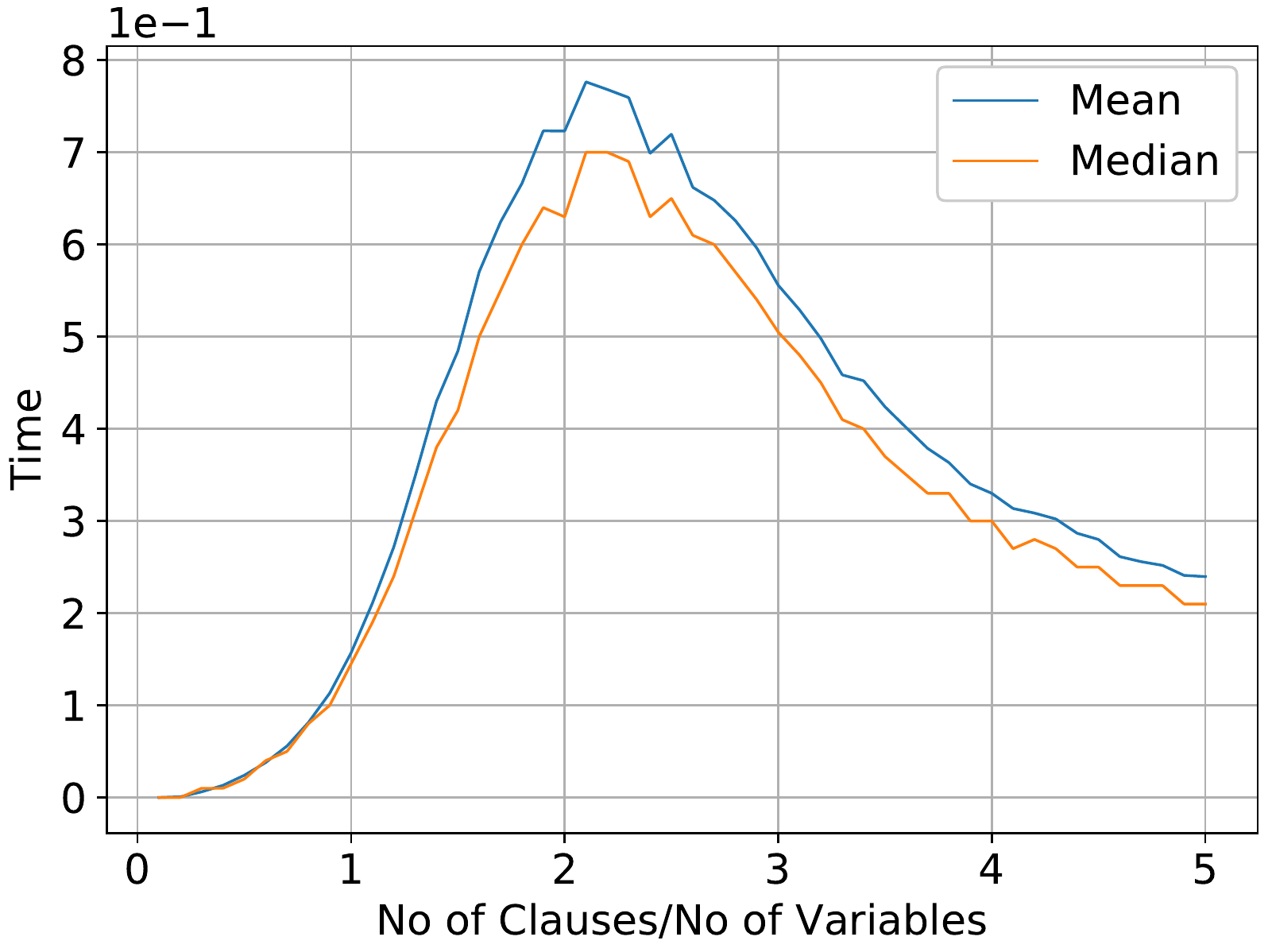}
		\caption{30 vars}
		\label{fig:appendix:sdd_tVc_3CNF_30vars}
	\end{subfigure}
	\hfill
	\begin{subfigure}[b]{0.30\textwidth}
		\centering
		\includegraphics[width=\linewidth]{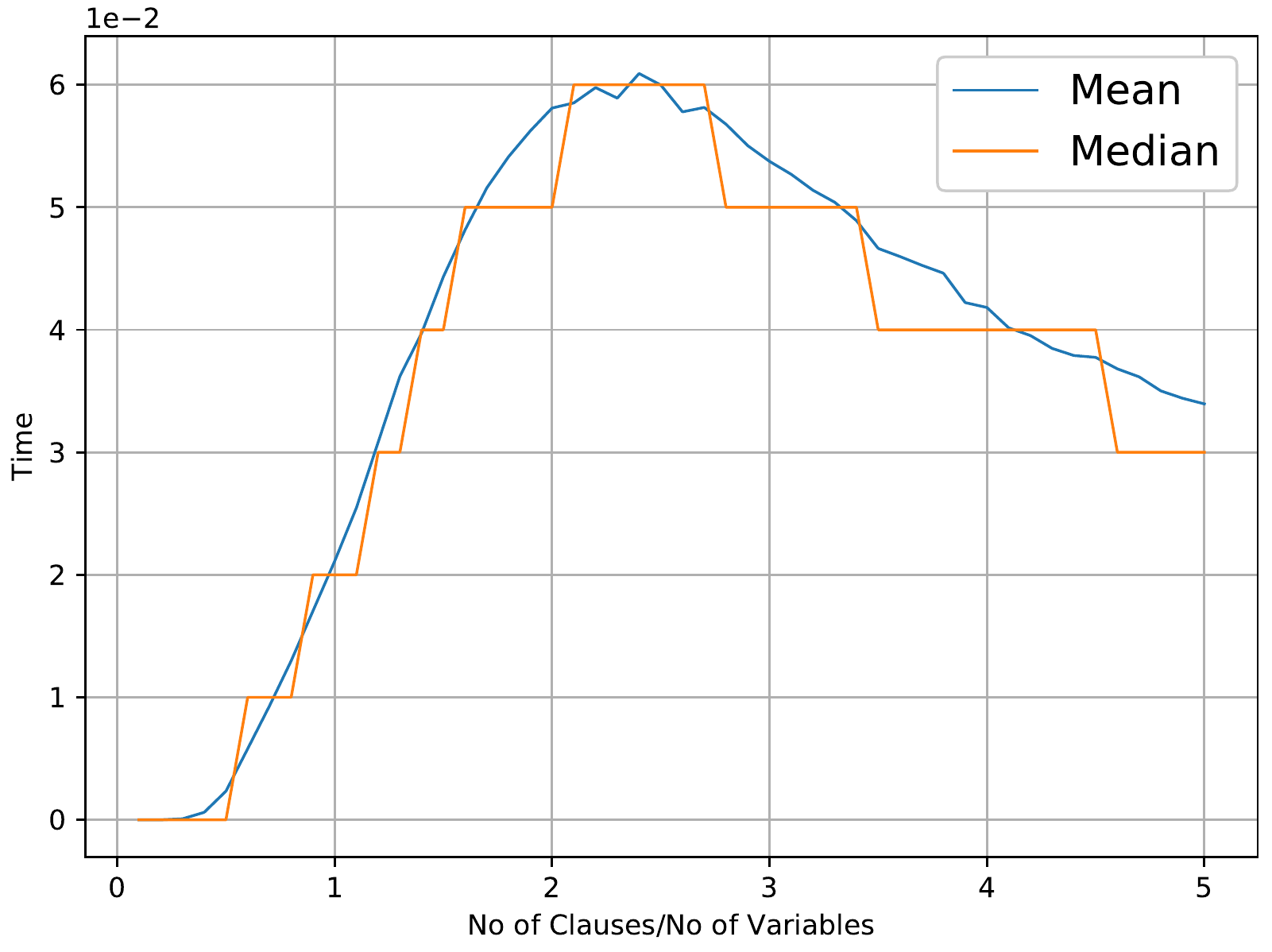}
		\caption{20 vars}
		\label{fig:appendix:sdd_tVc_3CNF_20vars}
	\end{subfigure}
	\caption{Compile-time for SDD vs clause density with different number of variables}
	\label{fig:appendix:sdd_tVc_3CNF}
\end{figure*}

\begin{figure*}[!h]
	\centering
	\begin{subfigure}[b]{0.30\textwidth}
		\centering
		\includegraphics[width=\linewidth]{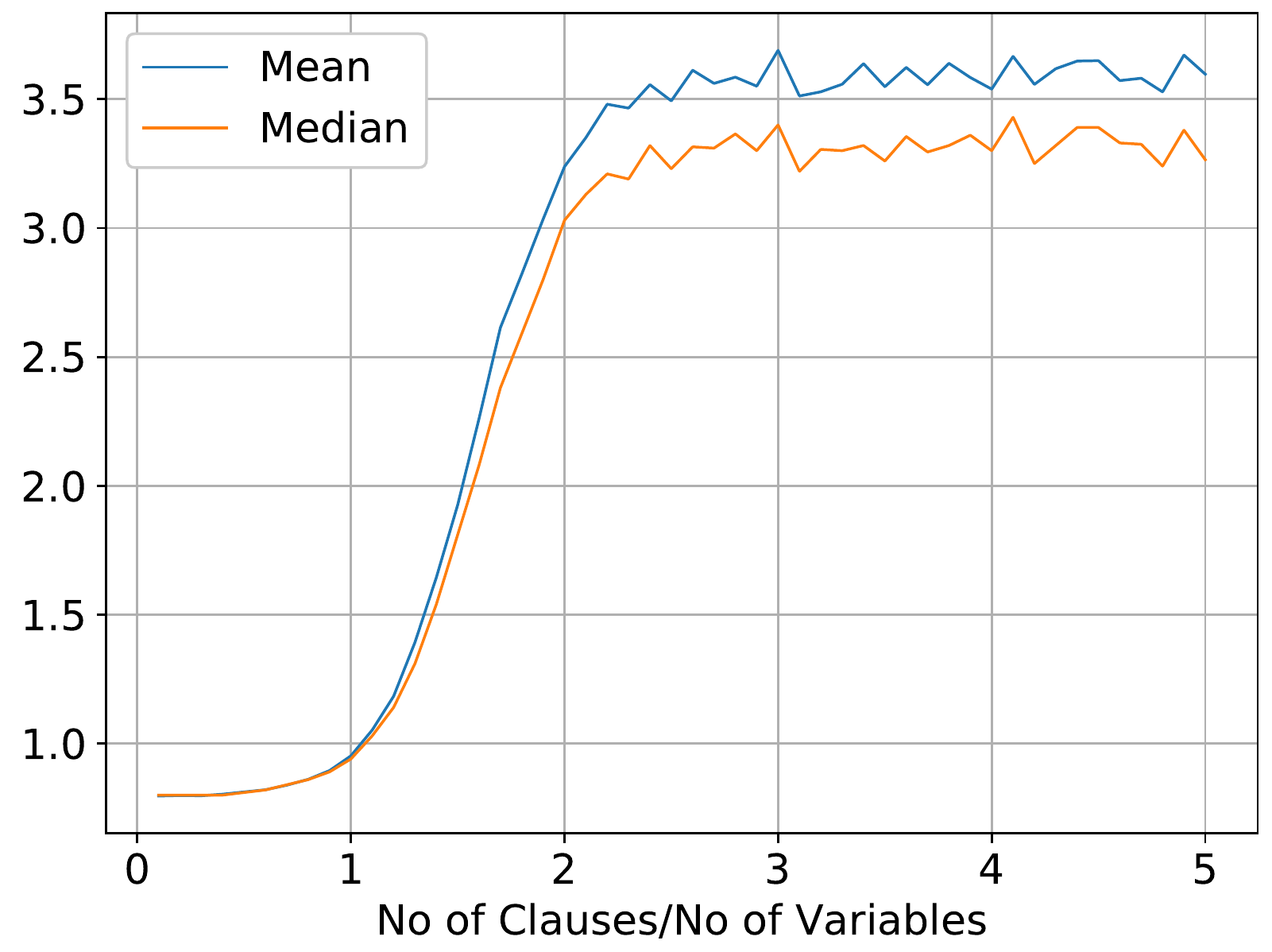}
		\caption{40 vars}
		\label{fig:appendix:CUDD_tVc_3CNF_40vars}
	\end{subfigure}
	\hfill
	\begin{subfigure}[b]{0.30\textwidth}
		\centering
		\includegraphics[width=\linewidth]{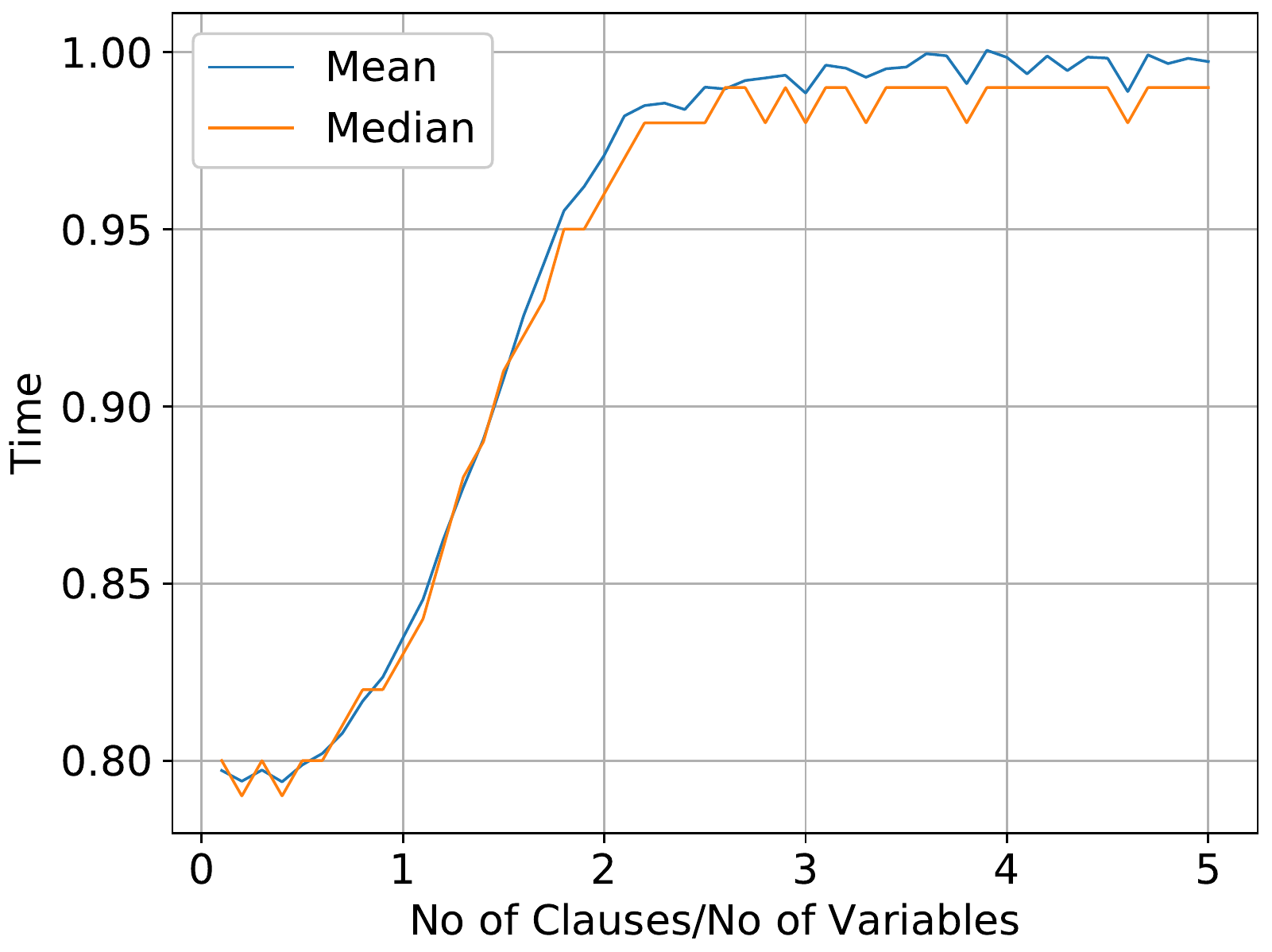}
		\caption{30 vars}
		\label{fig:appendix:CUDD_tVc_3CNF_30vars}
	\end{subfigure}
	\hfill
	\begin{subfigure}[b]{0.30\textwidth}
		\centering
		\includegraphics[width=\linewidth]{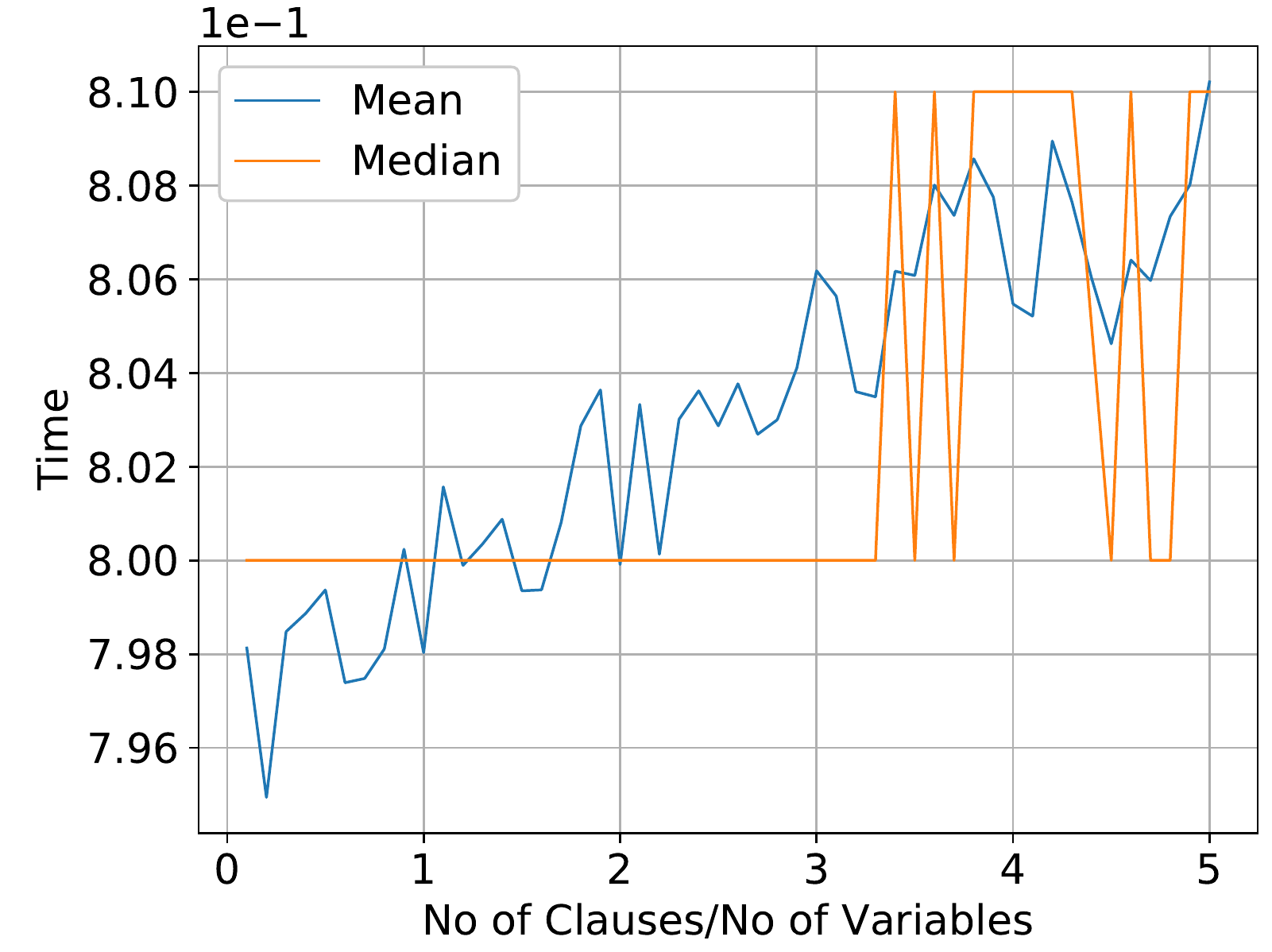}
		\caption{20 vars}
		\label{fig:appendix:CUDD_tVc_3CNF_20vars}
	\end{subfigure}
	\caption{Compile-time for OBDD vs clause density with different number of variables}
	\label{fig:appendix:CUDD_tVc_3CNF}
\end{figure*}

Figures~\ref{fig:appendix:d4_tVc_3CNF}, \ref{fig:appendix:sdd_tVc_3CNF}, \ref{fig:appendix:CUDD_tVc_3CNF} shows the variation in average runtimes against clause density for d-DNNF, SDD and OBDD for different number of variables. 

\begin{figure*}[!th]
	\centering
	\begin{subfigure}[b]{0.30\linewidth}
		\centering
		\includegraphics[width=\linewidth]{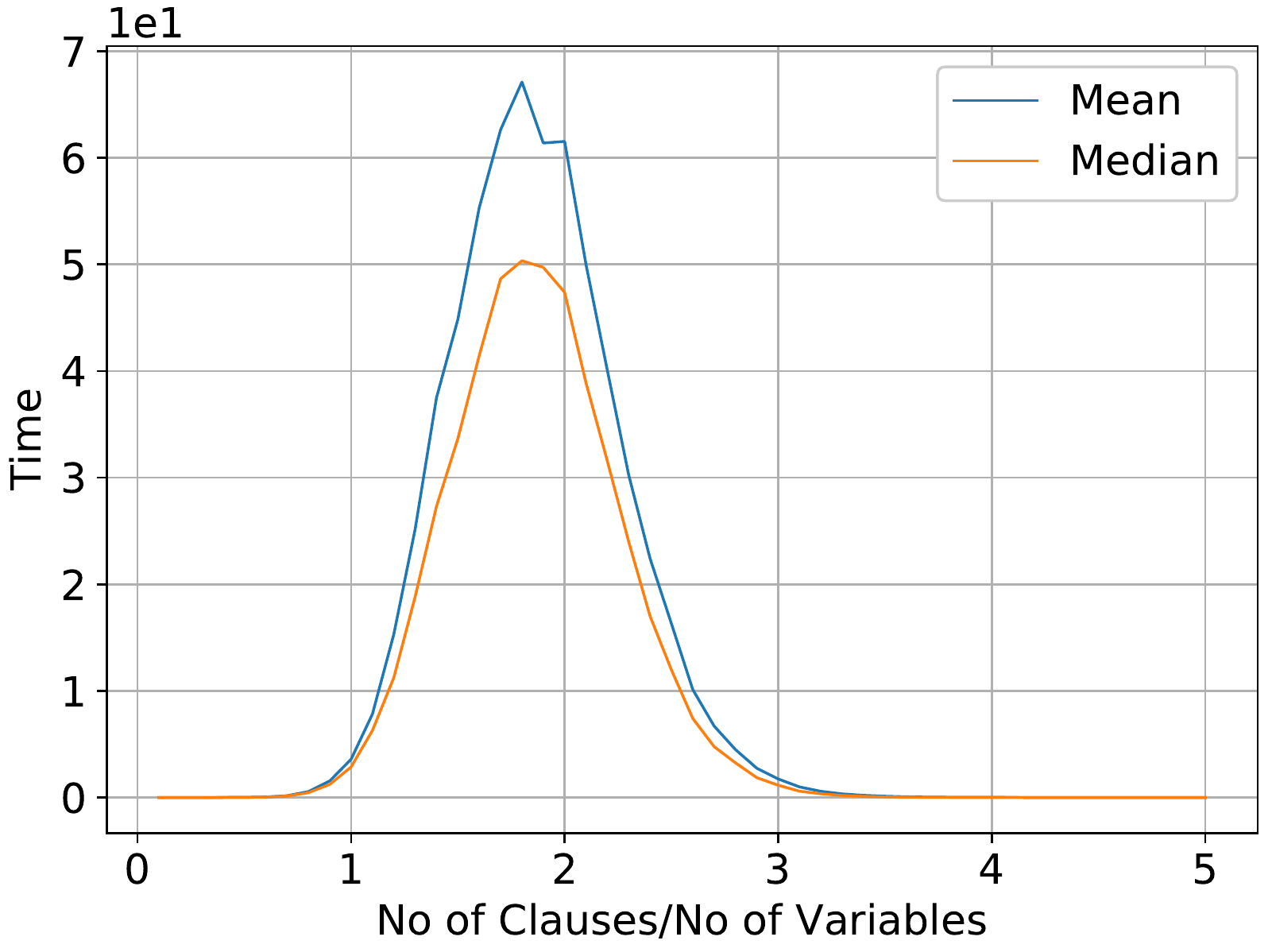}
		\caption{d-DNNF, 70 vars}
		\label{fig:appendix:d4_tVc_3CNF_70vars}
	\end{subfigure}
	\hfill
	\begin{subfigure}[b]{0.30\linewidth}
		\centering
		\includegraphics[width=\linewidth]{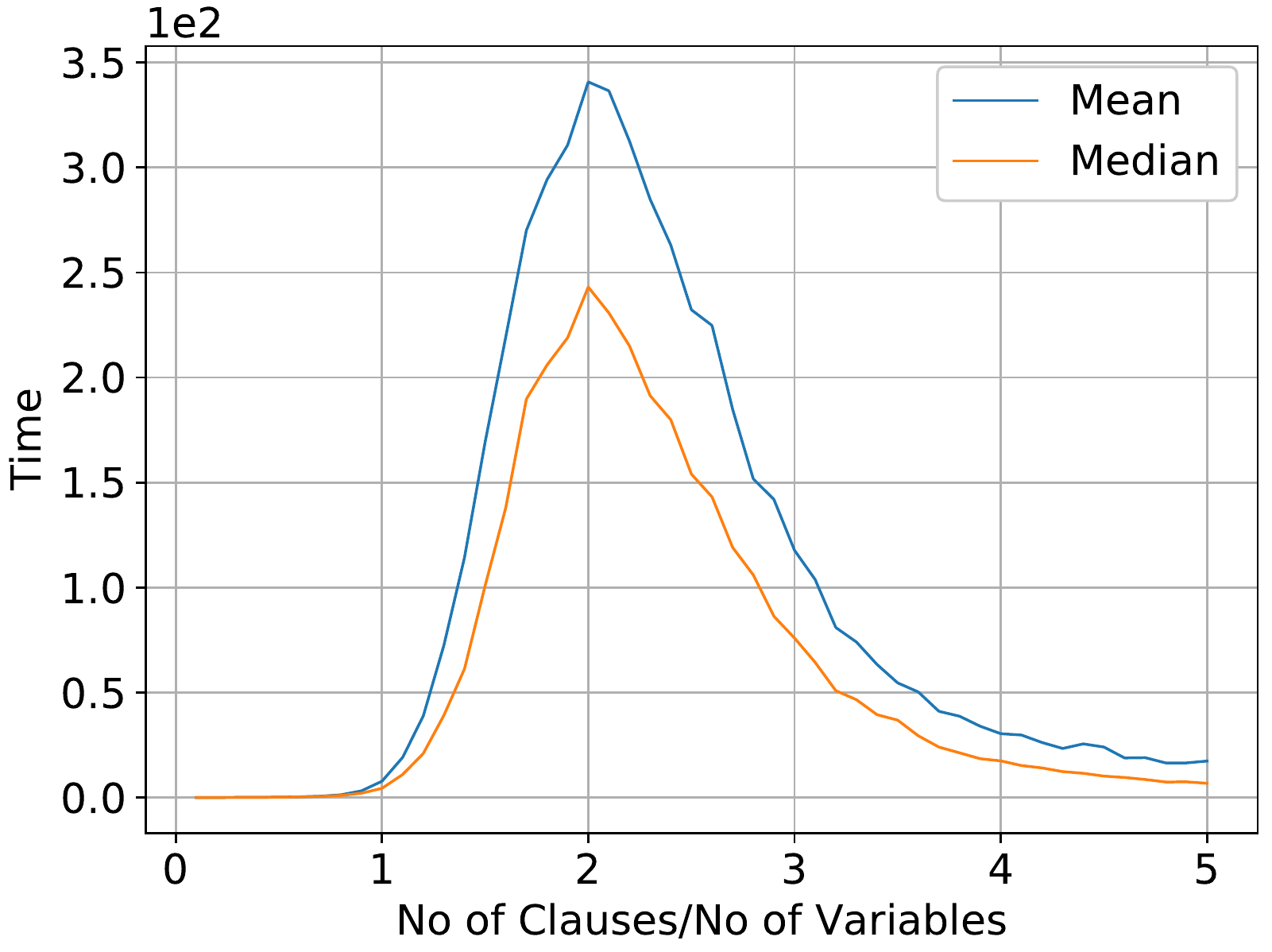}
		\caption{SDD, 50 vars}
		\label{fig:appendix:sdd_tVc_50vars_3cnf}
	\end{subfigure}
	\hfill
	\begin{subfigure}[b]{0.30\linewidth}
		\centering
		\includegraphics[width=\linewidth]{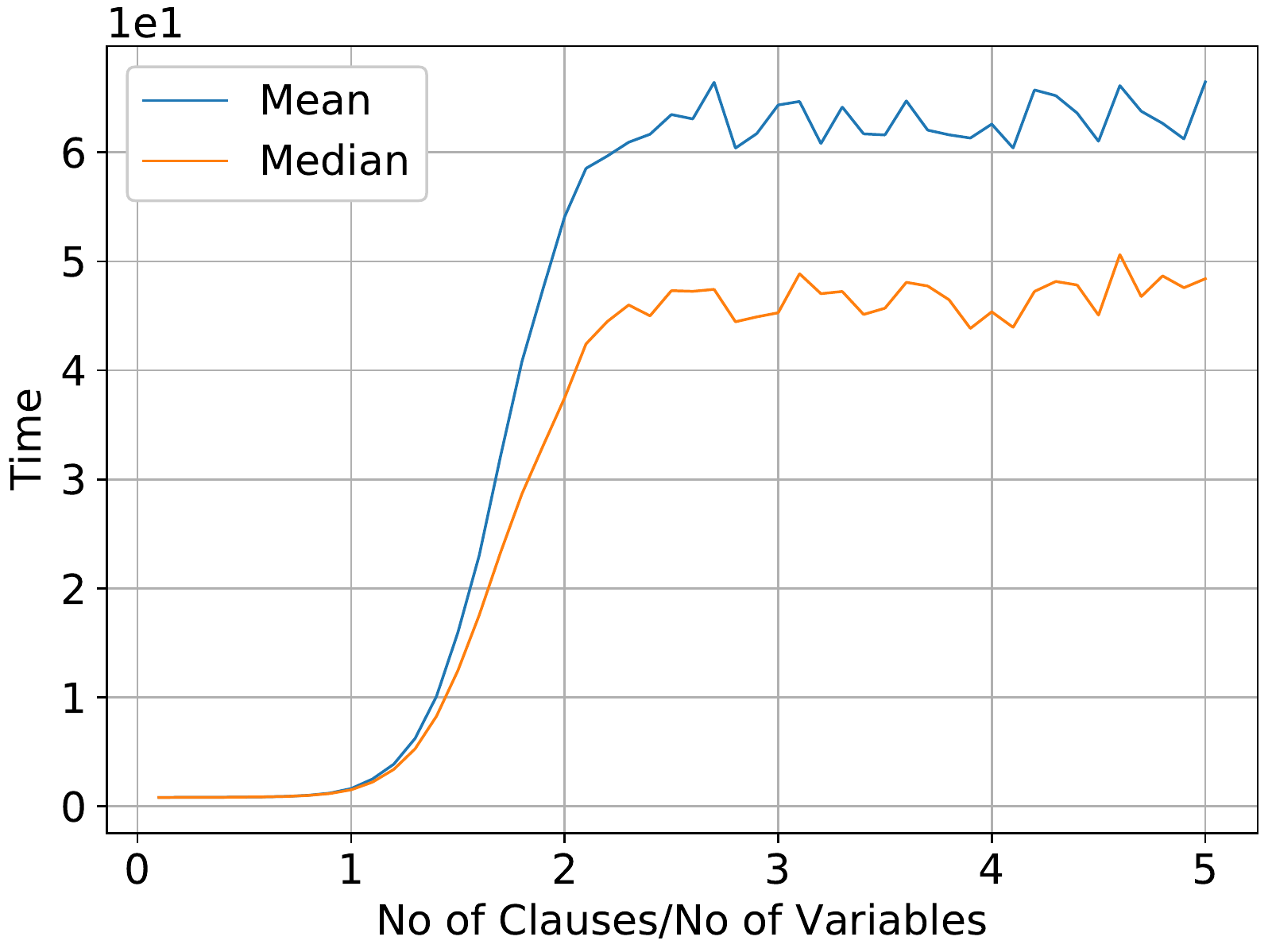}
		\caption{OBDD, 50 vars}
		\label{fig:appendix:CUDD_tVc_3CNF_50vars}
	\end{subfigure}
	\caption{Average runtime for 3-CNF}
	\label{fig:appendix:tVc_3CNF}
\end{figure*}

\clearpage
\subsection{Diving deep into nature of complexity}

\begin{figure*}[!th]
	\centering
	\begin{subfigure}[b]{0.495\linewidth}
		\centering
		\includegraphics[width=\linewidth]{MainFigures/dDNNF/d4_log_mean_nodesVvars_3cnf0d4-0d6clD1exp}
		\caption{clause densities 0.6 and 0.4}
		\label{fig:appendix:d4_log_nVv_0.4-0.6cl_3cnf}
	\end{subfigure}
	\hfill
	\begin{subfigure}[b]{0.495\linewidth}
		\centering
		\includegraphics[width=\linewidth]{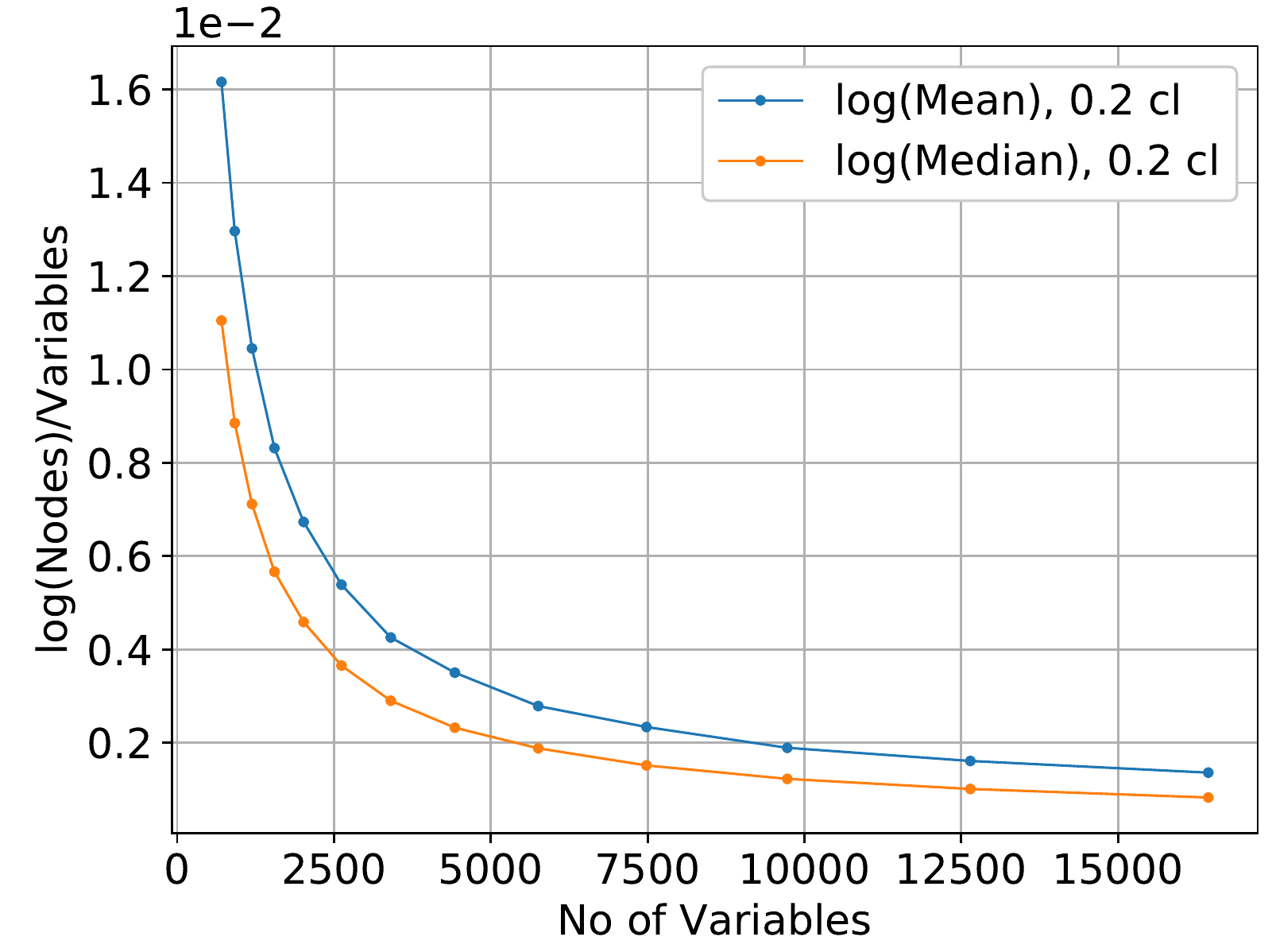}
		\caption{clause density 0.2}
		\label{fig:appendix:d4_log_nVv_0.2cl_3cnf}
	\end{subfigure}
    \caption{log(nodes in \dDNNF)/variables vs variables for small $r$}
	\label{fig:appendix:d4_log_nVv_3CNF}
\end{figure*}

\begin{figure}
	\centering
	\centering
	\includegraphics[width=0.6\linewidth]{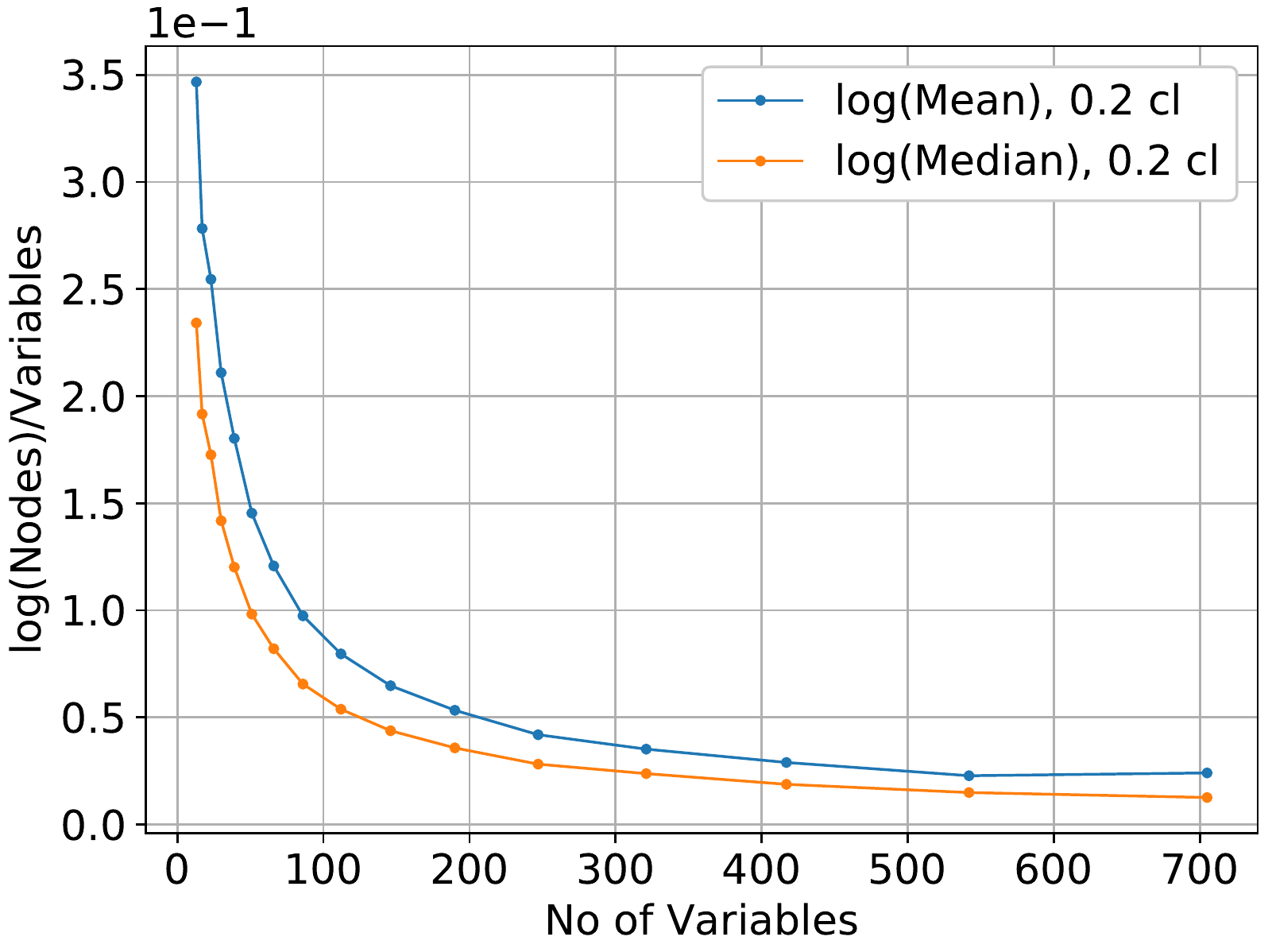}
	\caption{log(nodes in SDD)/variables vs variables for $r=0.2$}
	\label{fig:appendix:sdd_log_nVv_3CNF}
\end{figure}

\begin{figure}
	\centering
		\centering
		\includegraphics[width=0.6\linewidth]{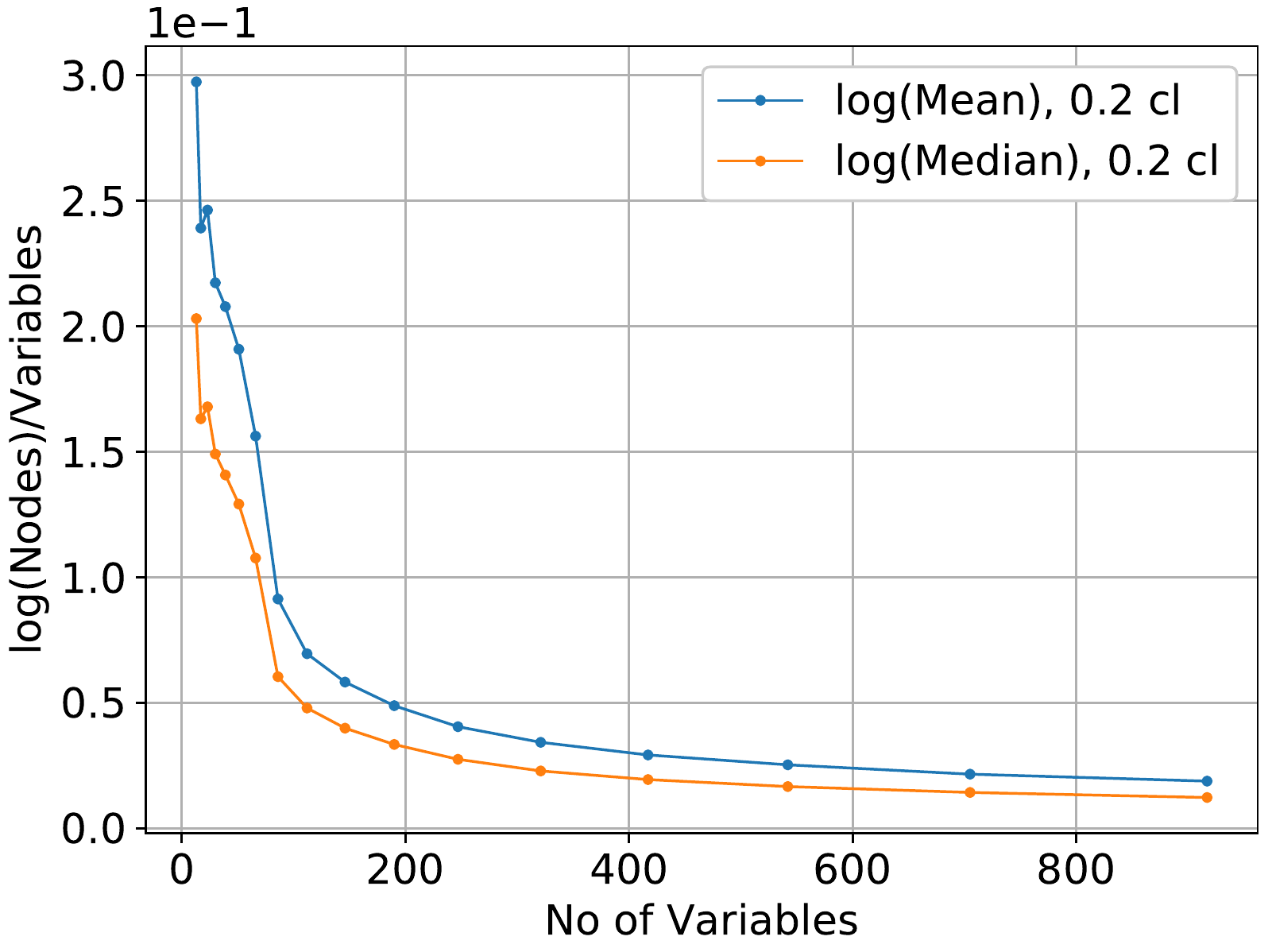}
    \caption{log(nodes in OBDD)/variables vs variables for small $r = 0.2$ }
	\label{fig:appendix:CUDD_log_nVv_3CNF}
\end{figure}

We plot $\log(\#Nodes)/variables$ for different compilations against varying number of variables for small clause densities in Figures \ref{fig:appendix:d4_log_nVv_3CNF}, \ref{fig:appendix:sdd_log_nVv_3CNF} and \ref{fig:appendix:CUDD_log_nVv_3CNF}.

\begin{figure}
	\centering
		\includegraphics[width=0.6\linewidth]{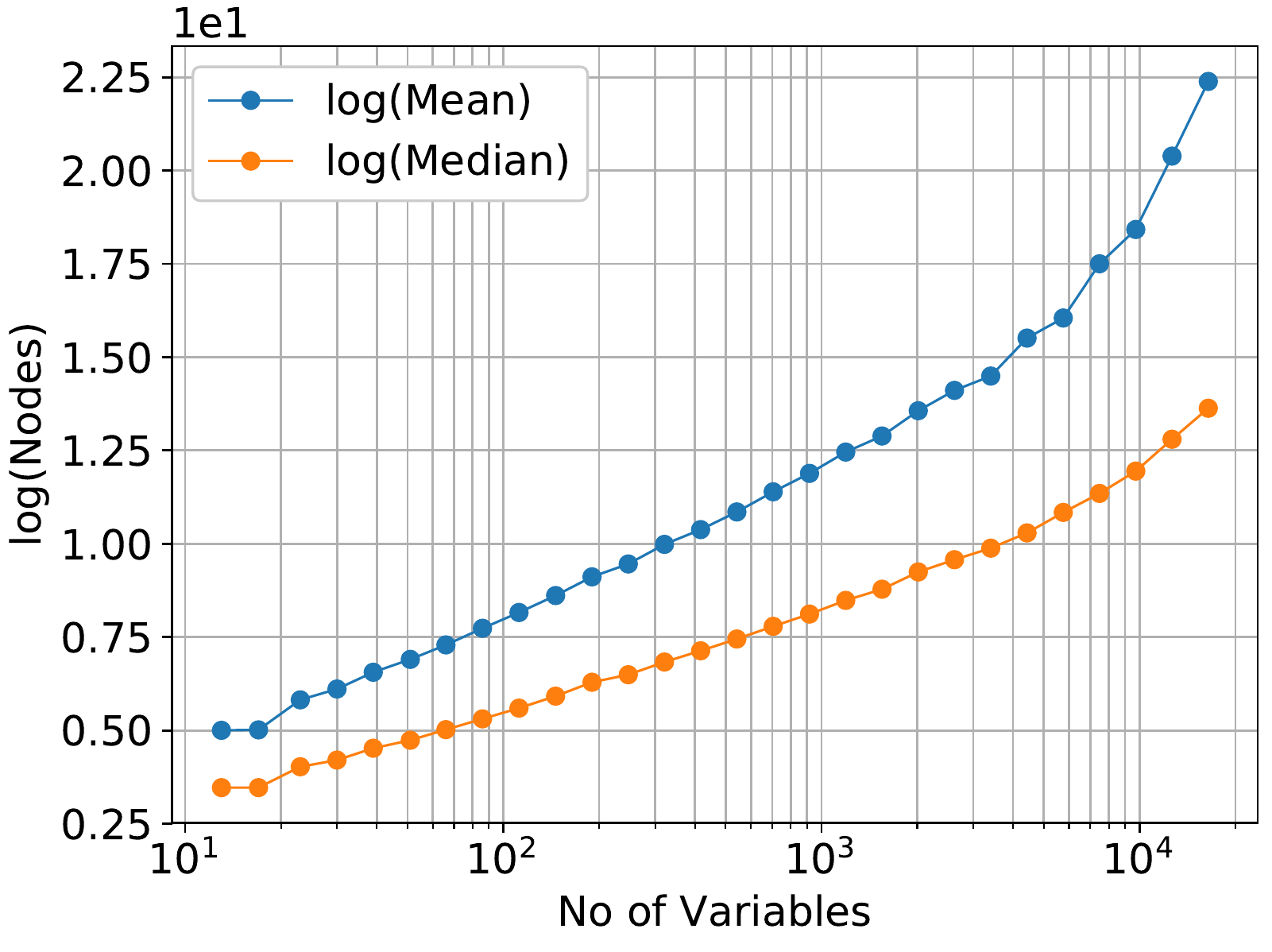}
        \caption{log-log graph for  nodes in d-DNNF compilation with clause density 0.2}
		\label{fig:appendix:d4_loglog_nVv_0.2cl_3cnf}
\end{figure}

\begin{figure}
	\centering
		\includegraphics[width=0.6\linewidth]{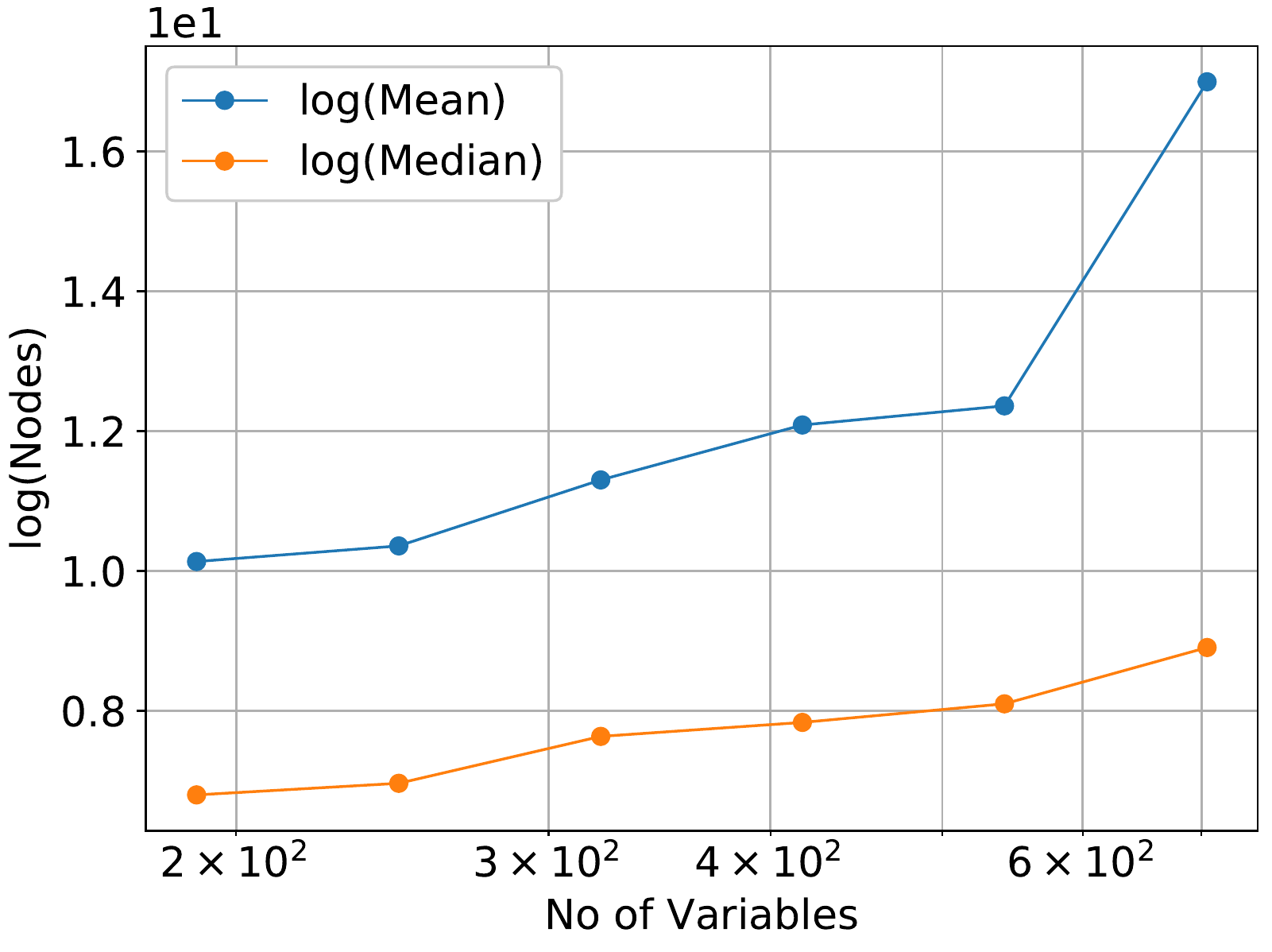}
		\caption{log-log graph for  nodes in SDD compilation with clause density 0.2}
		\label{fig:appendix:sdd_loglog_nVv_0.2cl_3cnf}
\end{figure}

\begin{figure}
	\centering
		\includegraphics[width=0.6\linewidth]{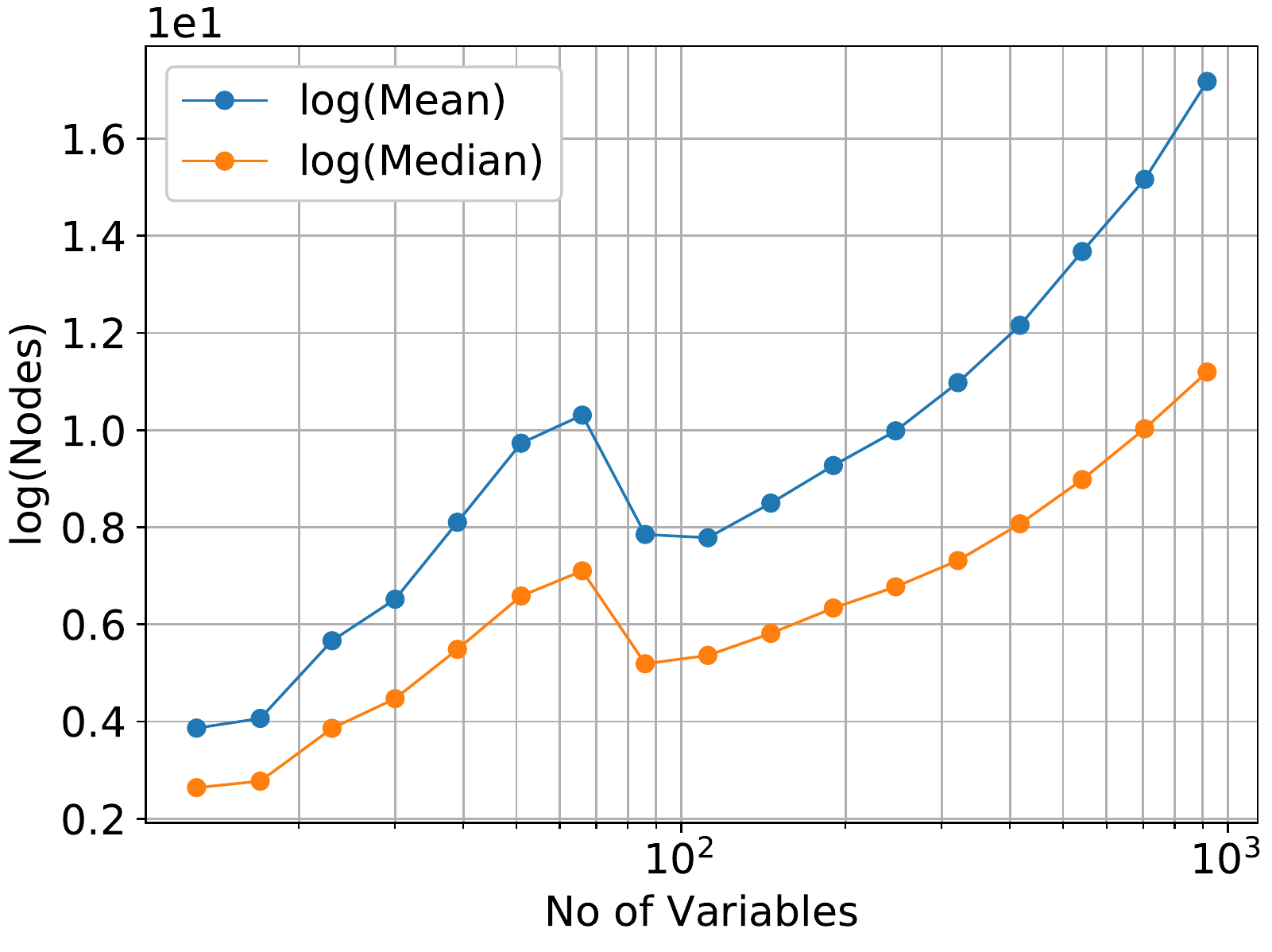}
		\caption{log-log graph for  nodes in OBDD compilation with clause density 0.2}
		\label{fig:appendix:CUDD_loglog_nVv_0.2cl_3cnf}
\end{figure}

We compare $\log(nodes\,in\,compilations)$ against $\log(number\,of\,variables)$ for different compilations in Figures \ref{fig:appendix:d4_loglog_nVv_0.2cl_3cnf}, \ref{fig:appendix:sdd_loglog_nVv_0.2cl_3cnf}, \ref{fig:appendix:CUDD_loglog_nVv_0.2cl_3cnf}.

\begin{figure*}[!th]
	\centering
	\begin{subfigure}[b]{0.495\linewidth}
		\centering
		\includegraphics[width=\linewidth]{MainFigures/dDNNF/d4_log_mean_timesVvars_3cnf0d4-0d6clD1exp}
		\caption{clause densities 0.6 and 0.4}
		\label{fig:appendix:d4_log_tVv_0.6cl_3cnf}
	\end{subfigure}
	\hfill
	\begin{subfigure}[b]{0.495\linewidth}
		\centering
		\includegraphics[width=\linewidth]{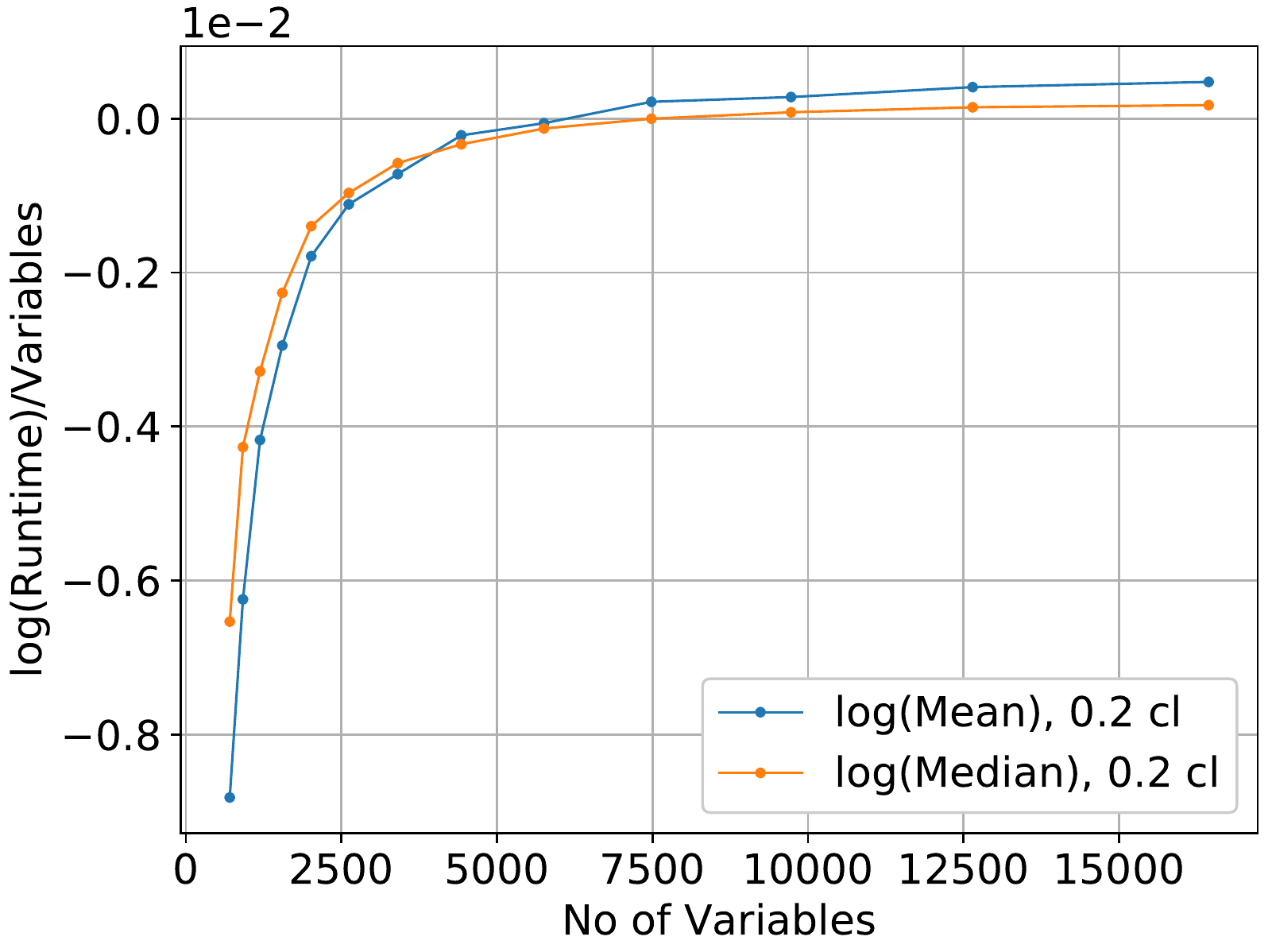}
		\caption{clause density 0.2}
		\label{fig:appendix:d4_log_tVv_0.2cl_3cnf}
	\end{subfigure}
    \caption{log(runtime for \dDNNF)/variables vs variables for small $r$}
	\label{fig:appendix:d4_log_tVv_3CNF}
\end{figure*}

\begin{figure}
	\centering
	\includegraphics[width=0.6\linewidth]{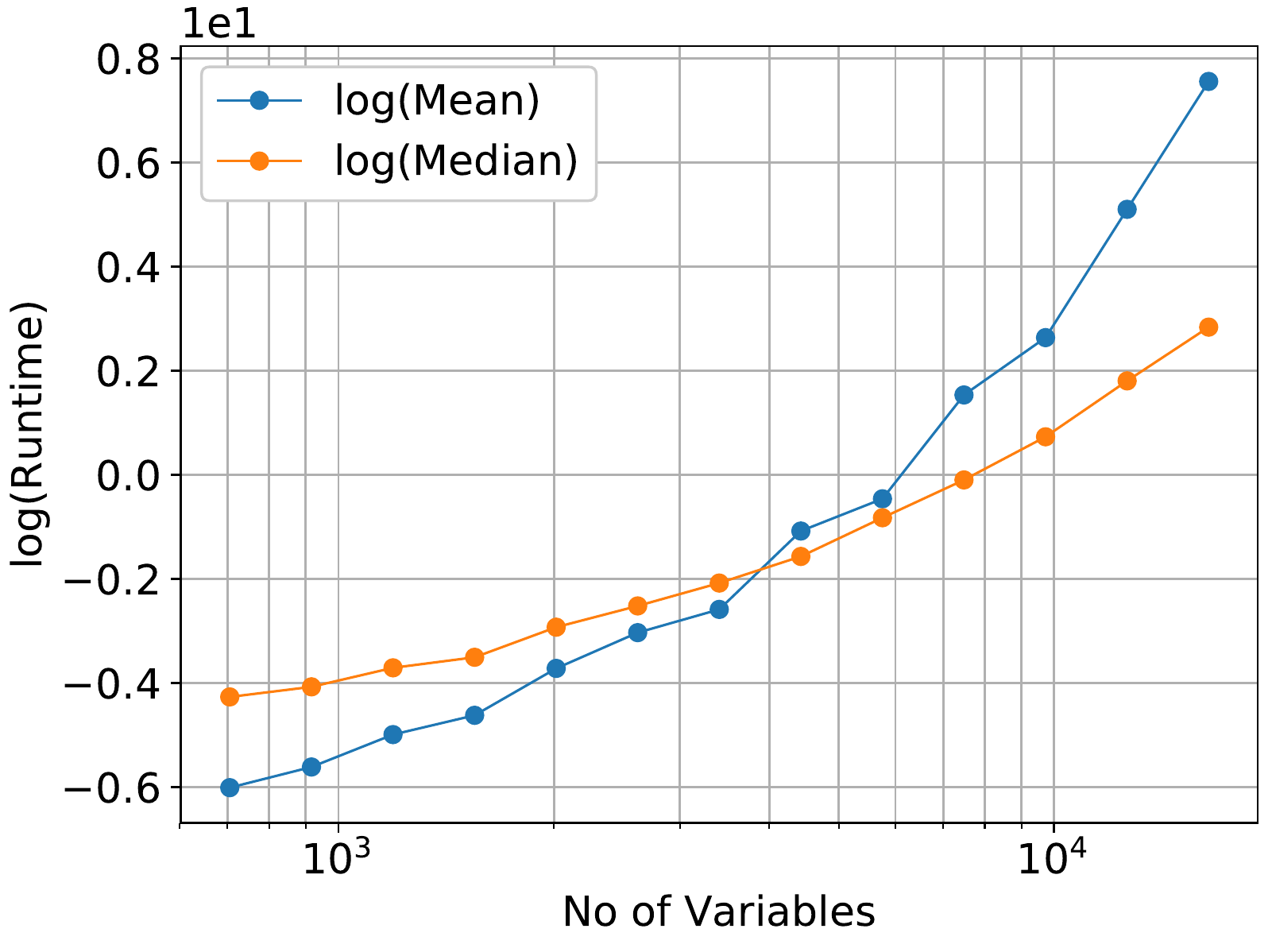}
	\caption{log-log graph for runtime of d-DNNF compilation with clause density~0.2}
	\label{fig:appendix:d4_loglog_tVv_0.2cl_3cnf}
\end{figure}

\begin{figure}
	\centering
		\centering
		\includegraphics[width=0.6\linewidth]{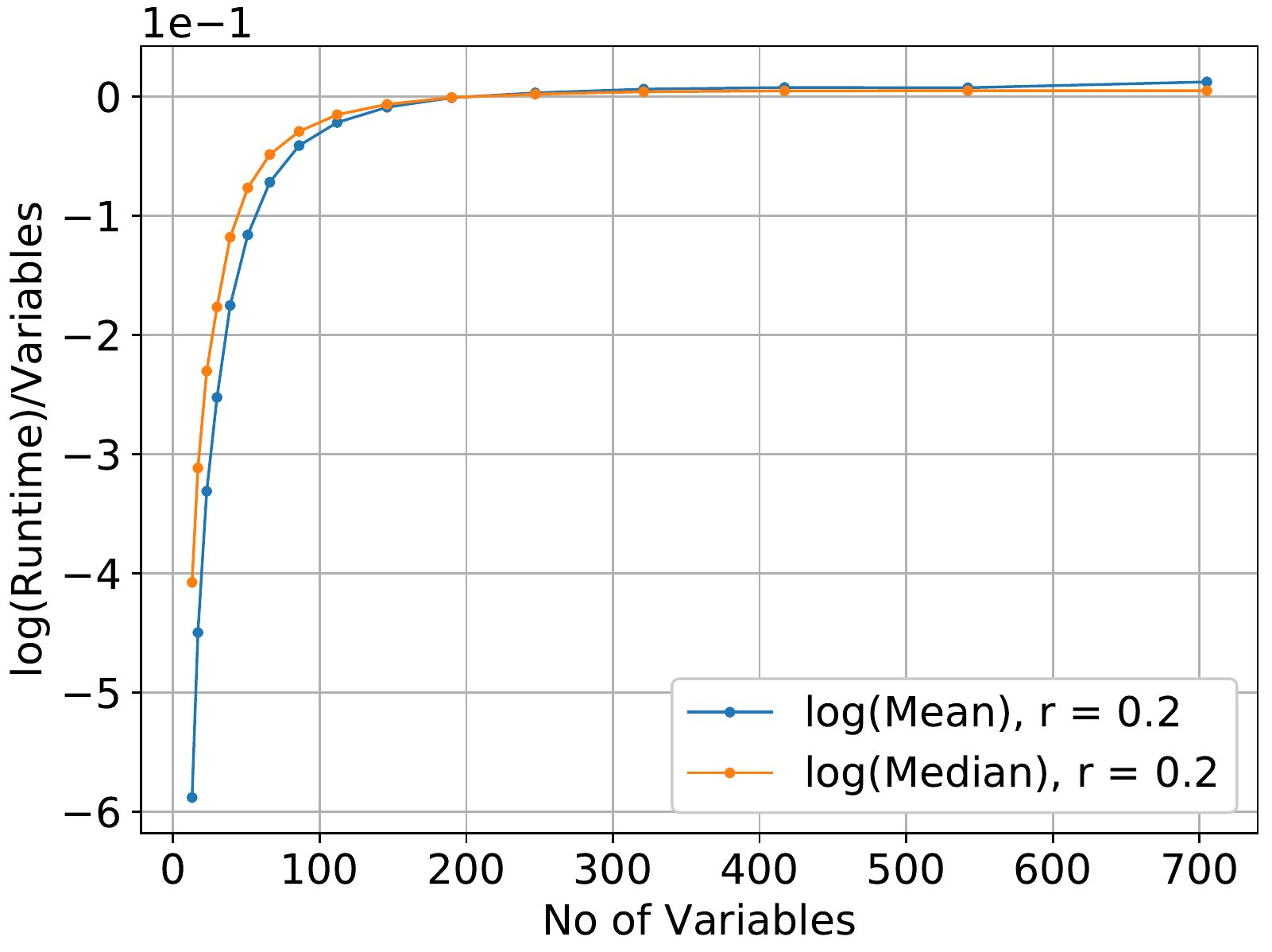}
	\caption{log(runtime for SDD)/variables vs variables for $r=0.2$}
	\label{fig:appendix:sdd_log_tVv_3CNF}
\end{figure}
\begin{figure}
	\centering
	\includegraphics[width=0.6\linewidth]{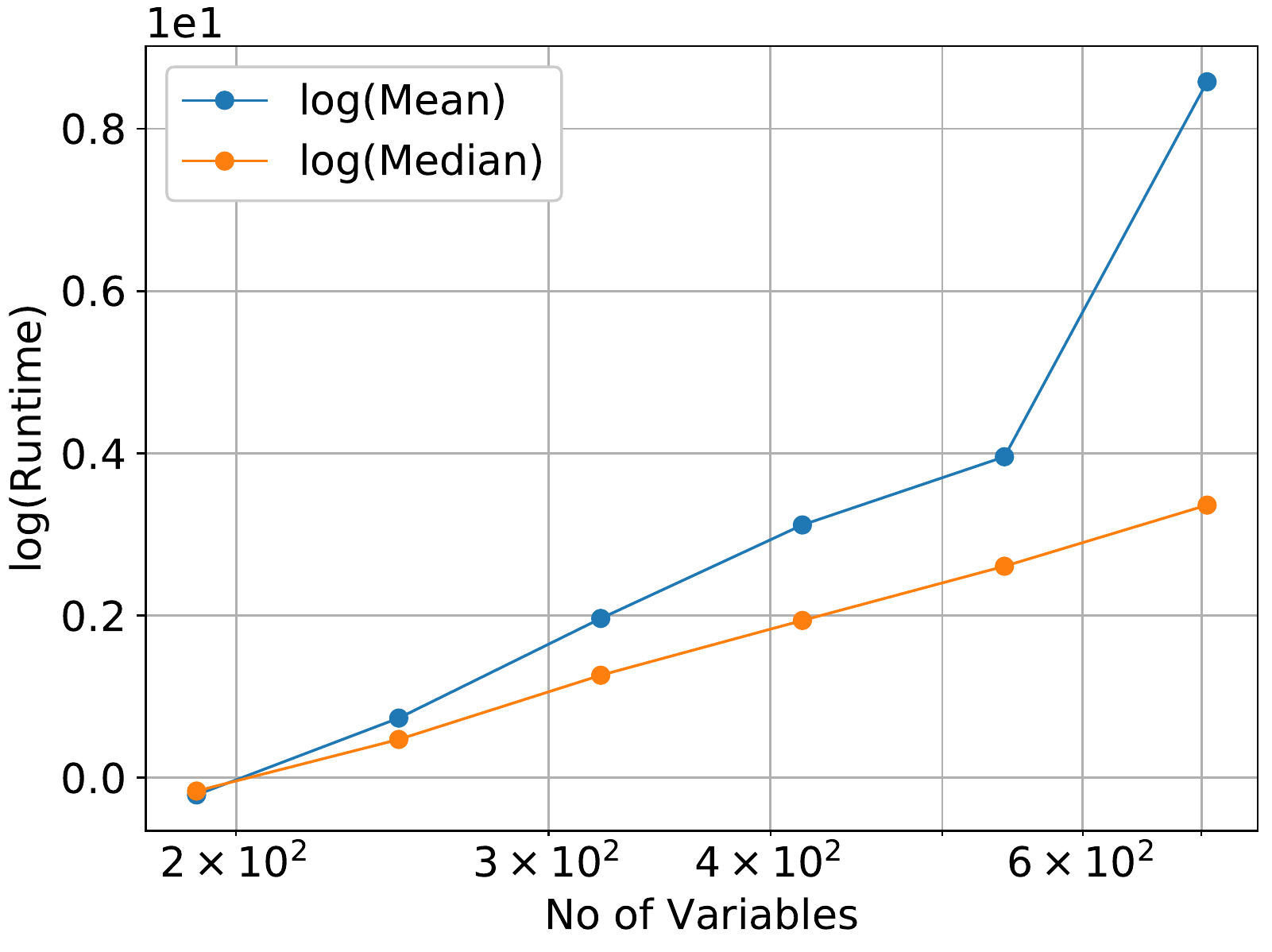}
	\caption{log-log graph for runtime of SDD compilation with clause density 0.2}
	\label{fig:appendix:sdd_loglog_tVv_0.2cl_3cnf}
\end{figure}


\begin{figure}
	\centering
		\centering
		\includegraphics[width=0.6\linewidth]{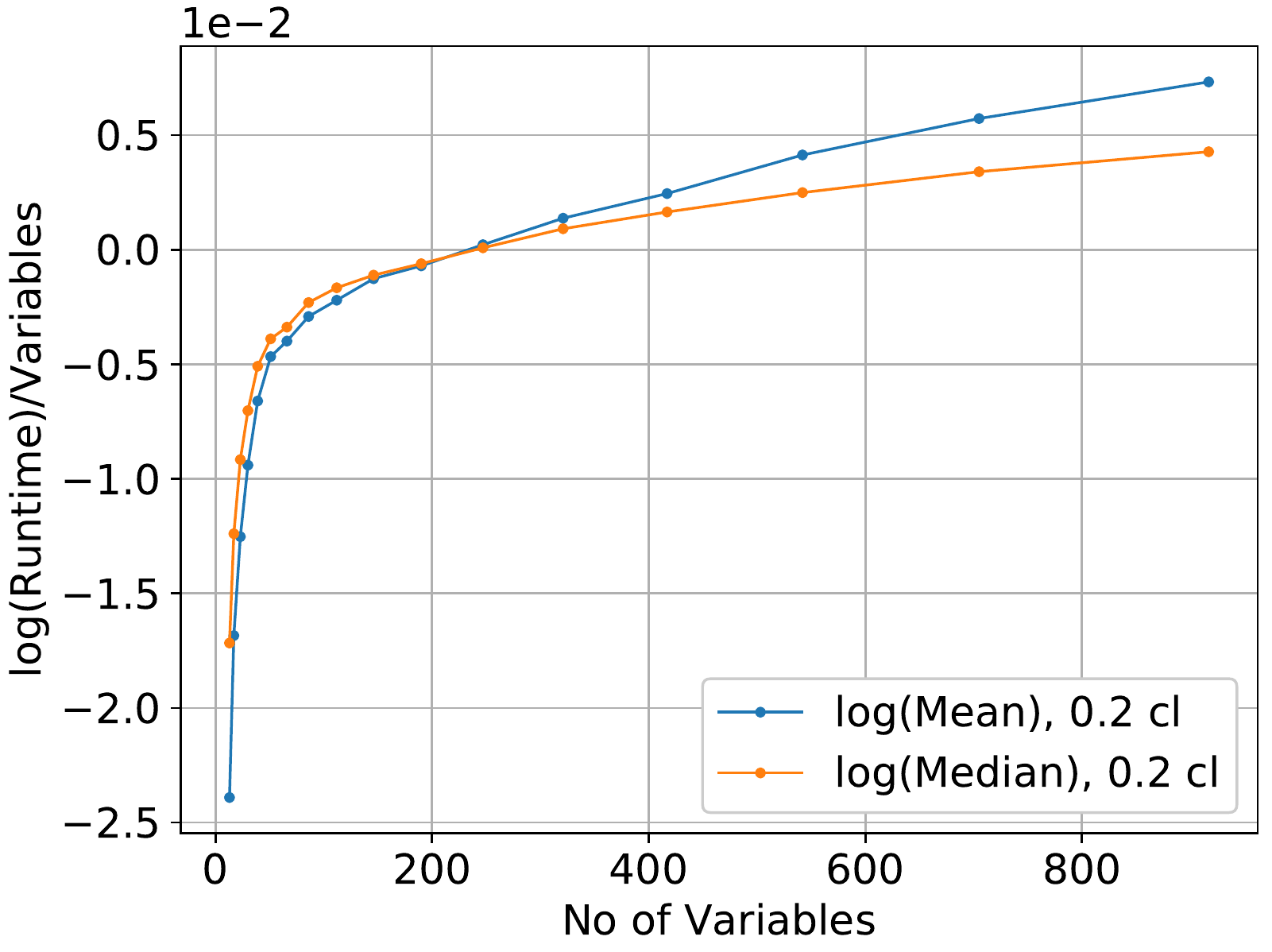}
    \caption{log(runtime for OBDD)/variables vs variables for small $r = 0.2$ }
	\label{fig:appendix:CUDD_log_tVv_3CNF}
\end{figure}

\begin{figure}
	\centering
	\includegraphics[width=0.6\linewidth]{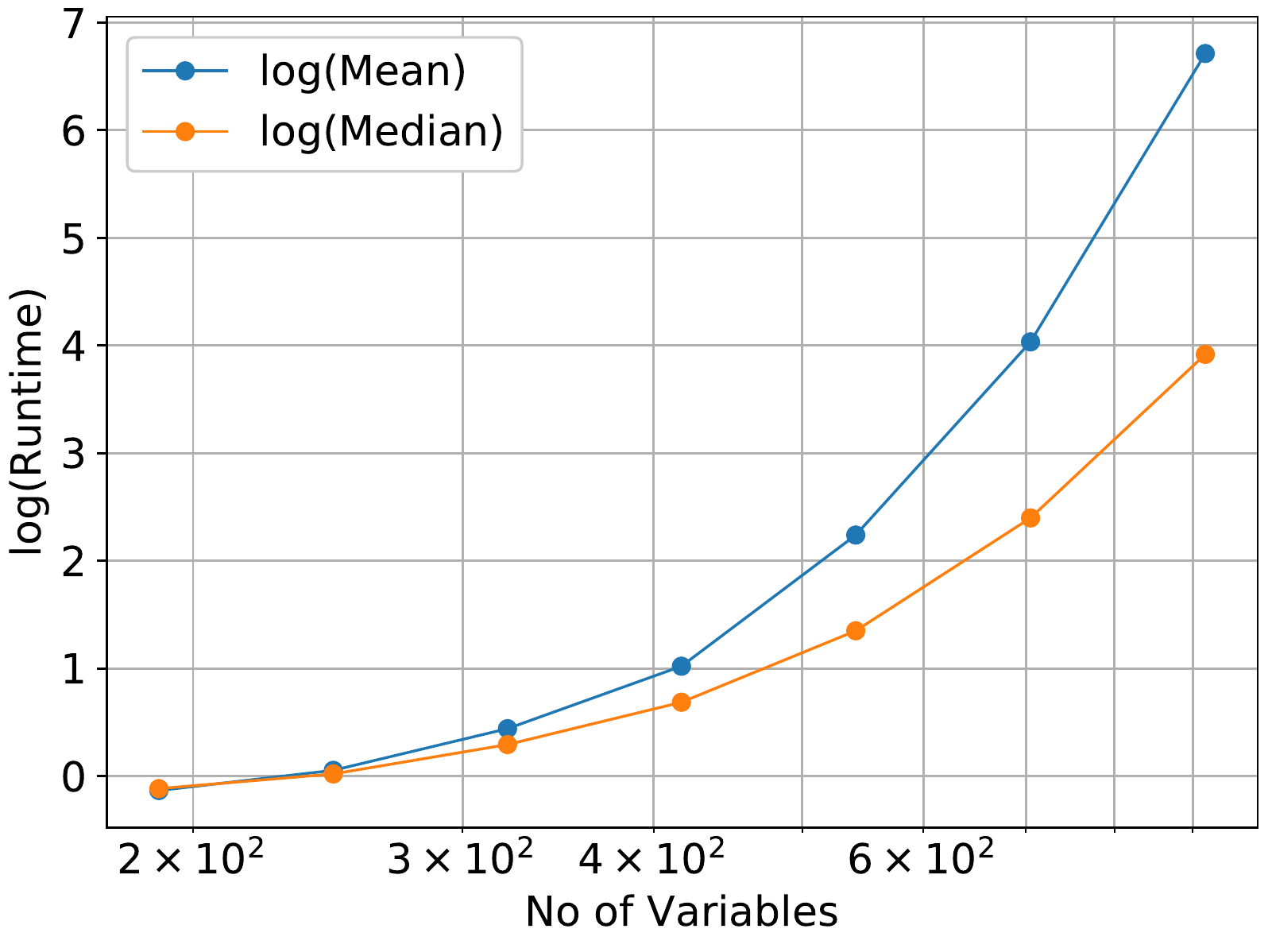}
	\caption{log-log graph for runtime of OBDD compilation with clause density 0.2}
	\label{fig:appendix:CUDD_loglog_tVv_0.2cl_3cnf}
\end{figure}

We plot $\log(runtime\,for\,compilations)/variables$ against varying number of variables for smaller clause densities in Figures \ref{fig:appendix:d4_log_tVv_3CNF}, \ref{fig:appendix:sdd_log_tVv_3CNF} and \ref{fig:appendix:CUDD_log_tVv_3CNF}. 

We compare $\log(runtime\,for\,compilations)$ against $\log(number\,of\,variables)$ for different compilations in Figures \ref{fig:appendix:d4_loglog_tVv_0.2cl_3cnf}, \ref{fig:appendix:sdd_loglog_tVv_0.2cl_3cnf}, \ref{fig:appendix:CUDD_loglog_tVv_0.2cl_3cnf}.

\newpage

\subsection{Impact of clause length on phase transition}

\begin{figure*}[!th]
	\centering
	\begin{subfigure}[b]{0.30\linewidth}
		\centering
		\includegraphics[width=\textwidth]{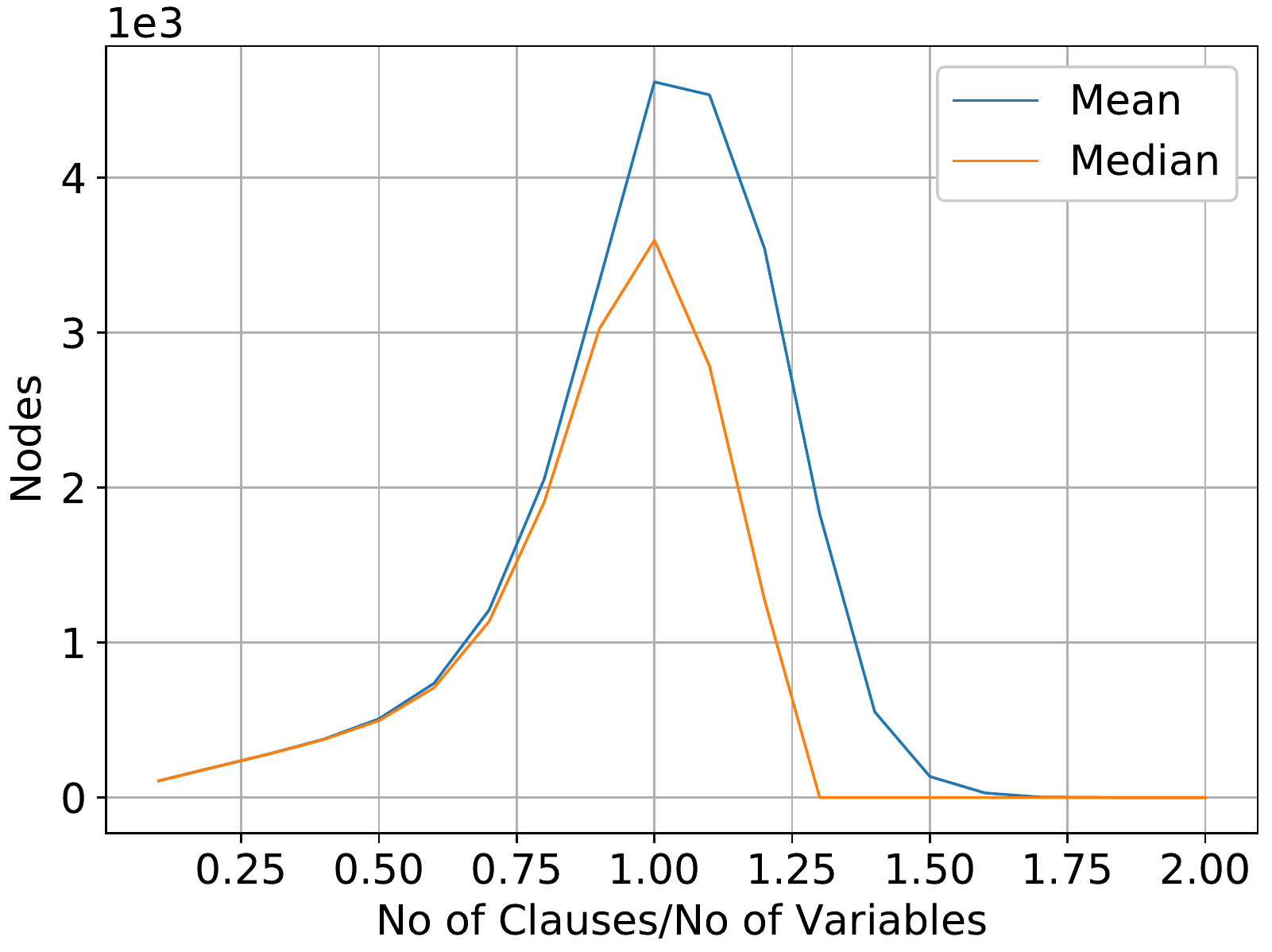}
		\caption{2-CNF and 200 vars}
		\label{fig:appendix:d4_nVc_2CNF_200vars}
	\end{subfigure}
	\hfill
	\begin{subfigure}[b]{0.30\linewidth}
		\centering
		\includegraphics[width=\textwidth]{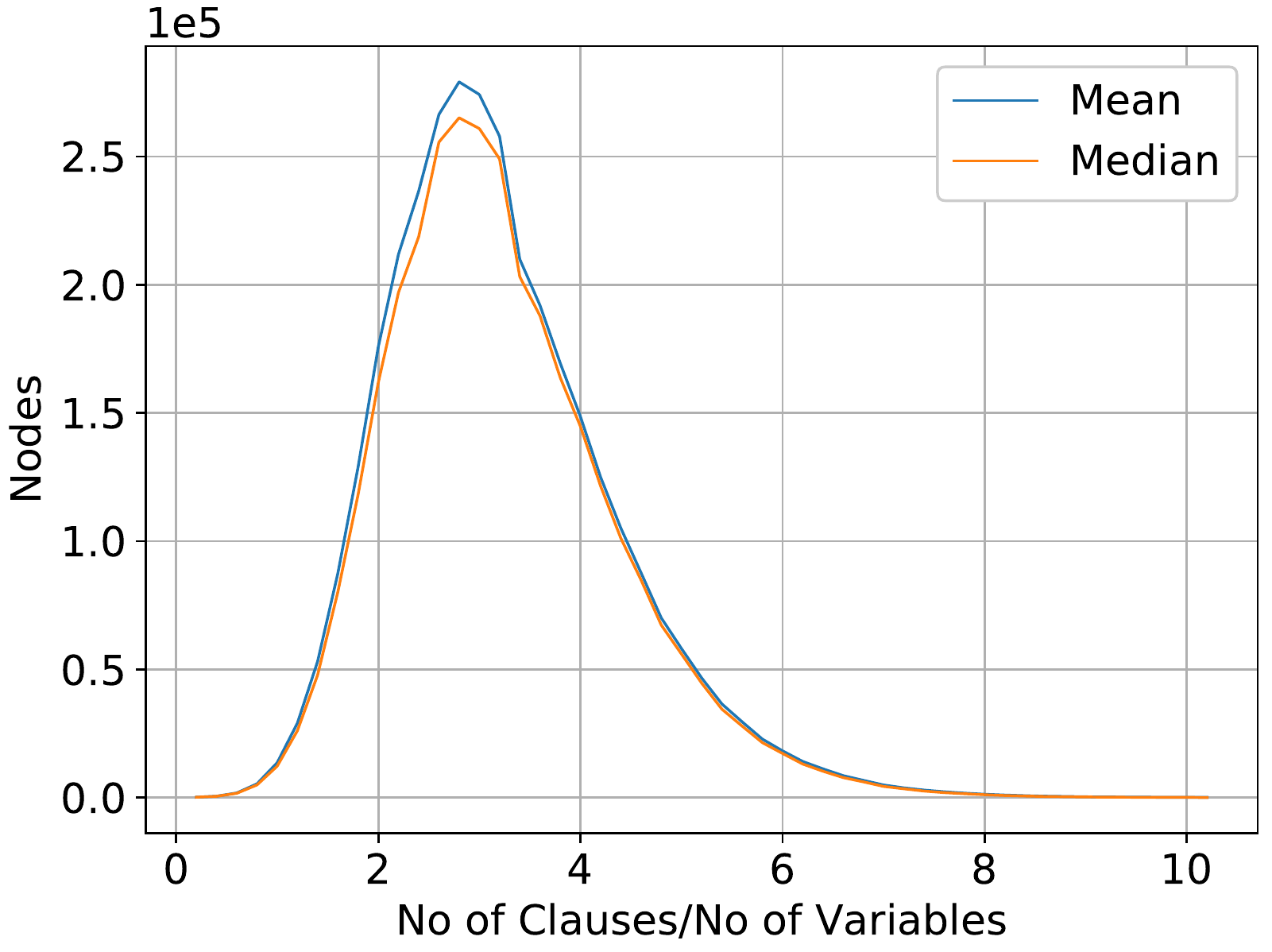}
		\caption{4-CNF and 30 vars}
		\label{fig:appendix:d4_nVc_4CNF_30vars}
	\end{subfigure}
	\hfill
	\begin{subfigure}[b]{0.30\linewidth}
		\centering
		\includegraphics[width=\textwidth]{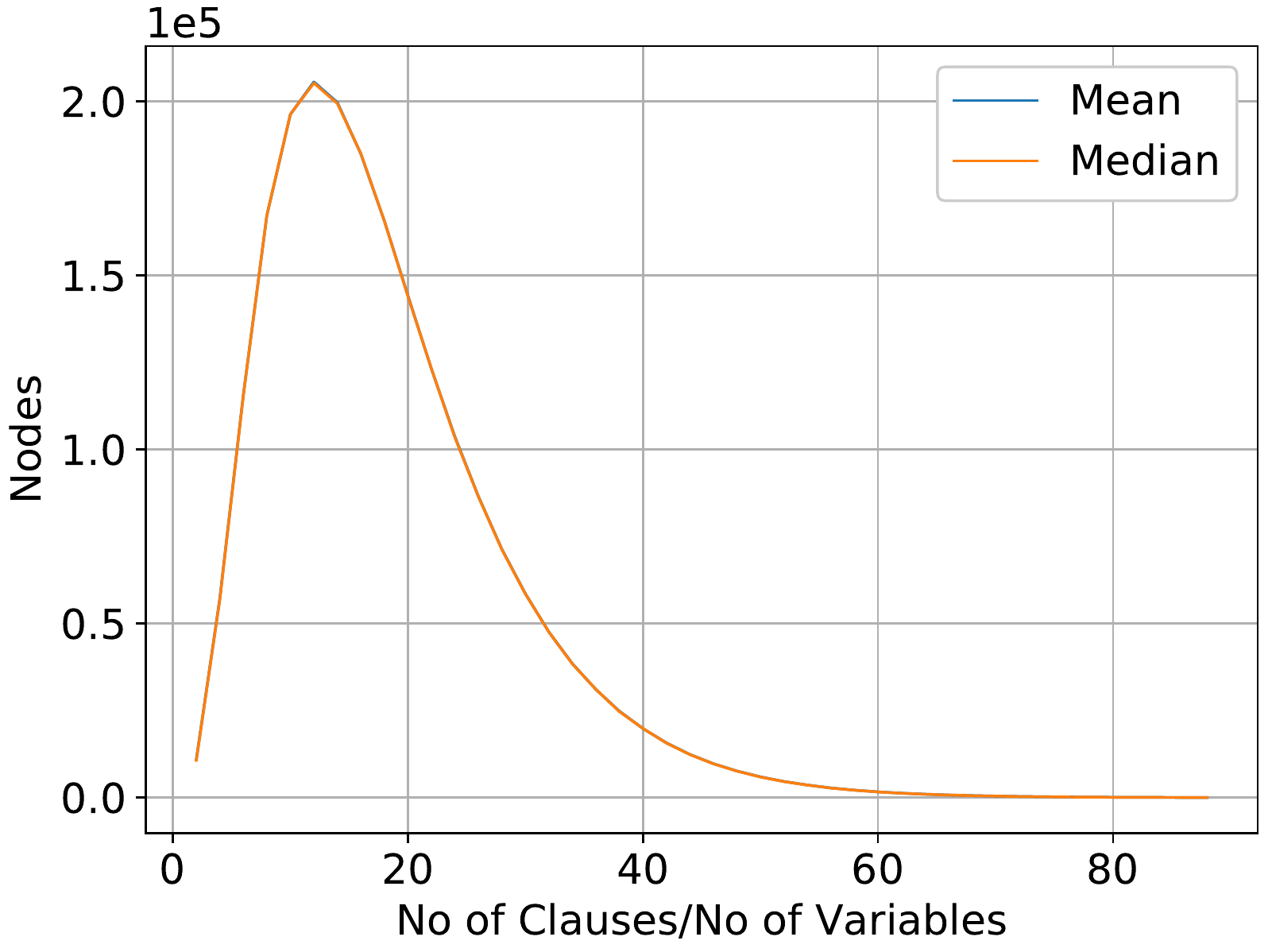}
		\caption{7-CNF and 20 vars}
		\label{fig:appendix:d4_nVc_7CNF_20vars}
	\end{subfigure}	
	\caption{\label{fig:appendix:d4_nVc_CNF}Nodes in d-DNNF vs clause density for different clause lengths($k$)}
\end{figure*}

\begin{figure*}[!th]
	\centering
	\begin{subfigure}[b]{0.30\linewidth}
		\centering
		\includegraphics[width=\textwidth]{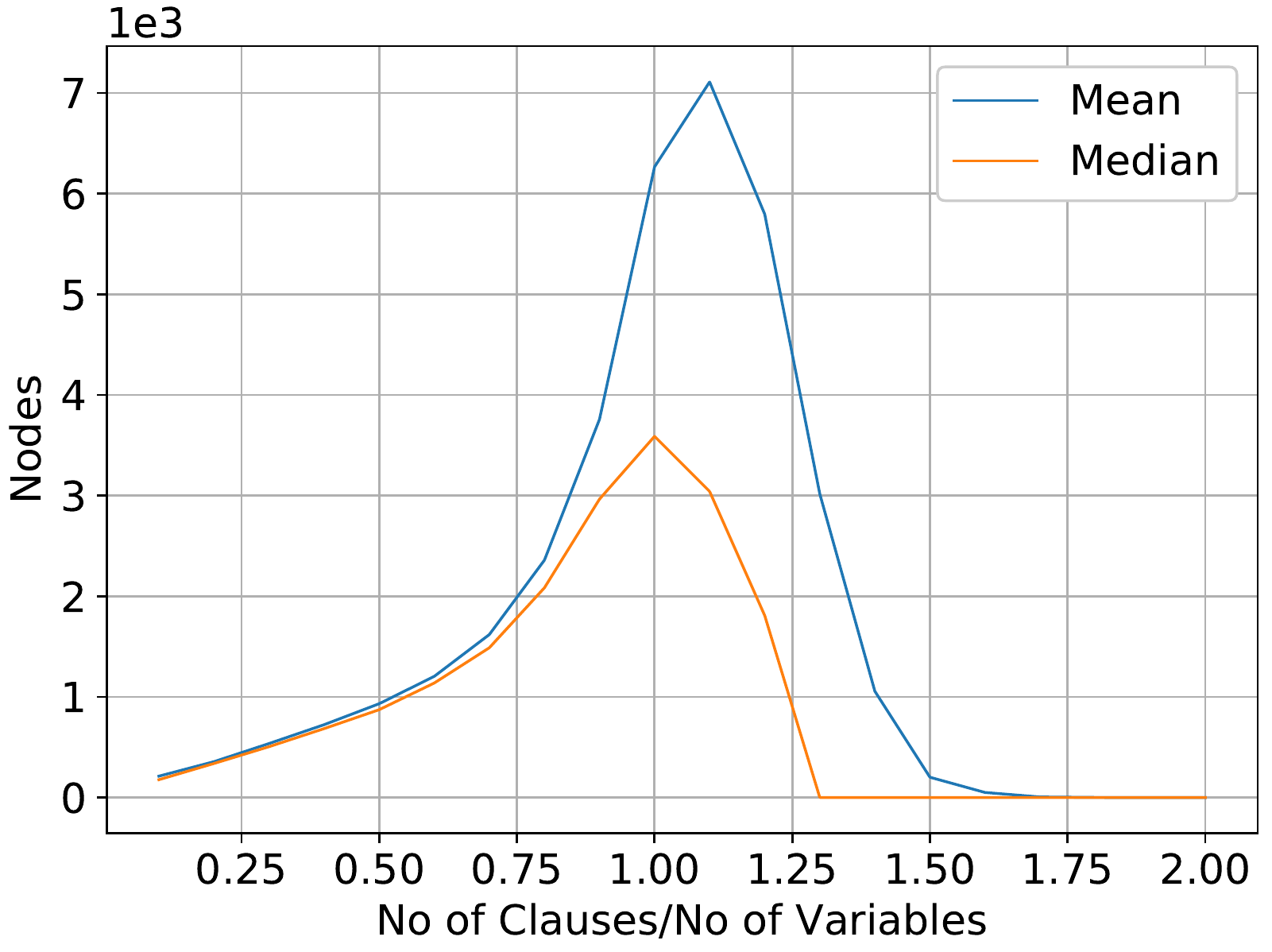}
		\caption{2-CNF and 200 vars}
		\label{fig:appendix:sdd_nVc_2CNF_200vars}
	\end{subfigure}
	\hfill
	\begin{subfigure}[b]{0.30\linewidth}
		\centering
		\includegraphics[width=\textwidth]{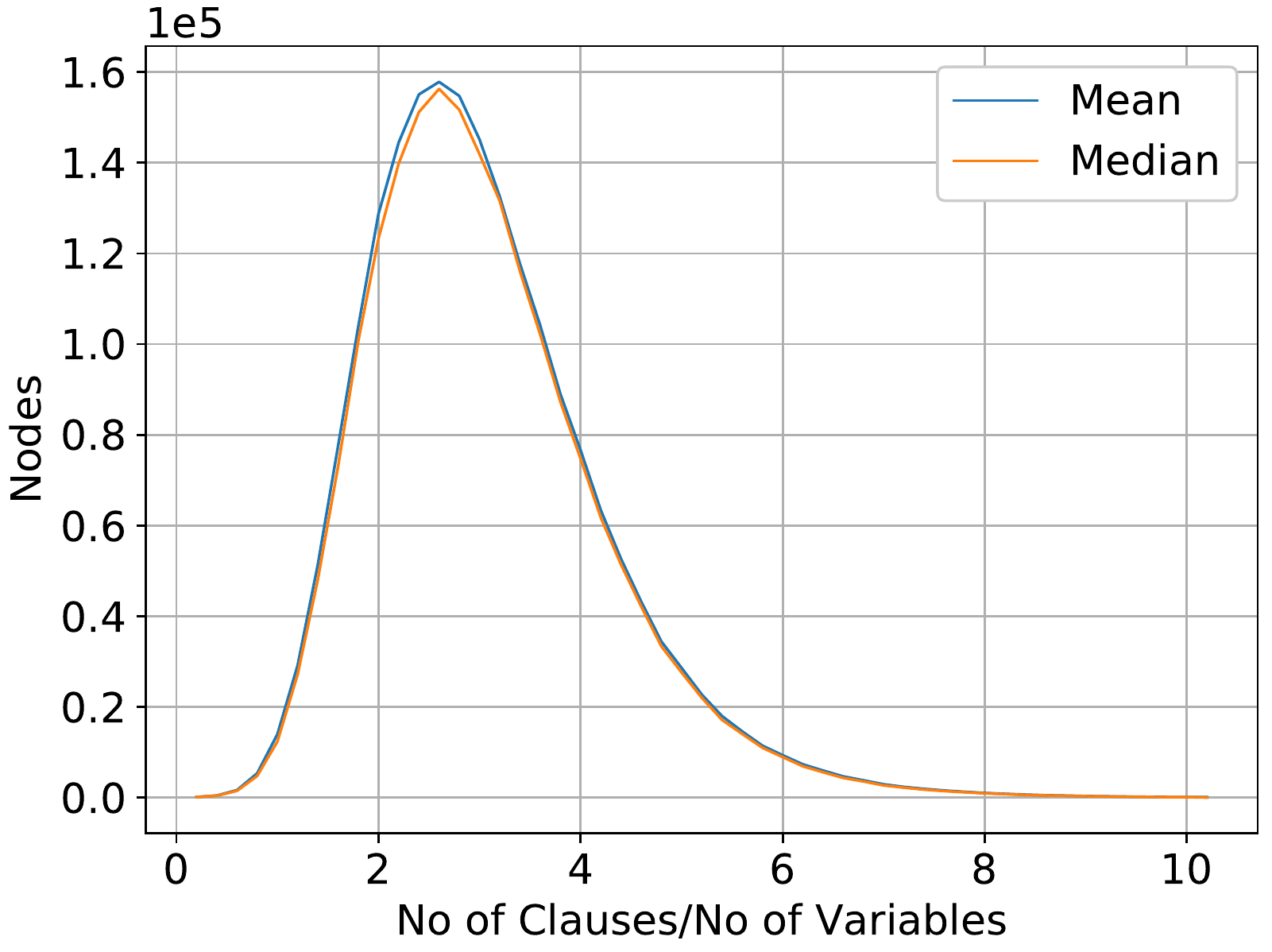}
		\caption{4-CNF and 30 vars}
		\label{fig:appendix:sdd_nVc_4CNF_30vars}
	\end{subfigure}
	\hfill
	\begin{subfigure}[b]{0.30\linewidth}
		\centering
		\includegraphics[width=\textwidth]{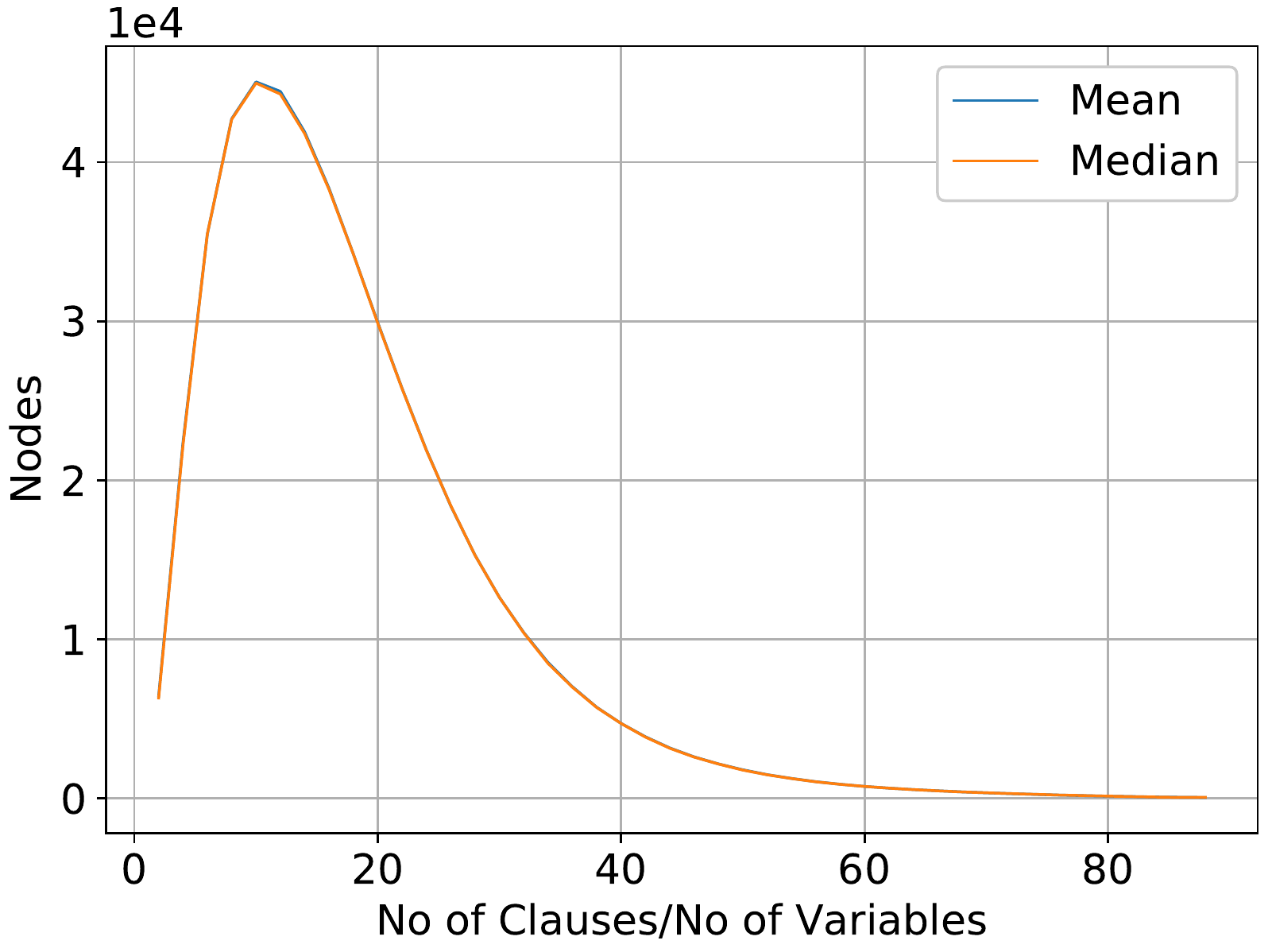}
		\caption{7-CNF and 20 vars}
		\label{fig:appendix:sdd_nVc_7CNF_20vars}
	\end{subfigure}	
	\caption{\label{fig:appendix:sdd_nVc_CNF}Nodes in SDD vs clause density for different clause lengths($k$)}
\end{figure*}

\begin{figure*}[!th]
	\centering
	\begin{subfigure}[b]{0.30\linewidth}
		\centering
		\includegraphics[width=\textwidth]{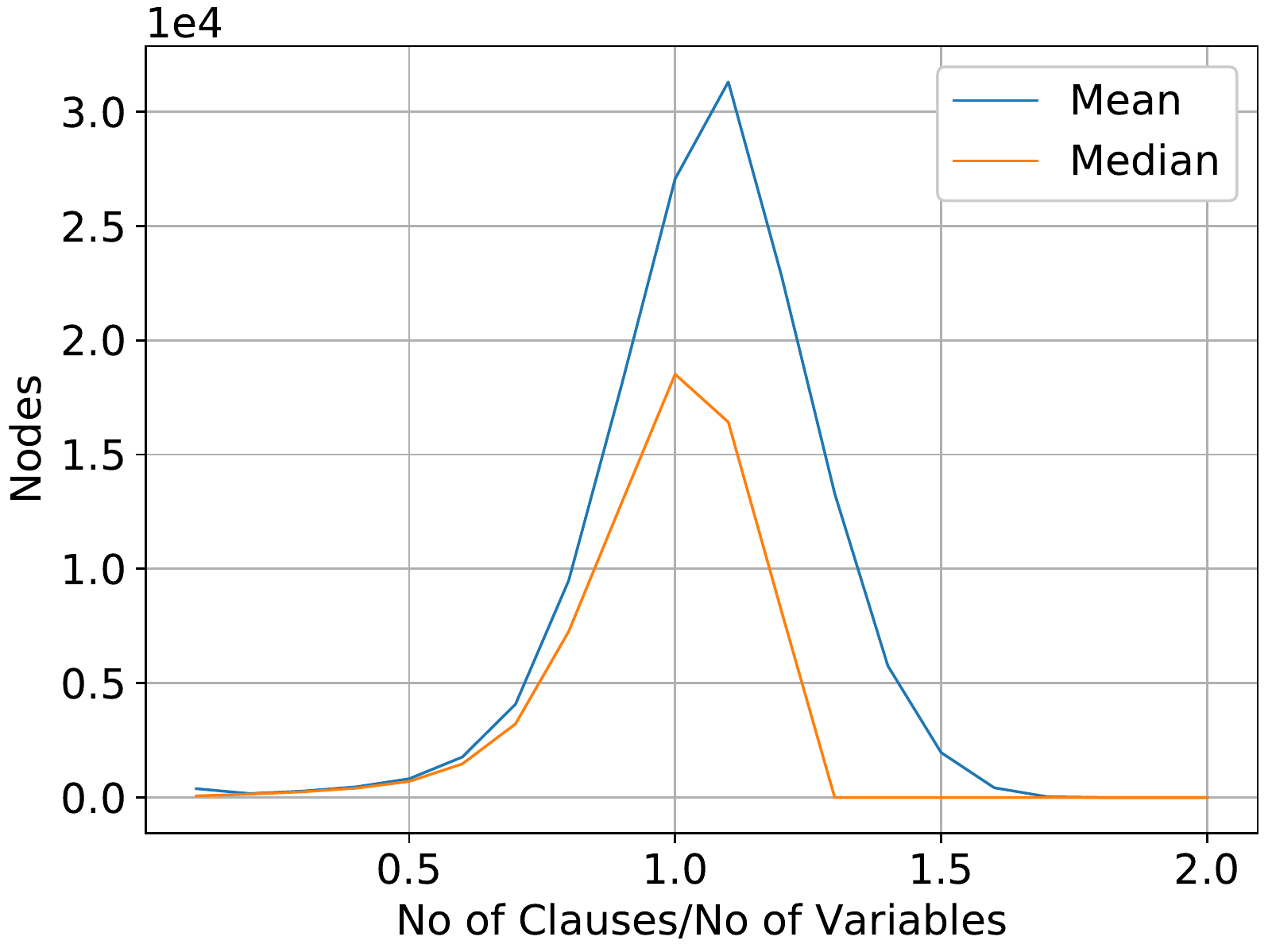}
		\caption{2-CNF and 200 vars}
		\label{fig:appendix:CUDD_nVc_2CNF_200vars}
	\end{subfigure}
	\hfill
	\begin{subfigure}[b]{0.30\linewidth}
		\centering
		\includegraphics[width=\textwidth]{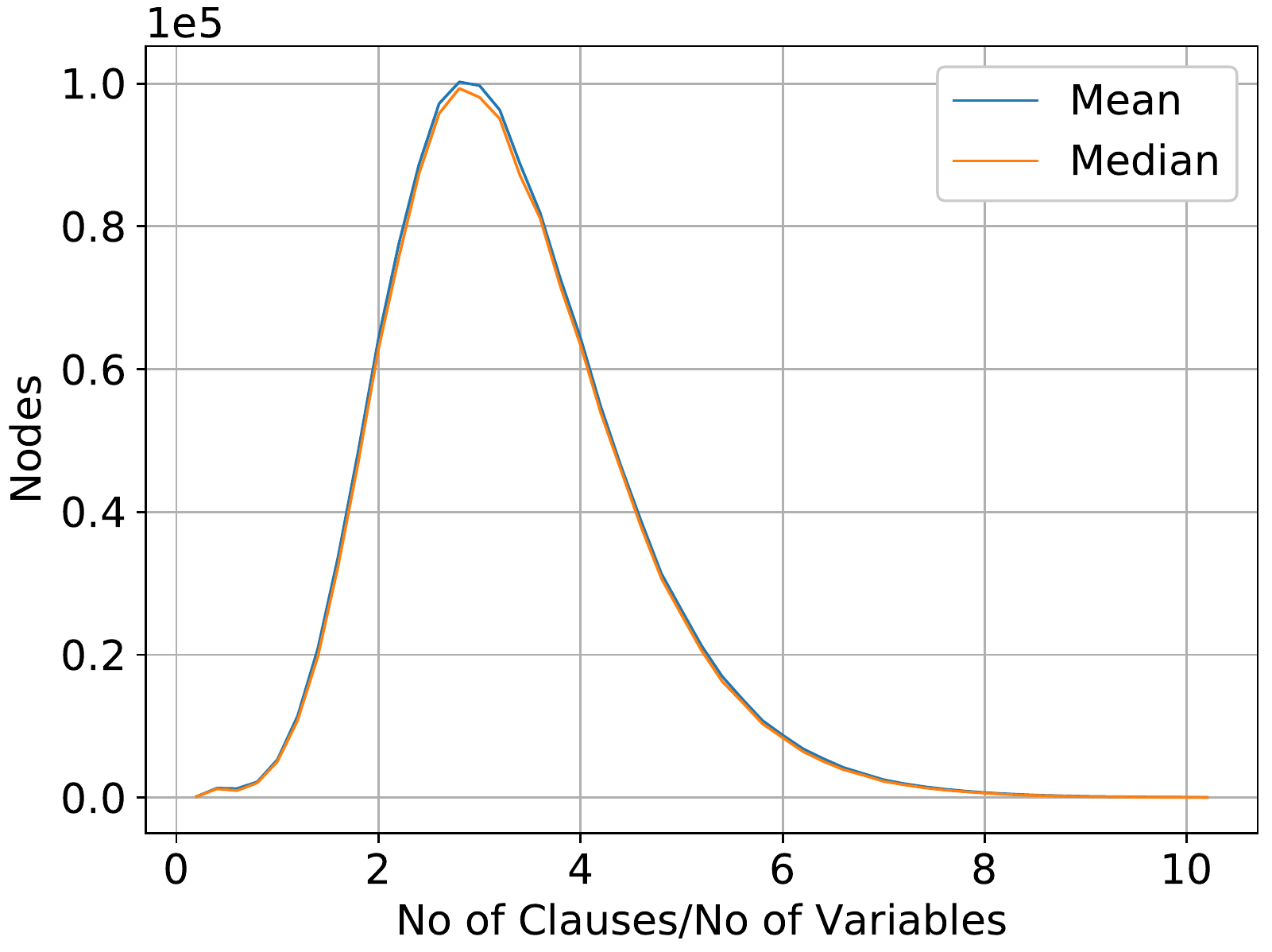}
		\caption{4-CNF and 30 vars}
		\label{fig:appendix:CUDD_nVc_4CNF_30vars}
	\end{subfigure}
	\hfill
	\begin{subfigure}[b]{0.30\linewidth}
		\centering
		\includegraphics[width=\textwidth]{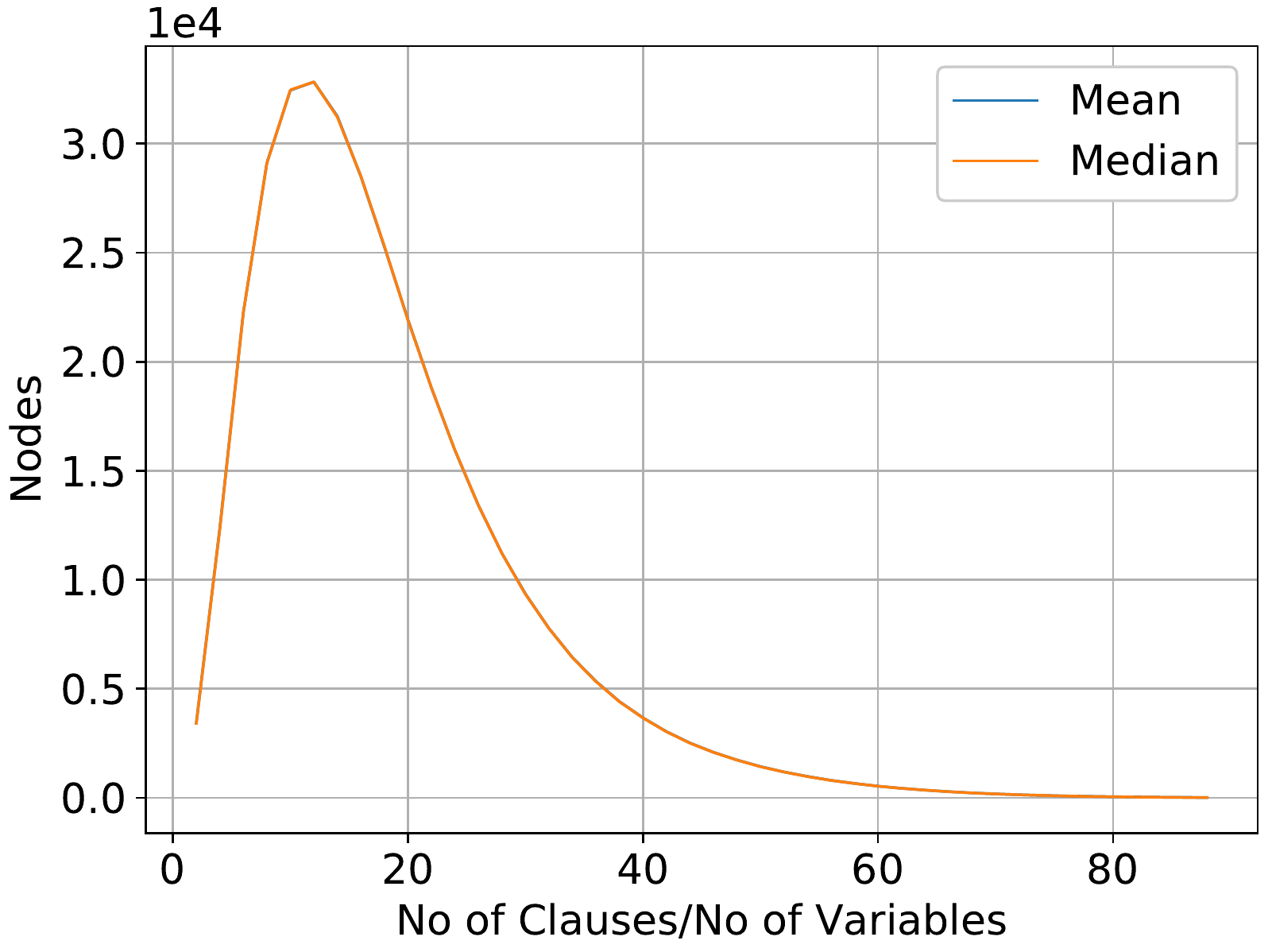}
		\caption{7-CNF and 20 vars}
		\label{fig:appendix:CUDD_nVc_7CNF_20vars}
	\end{subfigure}	
	\caption{\label{fig:appendix:CUDD_nVc_CNF} Nodes in BDD vs clause density for different clause lengths($k$)}
\end{figure*}

\begin{figure*}[!th]
	\centering
	\begin{subfigure}[b]{0.30\linewidth}
		\centering
		\includegraphics[width=\textwidth]{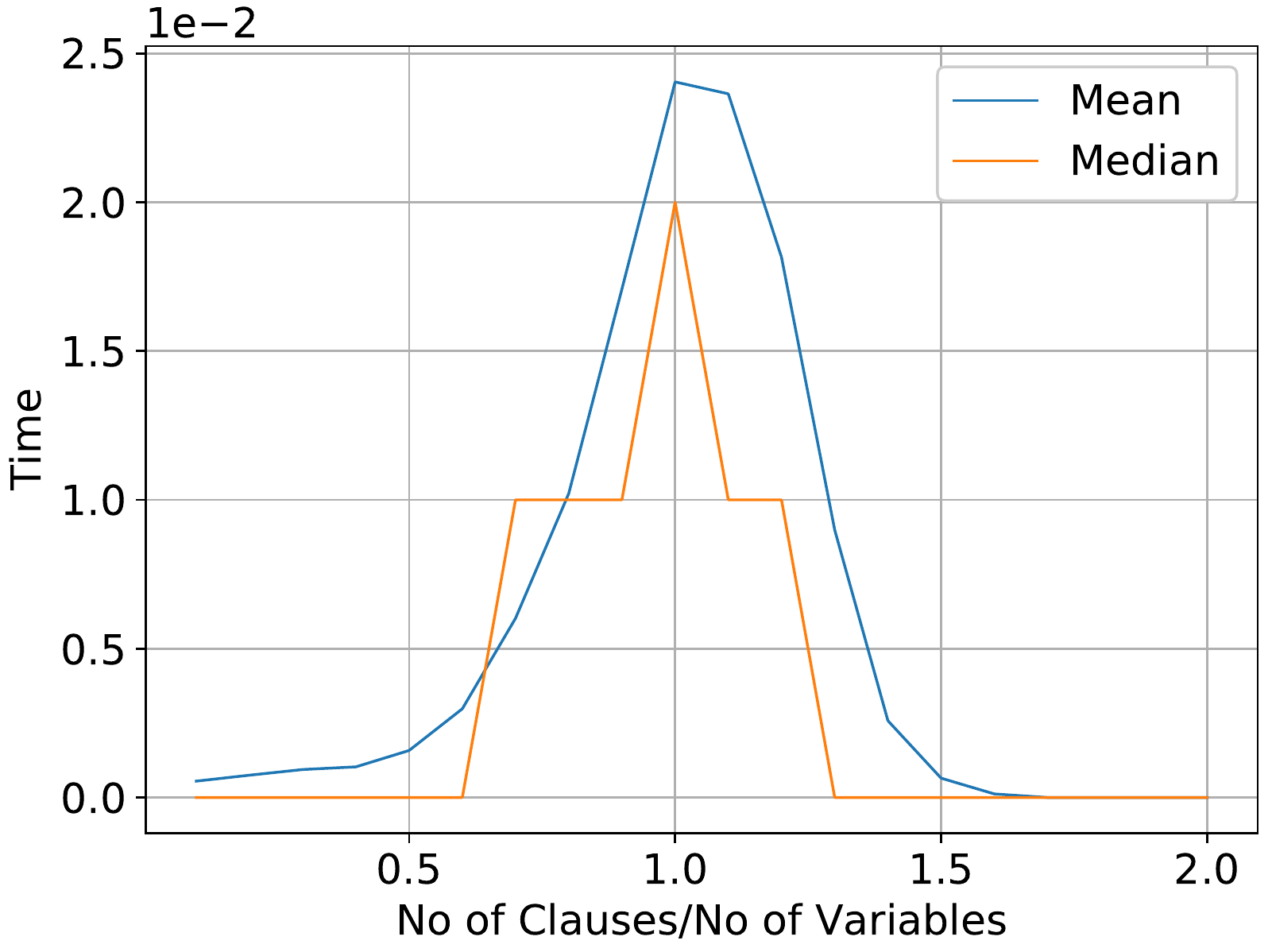}
		\caption{2-CNF and 200 vars}
		\label{fig:appendix:d4_tVc_2CNF_200vars}
	\end{subfigure}
	\hfill
	\begin{subfigure}[b]{0.30\linewidth}
		\centering
		\includegraphics[width=\textwidth]{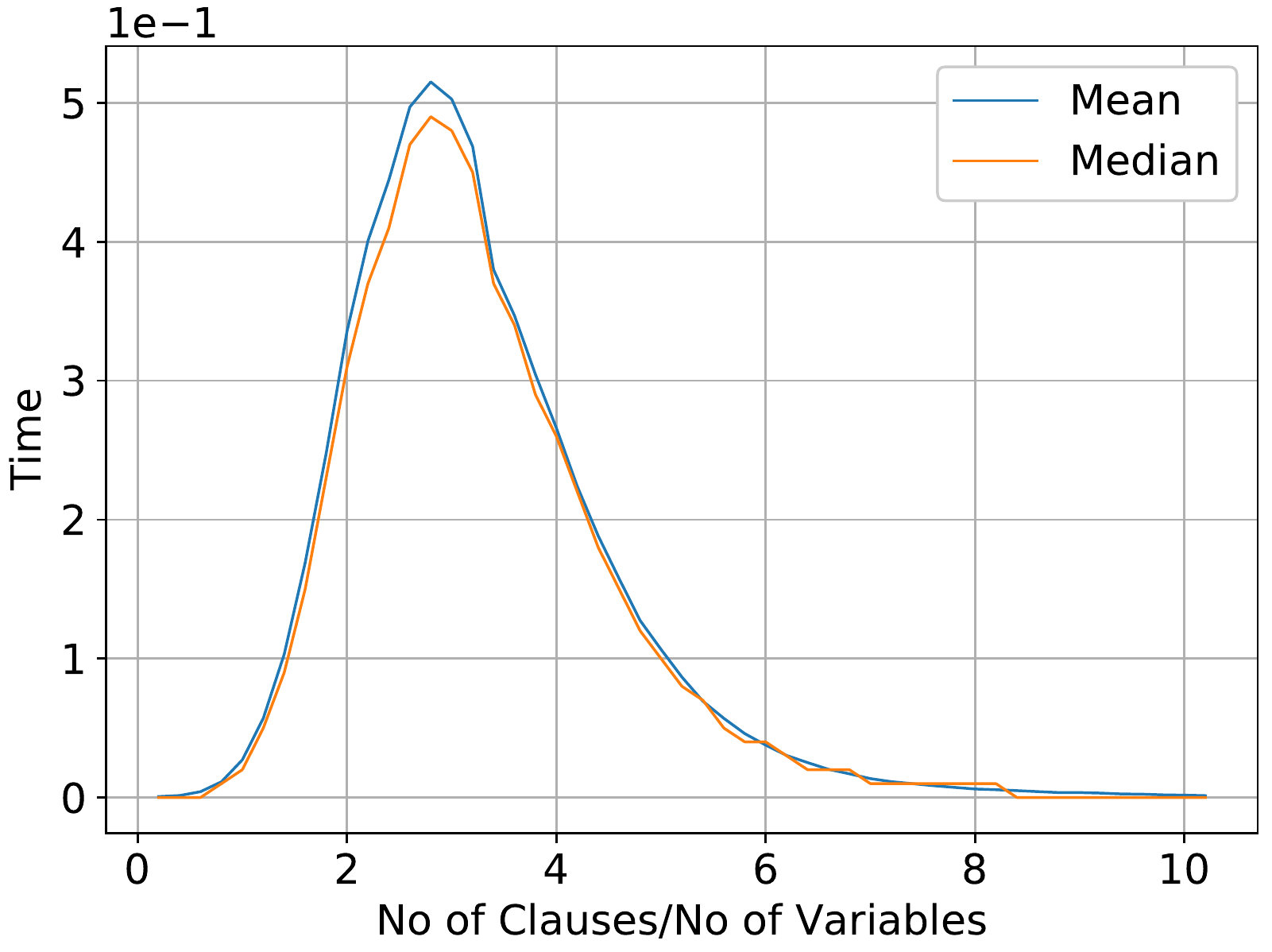}
		\caption{4-CNF and 30 vars}
		\label{fig:appendix:d4_tVc_4CNF_30vars}
	\end{subfigure}
	\hfill
	\begin{subfigure}[b]{0.30\linewidth}
		\centering
		\includegraphics[width=\textwidth]{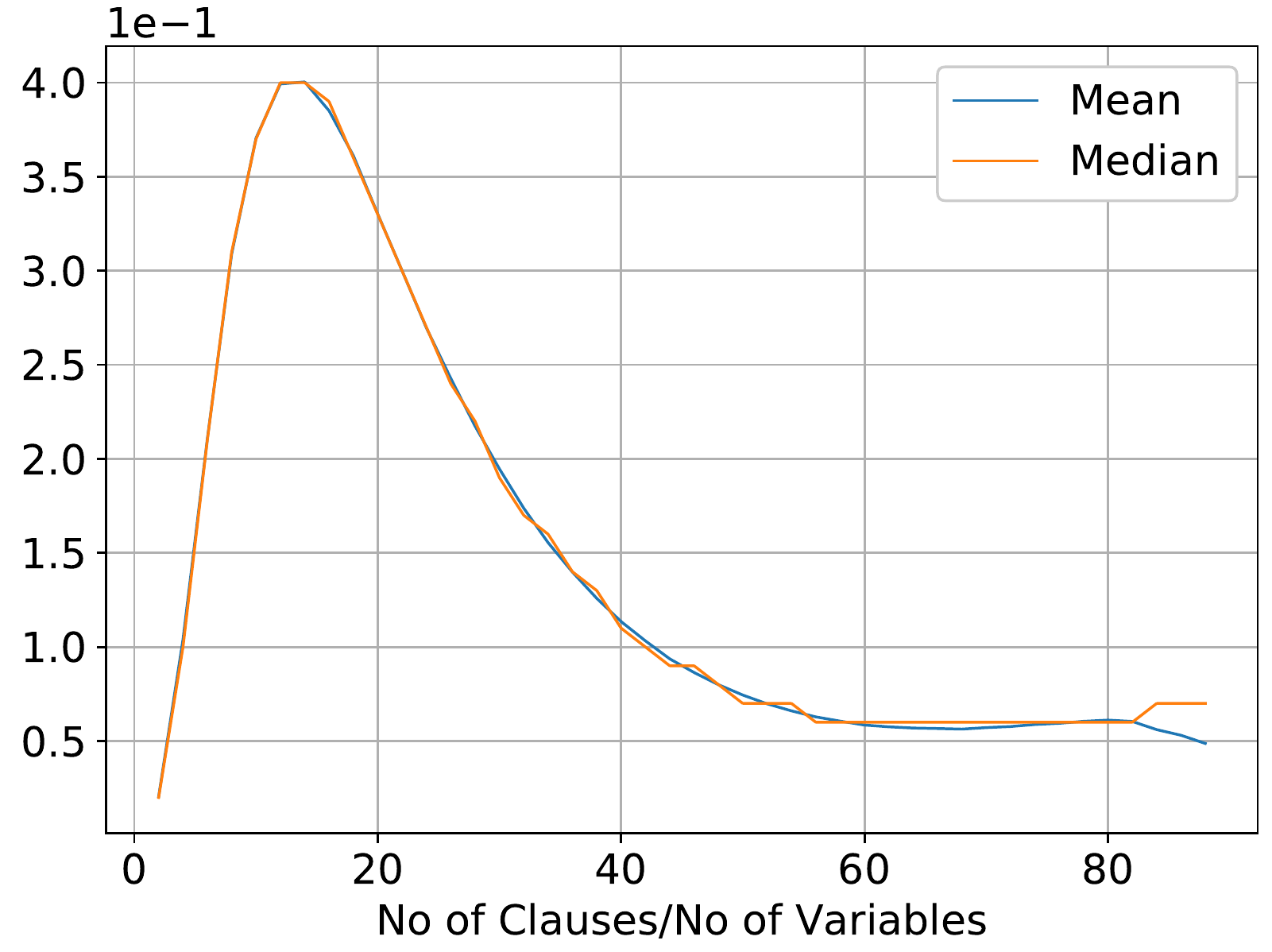}
		\caption{7-CNF and 20 vars}
		\label{fig:appendix:d4_tVc_7CNF_20vars}
	\end{subfigure}	
	\caption{\label{fig:appendix:d4_tVc_CNF}Compile-time for d-DNNF vs clause density for different clause lengths($k$)}
\end{figure*}

\begin{figure*}[!th]
	\centering
	\begin{subfigure}[b]{0.30\linewidth}
		\centering
		\includegraphics[width=\textwidth]{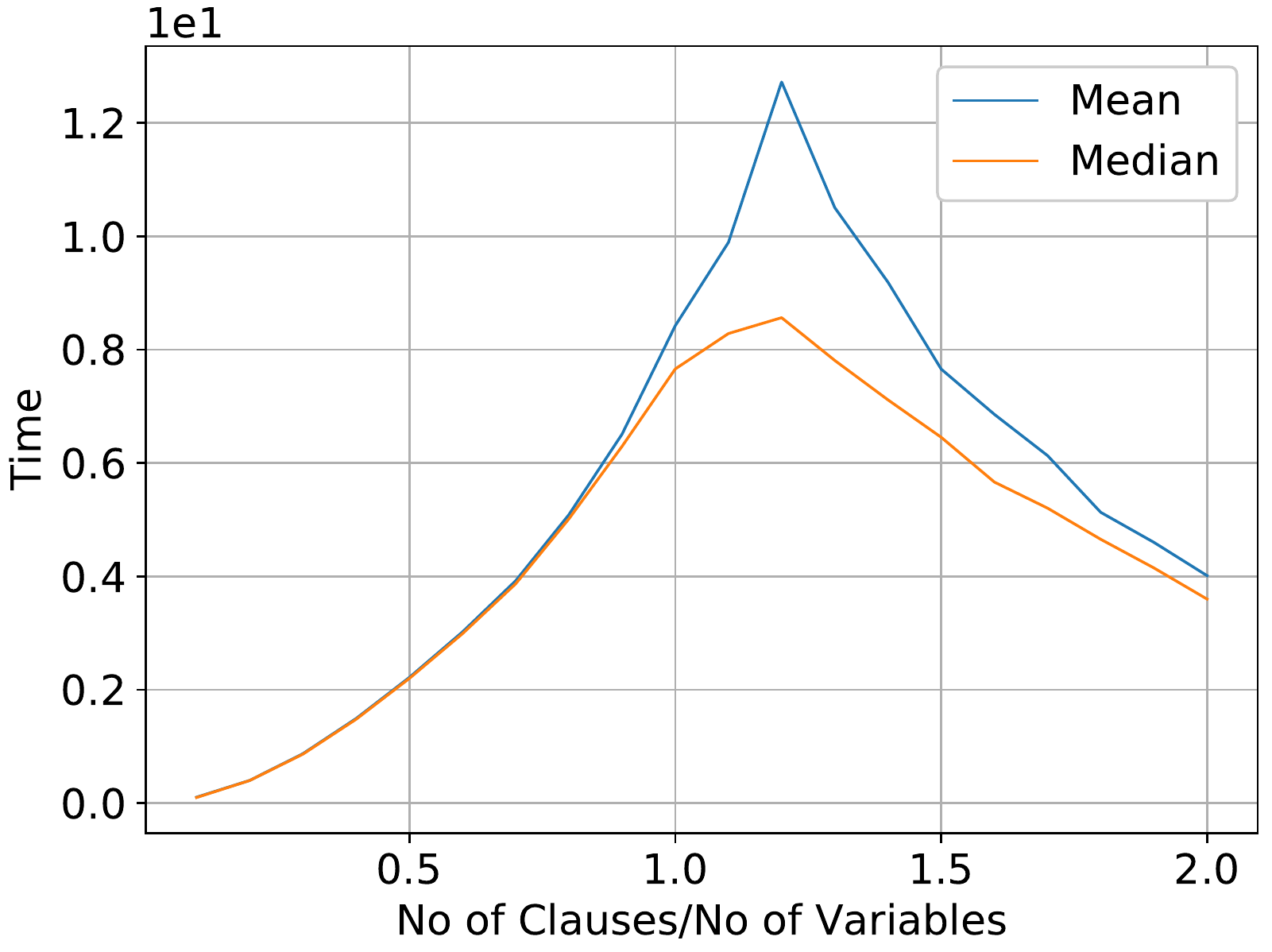}
		\caption{2-CNF and 200 vars}
		\label{fig:appendix:sdd_tVc_2CNF_200vars}
	\end{subfigure}
	\hfill
	\begin{subfigure}[b]{0.30\linewidth}
		\centering
		\includegraphics[width=\textwidth]{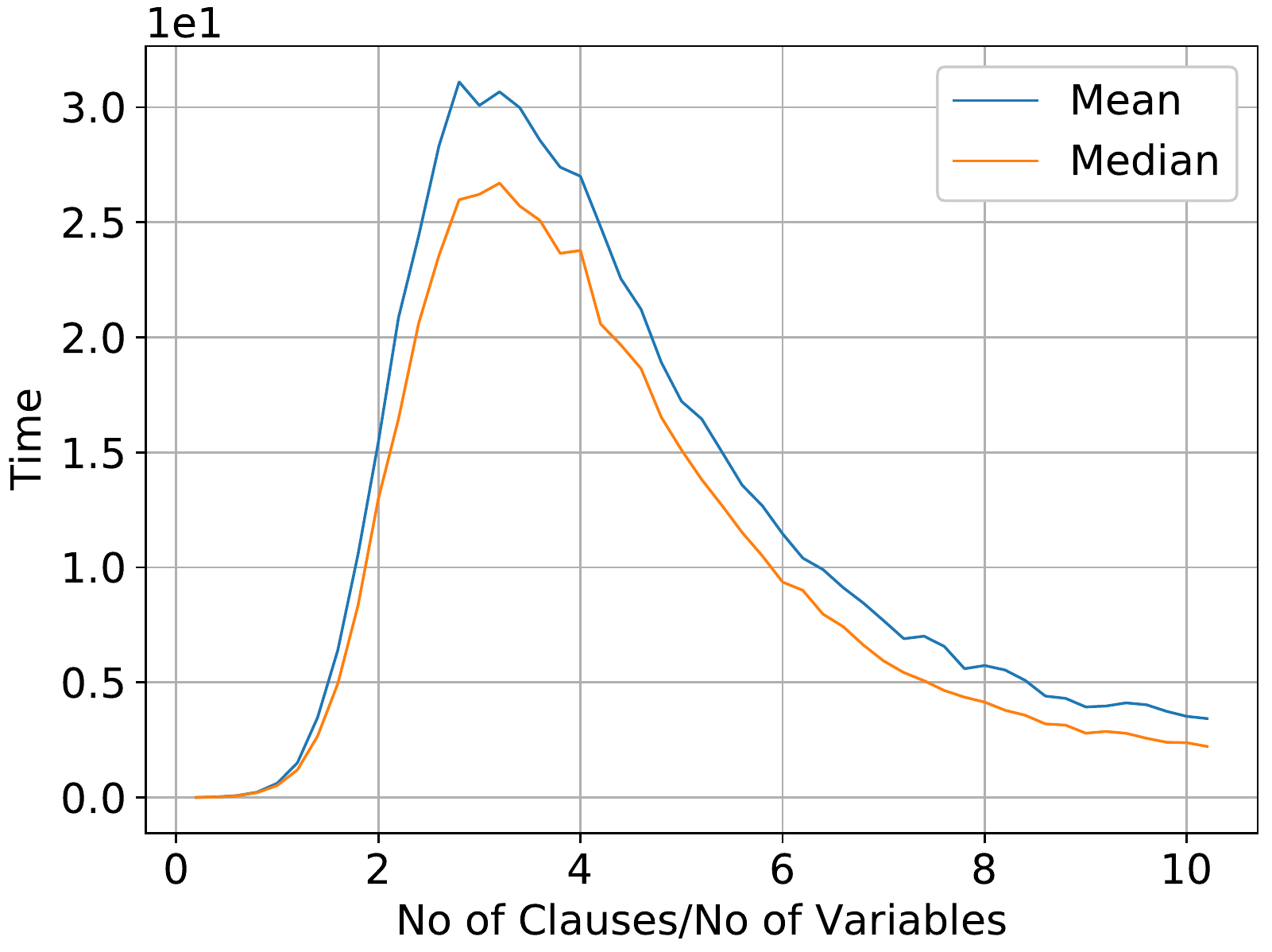}
		\caption{4-CNF and 30 vars}
		\label{fig:appendix:sdd_tVc_4CNF_30vars}
	\end{subfigure}
	\hfill
	\begin{subfigure}[b]{0.30\linewidth}
		\centering
		\includegraphics[width=\textwidth]{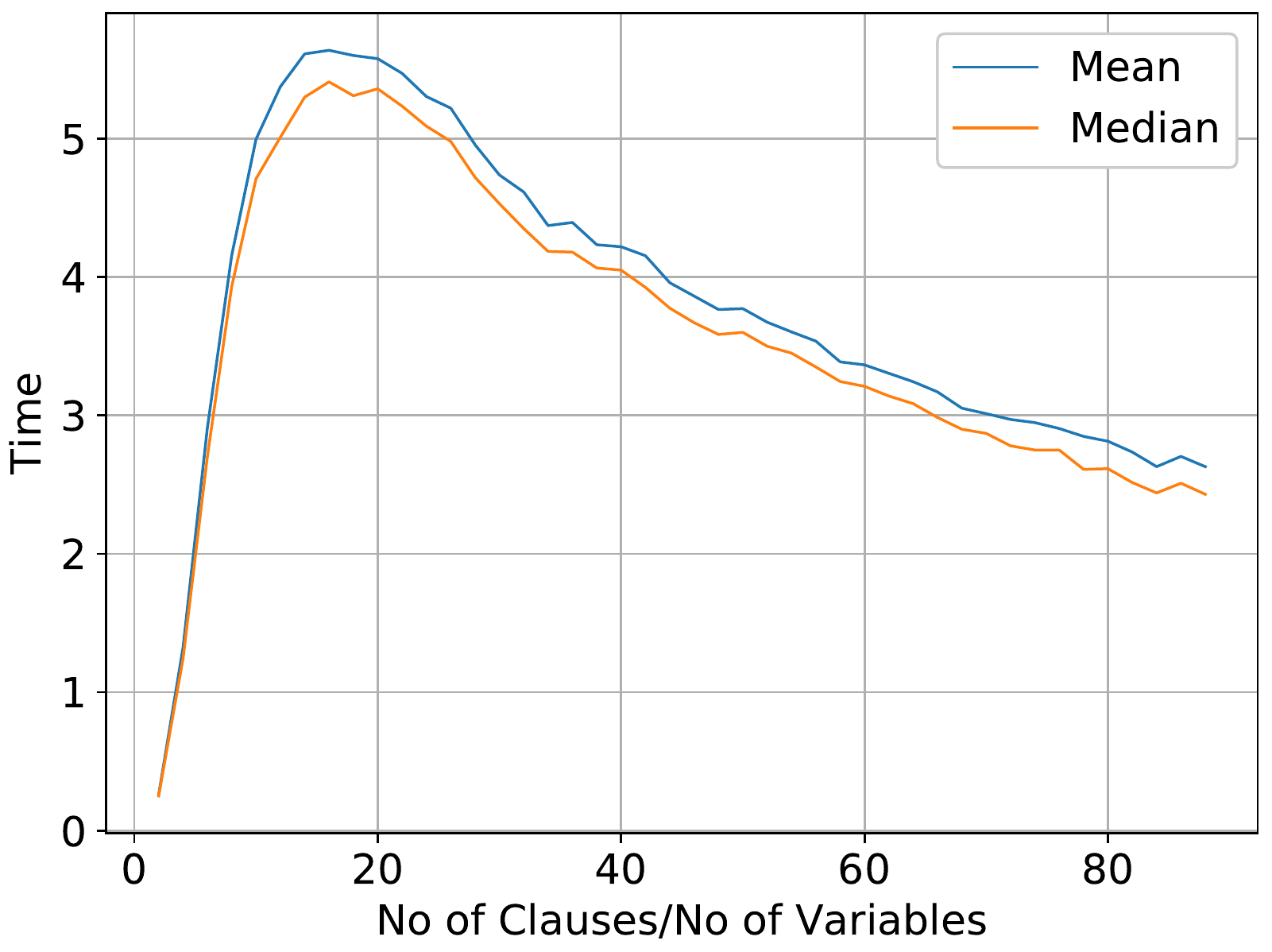}
		\caption{7-CNF and 20 vars}
		\label{fig:appendix:sdd_tVc_7CNF_20vars}
	\end{subfigure}	
	\caption{\label{fig:appendix:sdd_tVc_CNF}Compile-time for SDD vs clause density for different clause lengths($k$)}
\end{figure*}

\begin{figure*}[!th]
	\centering
	\begin{subfigure}[b]{0.30\linewidth}
		\centering
		\includegraphics[width=\textwidth]{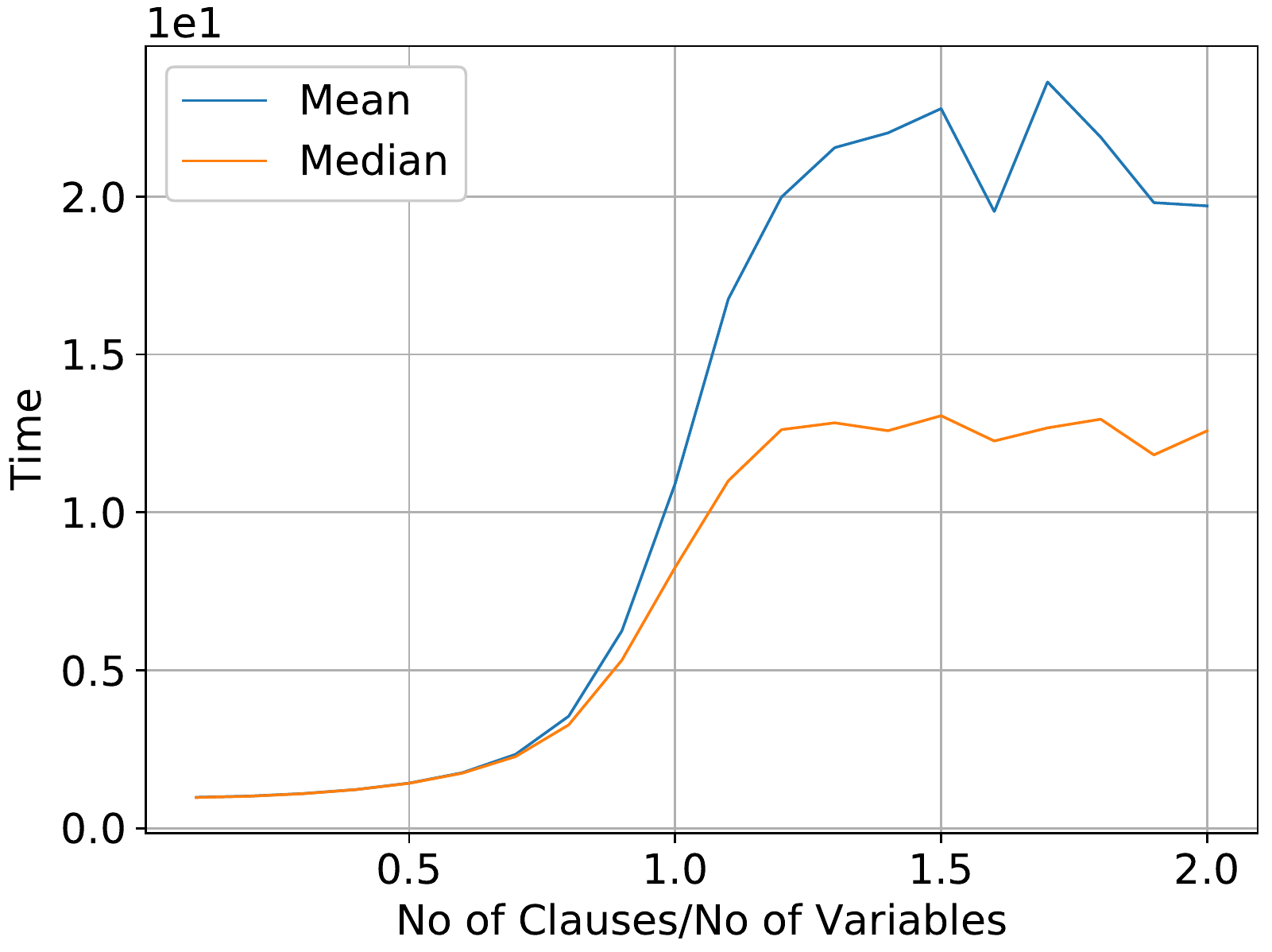}
		\caption{2-CNF and 200 vars}
		\label{fig:appendix:CUDD_tVc_2CNF_200vars}
	\end{subfigure}
	\hfill
	\begin{subfigure}[b]{0.30\linewidth}
		\centering
		\includegraphics[width=\textwidth]{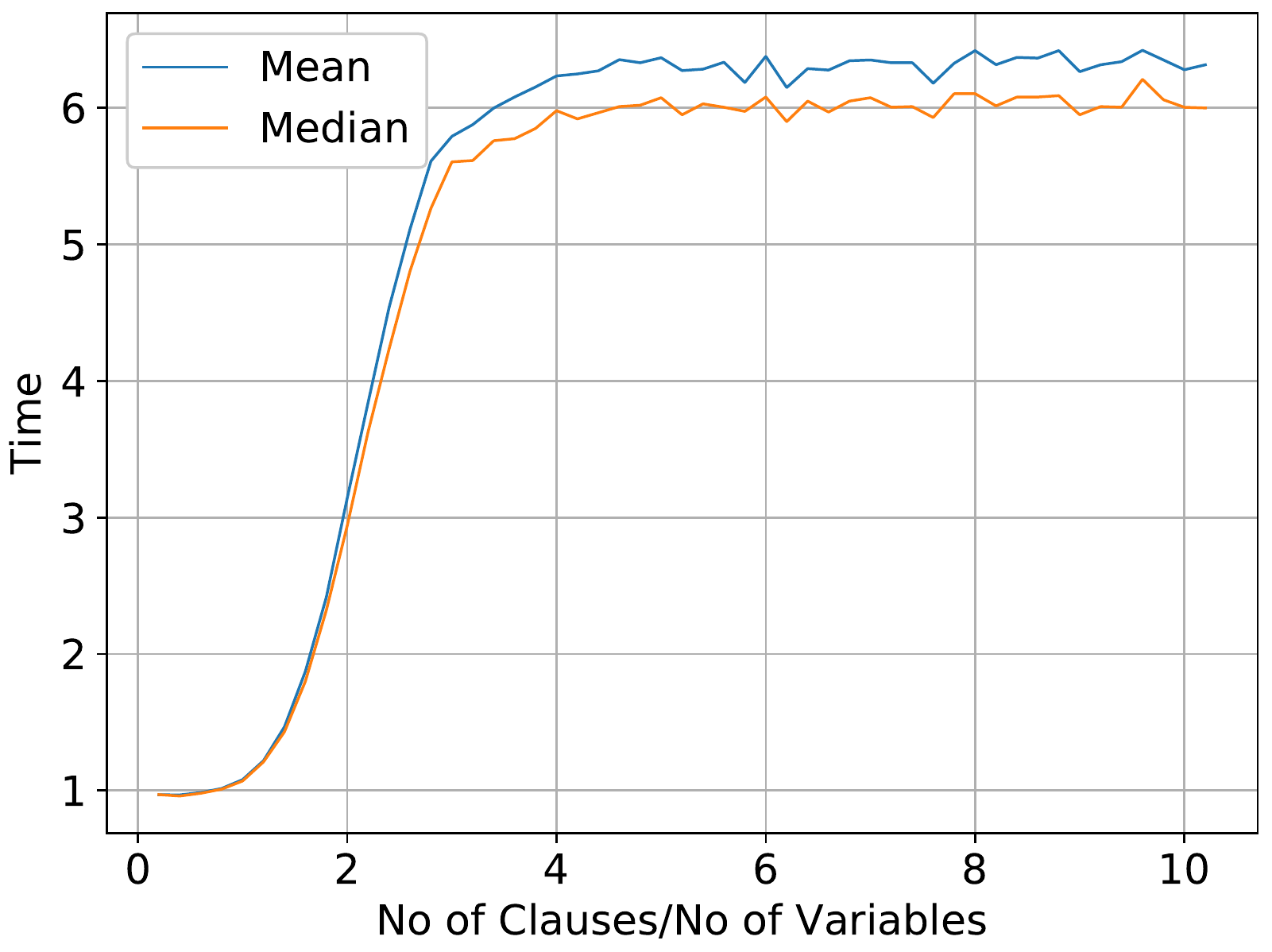}
		\caption{4-CNF and 30 vars}
		\label{fig:appendix:CUDD_tVc_4CNF_30vars}
	\end{subfigure}
	\hfill
	\begin{subfigure}[b]{0.30\linewidth}
		\centering
		\includegraphics[width=\textwidth]{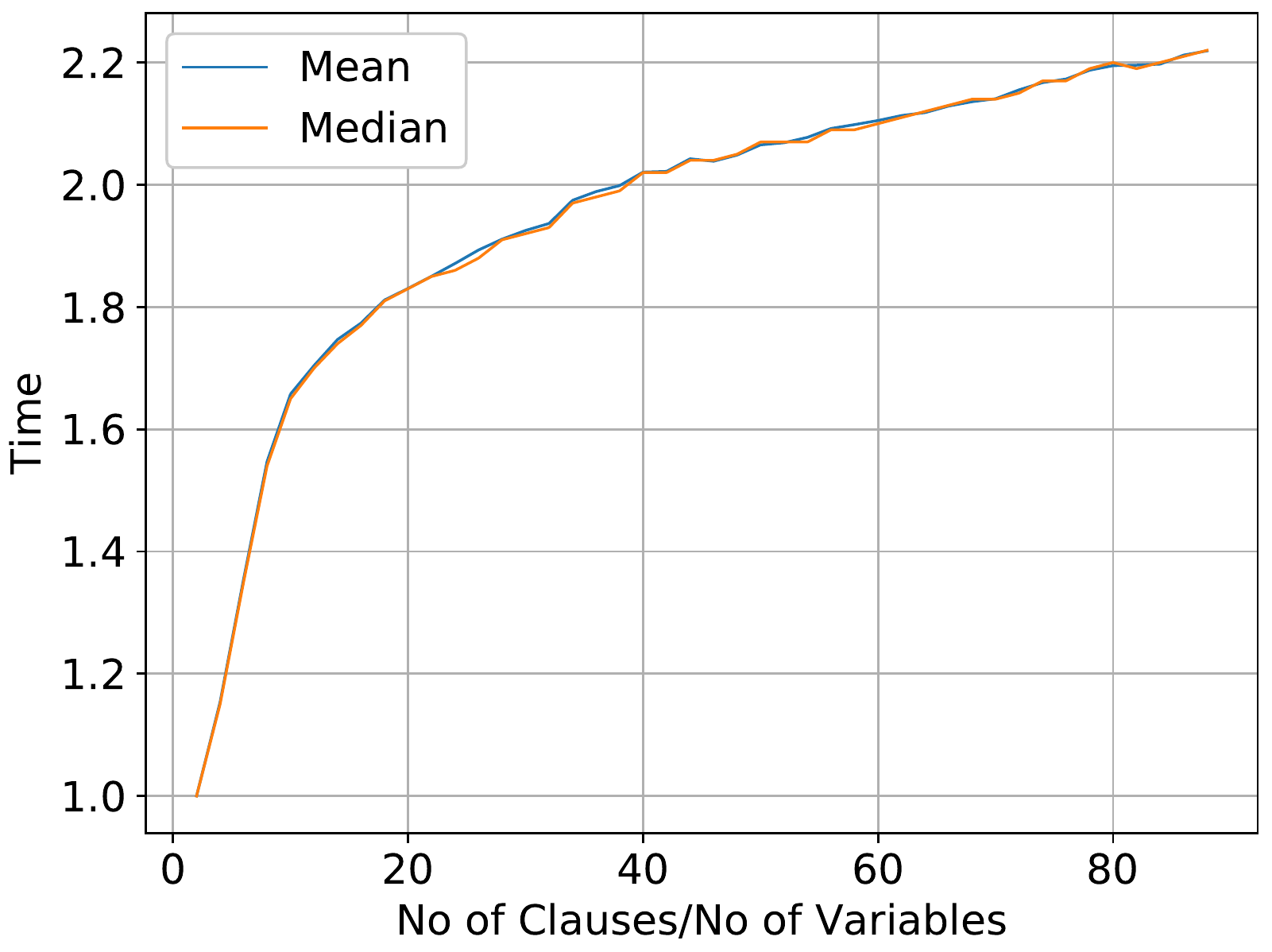}
		\caption{7-CNF and 20 vars}
		\label{fig:appendix:CUDD_tVc_7CNF_20vars}
	\end{subfigure}	
	\caption{\label{fig:appendix:CUDD_tVc_CNF}Compile-time for BDD vs clause density for different clause lengths($k$)}
\end{figure*}

Figures~\ref{fig:appendix:d4_nVc_CNF},~\ref{fig:appendix:sdd_nVc_CNF} and~\ref{fig:appendix:CUDD_nVc_CNF} show the variation in size  with clause density for different clause lengths. Figures~\ref{fig:appendix:d4_tVc_CNF},~\ref{fig:appendix:sdd_tVc_CNF} and ~\ref{fig:appendix:CUDD_tVc_CNF} show the variation in runtime  with clause density for different clause lengths.

\clearpage
\section{Phase transtion with solution density}
In this section, we show the variations in size and compile-times obtained empirically for d-DNNFs, SDDs and OBDDs with solution density.

\subsection{Observing the phase transition}

\subsubsection{Size with number of variables} 
\begin{figure*}[!th]
	\centering
	\begin{subfigure}[b]{0.30\linewidth}
		\centering
		\includegraphics[width=\textwidth]{MainFigures/dDNNF/d4act_10inst_nodesValpha_3cnf60vars.png}
		\caption{60 vars}
		\label{fig:appendix:d4_nVa_3CNF_60vars}
	\end{subfigure}
	\hfill
	\begin{subfigure}[b]{0.30\linewidth}
		\centering
		\includegraphics[width=\textwidth]{MainFigures/dDNNF/d4act_10inst_nodesValpha_3cnf40vars.png}
		\caption{40 vars}
		\label{fig:appendix:d4_nVa_3CNF_40vars}
	\end{subfigure}
	\hfill
	\begin{subfigure}[b]{0.30\linewidth}
		\centering
		\includegraphics[width=\textwidth]{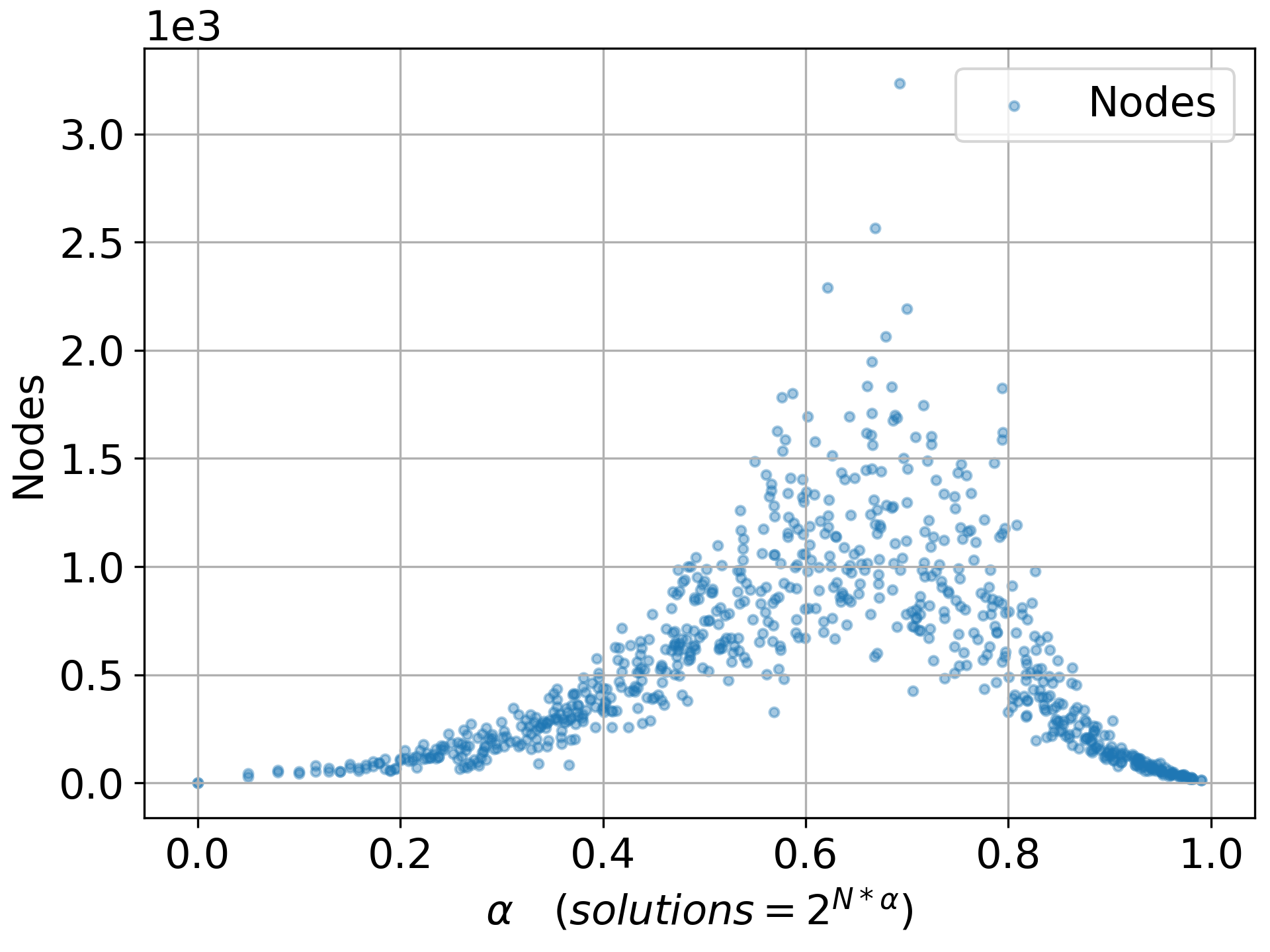}
		\caption{20 vars}
		\label{fig:appendix:d4_nVa_3CNF_20vars}
	\end{subfigure}
	\caption{Nodes in d-DNNF vs solution density($\alpha$) for different number of variables}
	\label{fig:appendix:d4_nVa_3CNF_vars}
\end{figure*}

\begin{figure*}[!th]
	\centering
	\begin{subfigure}[b]{0.30\linewidth}
		\centering
		\includegraphics[width=\textwidth]{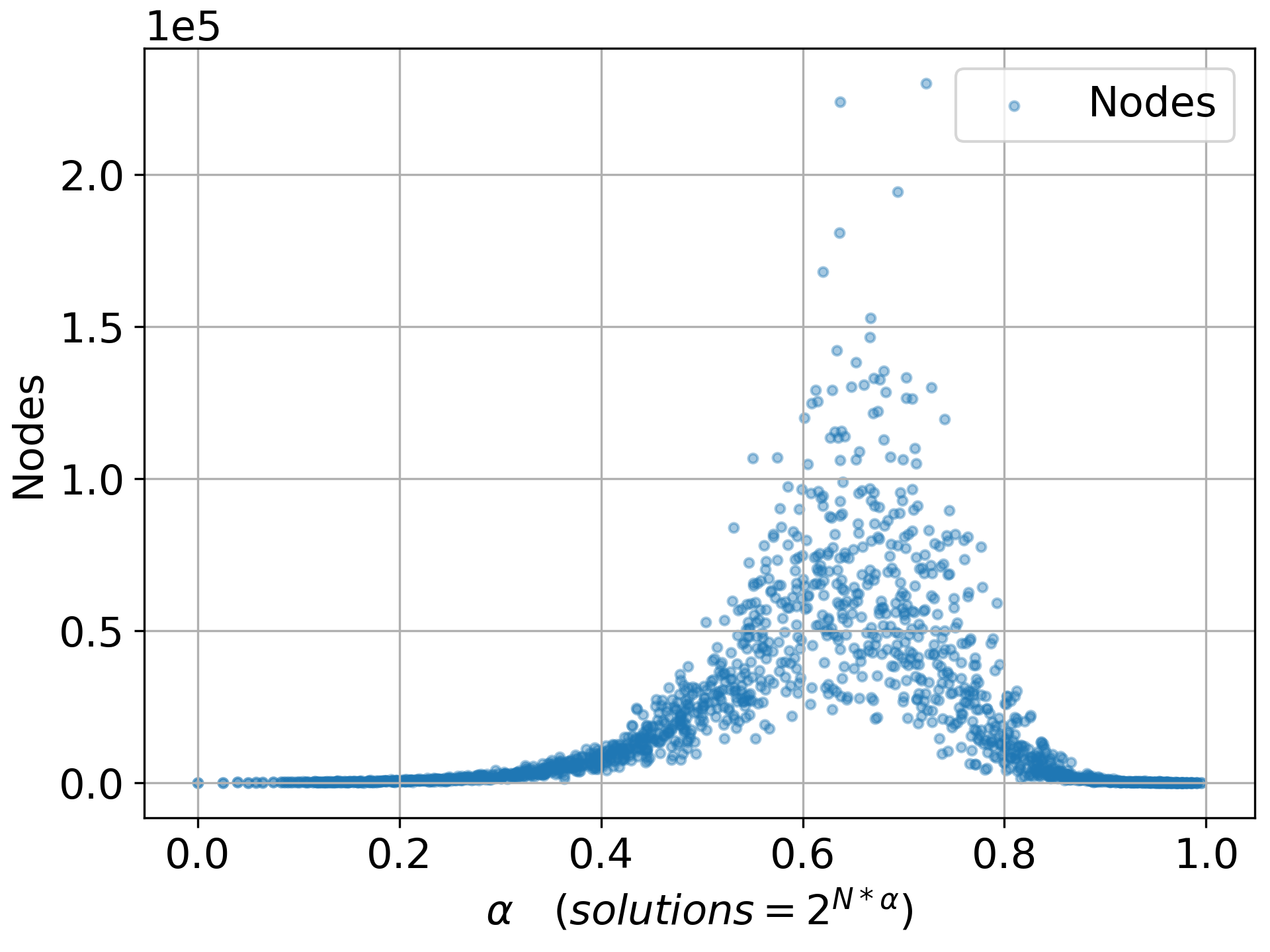}
		\caption{40 vars}
		\label{fig:appendix:sdd_nVa_3CNF_40vars}
	\end{subfigure}
	\hfill
	\begin{subfigure}[b]{0.30\linewidth}
		\centering
		\includegraphics[width=\textwidth]{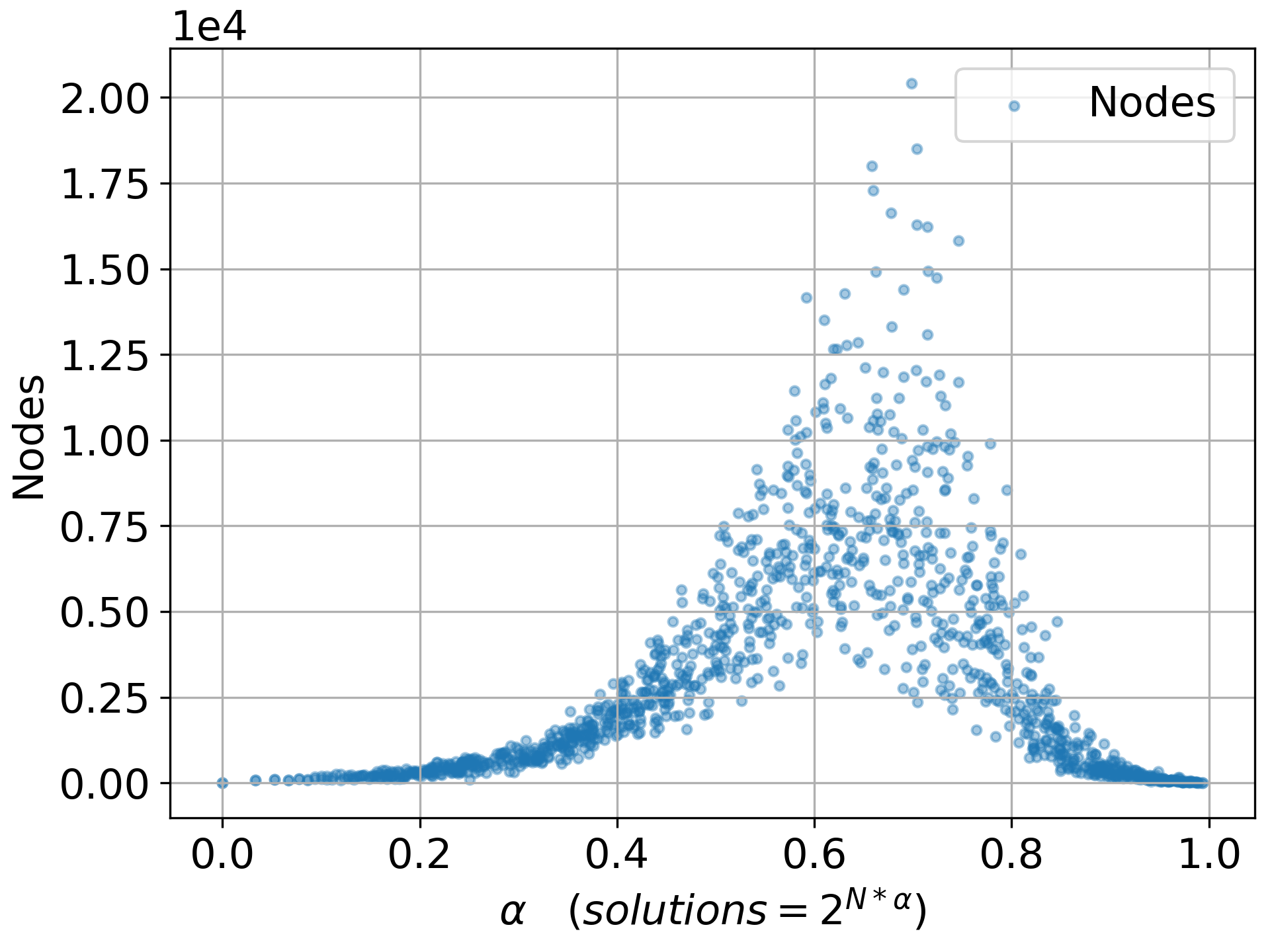}
		\caption{30 vars}
		\label{fig:appendix:sdd_nVa_3CNF_30vars}
	\end{subfigure}
	\hfill
	\begin{subfigure}[b]{0.30\linewidth}
		\centering
		\includegraphics[width=\textwidth]{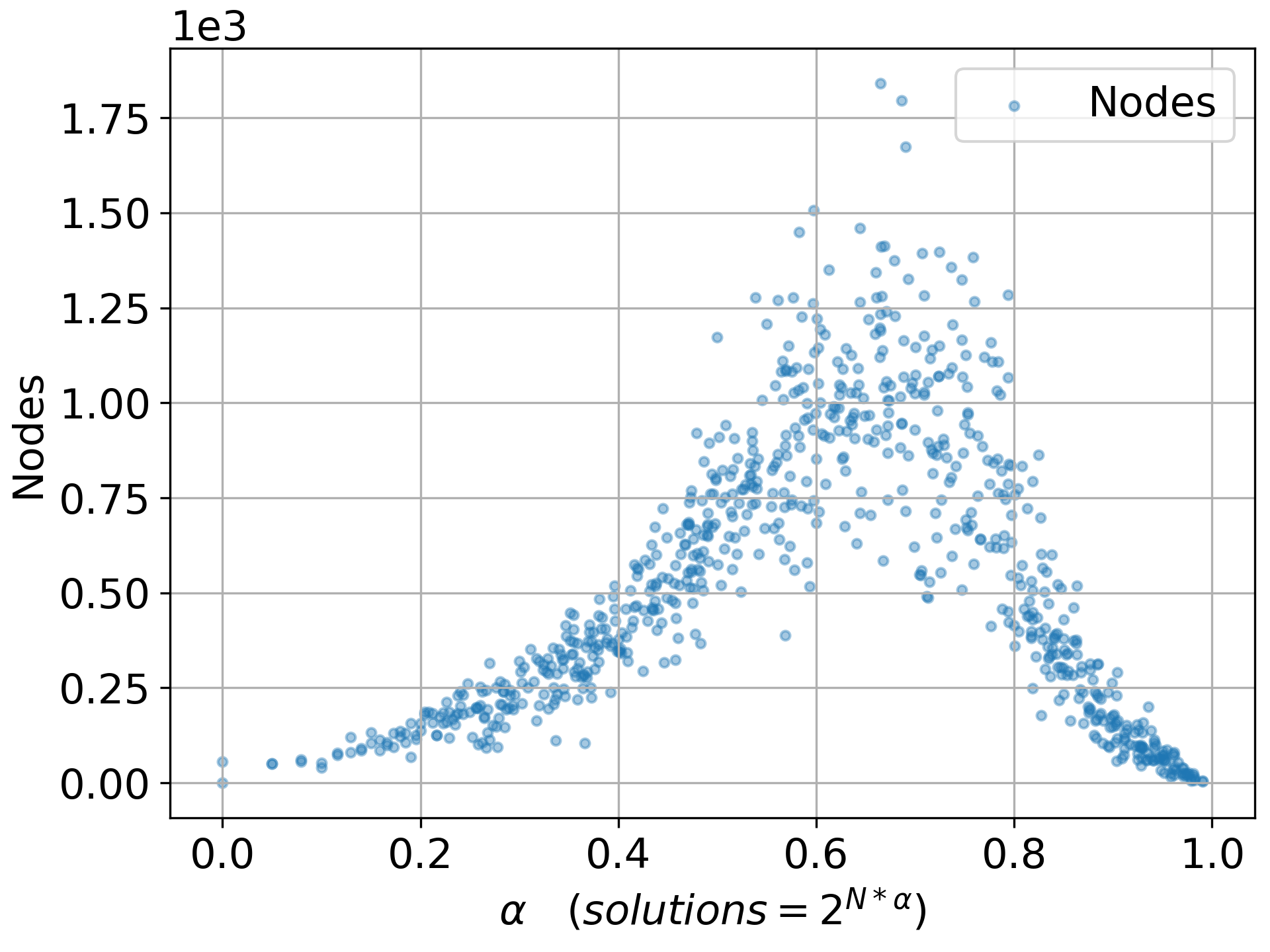}
		\caption{20 vars}
		\label{fig:appendix:sdd_nVa_3CNF_20vars}
	\end{subfigure}
	\caption{Nodes in SDD vs solution density($\alpha$) for different number of variables}
	\label{fig:appendix:sdd_nVa_3CNF_vars}
\end{figure*}

\begin{figure*}[!th]
	\centering
	\begin{subfigure}[b]{0.30\linewidth}
		\centering
		\includegraphics[width=\textwidth]{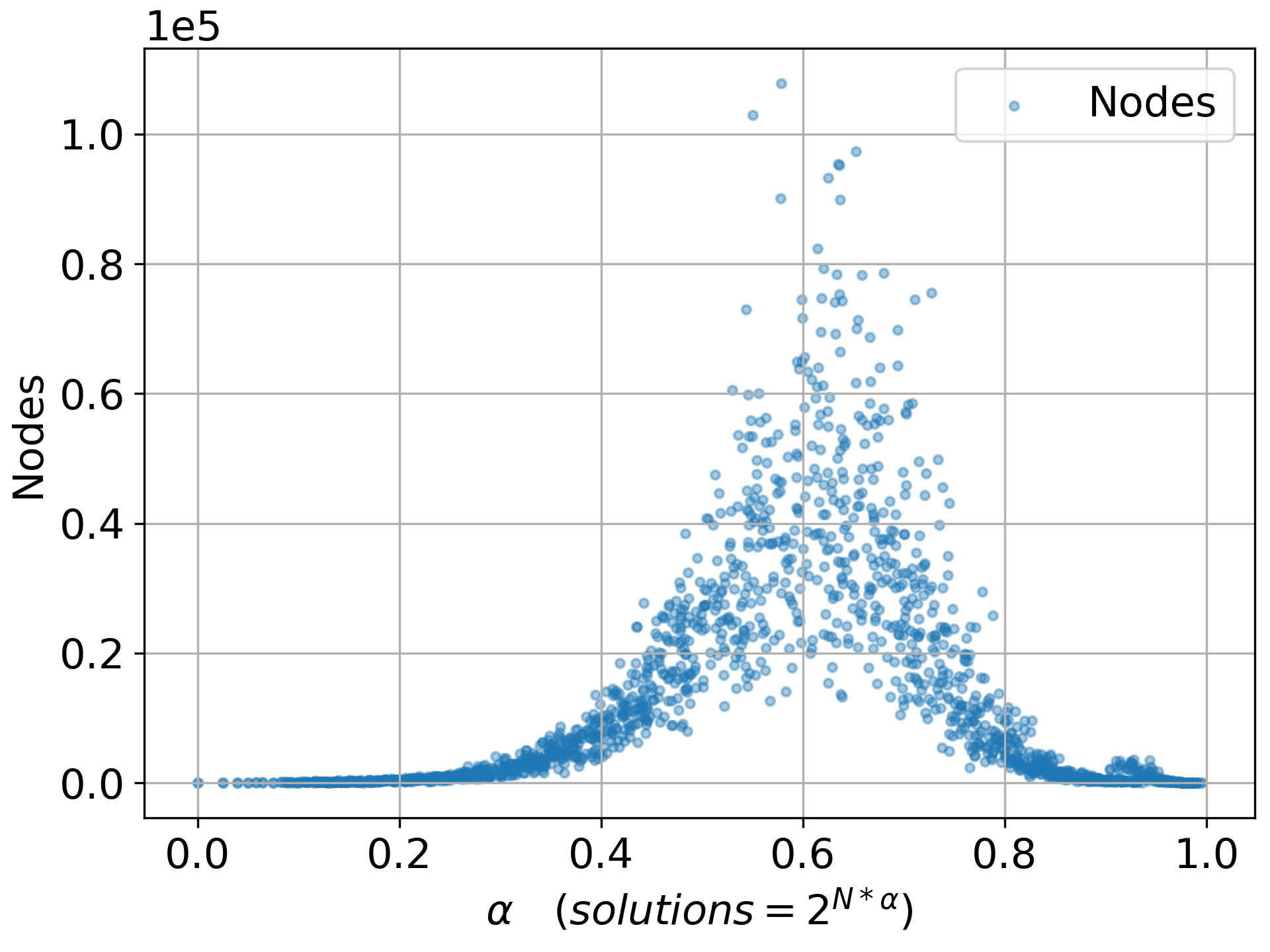}
		\caption{40 vars}
		\label{fig:appendix:CUDD_nVa_3CNF_40vars}
	\end{subfigure}
	\hfill
	\begin{subfigure}[b]{0.30\linewidth}
		\centering
		\includegraphics[width=\textwidth]{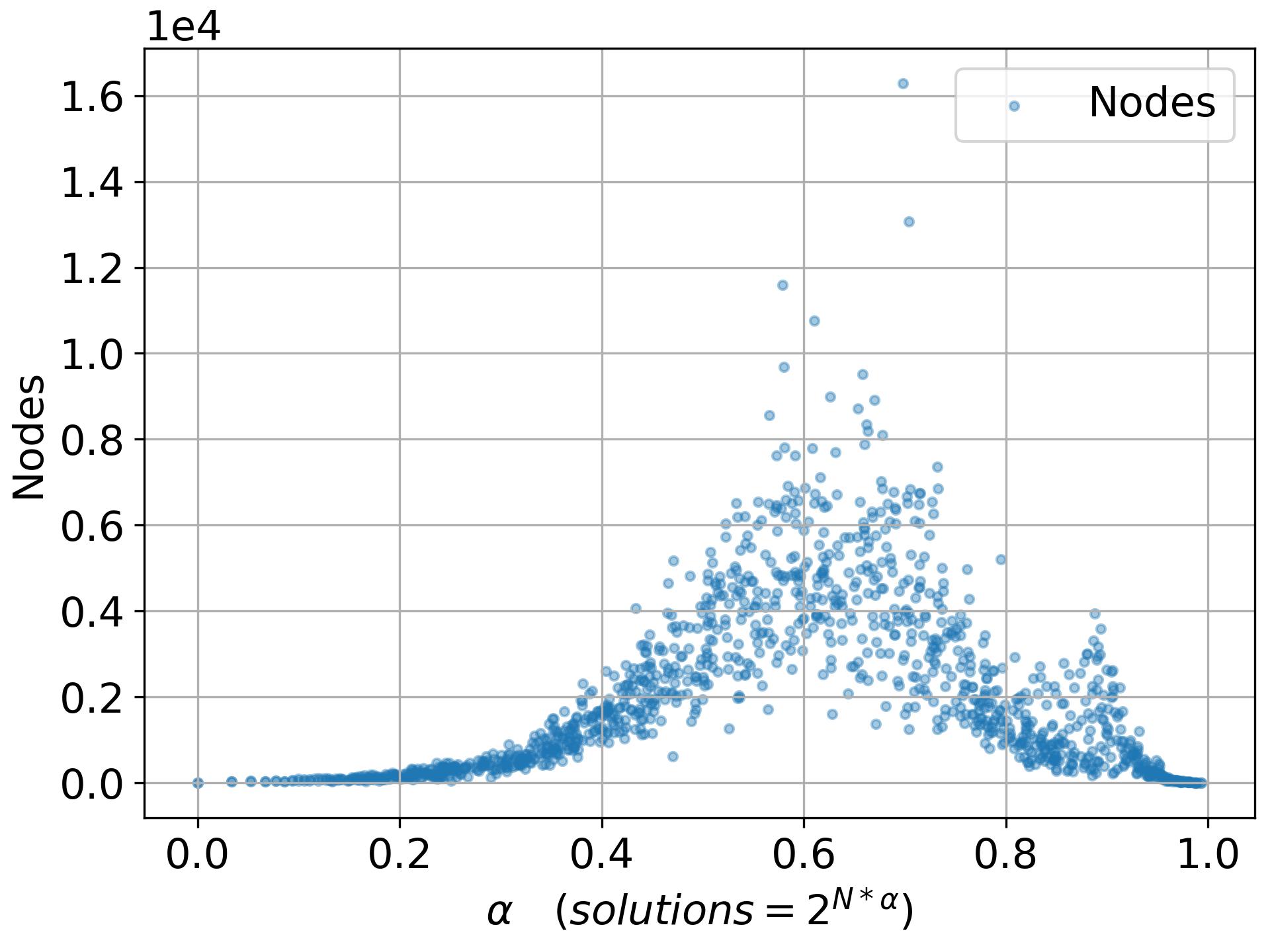}
		\caption{30 vars}
		\label{fig:appendix:CUDD_nVa_3CNF_30vars}
	\end{subfigure}
	\hfill
	\begin{subfigure}[b]{0.30\linewidth}
		\centering
		\includegraphics[width=\textwidth]{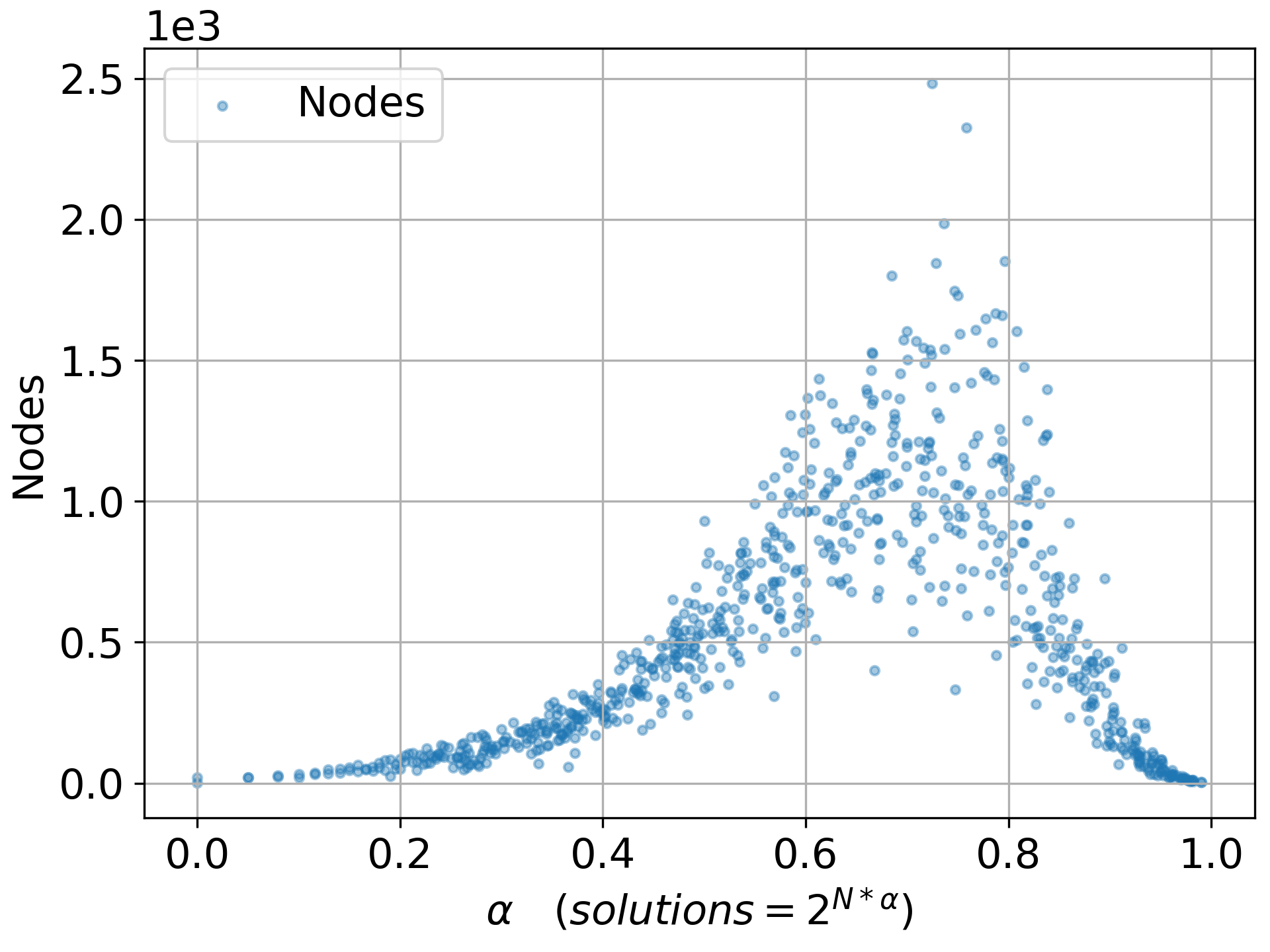}
		\caption{20 vars}
		\label{fig:appendix:CUDD_nVa_3CNF_20vars}
	\end{subfigure}
	\caption{Nodes in OBDD vs solution density($\alpha$) for different number of variables}
	\label{fig:appendix:CUDD_nVa_3CNF_vars}
\end{figure*}

Figures~\ref{fig:appendix:d4_nVa_3CNF_vars}, \ref{fig:appendix:sdd_nVa_3CNF_vars} and \ref{fig:appendix:CUDD_nVa_3CNF_vars} show the variation of nodes in d-DNNF, SDD and OBDD compilations with solution density for different number of variables.


\subsubsection{Runtime of compilation} 
\begin{figure*}[!th]
	\centering
	\begin{subfigure}[b]{0.30\linewidth}
		\centering
		\includegraphics[width=\textwidth]{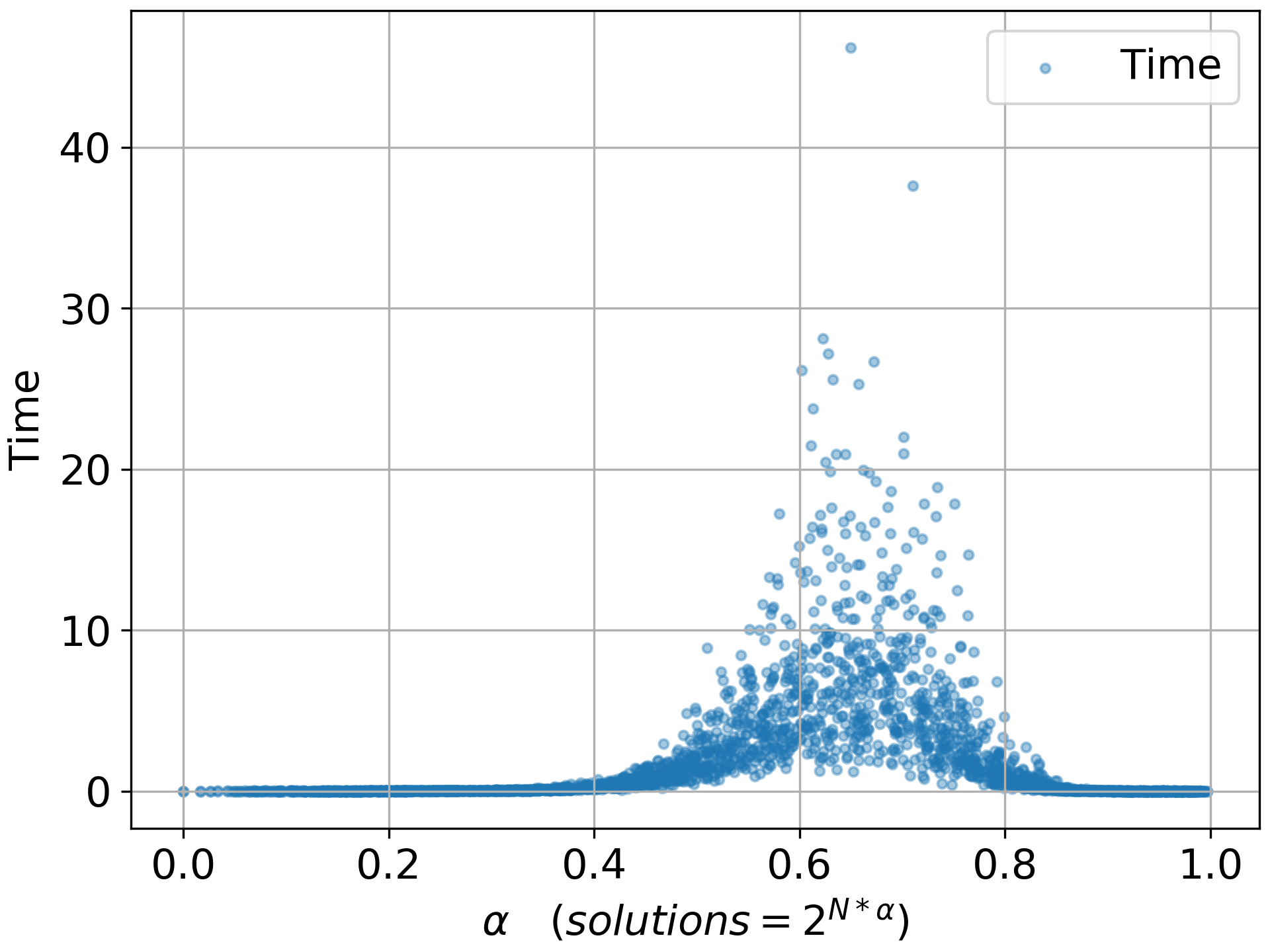}
		\caption{60 vars}
		\label{fig:appendix:d4_tVa_3CNF_60vars}
	\end{subfigure}
	\hfill
	\begin{subfigure}[b]{0.30\linewidth}
		\centering
		\includegraphics[width=\textwidth]{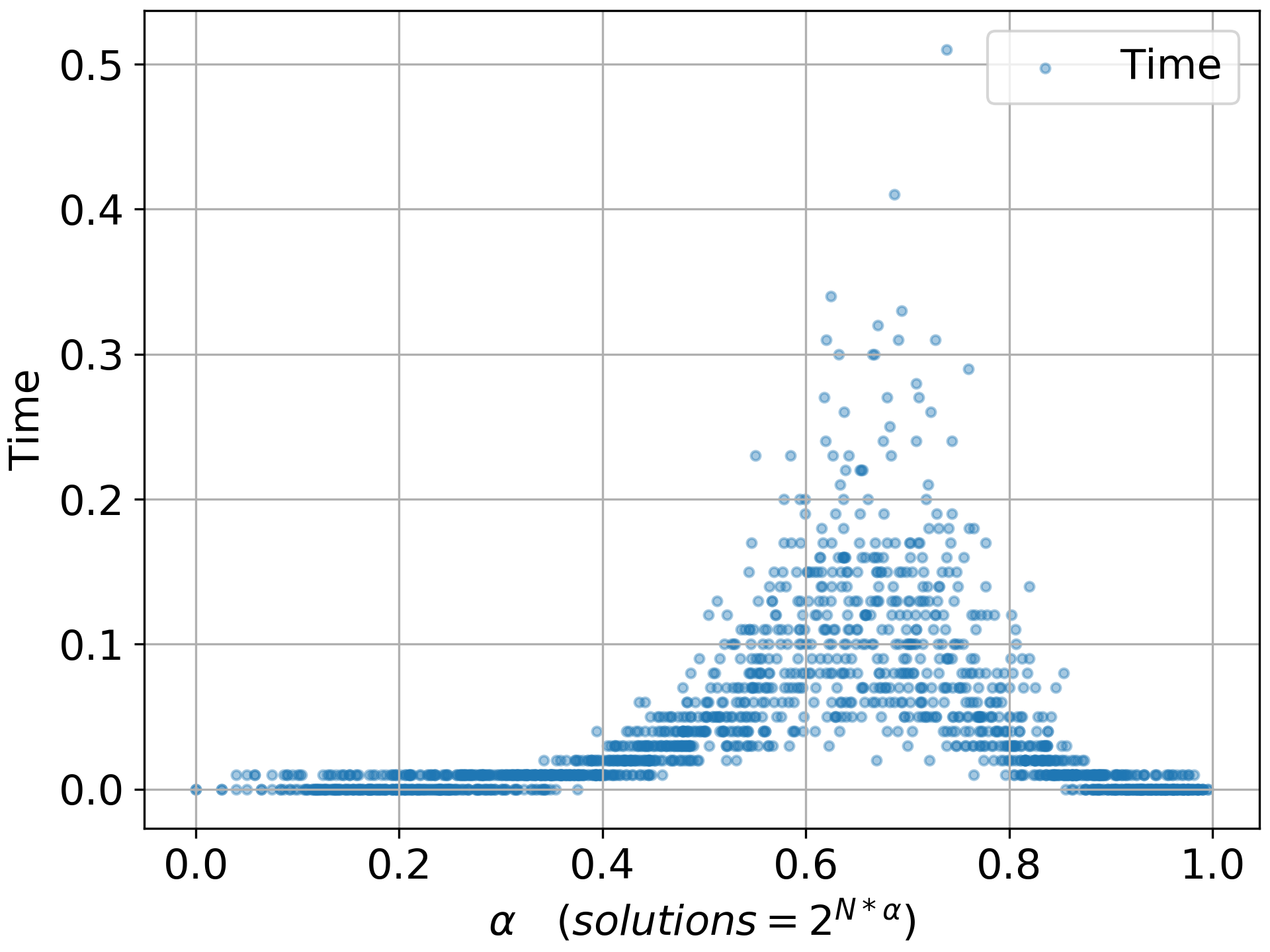}
		\caption{40 vars}
		\label{fig:appendix:d4_tVa_3CNF_40vars}
	\end{subfigure}
	\hfill
	\begin{subfigure}[b]{0.30\linewidth}
		\centering
		\includegraphics[width=\textwidth]{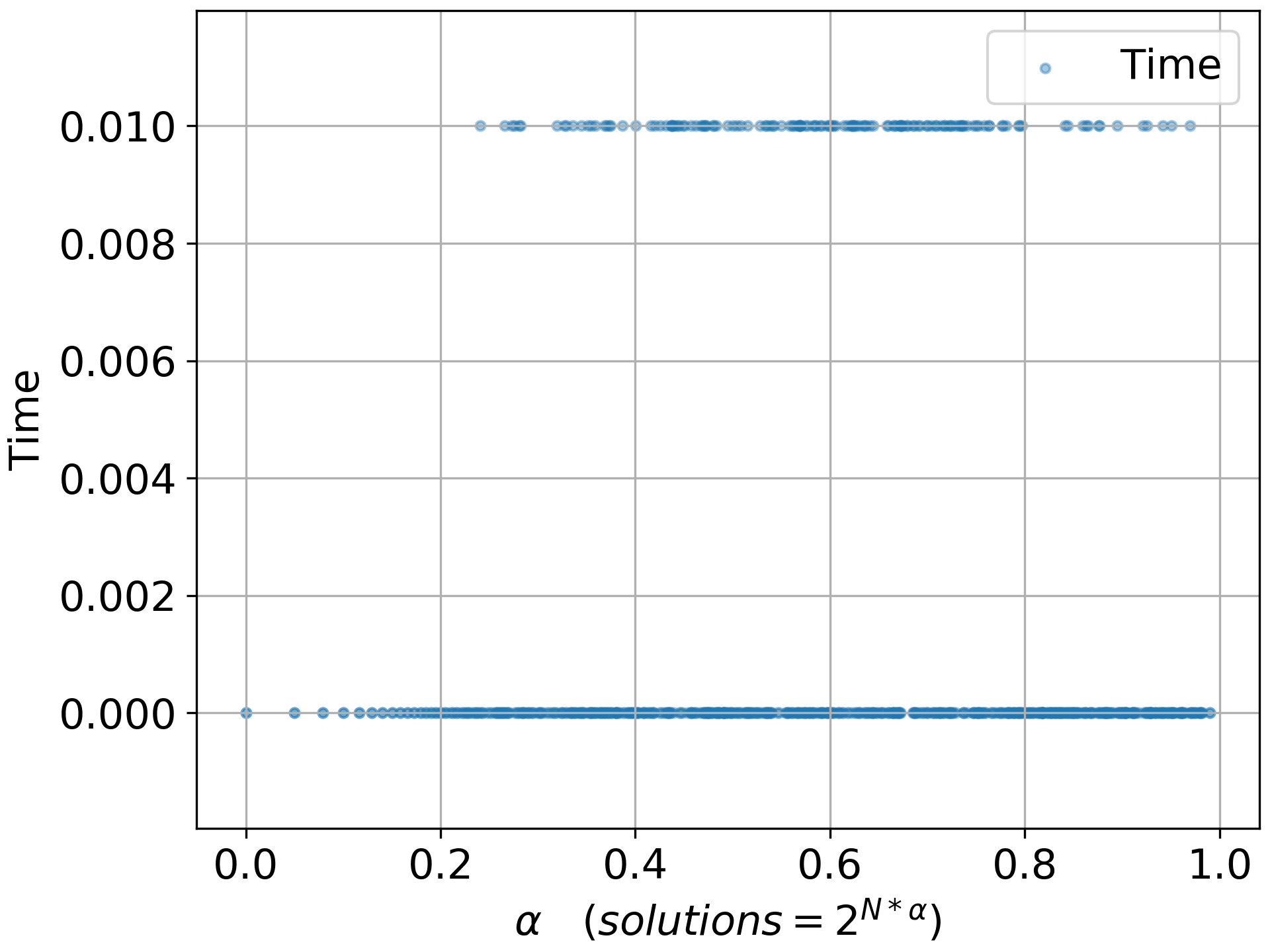}
		\caption{20 vars}
		\label{fig:appendix:d4_tVa_3CNF_20vars}
	\end{subfigure}
	\caption{Compile-time for d-DNNF vs solution density($\alpha$) for different number of variables}
	\label{fig:appendix:d4_tVa_3CNF_vars}
\end{figure*}

\begin{figure*}[!th]
	\centering
	\begin{subfigure}[b]{0.30\linewidth}
		\centering
		\includegraphics[width=\textwidth]{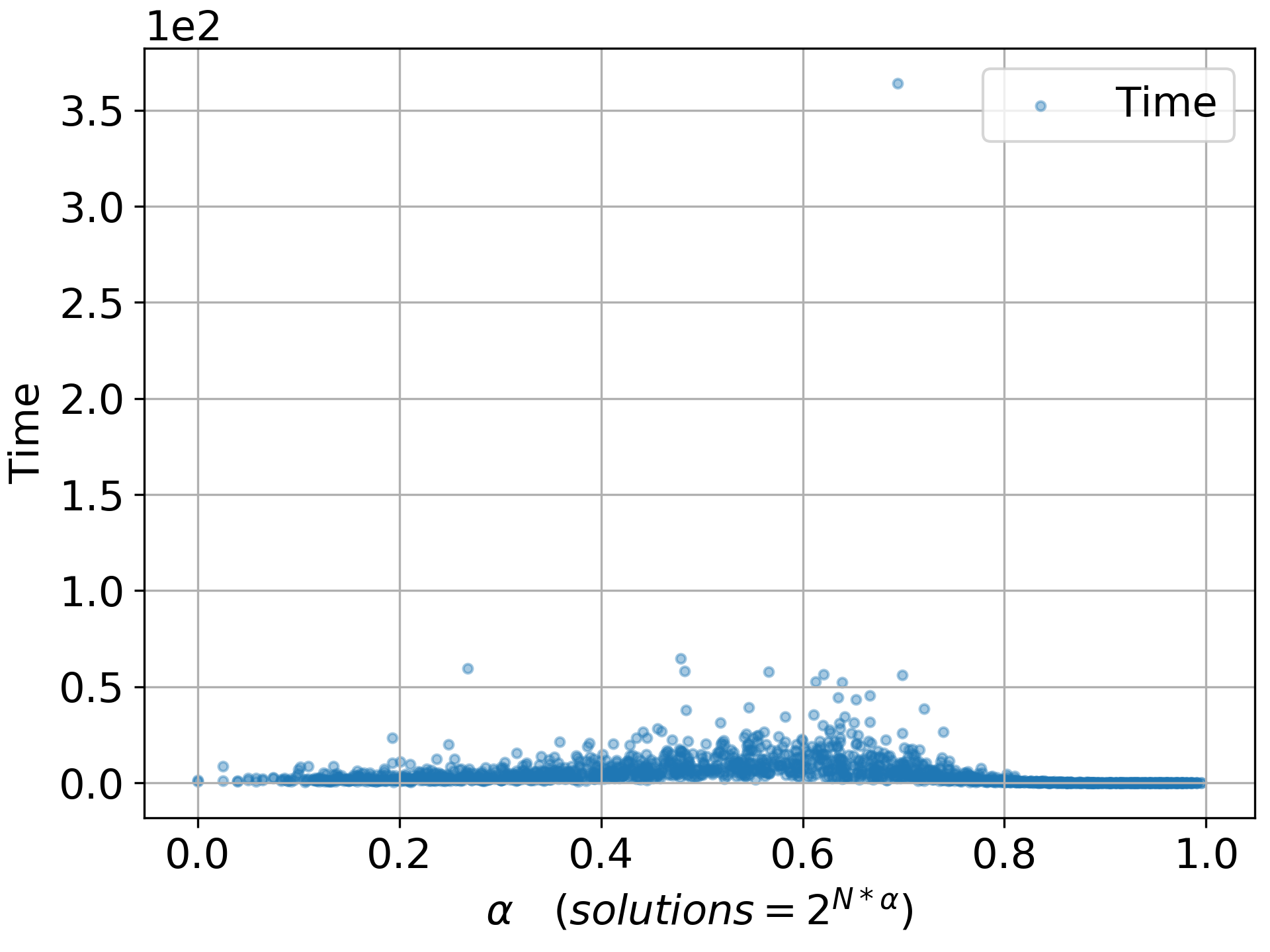}
		\caption{40 vars}
		\label{fig:appendix:sdd_tVa_3CNF_40vars}
	\end{subfigure}
	\hfill
	\begin{subfigure}[b]{0.30\linewidth}
		\centering
		\includegraphics[width=\textwidth]{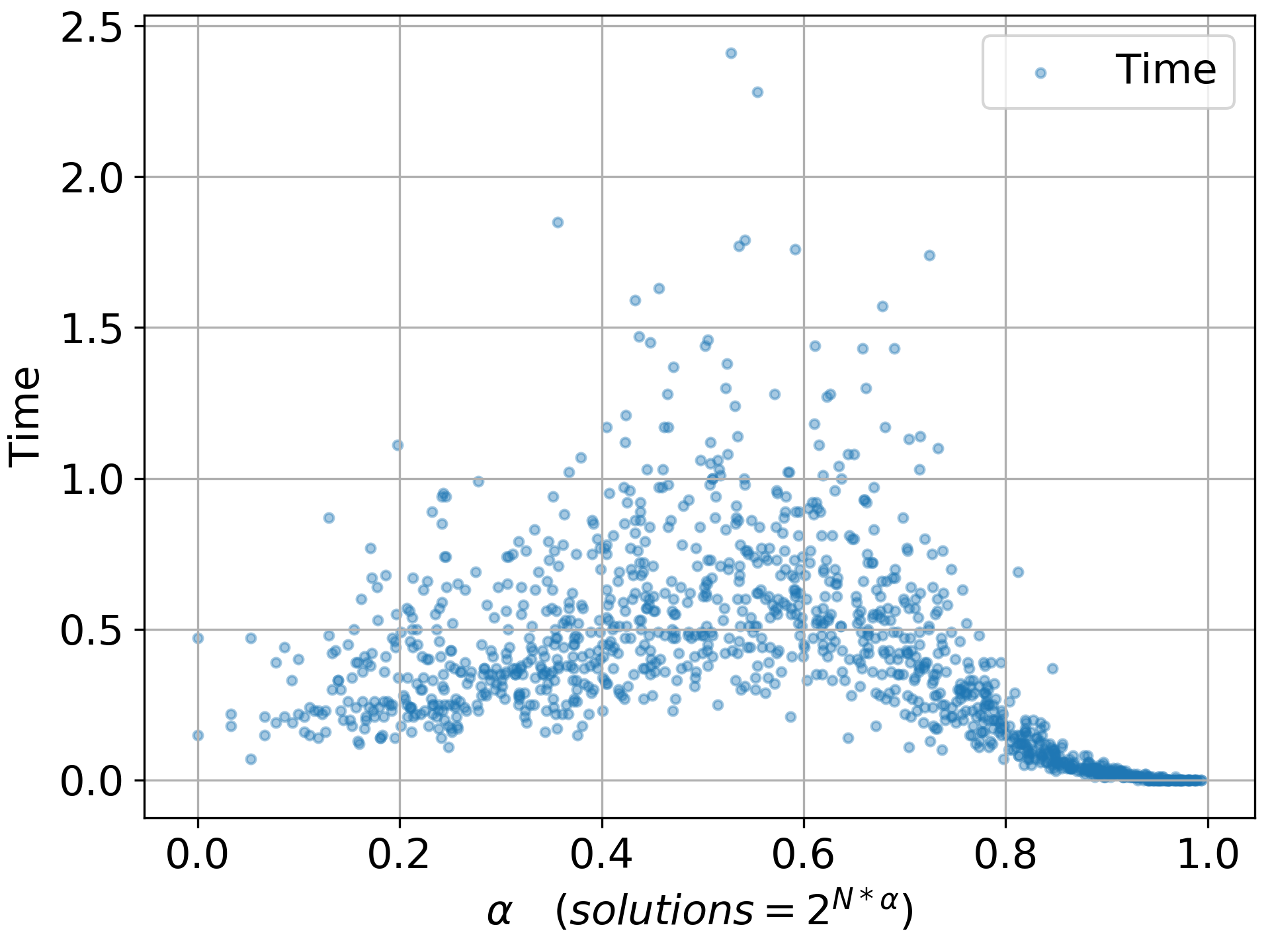}
		\caption{30 vars}
		\label{fig:appendix:sdd_tVa_3CNF_30vars}
	\end{subfigure}
	\hfill
	\begin{subfigure}[b]{0.30\linewidth}
		\centering
		\includegraphics[width=\textwidth]{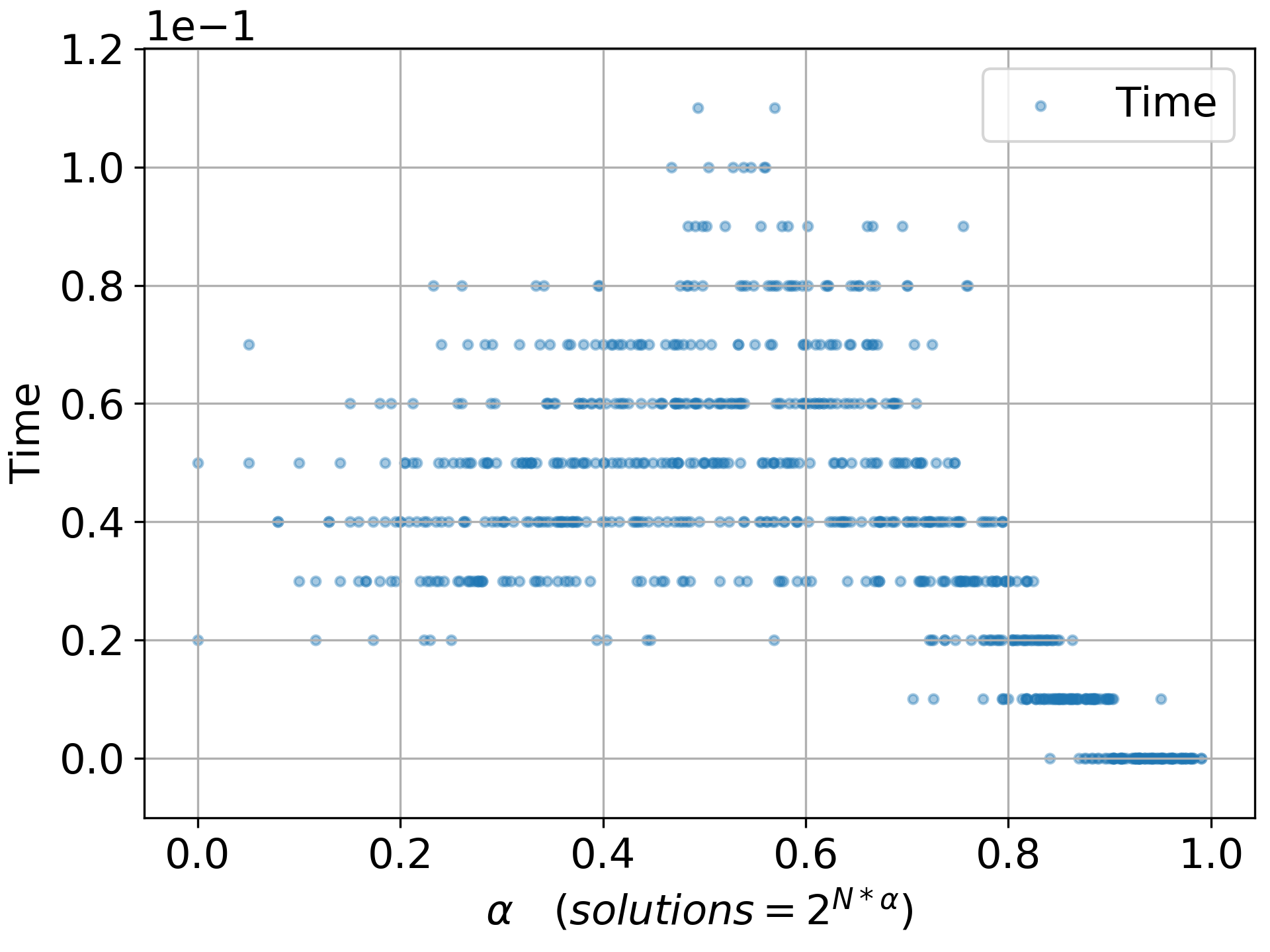}
		\caption{20 vars}
		\label{fig:appendix:sdd_tVa_3CNF_20vars}
	\end{subfigure}
	\caption{Compile-time for SDD vs solution density($\alpha$) for different number of variables}
	\label{fig:appendix:sdd_tVa_3CNF_vars}
\end{figure*}

\begin{figure*}[!th]
	\centering
	\begin{subfigure}[b]{0.30\linewidth}
		\centering
		\includegraphics[width=\textwidth]{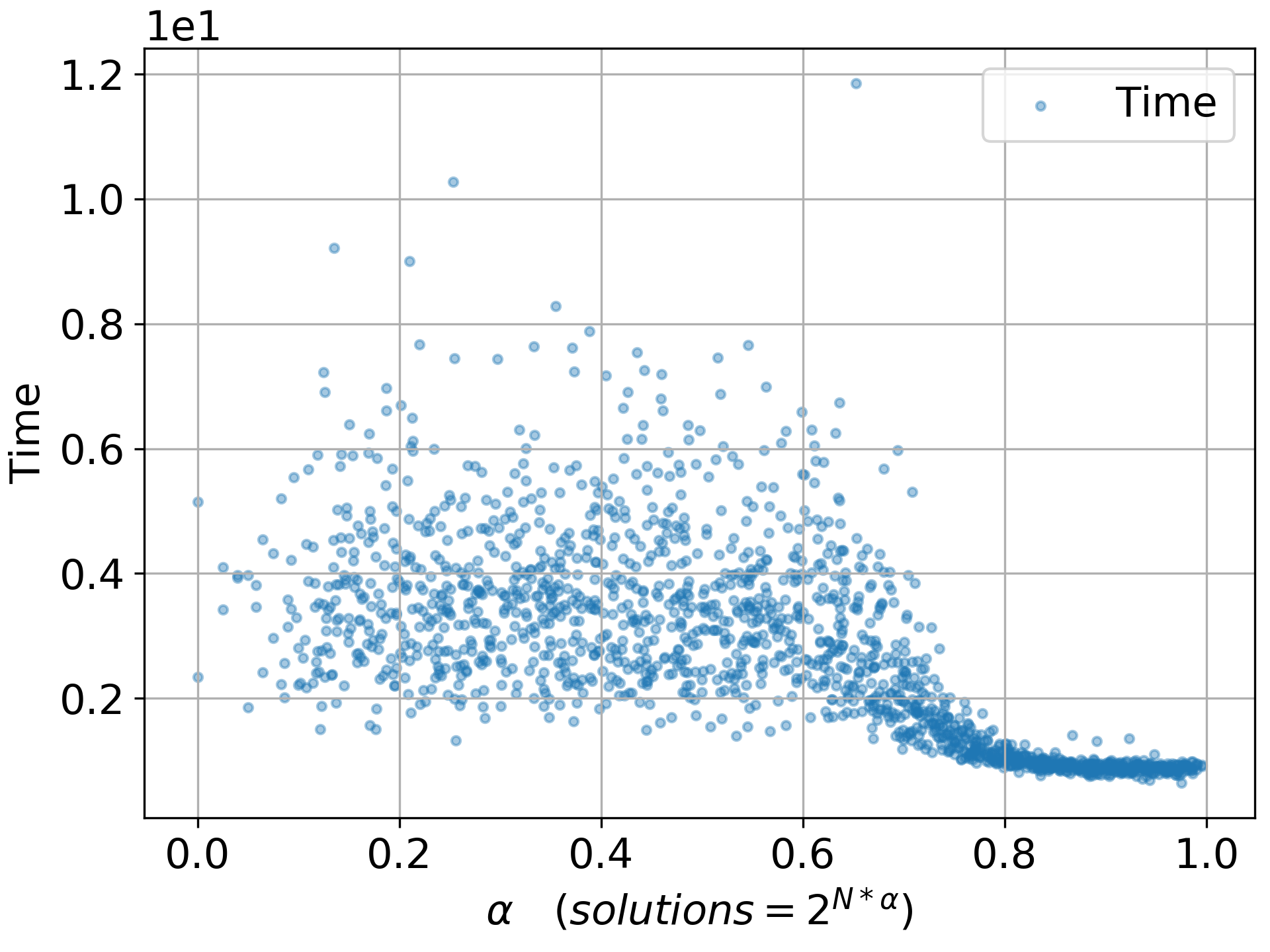}
		\caption{40 vars}
		\label{fig:appendix:CUDD_tVa_3CNF_40vars}
	\end{subfigure}
	\hfill
	\begin{subfigure}[b]{0.30\linewidth}
		\centering
		\includegraphics[width=\textwidth]{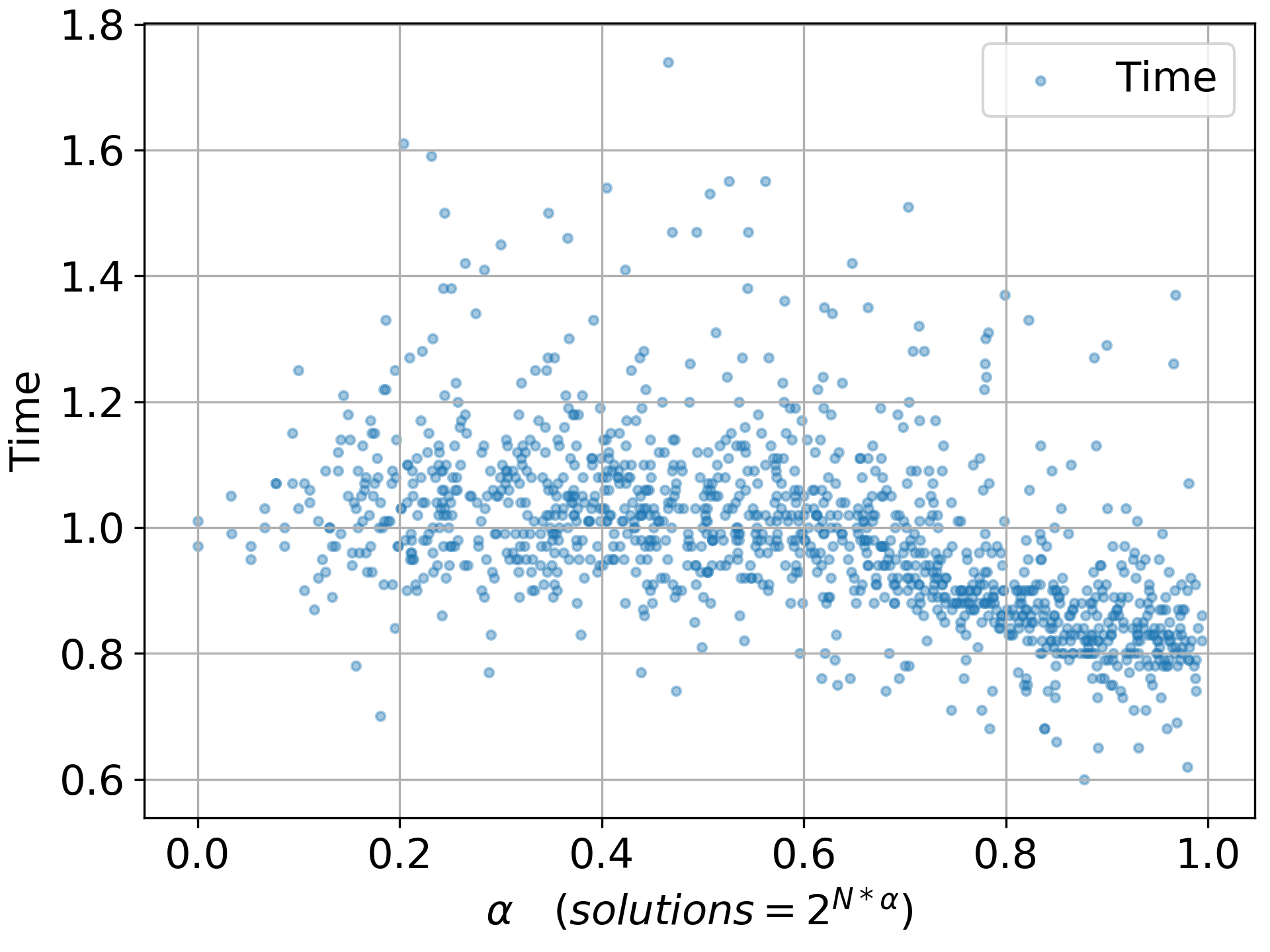}
		\caption{30 vars}
		\label{fig:appendix:CUDD_tVa_3CNF_30vars}
	\end{subfigure}
	\hfill
	\begin{subfigure}[b]{0.30\linewidth}
		\centering
		\includegraphics[width=\textwidth]{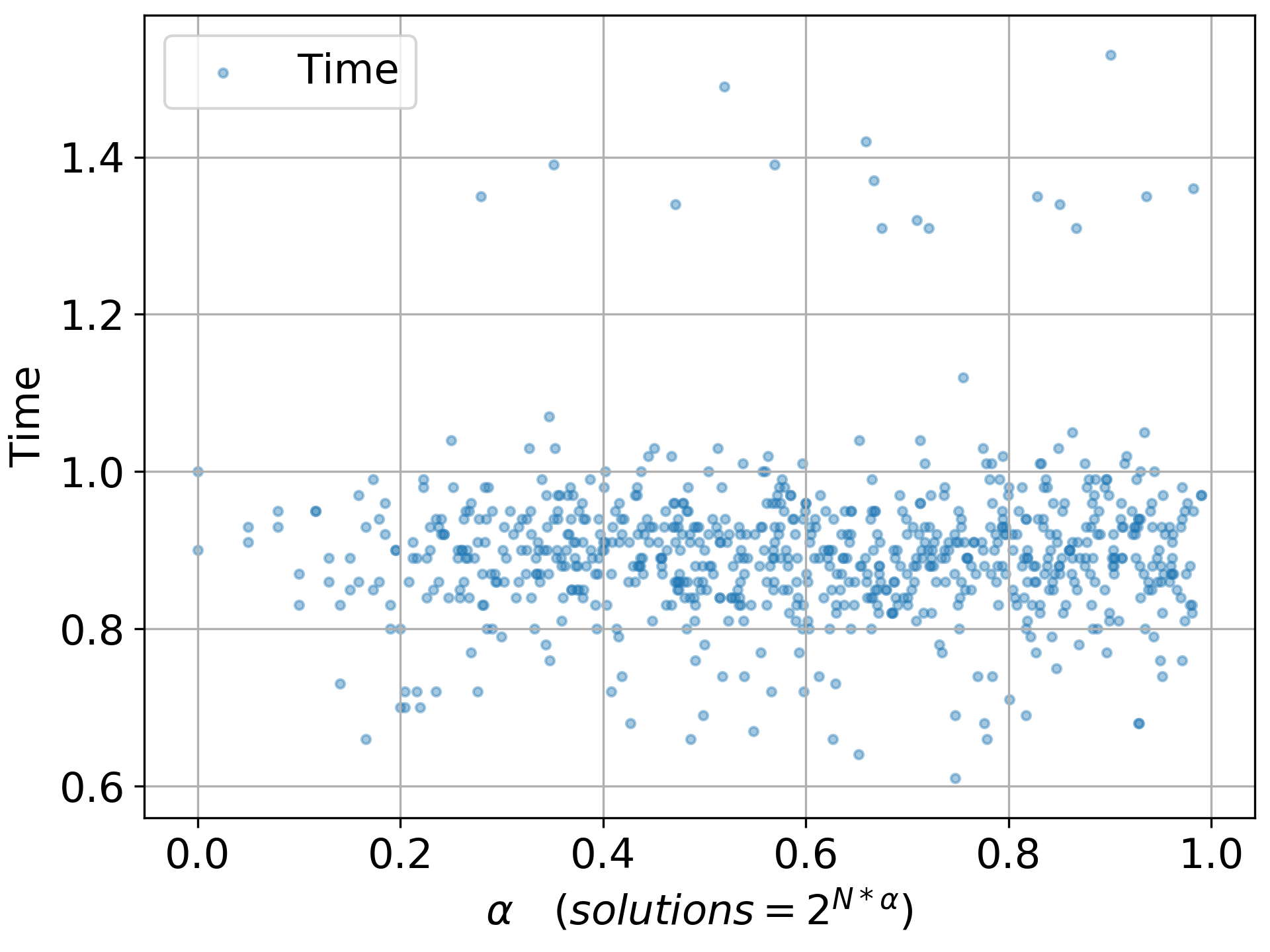}
		\caption{20 vars}
		\label{fig:appendix:CUDD_tVa_3CNF_20vars}
	\end{subfigure}
	\caption{Compile-time for OBDD vs solution density($\alpha$) for different number of variables}
	\label{fig:appendix:CUDD_tVa_3CNF_vars}
\end{figure*}

Figures~\ref{fig:appendix:d4_tVa_3CNF_vars}, \ref{fig:appendix:sdd_tVa_3CNF_vars} and \ref{fig:appendix:CUDD_tVa_3CNF_vars} show the distribution in runtimes of compilation for d-DNNFs, SDDs and OBDDs with different number of variables.

\begin{figure*}[!th]
	\centering
	\begin{subfigure}[b]{0.30\linewidth}
		\centering
		\includegraphics[width=\linewidth]{MainFigures/dDNNF/d4act_10inst_timesValpha_3cnf70vars.png}
		\caption{d-DNNF, 70 vars}
		\label{fig:appendix:d4_tVa_3CNF_70vars}
	\end{subfigure}
	\hfill
	\begin{subfigure}[b]{0.30\linewidth}
		\centering
		\includegraphics[width=\linewidth]{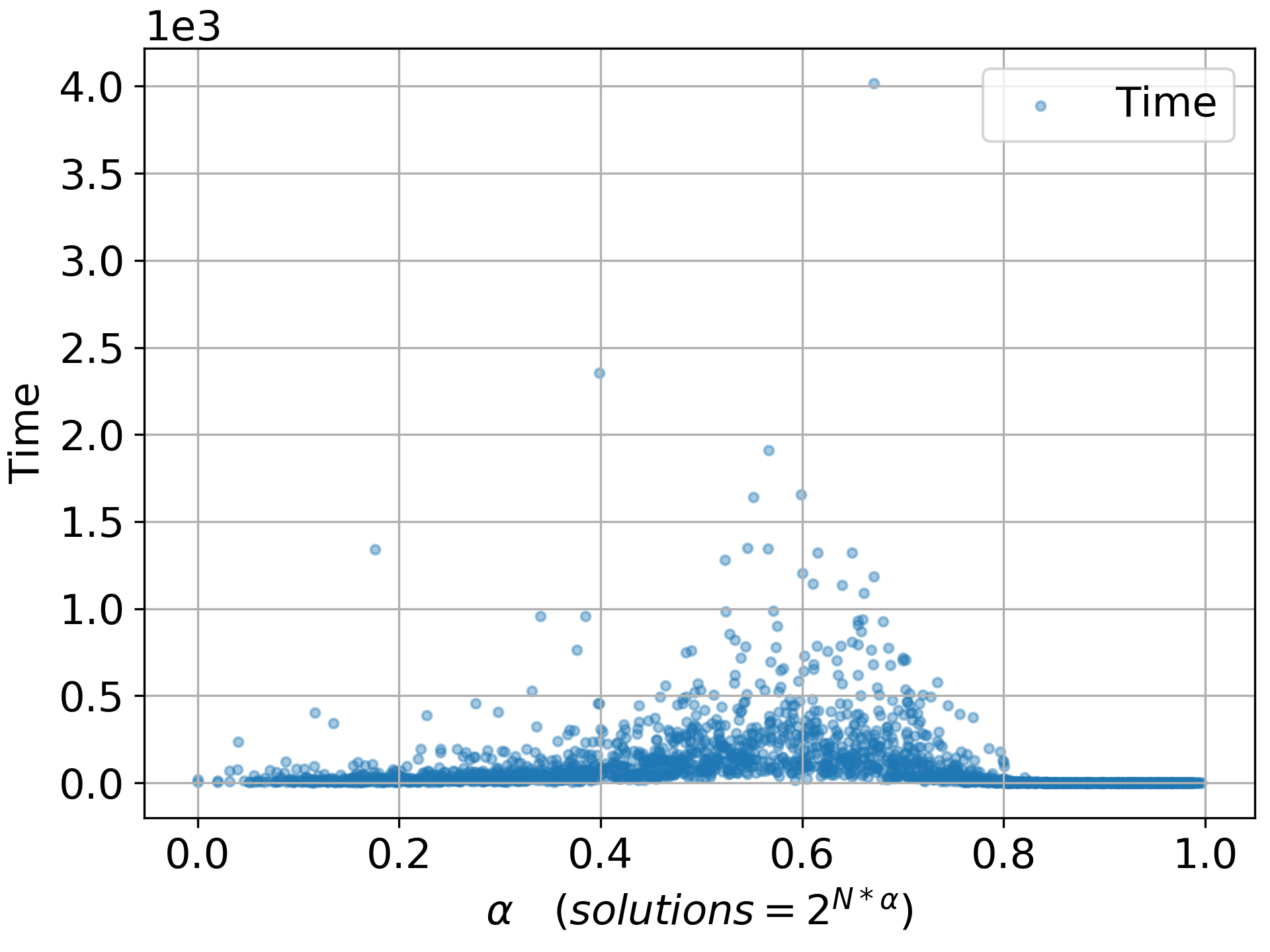}
		\caption{SDD, 50 vars}
		\label{fig:appendix:sdd_tVa_50vars_3cnf}
	\end{subfigure}
	\hfill
	\begin{subfigure}[b]{0.30\linewidth}
		\centering
		\includegraphics[width=\linewidth]{MainFigures/bdd/CUDD_SIFTreordering_10inst_timesValpha_3cnf50vars.png}
		\caption{OBDD, 50 vars}
		\label{fig:appendix:CUDD_tVa_3CNF_50vars}
	\end{subfigure}
	\caption{Compile-time for individual instances of 3-CNF against solution density}
	\label{fig:appendix:tVa_3CNF}
\end{figure*}


\clearpage
\subsection{Impact of clause length on phase transition} 
\begin{figure*}[!th]
	\centering
	\begin{subfigure}[b]{0.30\linewidth}
		\centering
		\includegraphics[width=\textwidth]{MainFigures/dDNNF/d4act_10inst_nodesValpha_2cnf600vars.png}
		\caption{2-CNF and 600 vars}

	\end{subfigure}
	\hfill
	\begin{subfigure}[b]{0.30\linewidth}
		\centering
		\includegraphics[width=\textwidth]{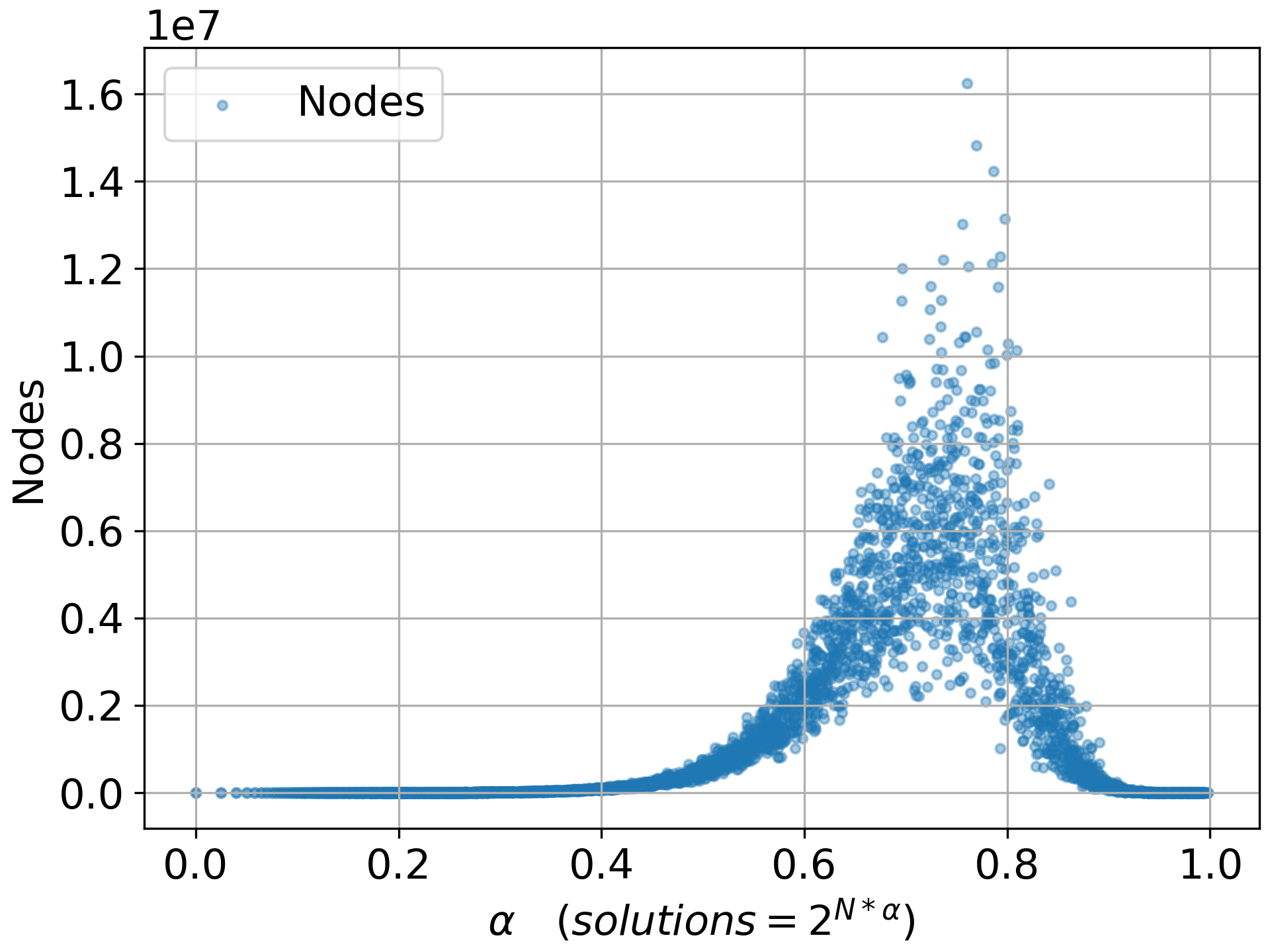}
		\caption{4-CNF and 40 vars}

	\end{subfigure}
	\hfill
	\begin{subfigure}[b]{0.30\linewidth}
		\centering
		\includegraphics[width=\textwidth]{MainFigures/dDNNF/d4act_10inst_nodesValpha_7cnf25vars.png}
		\caption{7-CNF and 25 vars}
	\end{subfigure}	
	\caption{\label{fig:appendix:d4_nVa_CNF}Nodes in d-DNNF vs solution density for different clause lengths($k$)}
\end{figure*}

\begin{figure*}[!th]
	\centering
	\begin{subfigure}[b]{0.32\textwidth}
		\centering
		\includegraphics[width=\textwidth]{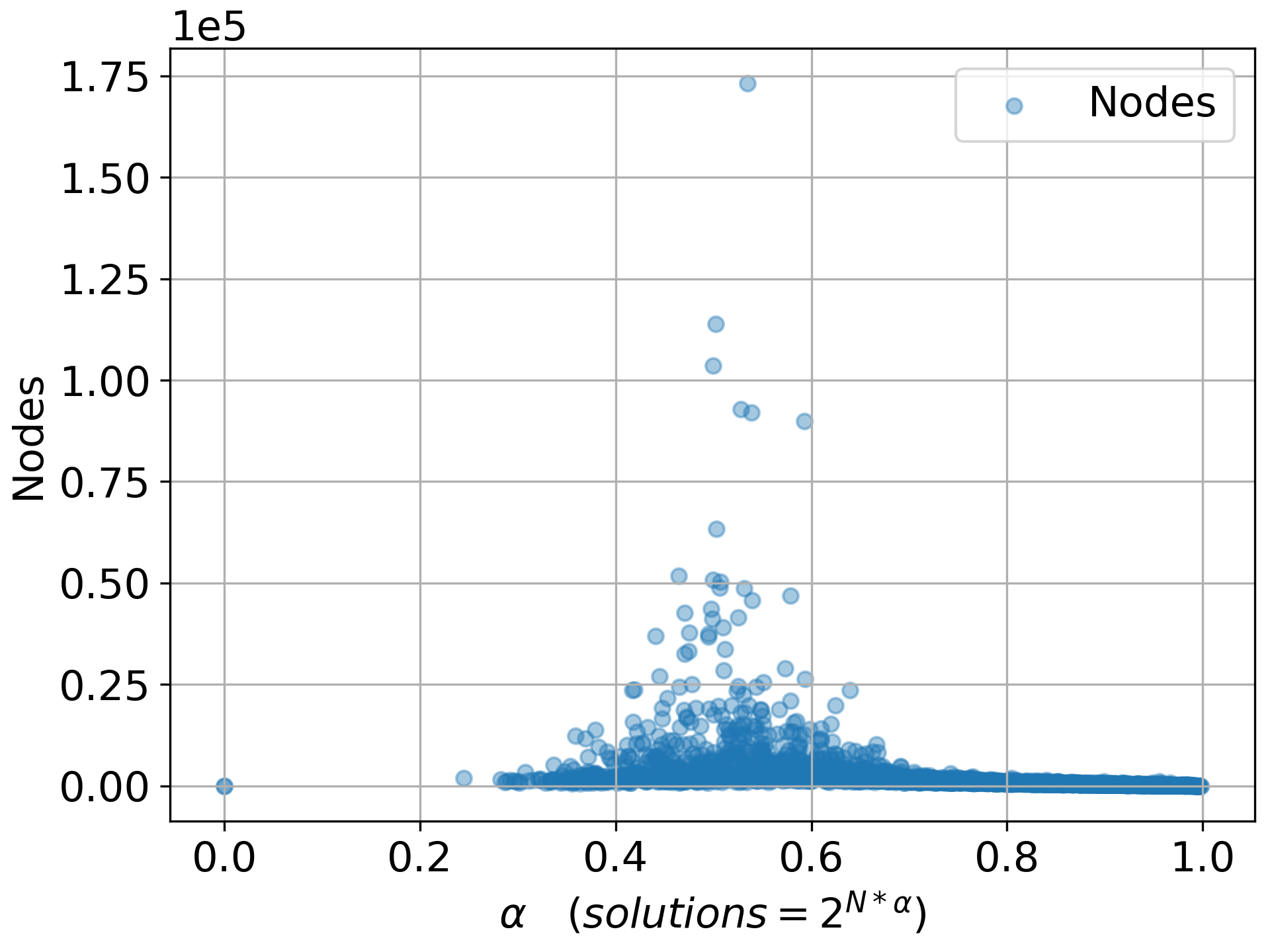}
		\caption{2-CNF and 200 vars}

	\end{subfigure}
	\hfill
	\begin{subfigure}[b]{0.32\textwidth}
		\centering
		\includegraphics[width=\textwidth]{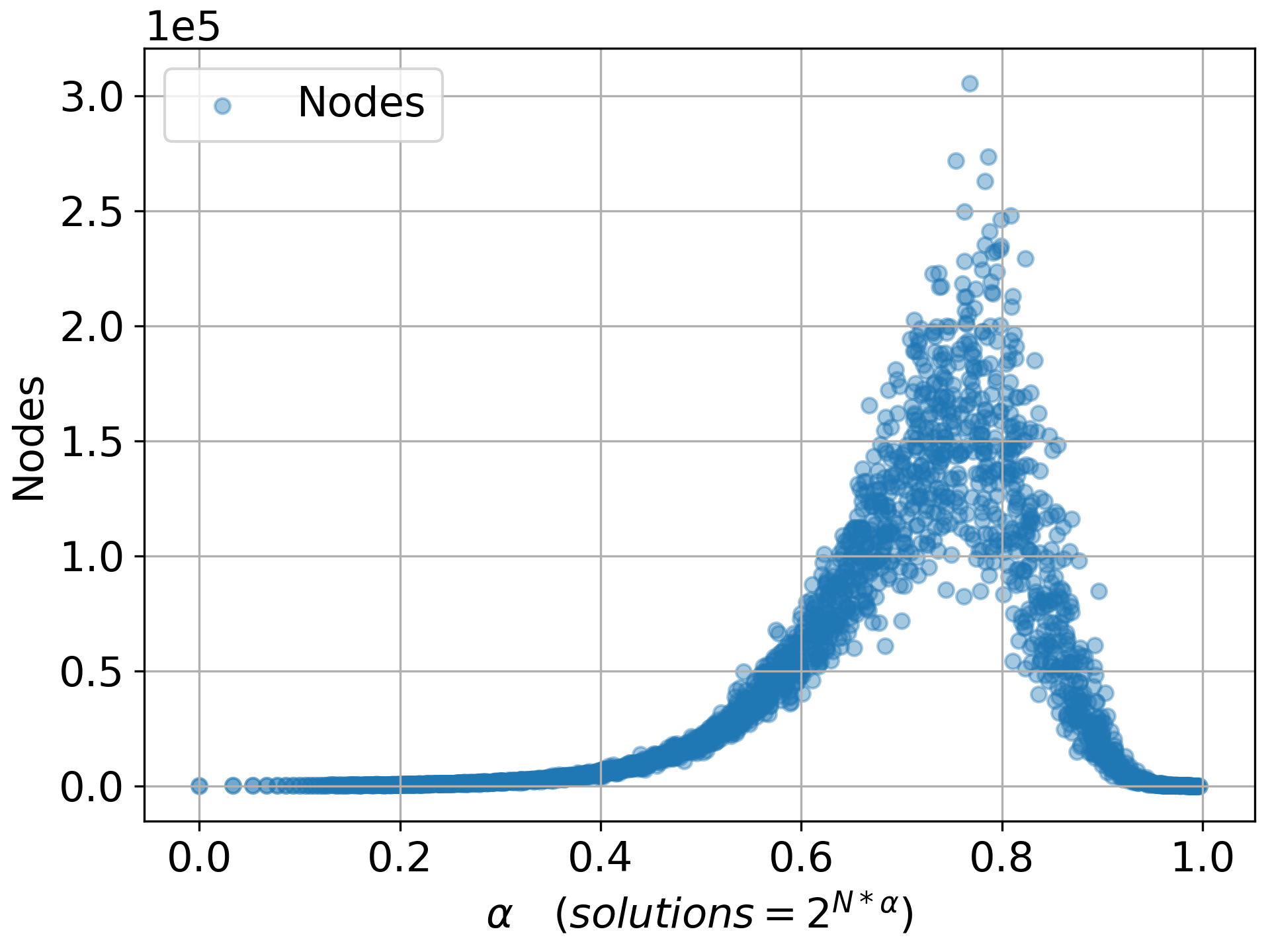}
		\caption{4-CNF and 30 vars}

	\end{subfigure}
	\hfill
	\begin{subfigure}[b]{0.32\textwidth}
		\centering
		\includegraphics[width=\textwidth]{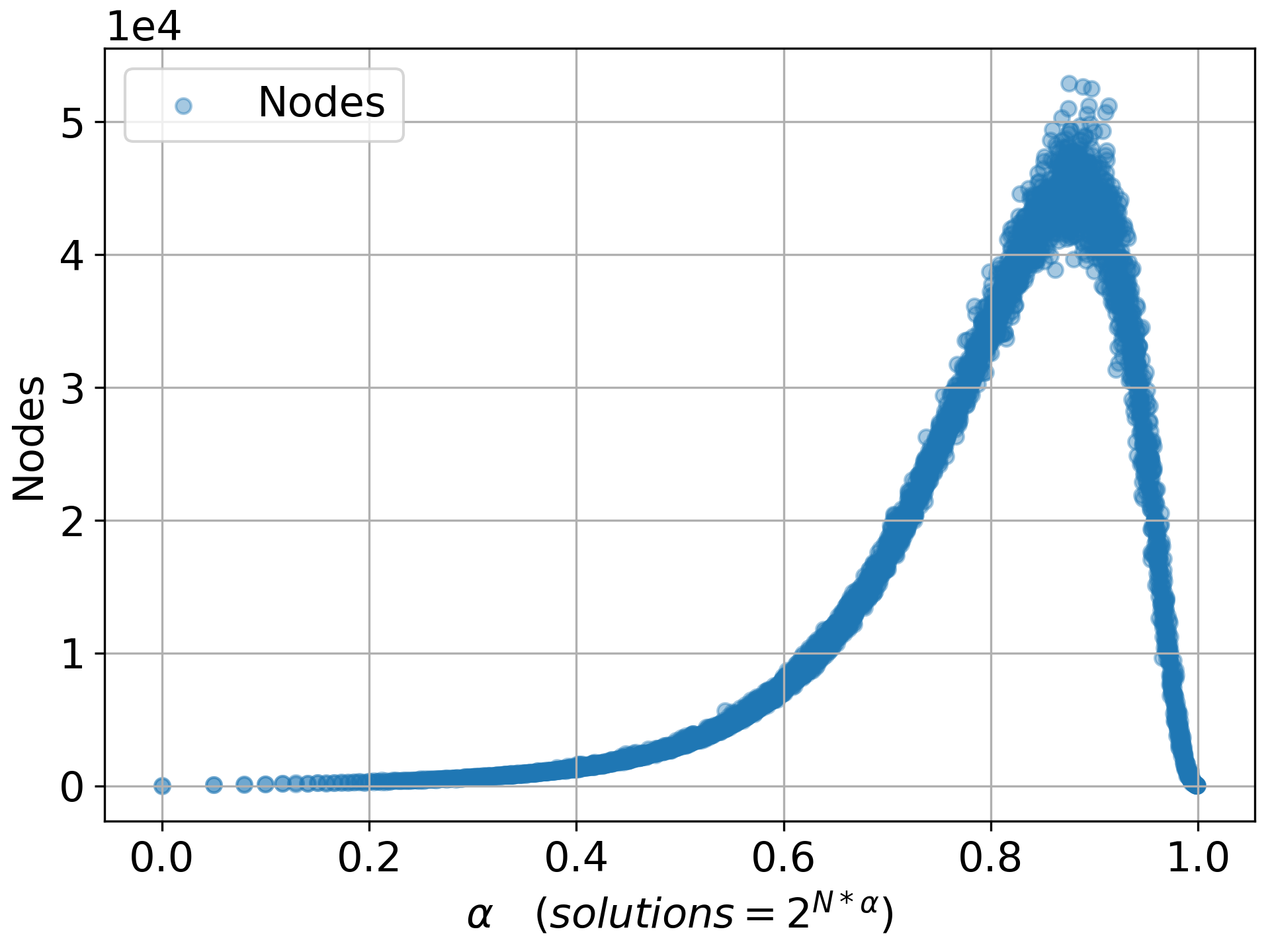}
		\caption{7-CNF and 20 vars}

	\end{subfigure}	
	\caption{\label{fig:appendix:sdd_nVa_CNF}Nodes in SDD vs solution density for different clause lengths($k$)}
\end{figure*}

\begin{figure*}[t]
	\centering
	\begin{subfigure}[b]{0.32\textwidth}
		\centering
		\includegraphics[width=\textwidth]{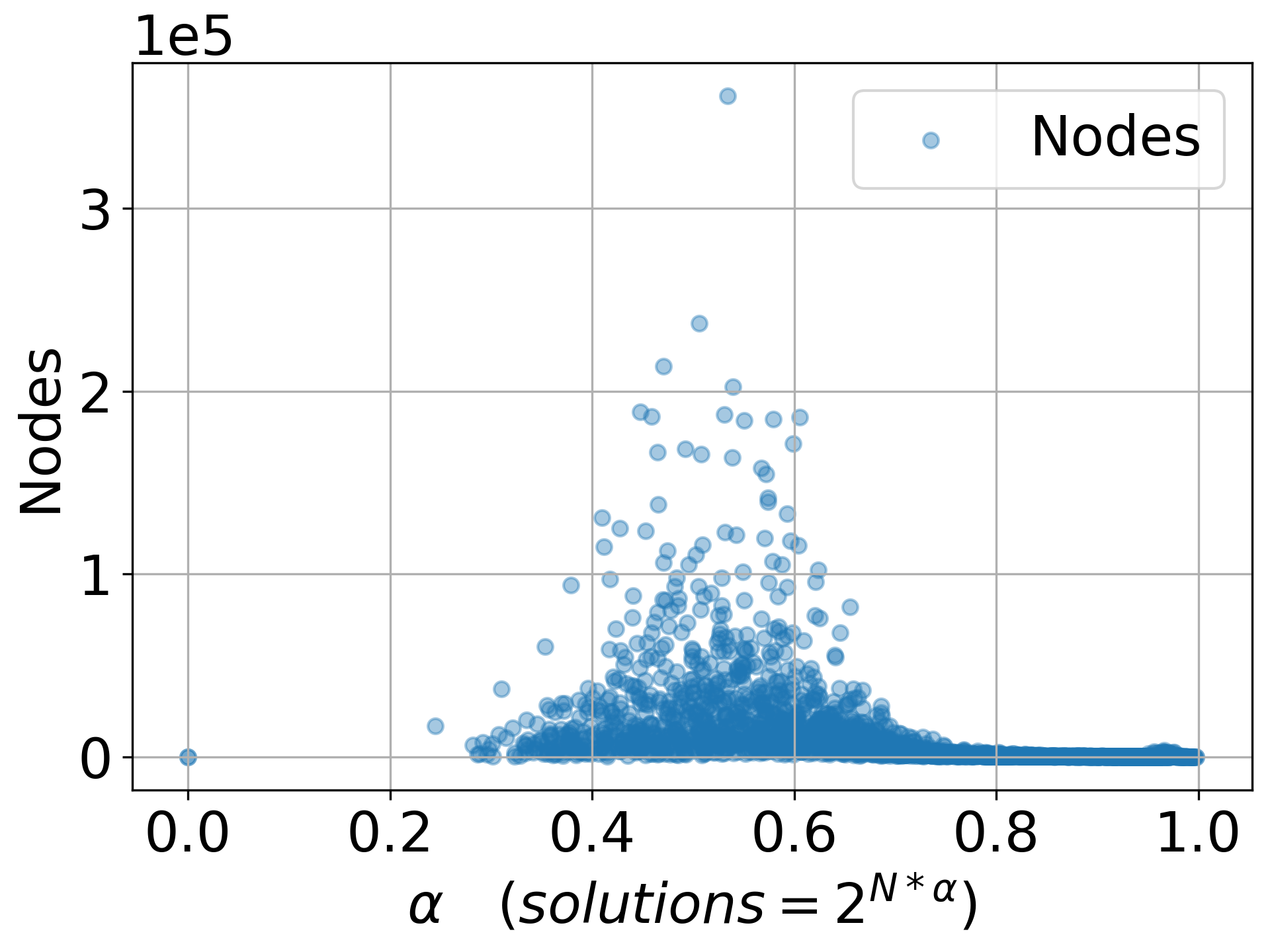}
		\caption{2-CNF and 200 vars}
	\end{subfigure}
	\hfill
	\begin{subfigure}[b]{0.32\textwidth}
		\centering
		\includegraphics[width=\textwidth]{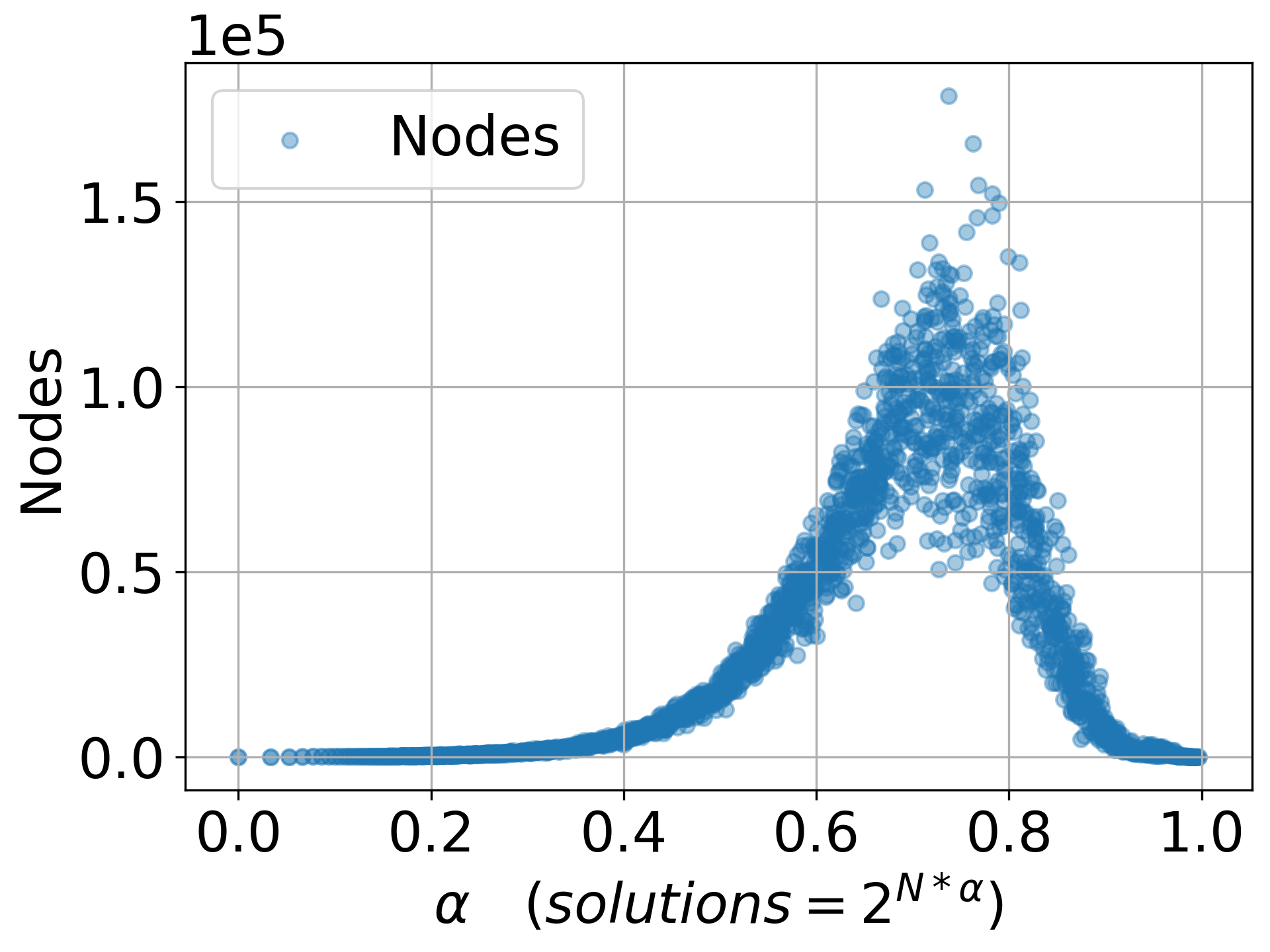}
		\caption{4-CNF and 40 vars}
	\end{subfigure}
	\hfill
	\begin{subfigure}[b]{0.32\textwidth}
		\centering
		\includegraphics[width=\textwidth]{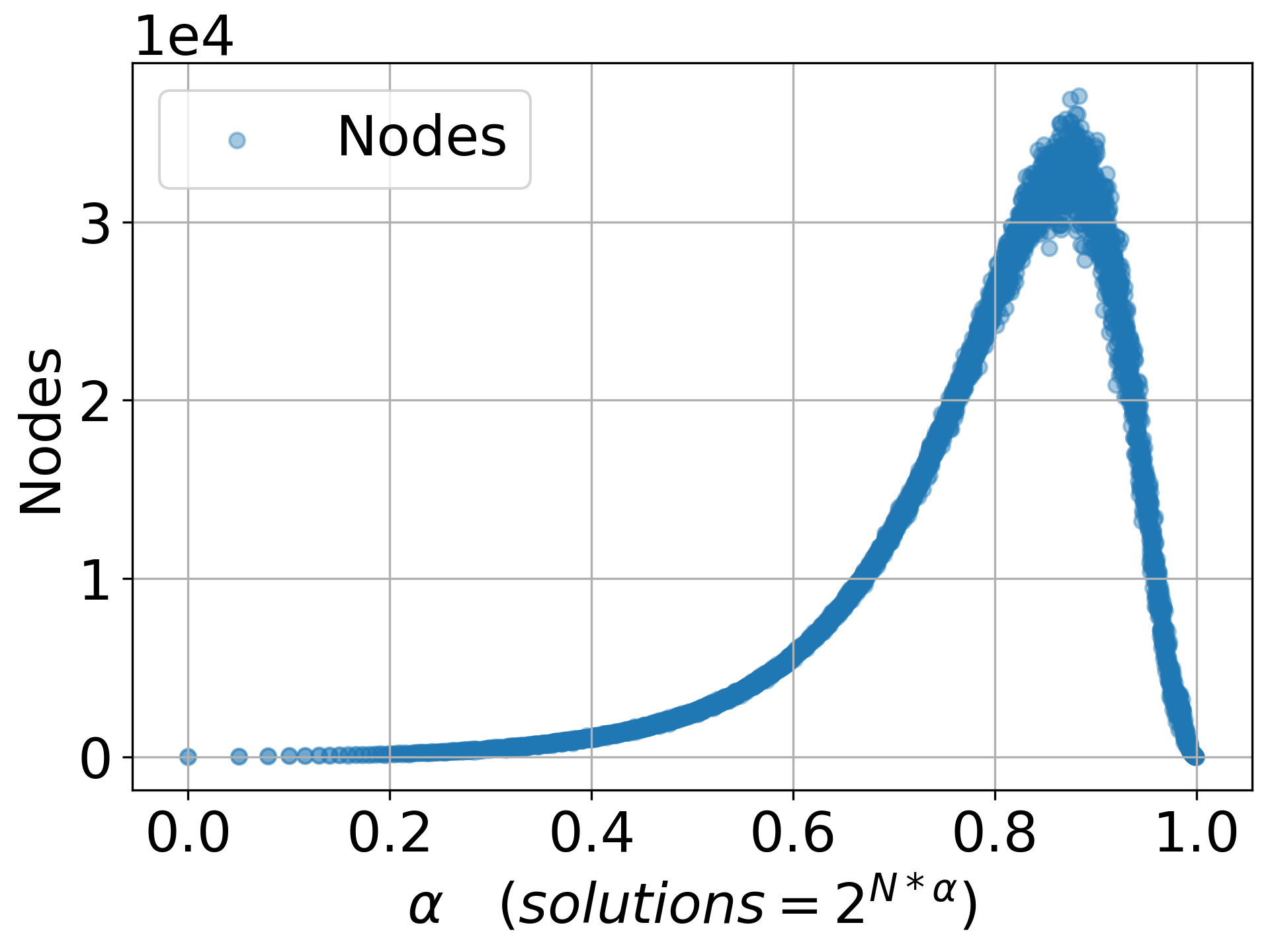}
		\caption{7-CNF and 20 vars}
	\end{subfigure}	
	\caption{\label{fig:appendix:CUDD_nVa_CNF}Nodes in OBDD vs solution density for different clause lengths($k$)}
\end{figure*}

\begin{figure*}[!th]
	\centering
	\begin{subfigure}[b]{0.30\linewidth}
		\centering
		\includegraphics[width=\textwidth]{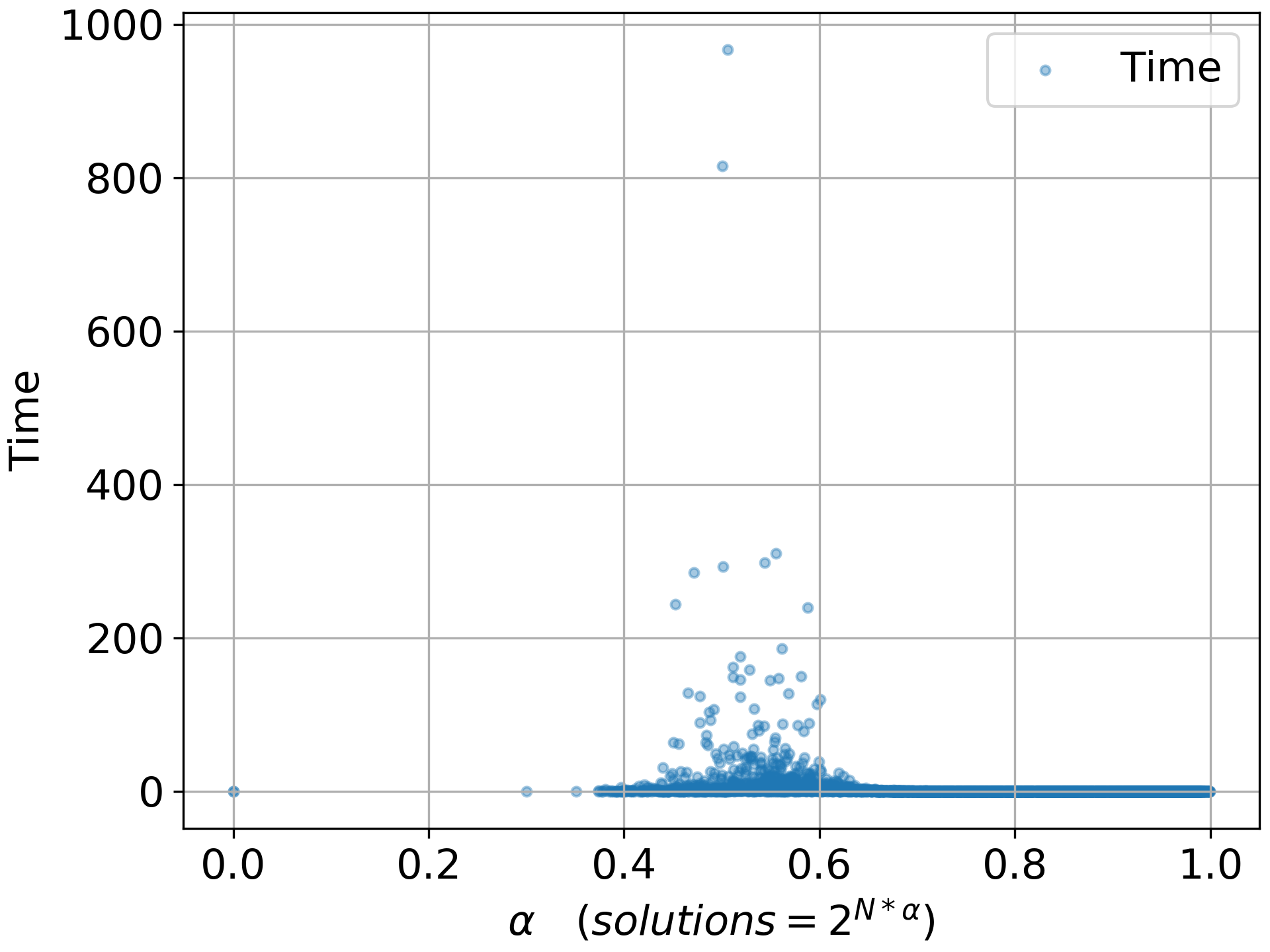}
		\caption{2-CNF and 600 vars}
		\label{fig:appendix:d4_nVa_2CNF_600vars}
	\end{subfigure}
	\hfill
	\begin{subfigure}[b]{0.30\linewidth}
		\centering
		\includegraphics[width=\textwidth]{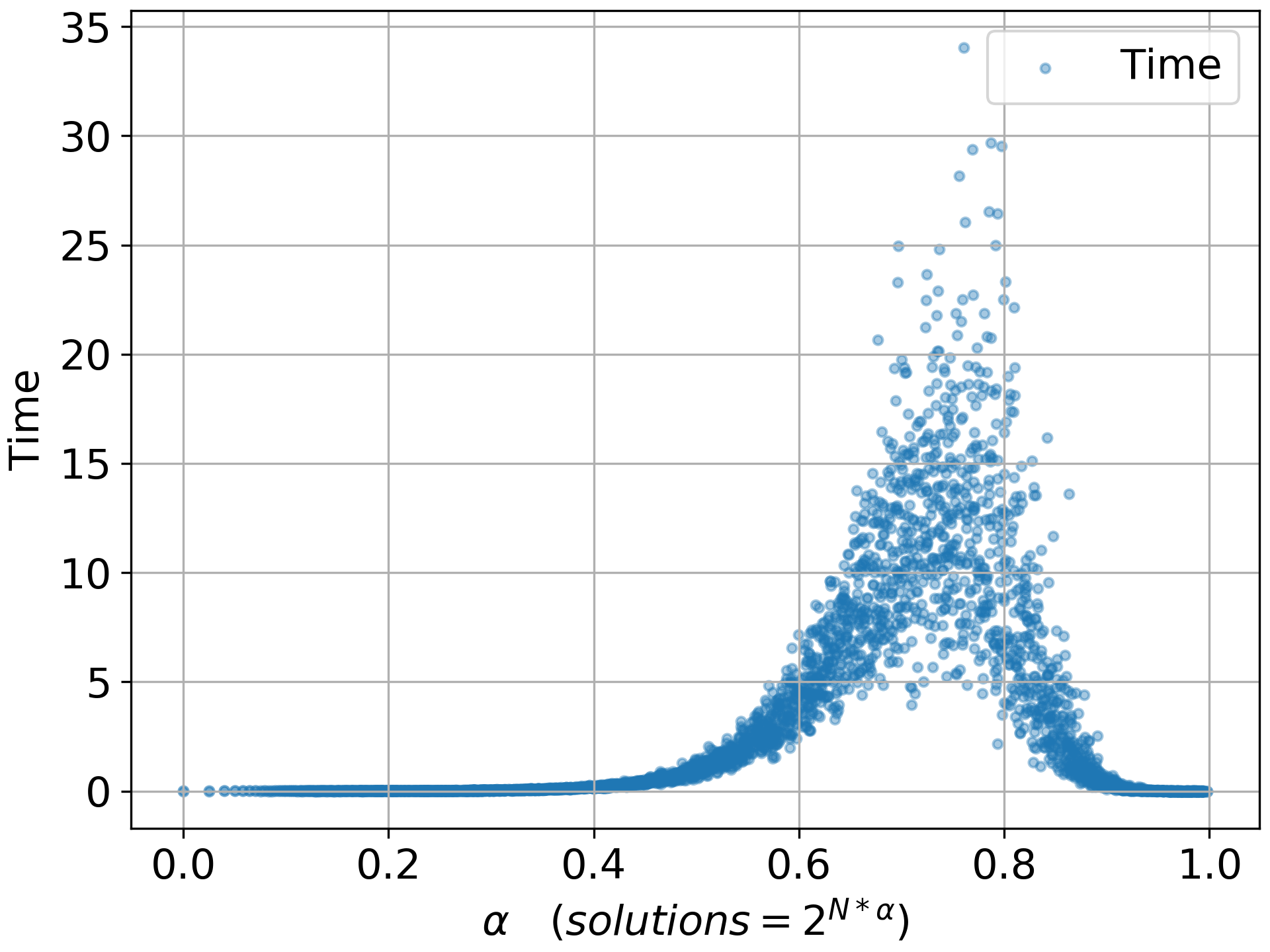}
		\caption{4-CNF and 40 vars}
		\label{fig:appendix:d4_nVa_4CNF_40vars}
	\end{subfigure}
	\hfill
	\begin{subfigure}[b]{0.30\linewidth}
		\centering
		\includegraphics[width=\textwidth]{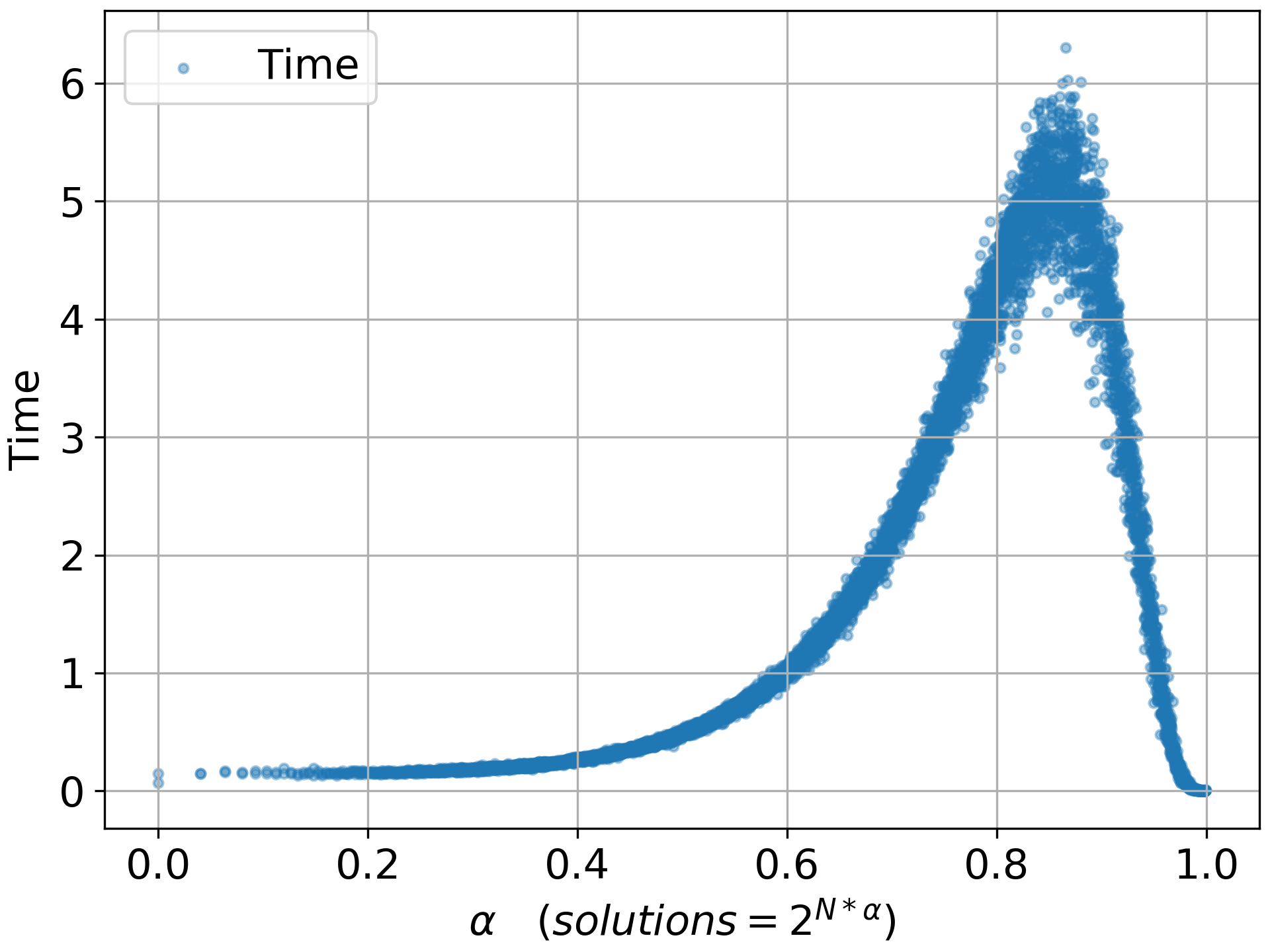}
		\caption{7-CNF and 25 vars}
		\label{fig:appendix:d4_nVa_7CNF_30vars}
	\end{subfigure}	
	\caption{\label{fig:appendix:d4_tVa_CNF}Runtime of d-DNNF compilation vs solution density for different clause lengths($k$)}
\end{figure*}

\begin{figure*}[!th]
	\centering
	\begin{subfigure}[b]{0.32\textwidth}
		\centering
		\includegraphics[width=\textwidth]{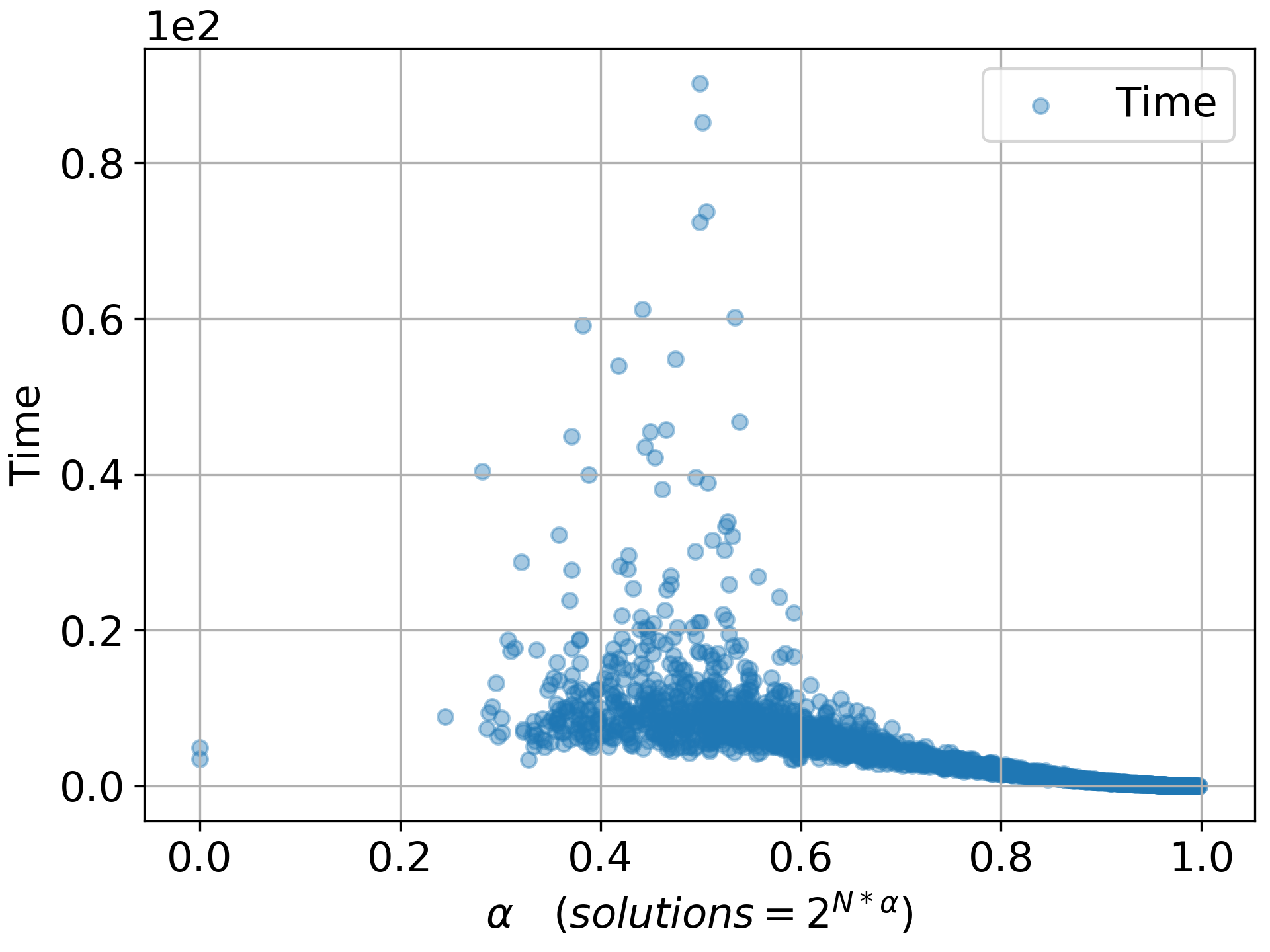}
		\caption{2-CNF and 200 vars}
		\label{fig:appendix:d4_nVa_2CNF_200vars}
	\end{subfigure}
	\hfill
	\begin{subfigure}[b]{0.32\textwidth}
		\centering
		\includegraphics[width=\textwidth]{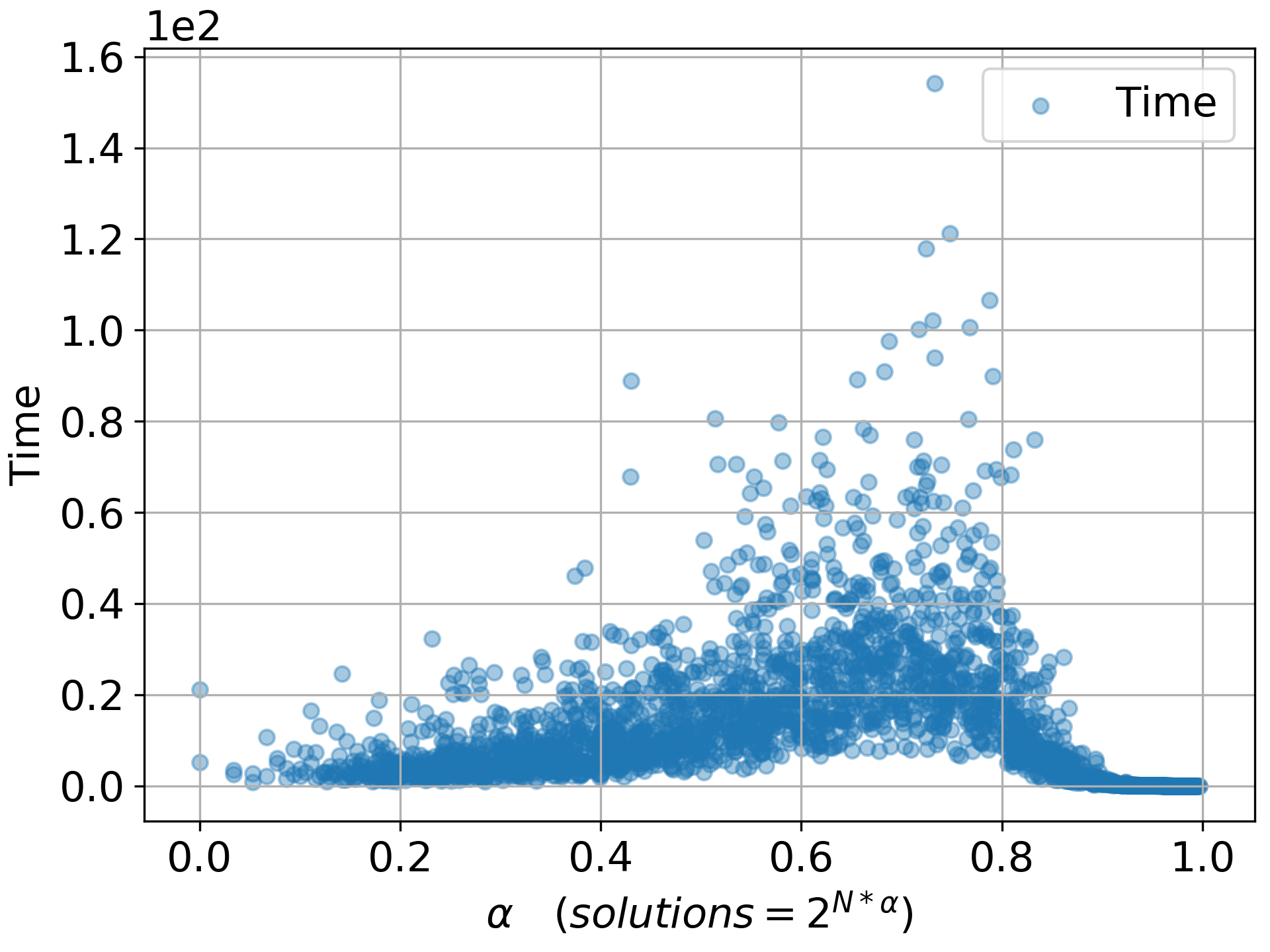}
		\caption{4-CNF and 30 vars}
		\label{fig:appendix:d4_nVa_4CNF_30vars}
	\end{subfigure}
	\hfill
	\begin{subfigure}[b]{0.32\textwidth}
		\centering
		\includegraphics[width=\textwidth]{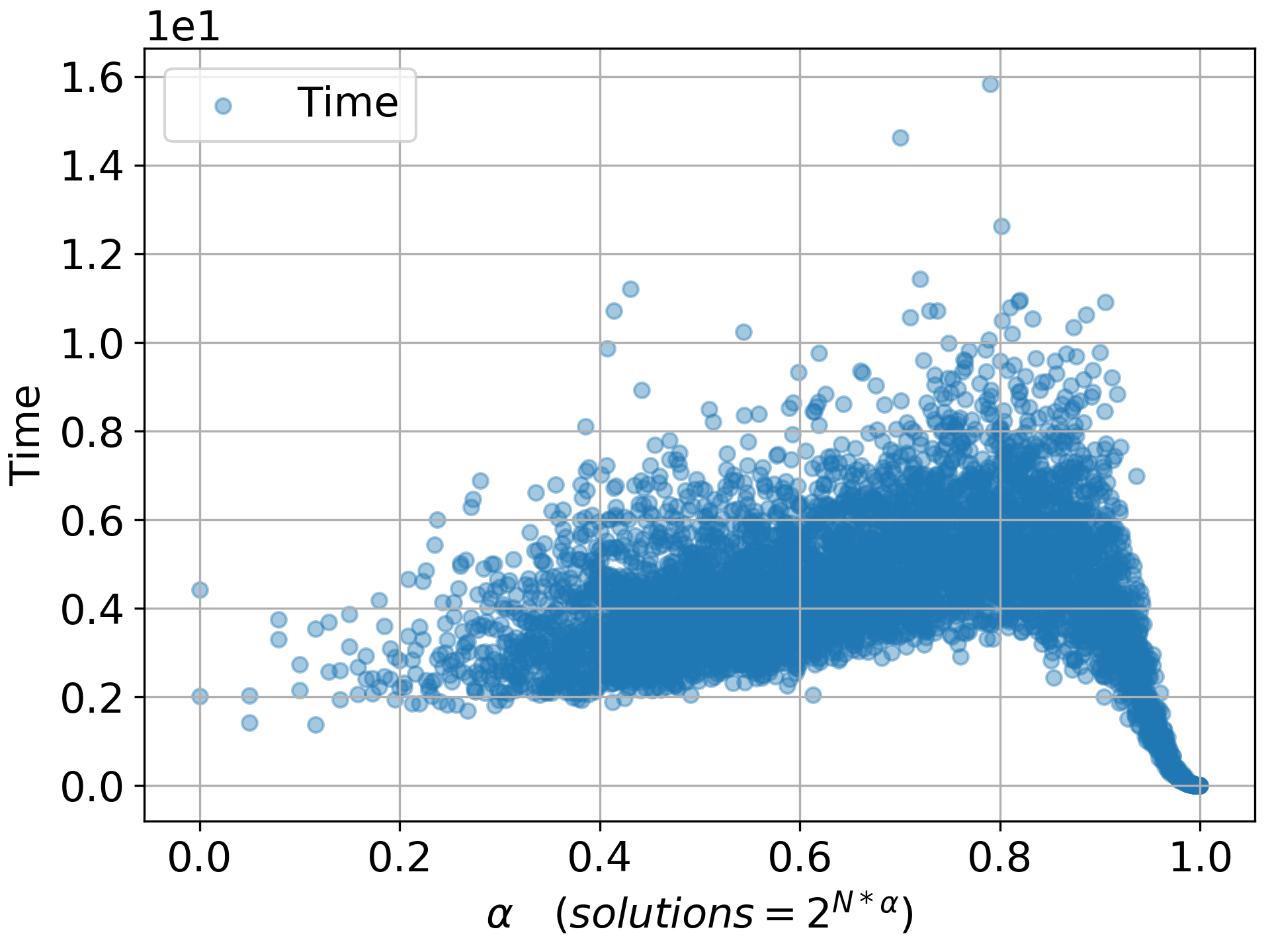}
		\caption{7-CNF and 20 vars}
		\label{fig:appendix:d4_nVa_7CNF_30vars}
	\end{subfigure}	
	\caption{\label{fig:appendix:sdd_tVa_CNF}Runtime of SDD compilation vs solution density for different clause lengths($k$)}
\end{figure*}

\begin{figure*}[t]
	\centering
	\begin{subfigure}[b]{0.32\textwidth}
		\centering
		\includegraphics[width=\textwidth]{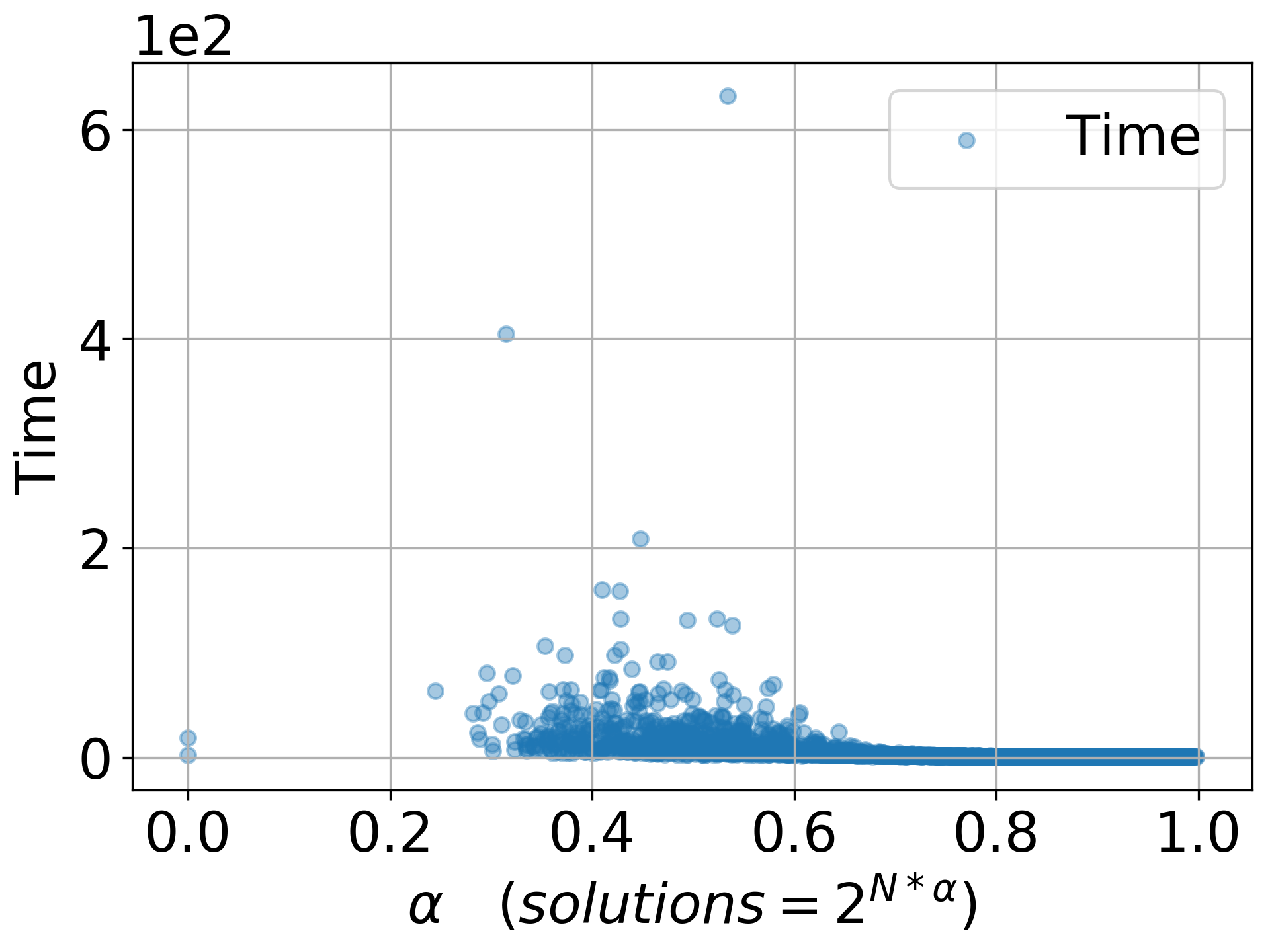}
		\caption{2-CNF and 200 vars}
		\label{fig:appendix:d4_tVa_2CNF_200vars}
	\end{subfigure}
	\hfill
	\begin{subfigure}[b]{0.32\textwidth}
		\centering
		\includegraphics[width=\textwidth]{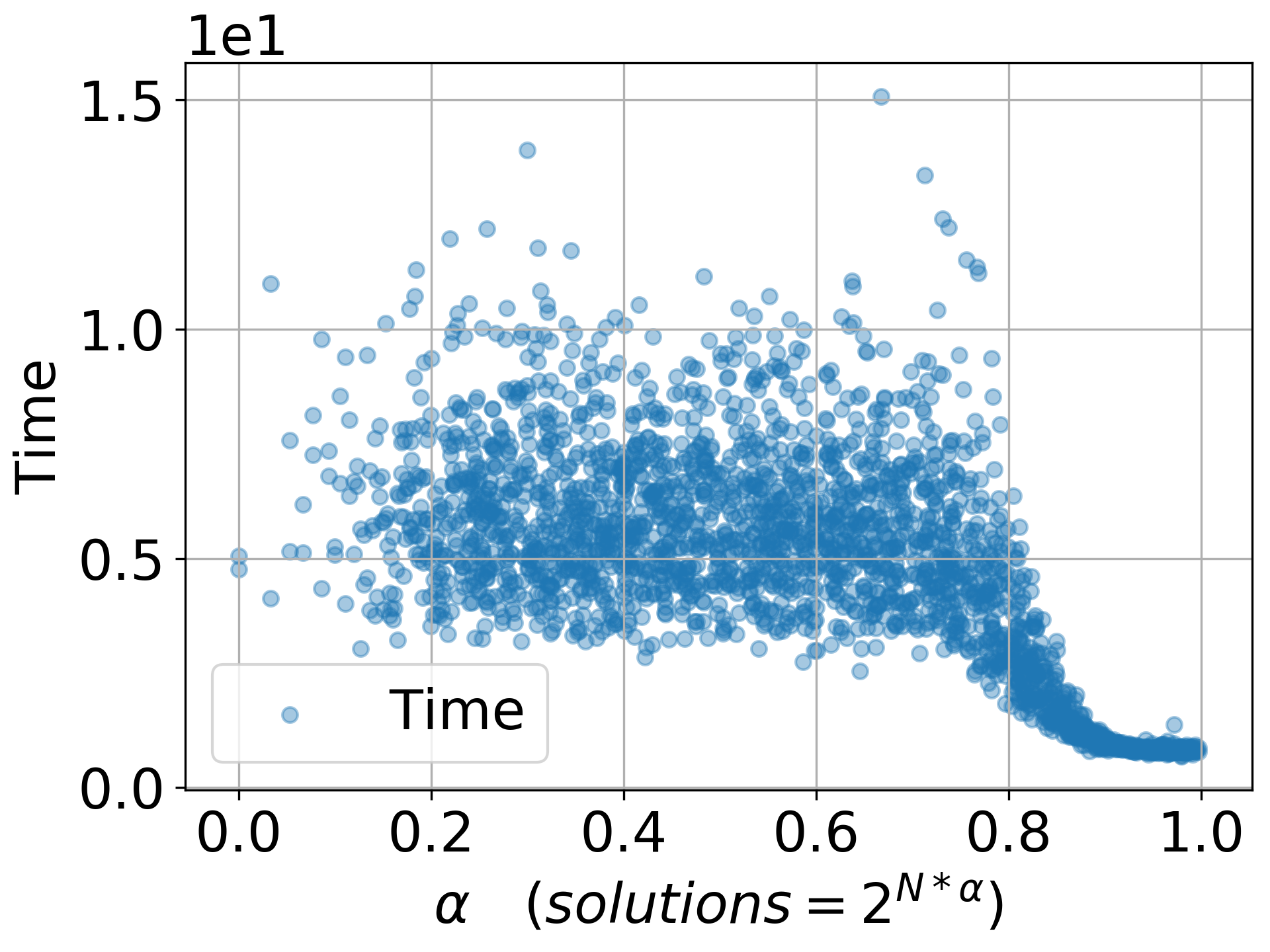}
		\caption{4-CNF and 30 vars}
		\label{fig:appendix:d4_tVa_4CNF_40vars}
	\end{subfigure}
	\hfill
	\begin{subfigure}[b]{0.32\textwidth}
		\centering
		\includegraphics[width=\textwidth]{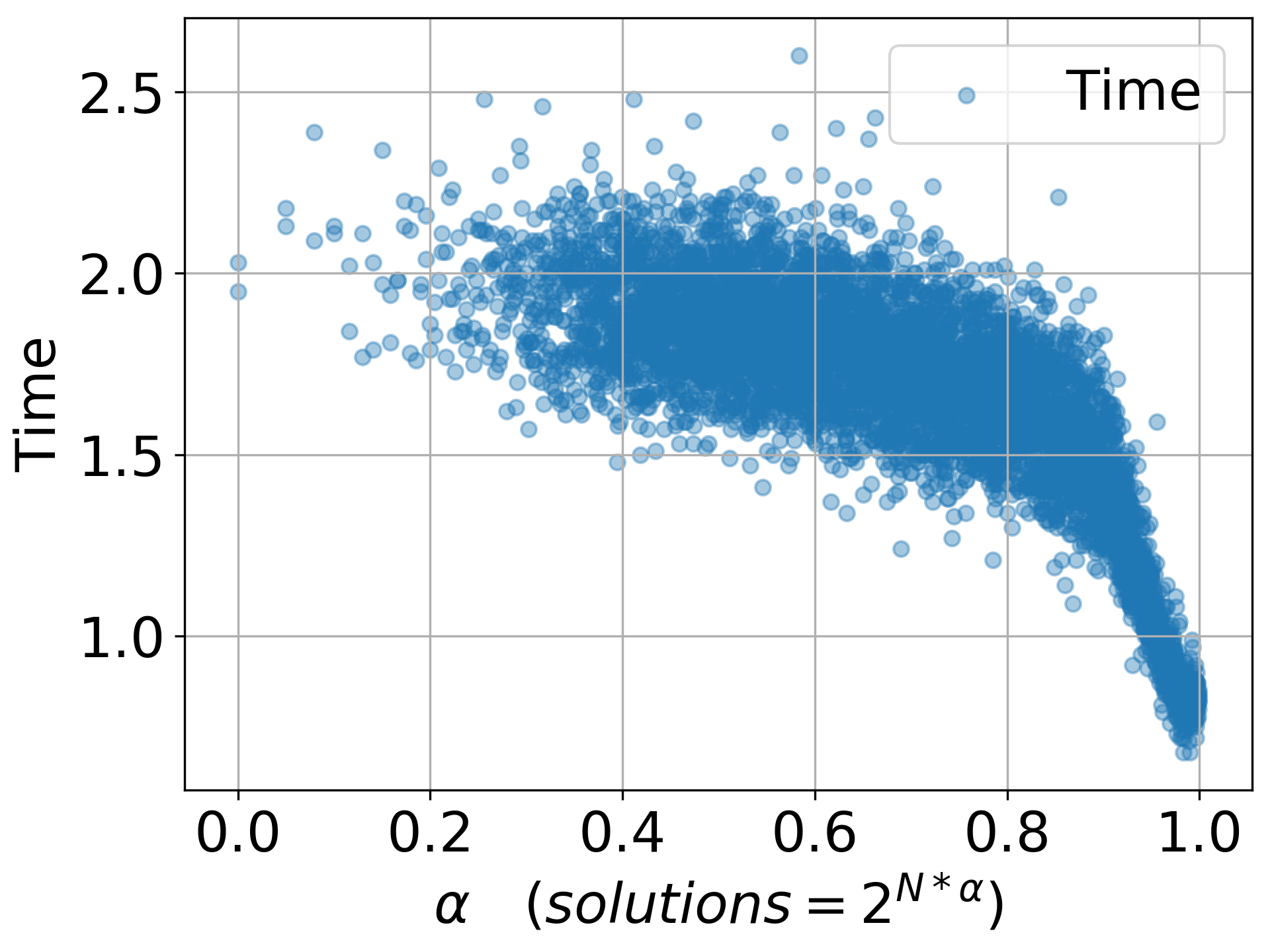}
		\caption{7-CNF and 20 vars}
		\label{fig:appendix:d4_tVa_7CNF_20vars}
	\end{subfigure}	
	\caption{\label{fig:appendix:CUDD_tVa_CNF}Runtime of OBDD compilation vs solution density for different clause lengths($k$)}
\end{figure*}

Figures~\ref{fig:appendix:d4_nVa_CNF},~\ref{fig:appendix:sdd_nVa_CNF} and~\ref{fig:appendix:CUDD_nVa_CNF} show the variation in the size with solution density for different clause lengths and number of variables. Figures~\ref{fig:appendix:d4_tVa_CNF},~\ref{fig:appendix:sdd_tVa_CNF} and~\ref{fig:appendix:CUDD_tVa_CNF} show the variation in the runtime  with solution density for different clause lengths and number of variables.

\clearpage

\subsection{Combined effect of clause and solution density}
\begin{figure*}[!th]
	\centering
	\begin{subfigure}[b]{0.4\textwidth}
		\centering
		\includegraphics[width=\textwidth]{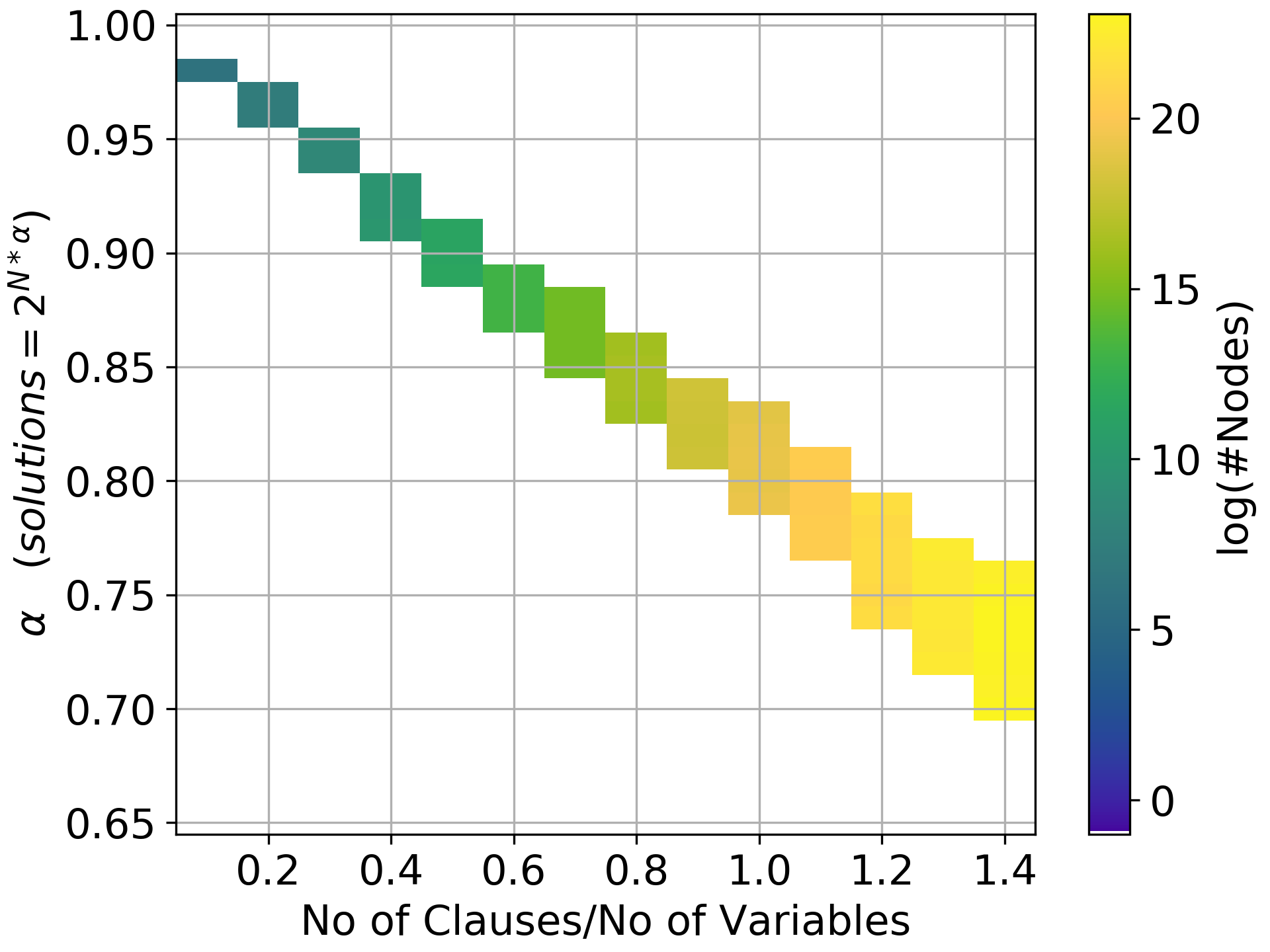}
		\caption{$r\in [0.1,1.4]$}
		\label{fig:appendix:d4_log_nVac_1d4}
	\end{subfigure}
	\hfill
	\begin{subfigure}[b]{0.4\textwidth}
		\centering
		\includegraphics[width=\textwidth]{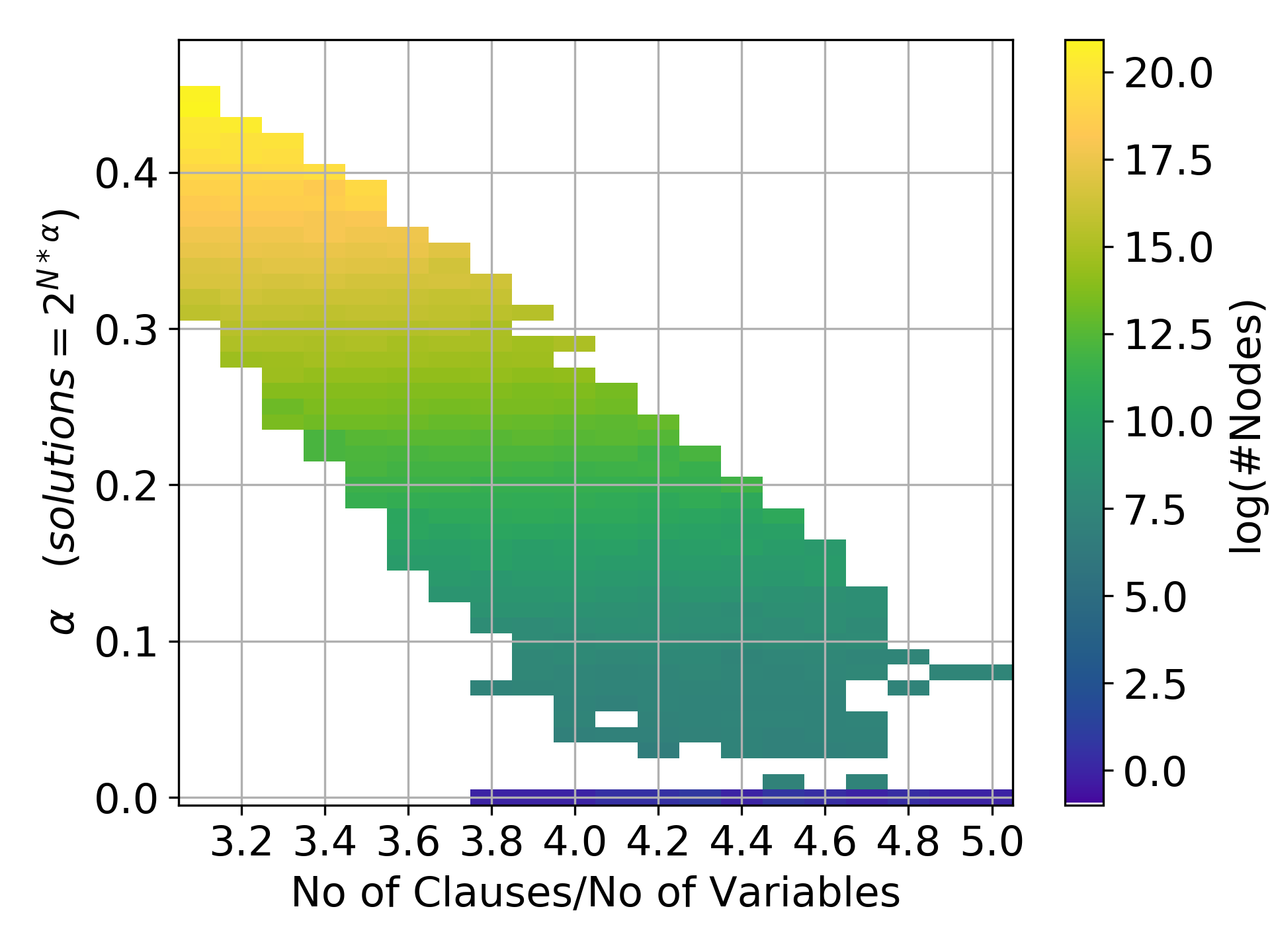}
		\caption{$r \in [3.1,5.0]$}
		\label{fig:appendix:d4_log_nVac_3d1}
	\end{subfigure}
	\caption{$\log(\#Nodes)$ in d-DNNF compilation in solution vs clause density grid for~3-CNF}
	\label{fig:appendix:d4_log_nVac}
\end{figure*}

\begin{figure*}[!th]
	\centering
	\begin{subfigure}[b]{0.4\textwidth}
		\centering
		\includegraphics[width=\textwidth]{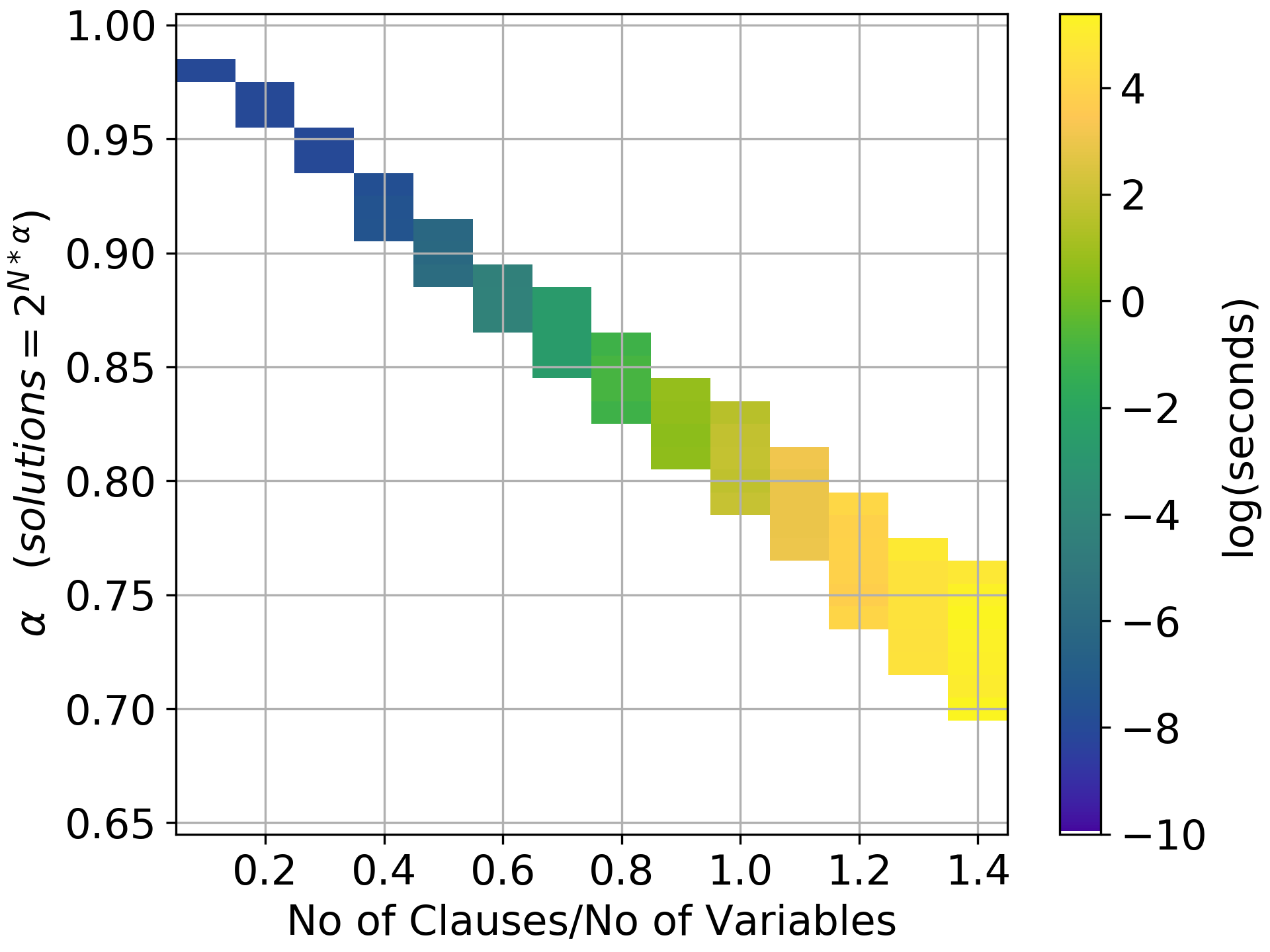}
		\caption{$r\in [0.1,1.4]$}
		\label{fig:appendix:d4_log_tVac_1d4}
	\end{subfigure}
	\hfill
	\begin{subfigure}[b]{0.4\textwidth}
		\centering
		\includegraphics[width=\textwidth]{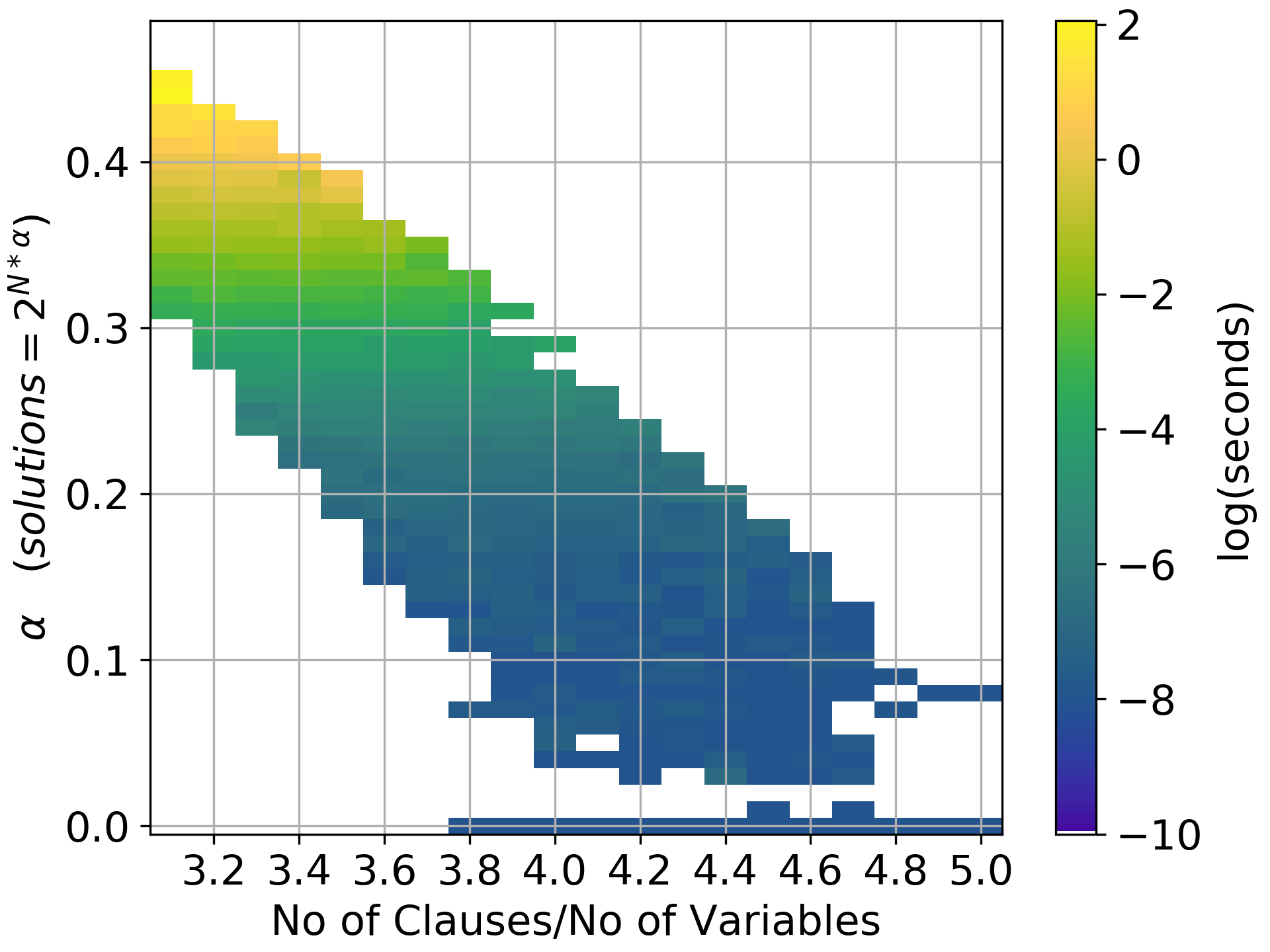}
		\caption{$r \in [3.1,5.0]$}
		\label{fig:appendix:d4_log_tVac_3d1}
	\end{subfigure}
	\caption{$\log(runtime)$ for d-DNNF compilation in solution vs clause density grid for 3-CNF}
	\label{fig:appendix:d4_log_tVac}
\end{figure*}

In Figures~\ref{fig:appendix:d4_log_nVac} and \ref{fig:appendix:d4_log_tVac}, we plot a heatmap on $\alpha \times r$ grid where the colours indicate the size and runtimes of compilations, respectively. For each cell in the grid, we take the average of instances with the corresponding clause density that lie within a small interval of solution density marked by the cell. Since the number of such instances can differ across the cells, we mark the average for a cell only if the number of instances is greater than $5$ (to minimize the variance to a feasible extent).

\end{document}